\documentclass{article}
\PassOptionsToPackage{numbers, compress}{natbib}

\usepackage[main, final]{neurips_2026}

\usepackage[utf8]{inputenc}
\usepackage[T1]{fontenc}
\usepackage{hyperref}
\usepackage{url}
\usepackage{booktabs}
\usepackage{amsfonts}
\usepackage{amsmath}
\usepackage{amssymb}
\usepackage{nicefrac}
\usepackage{microtype}
\usepackage{xcolor}
\usepackage{graphicx}
\usepackage{subcaption}
\usepackage{multirow}
\usepackage{algorithm}
\usepackage{algorithmic}
\usepackage{wrapfig}
\usepackage{tikz}
\usepackage{amsmath,amssymb}
\usepackage{fontawesome5}
\usepackage{pifont}
\usepackage{enumitem}

\newcommand{\cmark}{\ding{51}}

\newcommand{\method}{TrajLoc}

\newcommand{\RR}{\mathbb{R}}

\newcommand{\bz}{\mathbf{z}}
\newcommand{\bh}{\mathbf{h}}
\newcommand{\bQ}{\mathbf{Q}}
\newcommand{\bK}{\mathbf{K}}
\newcommand{\bV}{\mathbf{V}}
\newcommand{\bA}{\mathbf{A}}
\newcommand{\bO}{\mathbf{O}}

\title{TrajLoc: Trajectory-Attention Localization for \\ Multi-Object Motion Control}

\author{%
  Omer Sela$^{1,2}$ \quad
  Inbar Huberman-Spiegelglas$^{1}$ \\[3pt]
  \bfseries Michael Rotman$^{1}$ \quad
  Sagie Benaim$^{1,3}$ \quad
  Avi Ben-Cohen$^{1}$ \\[5pt]
  \normalfont $^{1}$Amazon Prime Video \qquad $^{2}$Tel Aviv University \qquad $^{3}$Hebrew University of Jerusalem
}

\begin{document}

\addtocontents{toc}{\protect\setcounter{tocdepth}{-1}}

\maketitle

\vspace{-0.333in}
{\centering
  \raisebox{-0.3em}{\includegraphics[height=1.35em]{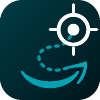}}~\href{https://sela-omer.github.io/traj-loc/}{\texttt{sela-omer.github.io/traj-loc}}\par}
\vspace{0.165in}

\begin{abstract}

Controlling the motion of multiple objects in image-to-video (I2V) generation requires preserving object identities while enforcing adherence to distinct target trajectories. This becomes particularly challenging as the number of objects increases and their paths intersect or occlude one another. Existing approaches entangle multiple trajectories within a shared, dense conditioning signal, making object-level correspondence difficult to preserve in crowded scenes. We depart from this paradigm and enforce a strict, per object spatial constraint that isolates instances independently. Our method, \method{}, achieves this directly within the attention layers by substituting the cross-attention weights of each object token with a Gaussian heatmap centered on its target location at every frame. The same per object token interface carries trajectory and depth through a learned embedding and preserves identity by encoding first frame appearance in place of an object token. Evaluations across six datasets, featuring up to 20 simultaneously controlled objects and out of distribution real world scenes, demonstrate that our method consistently improves both visual fidelity and trajectory adherence. Applied to two architecturally distinct backbones (CogVideoX 5B and WaN 2.1 14B), our approach achieves average gains of +4.3 dB PSNR and a 51\% reduction in trajectory end point error compared to the strongest baselines.

\end{abstract}
\section{Introduction}
\label{sec:intro}

Modern video diffusion models generate highly realistic videos~\citep{cogvideox,wan,opensora}, yet controlling the motion of \emph{multiple} interacting objects, each following a distinct trajectory while preserving identity, remains a significant challenge. Generating videos with precise object control is essential for applications such as simulation and synthetic data generation for autonomous driving~\citep{hu2023gaia1,wang2024drivedreamer,hassan2025gem}, robot learning~\citep{du2023unipi,liang2024dreamitate}, and professional video editing~\citep{xing2025motioncanvas,wang2025cinemaster}, where multiple objects must follow prescribed motion patterns while maintaining visual consistency. As the number of controlled objects grows and trajectories begin to interact, preserving object-level correspondence becomes increasingly difficult, often leading to identity confusion and motion drift.

\begin{figure*}
\vspace{-6pt}
\centering
\setlength{\tabcolsep}{0.8pt}
\renewcommand{\arraystretch}{0.2}
\newcommand{\teaserimg}[1]{\includegraphics[width=0.248\textwidth]{#1}}
\newcommand{\teaserframesarrow}{%
  \begin{tikzpicture}[baseline=(mid.base)]
    \draw[->, line width=0.5pt] (0,0) -- (0.735\textwidth, 0);
    \node[fill=white, inner sep=1pt] (mid) at (0.3675\textwidth, 0) {\normalsize frames};
  \end{tikzpicture}%
}
\newcommand{\teasercondition}{%
\makebox[0.248\textwidth]{\normalsize condition}%
}         
      
\begin{tabular}{@{}c@{\hspace{6pt}}ccc@{}} 
\teasercondition & \multicolumn{3}{c}{\teaserframesarrow} \\[1pt]
\teaserimg{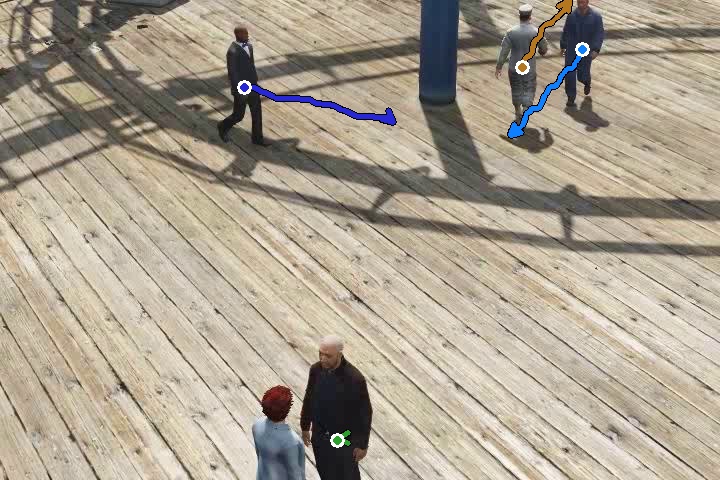} &
\teaserimg{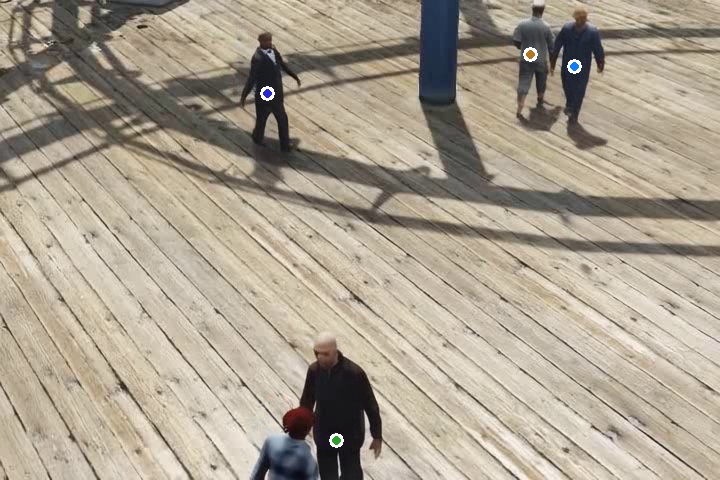} &
\teaserimg{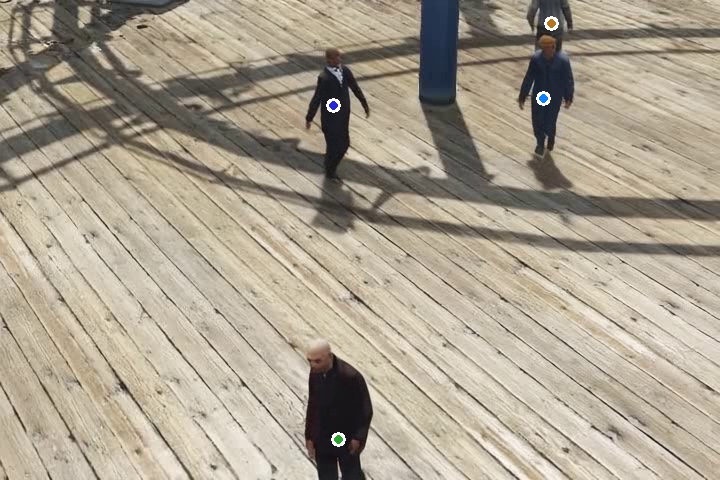} &
\teaserimg{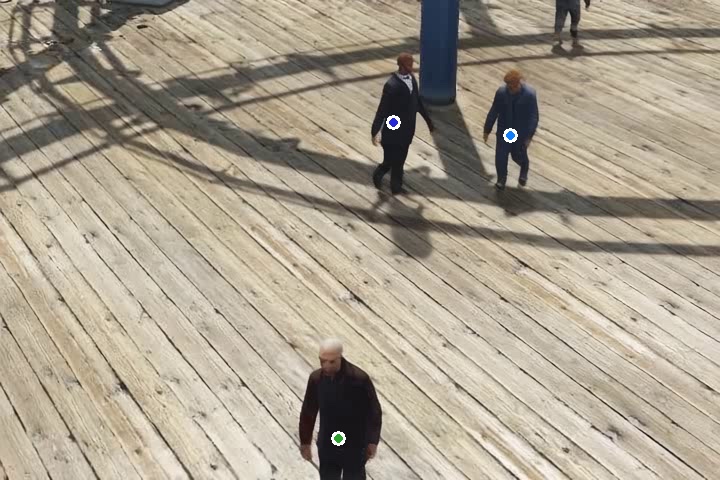} \\[0.8pt]
\teaserimg{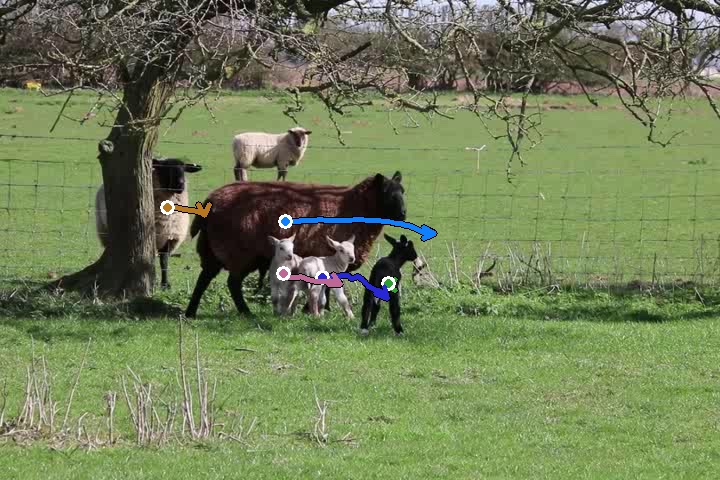} &
\teaserimg{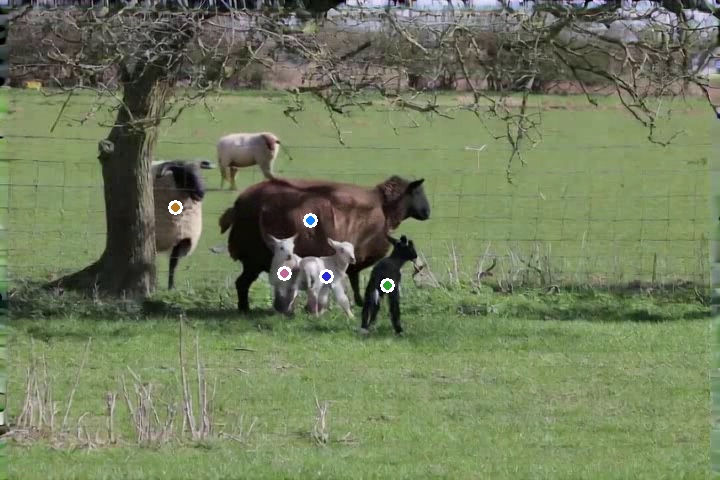} &
\teaserimg{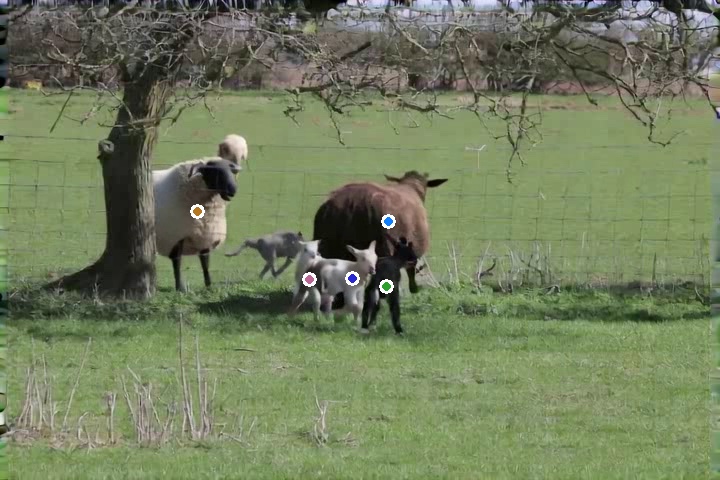} &
\teaserimg{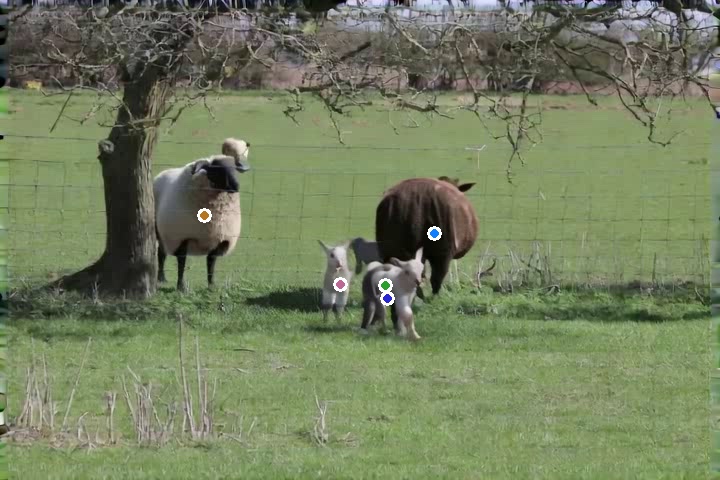} \\
\end{tabular}
\caption{\textbf{\method{}.} Given a first frame and a set of target trajectories (left column, with colored polylines), the goal is to generate a video that moves each object along its prescribed path while preserving its visual identity. Top: multiple pedestrians on a synthetic urban scene. Bottom: sheep in a natural outdoor scene. The remaining columns show three uniformly spaced generated frames with the ground-truth position of each object marked as a colored dot. In both scenes, our method places each object on its target dot, showing accurate trajectory adherence and consistent appearance throughout the video (best viewed in color).}  
\label{fig:teaser}
\vspace{-14pt}
\end{figure*}

While recent work has made progress on trajectory-conditioned generation~\citep{zhang2024tora,wanmove,ati,intragen,wu2024draganything,wang2024motionctrl}, existing methods struggle to scale to crowded scenes with many moving objects and are therefore typically evaluated on only one to three objects with limited interactions. This limitation is tied to how trajectory control is represented. The user input is inherently sparse: each object trajectory is specified by only a 2D coordinate per frame, requiring at most $2T$ scalar values for a video of $T$ frames. Existing methods inflate this into dense tensors of size up to $T \!\times\! H \!\times\! W \!\times\! C$, whether through rasterized motion patches~\citep{zhang2024tora}, latent feature propagation~\citep{wanmove,ati}, rendered trajectory videos~\citep{intragen}, or dense conditioning volumes~\citep{geng2024motion,wang2024levitor}. Because multiple trajectories are fused within the same dense signal, object-level correspondence degrades under occlusions and crossings. The dense representation further inherits the full spatiotemporal footprint of the video, requiring additional trainable modules such as ControlNet encoders~\citep{controlnet} or FiLM layers~\citep{zhang2024tora,motionpro}, and reducing portability across different diffusion backbones.

Our core insight is that sparse trajectory signals are sufficient to achieve precise multi-object control when incorporated directly into the model's existing conditioning spaces, without any dense control tensors or additional condition modules. To this end, we directly modify the cross-attention map so that each object's text token attends only to video locations along its prescribed trajectory and is attenuated everywhere else. Beyond spatial localization, the cross-attention heatmaps alone cannot convey depth ordering or visual identity. 
We address this by injecting learned per-object representations directly into the text conditioning space.
  
Specifically, \method{} replace the cross-attention weight map of each object's text token with Gaussian heatmaps centered on the target locations at every frame, providing a hard, per-object spatial constraint that scales naturally since each object is handled independently. We further enrich the same conditioning spaces with information beyond 2D position, each object is represented by two dedicated text tokens: a trajectory token encoding motion and depth, and an appearance token encoding first-frame visual identity in place of the generic category word. Our method applies to both separate cross-attention architectures (WaN) and joint self-attention designs (CogVideoX) without architecture-specific modifications.

We evaluate the approach on CogVideoX-5B-I2V and WaN~2.1-14B-I2V across six datasets spanning synthetic objects, billiard scenes, pedestrian crowds with up to 20 interacting objects, and real-world surveillance footage (MOT17) and diverse real-world scenes (DAVIS) outside the training distribution. Across all six datasets and both backbone families, our method sets a new state of the art, leading on both reconstruction quality (PSNR) and trajectory fidelity (end-point error) in every backbone group on every dataset. Average gains are $+4.3$~dB PSNR ($0.7$--$10$~dB range) and a $51\%$ average reduction in end-point error ($11$--$96\%$ range) relative to the strongest same-backbone baseline per metric and dataset.
In summary, our contributions are:
\begin{enumerate}[label=(\roman*), leftmargin=2em, nosep]
\item We introduce attention localization, a mechanism that replaces cross-attention weight columns with trajectory-aligned Gaussian heatmaps, providing a hard per-object spatial constraint without dense conditioning tensors or additional modules.
\item We encode each object's trajectory and appearance as dedicated tokens in the text prompt, conveying depth and visual identity through the same conditioning space the attention mechanism already operates on.
\item Since each object's motion, identity, and spatial attention are handled independently, our method scales naturally to crowded scenes with many interacting objects and preserves per-object control under occlusions and trajectory crossings.
\item We demonstrate state-of-the-art results on six datasets with up to 20 simultaneous objects across two architecturally distinct backbones, trained exclusively on synthetic data yet generalizing to real-world scenes.
\end{enumerate}

\section{Related Work}
\label{sec:related}

\noindent\textbf{Dense trajectory conditioning for video generation} \quad The dominant approach converts sparse point inputs into dense spatiotemporal tensors. Tora~\citep{zhang2024tora} rasterizes trajectories into motion patches via a 3D VAE,
injected via adaptive-norm (FiLM-style) modulation. MotionCtrl~\citep{wang2024motionctrl} and MotionPro~\citep{motionpro} construct dense displacement fields processed by convolutional modules. A parallel family uses ControlNet-style encoders:
Motion Prompting~\citep{geng2024motion} conditions on sinusoidal track volumes, LeviTor~\citep{wang2024levitor} on dense heatmap volumes with depth and instance channels, MagicMotion~\citep{li2025magicmotion} on VAE-encoded
trajectory maps, and PoseTraj~\citep{posetraj} on rendered pose-aware trajectory maps with 3D bounding-box supervision. DragAnything~\citep{wu2024draganything} and DragEntity~\citep{dragentity} inject entity representations via ControlNet-style encoders, while
InTraGen~\citep{intragen} renders and VAE-encodes two full-resolution RGB videos. Wan-Move~\citep{wanmove} and ATI~\citep{ati} propagate first-frame features along trajectories within the existing I2V condition tensor,
adding no new parameters but still producing a video-latent-sized signal. In all cases, multiple trajectories are fused within a shared dense signal, making object-level correspondence difficult to preserve under occlusions.
Our method avoids this entirely by operating at the object-token level, where each object's motion and identity are encoded independently.

\noindent\textbf{Token-based control} \quad Boximator~\citep{wang2024boximator} uses bounding-box sequences as input, injecting them as tokens via GLIGEN-style~\citep{gligen} self-attention layers, and relies on the model to learn box-to-object correspondence from data rather than enforcing it, a task that becomes harder as the number of overlapping objects grows. FlexTraj~\citep{flextraj} concatenates VAE-encoded condition videos as tokens, though these are still derived from dense video-resolution maps. SG-I2V~\citep{namekata2024sg} is a zero-shot approach that enforces cross-frame feature similarity within bounding-box regions by replacing self-attention keys and values with first-frame features, requiring no training but limited to coarse bounding-box control. In contrast, our method provides fine-grained point-level trajectory control through learned per-object tokens and hard attention replacement, scaling to tens of simultaneous objects.

\noindent\textbf{Cross-attention manipulation in diffusion models} \quad Editing attention maps for spatial control has been explored in the image domain. Prompt-to-Prompt~\citep{prompt2prompt} edits cross-attention for image editing, Attend-and-Excite~\citep{attendandexcite} optimizes latents to strengthen cross-attention at subject positions, and Peekaboo~\citep{jain2024peekaboo} masks spatiotemporal attention for video region control. Training-free video methods follow a similar principle. TrailBlazer~\citep{ma2024trailblazer} scales cross-attention maps over bounding-box keyframes, FreeTraj~\citep{qiu2024freetraj} combines trajectory-shaped noise with attention guidance, and Motion-Zero~\citep{motionzero} optimizes latents against box-alignment losses on cross-attention. For temporal rather than spatial control, Mind the Time~\citep{wu2025mindthetime} learns a time-aware positional encoding that biases cross-attention toward the event active in each segment of a text-to-video generation. All of these provide soft guidance signals. Our approach differs in that we replace attention weight columns with spatial distributions, providing a hard constraint applied at every layer during both training and inference.

\section{Method}
\label{sec:method}

Given a first frame containing $N$ objects and a target trajectory for each, we generate a video such that every object moves along its prescribed path while preserving its visual identity. Each object $i$ is identified by a semantic category $o_i$ (e.g., \texttt{ball}, \texttt{girl}) and a discretized trajectory $\tau_i(t) = (x_i(t), y_i(t), d_i(t))$ for $t = 1, \dots, T$ frames, where $(x_i(t), y_i(t))$ denotes the center-of-mass location at frame $t$ and $d_i(t)$ the depth relative to the screen plane under a fixed camera geometry.
   
We build upon a pretrained image-to-video (I2V) model conditioned on text and the first frame.
In a standard I2V pipeline, video tokens attend to all text tokens without spatial bias, providing no explicit signal about \textit{where} each referenced object should appear at each frame. To address this, we express trajectory control entirely within the existing conditioning spaces of the pretrained model, avoiding any external dense signals.                     
We achieve this through two levels of intervention. Spatially, we substitute the learned cross-attention weights of each object token $o_i$ with spatiotemporal Gaussian heatmaps derived from $\tau_i(t)$, imposing a hard per-object spatial constraint at every frame (Sec.~\ref{sec:xattn_replace}). While this forces the model to attend to the correct regions, spatial heatmaps alone cannot convey depth or visual identity. Semantically, we address this gap by modifying the text conditioning space. We organize the inputs into a structured textual instruction such that each object $i$ is described with its semantic category $o_i$, a verb, and a trajectory token, 
 \begin{flalign}                                             
 & \texttt{"Scene where } o_0 \texttt{ moves [traj}_0\texttt{] and } o_1 \texttt{ moves [traj}_1\texttt{] ..."} & \label{eq:prompt}                                    
 \end{flalign}
  
To capture motion and depth $d_i(t)$ we replace the placeholder tokens [traj$_i$] with trajectory embeddings (Sec.~\ref{sec:traj_encoding}). To preserve the appearance of each specific instance in the input image we replace the semantic category $o_i$ embedding with visual features extracted from the first frame (Sec.~\ref{sec:appear_encoding}). Both encoders are learned through the training procedure detailed in Sec.~\ref{sec:training}. Unlike dense conditioning approaches, our representation scales with the number of objects $N$ rather than the video resolution. The complete pipeline is illustrated in Figure~\ref{fig:pipeline}.

\subsection{Attention Localization} \label{sec:xattn_replace} 

The trajectory $\tau_i(t)$ of each object is used to steer the model's attention toward specific spatiotemporal locations. This is achieved by substituting the cross-attention weight column at the object token index with a spatiotemporal Gaussian heatmap centered on the target trajectory across all attention layers.

\begin{figure}[t]
    \vspace{-4pt}
    \centering
    \includegraphics[width=\textwidth]{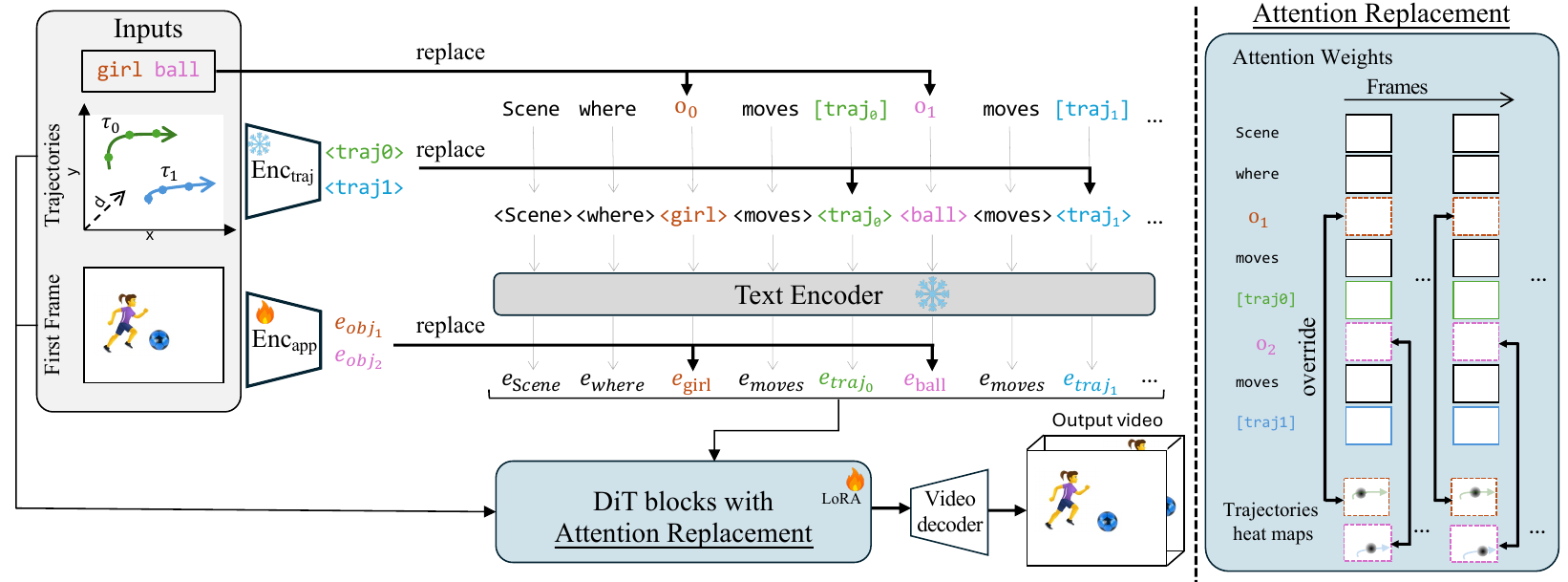}
    \vspace{-16pt}
        \caption{\textbf{An overview of \method{}.} A structured text prompt ``Scene where $o_0$ moves $[\text{traj}_0]$, $o_1$ moves $[\text{traj}_1]$, \ldots'' is constructed, where each $o_i$ is the object's given category name. The trajectory tokens $[\text{traj}_i]$ are replaced with learned embeddings from the pretrained (frozen) $\mathrm{Enc}_{\text{traj}}$, which independently encodes each target trajectory $(x_i(t), y_i(t), d_i(t))$.
        The learned appearance encoder $\mathrm{Enc}_{\text{app}}$ encodes each object's appearance from the first frame into an embedding that replaces the corresponding object token. The resulting conditioning is fed to an I2V DiT model, where attention localization is applied within each transformer block (right). \textbf{Right:} The standard cross-attention weights at each object category $o_i$ are overridden by trajectory-aligned Gaussian heatmaps, centered at the target coordinates $(x_i(t), y_i(t))$ at every frame.
         }
    \label{fig:pipeline}
    \vspace{-8pt}
\end{figure}

\noindent \textbf{Trajectory heatmap construction} \quad For each object $i$ we construct a spatiotemporal trajectory heatmap by placing a two dimensional isotropic Gaussian with standard deviation $\sigma$ centered at its trajectory location $(x_i(t), y_i(t))$ for every frame $t$ in which the object is visible. The resulting heatmap sequence is bilinearly resampled onto the latent spatiotemporal grid and normalized per frame to sum to one while frames in which the object is absent remain zero. Stacking these maps across all frames and vectorizing the result yields the per object heatmap $\mathbf{h}_i \in \mathbb{R}^M$ where $M$ is the total number of spatiotemporal latent tokens. As visualized in the right panel of Figure \ref{fig:pipeline} these Gaussian maps are created for every trajectory $\tau_i$ at each frame. Algorithm~\ref{alg:attn_loc} (Appendix~\ref{app:algorithm}) details this construction and the replacement below. 

\noindent \textbf{Column replacement} \quad We denote the cross-attention weight matrix $\mathbf{A}$ and the final attention output $\mathbf{O}$ as:
  \begin{equation}                                     
   \mathbf{A} = \text{softmax} \left( d^{-1/2}\,\mathbf{Q}_{\mathrm{vid}} \mathbf{K}_{\mathrm{txt}}^\top \right), \qquad \mathbf{O} = \mathbf{A}\mathbf{V}_{\mathrm{txt}}    
  \end{equation} 
  
where $\mathbf{A} \in \mathbb{R}^{M \times L}$ represents the alignment between the $M$ visual patches and $L$ text tokens. We modify this process by identifying the specific column index $q_i$ in $\mathbf{A}$ that corresponds to the object category $o_i$, i.e., the position of the token $o_i$ in the tokenized prompt of Eq.~\ref{eq:prompt}, and substituting its original weights with the spatiotemporal Gaussian heatmap $\mathbf{h}_i$. To maintain the probabilistic nature of the attention mechanism we then renormalize each row of the modified matrix $\mathbf{A}'$ such that:
  \begin{equation}                                      
   \mathbf{A}'[m, q_i] \leftarrow \mathbf{h}_i[m], \qquad \mathbf{A}'[m, :] \leftarrow \mathbf{A}'[m, :] \,/\, \sum_{k=1}^{L} \mathbf{A}'[m, k]                                                 
   \label{eq:col_replace}                                                                                                 
  \end{equation}   

where $m \in \{1, \ldots, M\}$ indexes video-token positions. The final attention output is then computed as $\mathbf{O}' = \mathbf{A}' \mathbf{V}_{\mathrm{txt}}$. This replacement ensures that the influence of the object token is strictly localized to the video regions defined by the prescribed trajectory while leaving other text tokens and their respective spatial alignments intact. As illustrated in Figure~\ref{fig:pipeline}, this mechanism overrides the learned attention weights with Gaussian heatmaps at every object token position, steering the generation toward the prescribed trajectories.

The replacement adds no new trainable parameters and no new dense conditioning tensors as it operates on values the attention mechanism already computes. The formulation above is exact for architectures with
explicit cross-attention (e.g., WaN~2.1), but infeasible for joint self-attention architectures such as CogVideoX, whose full attention matrix (approximately 30~GB per layer) exceeds GPU memory. For those architectures we derive an efficient approximation that preserves the replacement semantics, as described in Appendix~\ref{app:joint_attention}.

\subsection{Token Level Conditioning}
\label{sec:text_encoding}

While attention localization forces the model to look at the correct regions, it remains agnostic to depth and visual identity. We resolve this by modifying the text conditioning embeddings of the structured instruction described in Eq.~\ref{eq:prompt}. The prompt is passed through a tokenizer and then a frozen text encoder to produce the initial conditioning embeddings, as illustrated in Figure~\ref{fig:pipeline}. We apply two targeted replacements: the trajectory placeholder [traj$_i$] is replaced with a learned trajectory embedding  (Sec.~\ref{sec:traj_encoding}) and the semantic category $o_i$ is replaced with a learned visual encoding from the first frame (Sec.~\ref{sec:appear_encoding}).                                   
\begin{figure}[t]                                      
\vspace{-4pt}
\centering                                             
\resizebox{\linewidth}{!}{%
\begin{tikzpicture}[arrow/.style={->,thick,>=stealth}]                                                                                  
  \node[font=\normalsize](input)at(0,0){$\tau_i(t)$};                                  
  \node[font=\small,text=gray]at(0,-0.6){$(x_i,y_i,d_i,\mathrm{pos}_i)$};                                                                     
  \node[font=\normalsize](enclabel)at(2.8,0){$\mathrm{Enc}_{\text{traj}}$\,\textcolor{red!70!yellow}{\faFire}};                                                                                     
  \draw[thick,fill=white](1.8,-0.5)--(1.8,0.5)--(3.8,0.35)--(3.8,-0.35)--cycle;                                                            
  \node[font=\normalsize]at(2.8,0){$\mathrm{Enc}_{\text{traj}}$\,\textcolor{red!70!yellow}{\faFire}};                                                                                               
  \node[font=\normalsize](emb)at(5.2,0){$\langle traj_i \rangle$};                                                                      
  \node[draw,thick,fill=gray!15,rounded corners=4pt,minimum height=1cm,minimum width=2cm,align=center,font=\normalsize](te)at(7.8,0){Text Encoder\,\textcolor{cyan!70!blue}{\faSnowflake}};         
  \draw[thick,fill=white](9.8,-0.35)--(9.8,0.35)--(11.8,0.5)--(11.8,-0.5)--cycle;                                                        
  \node[font=\normalsize]at(10.8,0){$\mathrm{Dec}_{\text{traj}}$\,\textcolor{red!70!yellow}{\faFire}};                                                                                              
  \node[font=\normalsize](output)at(12.8,0){$\hat{\tau}_i(t)$};                                                                            
  \draw[arrow](input)--(1.8,0);                                                    
  \draw[arrow](3.8,0)--(emb);                                                      
  \draw[arrow](emb)--(te);                                                          
  \draw[arrow](te)--(9.8,0);                                                       
  \draw[arrow](11.8,0)--(output);                                                
  \end{tikzpicture}}                                                                                                                   
\caption{\textbf{Trajectory autoencoder pretraining.} The trajectory encoder maps each object trajectory $\tau_i(t) = (x_i(t), y_i(t), d_i(t))$ and a temporal position channel to a token embedding $\langle\text{traj}_i\rangle$ in the text encoder space. The embedding passes through the frozen text encoder before a decoder reconstructs the original trajectory, ensuring the representation remains informative after text encoder processing.}                           
\label{fig:traj_encoder}                                
\vspace{-10pt}
\end{figure}            

\subsubsection{Trajectory Encoding}
\label{sec:traj_encoding}

The trajectory encoder $\mathrm{Enc}_{\text{traj}}$ maps each trajectory $\tau_i$ to a single token embedding $\langle\text{traj}_i\rangle \in \mathbb{R}^{4096}$ that replaces the corresponding placeholder in the text prompt.         
Each trajectory is represented as a $T \times 4$ sequence of per-frame coordinates $(x_i(t), y_i(t), d_i(t))$ and a temporal position channel. A lightweight convolutional encoder maps this sequence to $\langle\text{traj}_i\rangle$, matching the text encoder's token dimension. A key requirement is that the trajectory information survives the frozen text encoder's self-attention layers. We therefore pre-train $\mathrm{Enc}_{\text{traj}}$ jointly with a decoder that reconstructs $\tau_i$ from the text encoder's output at the corresponding token position (Figure~\ref{fig:traj_encoder}). The reconstruction loss is      
\begin{equation}                                                                  \mathcal{L}_{\text{traj}} = \text{MSE}(\hat{\tau}_i, \tau_i) + \lambda_{\text{vel}}\,\text{MSE}(\Delta\hat{\tau}_i, \Delta\tau_i),               \label{eq:traj_vae_loss}                                                          \end{equation}                                                                    where $\hat{\tau}_i$ is the reconstructed trajectory and $\Delta\tau_i \equiv \tau_i^t - \tau_i^{t-1}$ penalizes velocity errors. After pretraining, the decoder is discarded and $\mathrm{Enc}_{\text{traj}}$ is frozen. Architectural details are provided in Appendix~\ref{app:traj_encoder}.

\subsubsection{Appearance Encoding}
\label{sec:appear_encoding}

A semantic category label such as \texttt{girl} in the text prompt does not distinguish between specific instances. To preserve each object's appearance as seen in the first frame, we replace the text token embedding of each category $o_i$ with a visual encoding extracted from the input image.
  
The appearance encoder $\mathrm{Enc}_{\text{app}}$ takes the first-frame VAE latent and extracts a feature at each object's initial position $(x_i^1, y_i^1)$, the first entry of its prescribed trajectory $\tau_i$, using a lightweight convolutional network, then projects it into the text embedding space to produce $e_{\text{o}_i}$. This embedding replaces the corresponding object token representation after the text encoder, as illustrated in Figure~\ref{fig:pipeline}. The output is normalized to match the statistics of the text encoder output space. Architectural details are provided in Appendix~\ref{app:app_encoder}.

\subsection{Training}
\label{sec:training}

We first pre-train the trajectory encoder using the reconstruction objective of Eq.~\ref{eq:traj_vae_loss}, which requires only trajectories and the first frame.
We then fine-tune a pretrained I2V diffusion backbone using LoRA~\citep{lora}, jointly training the appearance encoder while keeping the trajectory encoder frozen.
The diffusion training uses full video clips and combines the standard flow-matching loss with a trajectory-focused reconstruction term. Specifically, let $\mathcal{L}_\text{diff}$ denote the mean squared error of the noise prediction over all spatial positions, and $\mathcal{L}_\text{bbox}$ the corresponding error restricted to bounding boxes centered along each trajectory. The combined loss is:
\begin{equation}
    \mathcal{L}
    =
    (1-\alpha)\mathcal{L}_\text{diff}
    +
    \alpha\mathcal{L}_\text{bbox},
    \label{eq:diffusion_loss}
\end{equation}
with $\alpha$ balancing accurate object placement against global scene coherence. During training, attention columns are overwritten using heatmaps generated from the ground-truth trajectories, so the model learns to generate videos \emph{given} the forced attention pattern. At inference, the model needs only the first frame plus a category name and a trajectory for each object, with depth optional (Sec.~\ref{sec:ablations}).

\section{Experiments}
\label{sec:experiments}

 We evaluate \method{} on two architecturally distinct I2V backbones across six datasets, comparing against four state-of-the-art trajectory-conditioned baselines. We first describe the experimental setup (Sec.~\ref{sec:setup}), then report quantitative results (Sec.~\ref{sec:quant_results}), scaling with object count and trajectory crossings (Sec.~\ref{sec:scaling}), qualitative comparisons (Sec.~\ref{sec:qual_results}), and ablations (Sec.~\ref{sec:ablations}).

\subsection{Experimental setup}
\label{sec:setup}

\noindent  \textbf{Implementation details} \quad
We apply our method to two I2V backbones: WaN~2.1-14B, where attention localization is applied across all 40 transformer layers, and CogVideoX-5B, where we use our approximation (Appendix~\ref{app:two_sdpa}) across all 42 layers. In both cases we adapt the attention layers with LoRA (rank 64, $\alpha_{\mathrm{LoRA}}{=}64$) and train Eq.~\ref{eq:diffusion_loss} with bounding-box loss weight $\alpha{=}0.5$ jointly with the appearance encoder while all other parameters are frozen. Trajectory heatmaps use $\sigma{=}30$ pixels at video resolution. Additional implementation details, including diffusion fine-tuning hyperparameters and training budget, are provided in Appendix~\ref{app:additional_impl}. Code and our real-data checkpoints (Appendix~\ref{app:sav}) are available at \url{https://github.com/Sela-Omer/traj-loc}.
  
\noindent  \textbf{Training data} \quad
We train on ${\sim}$25K static-camera clips drawn from four publicly available synthetic datasets, as no real-world dataset provides static-camera multi-object trajectories at the required scale. The sources are MOTSynth~\citep{motsynth} (GTA~V urban scenes with up to 20 pedestrians), and three subsets from the ViN suite~\citep{intragen}: MoVi-Extended (Kubric-rendered scenes with 2--12 geometric objects), Pool (overhead billiard scenes with 2--3 balls), and Football (synthetic soccer scenes with up to 2 players and a ball). All datasets provide ground-truth trajectories. Depth is derived from world coordinates for MOTSynth and estimated with Depth Anything V2~\citep{yang2024depth} for ViN. All clips are 49 frames at $720 \!\times\! 480$ with a static camera. A detailed breakdown of the training data is provided in Appendix~\ref{app:train_details}.

\noindent  \textbf{Evaluation data} \quad
We evaluate on 446 held-out samples across six datasets. Four are in-distribution: 100 clips each from the MoVi-Extended, Pool, and Football ViN test splits, and 100 MOTSynth clips from 65 held-out scenes. Two are out-of-distribution (OOD): 33 real-world pedestrian clips from MOT17~\citep{mot17}, filtered to static-camera sequences, and 13 scenes from DAVIS 2017~\citep{pont2017davis} covering animals, vehicles, sports, and other categories absent from training, filtered for camera motion as detailed in Appendix~\ref{app:davis}. All evaluation clips use a static camera and match the training resolution and frame count. Test samples are filtered so that every tracked object is visible in the first frame, giving every method a well-defined first-frame appearance input.

\noindent  \textbf{Baselines} \quad
We compare against four recent state-of-the-art trajectory-conditioned methods, each of which turns the sparse trajectory input into a dense video-sized conditioning signal. On the Wan2.1-14B backbone, \textbf{Wan-Move}~\citep{wanmove} populates the I2V condition latent with first-frame VAE features at trajectory positions, and \textbf{ATI}~\citep{ati} does the same with Gaussian-weighted blending across each trajectory. On the CogVideoX-5B backbone, \textbf{Tora}~\citep{zhang2024tora} rasterizes trajectories into motion patches via a 3D VAE and convolutional encoder, and \textbf{MagicMotion}~\citep{li2025magicmotion} renders bounding-box trajectory maps and processes them through a ControlNet copy of the backbone. All baselines use their released default inference configurations and per-video captions (Appendix~\ref{app:inference},~\ref{app:captions}), providing richer scene descriptions than our template-based prompts. For ATI and Wan-Move, whose released models output 81 frames, we patch the frame count to 49 to match our evaluation protocol. Appendix~\ref{app:81f_sanity} confirms this patch does not disadvantage them relative to their native 81-frame output.

\noindent  \textbf{Evaluation metrics} \quad
We report four complementary metrics. PSNR measures pixel-level reconstruction quality. LPIPS~\citep{zhang2018perceptual} captures perceptual similarity. FVD~\citep{fvd} evaluates distributional quality of generated video. To assess motion fidelity, we measure the L2 distance between the ground-truth tracks and the tracks estimated from the generated videos, denoting this metric as end-point error (EPE), following Wan-Move~\citep{wanmove} and Motion Prompting~\citep{geng2024motion}.

\begin{table*}[t]
\vspace{-4pt}
\caption{\textbf{Evaluation on three synthetic ViN datasets}. 100 samples per dataset. Each baseline is evaluated on its native backbone. Our method leads on all metrics in both backbone groups. Best within each group in bold.}

\label{tab:vin_results}
\centering
\footnotesize
\setlength{\tabcolsep}{3pt}
\begin{tabular}{ll cccc cccc cccc}
\toprule
& & \multicolumn{4}{c}{\textbf{MoVi-Extended}} & \multicolumn{4}{c}{\textbf{Pool}} & \multicolumn{4}{c}{\textbf{Football}} \\
\cmidrule(lr){3-6} \cmidrule(lr){7-10} \cmidrule(lr){11-14}
& Method & PSNR$\uparrow$ & LPIPS$\downarrow$ & FVD$\downarrow$ & EPE$\downarrow$
         & PSNR$\uparrow$ & LPIPS$\downarrow$ & FVD$\downarrow$ & EPE$\downarrow$
         & PSNR$\uparrow$ & LPIPS$\downarrow$ & FVD$\downarrow$ & EPE$\downarrow$ \\
\midrule
\multirow{3}{*}{\rotatebox{90}{\scriptsize Wan 14B}}
& \textbf{Ours} & \textbf{29.24} & \textbf{.131} & \textbf{23.39} & \textbf{0.57} & \textbf{33.07} & \textbf{.049} & \textbf{31.34} & \textbf{0.08} & \textbf{24.27} & \textbf{.141} & \textbf{24.24} & \textbf{0.05} \\
& ATI & 20.33 & .355 & 37.50 & 2.01 & 21.13 & .303 & 57.78 & 0.37 & 19.68 & .250 & 46.32 & 1.33 \\
& Wan-Move & 19.73 & .331 & 40.38 & 2.18 & 22.94 & .191 & 33.98 & 0.29 & 18.36 & .296 & 69.84 & 3.81 \\
\midrule
\multirow{3}{*}{\rotatebox{90}{\scriptsize CogV}}
& \textbf{Ours} & \textbf{29.77} & \textbf{.108} & \textbf{14.98} & \textbf{0.53} & \textbf{35.64} & \textbf{.025} & \textbf{8.79} & \textbf{0.07} & \textbf{24.79} & \textbf{.090} & \textbf{13.87} & \textbf{0.03} \\
& MagicMotion & 19.79 & .293 & 44.67 & 1.92 & 29.52 & .068 & 18.93 & 0.13 & 22.28 & .161 & 50.56 & 0.33 \\
& Tora & 20.87 & .336 & 50.91 & 1.50 & 31.10 & .128 & 50.39 & 0.12 & 19.34 & .280 & 63.91 & 2.87 \\
\bottomrule
\end{tabular}
\vspace{-6pt}
\end{table*}

\begin{table*}[t]                                  
\caption{\textbf{Evaluation on MOTSynth and OOD datasets.} MOTSynth: 100 synthetic urban scenes. MOT17: 33 real-world surveillance clips. DAVIS: 13 diverse real-world scenes. MOT17 and DAVIS are never seen during training. All baselines are trained on real-world video data, while our method is trained on synthetic data only. Best within each backbone group in bold.} 

\label{tab:ped_results}                                        
\centering                                                     
\footnotesize                                                  
\setlength{\tabcolsep}{2.5pt}                                  
\begin{tabular}{ll cccc cccc cccc}                             
\toprule                                                       
& & \multicolumn{4}{c}{\textbf{MOTSynth}} & \multicolumn{4}{c}{\textbf{MOT17}} & \multicolumn{4}{c}{\textbf{DAVIS}} \\       
\cmidrule(lr){3-6} \cmidrule(lr){7-10} \cmidrule(lr){11-14}    
& Method & PSNR$\uparrow$ & LPIPS$\downarrow$ & FVD$\downarrow$ & EPE$\downarrow$                              
& PSNR$\uparrow$ & LPIPS$\downarrow$ & FVD$\downarrow$ & EPE$\downarrow$                                                
& PSNR$\uparrow$ & LPIPS$\downarrow$ & FVD$\downarrow$ & EPE$\downarrow$ \\                                             
\midrule                                                       
\multirow{3}{*}{\rotatebox{90}{\scriptsize Wan 14B}}           
& \textbf{Ours} & \textbf{21.39} & \textbf{.244} & \textbf{41.43} & \textbf{2.11} & \textbf{19.94} & \textbf{.277} & \textbf{51.46} & \textbf{2.17} & \textbf{18.30} & \textbf{.301} & 88.37 &            
\textbf{2.31} \\                                               
& ATI & 17.76 & .395 & 62.32 & 8.61 & 17.05 & .416 & 65.35 & 4.28 & 14.68 & .443 & 104.90 & 14.02 \\                        
& Wan-Move & 18.29 & .354 & 54.29 & 3.60 & 16.27 & .405 & 60.20 & 3.41 & 15.86 & .352 & \textbf{85.51} & 4.96 \\         
\midrule                                                       
\multirow{3}{*}{\rotatebox{90}{\scriptsize CogV}}              
& \textbf{Ours} & \textbf{22.34} & \textbf{.232} & \textbf{39.70} & \textbf{2.32} & \textbf{21.59} & \textbf{.265} & \textbf{53.92} & \textbf{2.37} & \textbf{18.39} & \textbf{.267} & 89.39 &            
  \textbf{2.43} \\                                             
& MagicMotion & 21.17 & .258 & 43.07 & 2.61 & 19.88 & .307 & 56.13 & 3.38 & 17.74 & .294 & \textbf{70.77} & 3.68 \\         
& Tora & 19.20 & .346 & 70.84 & 2.91 & 17.76 & .400 & 113.9 & 2.70 & 17.25 & .329 & 75.78 & 3.16 \\                          
\bottomrule                                                    
\end{tabular}                                                  
\vspace{-14pt}
\end{table*}

\begin{table*}[t]

\caption{\textbf{Scaling with object count on MOTSynth-Multi.} 20 held-out clips per bin with 5, 10, 15, and 17--20 controlled pedestrians. Best within each backbone group in bold.}
\label{tab:scaling}
\centering
\scriptsize
\setlength{\tabcolsep}{1.95pt}
\renewcommand{\arraystretch}{1.15}
\begin{tabular}{ll cccc@{\hspace{8pt}}cccc@{\hspace{8pt}}cccc@{\hspace{8pt}}cccc}
\toprule
& & \multicolumn{4}{c}{\textbf{5 pedestrians}} & \multicolumn{4}{c}{\textbf{10 pedestrians}} & \multicolumn{4}{c}{\textbf{15 pedestrians}} & \multicolumn{4}{c}{\textbf{17--20 pedestrians}} \\
\cmidrule(lr){3-6} \cmidrule(lr){7-10} \cmidrule(lr){11-14} \cmidrule(lr){15-18}
& Method & PSNR & LPIPS & FVD & EPE & PSNR & LPIPS & FVD & EPE & PSNR & LPIPS & FVD & EPE & PSNR & LPIPS & FVD & EPE \\
\midrule
\multirow{3}{*}{\rotatebox{90}{\scriptsize Wan 14B}}
& \textbf{Ours} & \textbf{20.40} & \textbf{.293} & \textbf{56.65} & \textbf{4.39} & \textbf{20.46} & \textbf{.286} & \textbf{59.45} & \textbf{2.63} & \textbf{20.66} & \textbf{.257} & \textbf{62.14} & \textbf{2.08} & \textbf{21.10} & \textbf{.240} & \textbf{48.47} & \textbf{0.99} \\
& ATI & 16.77 & .440 & 90.35 & 9.15 & 17.45 & .418 & 77.19 & 4.93 & 18.38 & .358 & 70.64 & 3.15 & 19.31 & .304 & 57.11 & 1.79 \\
& Wan-Move & 17.27 & .423 & 78.86 & 6.82 & 16.92 & .426 & 83.83 & 4.69 & 18.22 & .355 & 77.42 & 2.98 & 19.22 & .300 & 54.65 & 1.67 \\
\midrule
\multirow{3}{*}{\rotatebox{90}{\scriptsize CogV}}
& \textbf{Ours} & \textbf{21.12} & \textbf{.290} & \textbf{54.04} & 5.51 & \textbf{21.26} & \textbf{.268} & \textbf{49.98} & \textbf{2.75} & \textbf{21.67} & \textbf{.240} & \textbf{54.41} & \textbf{1.92} & \textbf{21.62} & \textbf{.237} & 49.04 & \textbf{1.11} \\
& MagicMotion & 20.03 & .315 & 68.65 & \textbf{5.30} & 20.93 & .285 & 67.13 & 3.08 & 21.17 & .250 & 55.95 & 2.33 & 21.40 & \textbf{.237} & \textbf{42.61} & 1.24 \\
& Tora & 18.79 & .386 & 104.5 & 5.81 & 18.00 & .417 & 127.4 & 3.55 & 16.47 & .463 & 152.4 & 2.79 & 17.66 & .454 & 162.9 & 1.64 \\
\bottomrule
\end{tabular}

\vspace{-6pt}
\end{table*}

\begin{table*}[t]

\caption{\textbf{Trajectory crossings on MoVi-Extended.} Test clips binned by the number of object pairs whose paths cross in the image plane. Best within each backbone group in bold.}
\label{tab:cross_movi}
\centering
\footnotesize
\setlength{\tabcolsep}{2.5pt}
\begin{tabular}{ll cccc cccc cccc}
\toprule
& & \multicolumn{4}{c}{\textbf{0 crossings (56 clips)}} & \multicolumn{4}{c}{\textbf{1 crossing (28 clips)}} & \multicolumn{4}{c}{\textbf{2+ crossings (16 clips)}} \\
\cmidrule(lr){3-6} \cmidrule(lr){7-10} \cmidrule(lr){11-14}
& Method & PSNR$\uparrow$ & LPIPS$\downarrow$ & FVD$\downarrow$ & EPE$\downarrow$ & PSNR$\uparrow$ & LPIPS$\downarrow$ & FVD$\downarrow$ & EPE$\downarrow$ & PSNR$\uparrow$ & LPIPS$\downarrow$ & FVD$\downarrow$ & EPE$\downarrow$ \\
\midrule
\multirow{3}{*}{\rotatebox{90}{\scriptsize Wan 14B}}
& \textbf{Ours} & \textbf{30.26} & \textbf{.120} & \textbf{25.53} & \textbf{0.48} & \textbf{29.30} & \textbf{.132} & \textbf{33.54} & \textbf{0.58} & \textbf{25.56} & \textbf{.169} & \textbf{35.57} & \textbf{0.90} \\
& ATI & 20.76 & .344 & 46.09 & 1.86 & 20.61 & .336 & 54.56 & 1.91 & 18.33 & .425 & 67.56 & 2.72 \\
& Wan-Move & 20.12 & .316 & 49.48 & 1.79 & 19.88 & .327 & 56.36 & 2.60 & 18.08 & .390 & 65.75 & 2.84 \\
\midrule
\multirow{3}{*}{\rotatebox{90}{\scriptsize CogV}}
& \textbf{Ours} & \textbf{30.61} & \textbf{.101} & \textbf{17.83} & \textbf{0.46} & \textbf{29.62} & \textbf{.107} & \textbf{23.09} & \textbf{0.51} & \textbf{27.13} & \textbf{.131} & \textbf{22.88} & \textbf{0.84} \\
& MagicMotion & 20.43 & .272 & 53.35 & 1.63 & 19.99 & .290 & 61.71 & 1.91 & 17.18 & .374 & 71.23 & 2.97 \\
& Tora & 21.73 & .315 & 54.88 & 1.42 & 20.07 & .356 & 84.36 & 1.33 & 19.26 & .374 & 77.62 & 2.07 \\
\bottomrule
\end{tabular}

\vspace{-10pt}
\end{table*}

\subsection{Quantitative evaluation}
\label{sec:quant_results}

Table~\ref{tab:vin_results} reports results on the three synthetic ViN datasets. Table~\ref{tab:ped_results} reports results on MOTSynth, MOT17, and DAVIS, where MOT17 and DAVIS are never seen during training. Our method leads on PSNR, LPIPS, and EPE across both backbones on all datasets, with PSNR gains of 4--10~dB on Pool and Football and 1--3~dB on the crowded multi-object settings (MOTSynth, MOT17). The improvements are particularly pronounced in trajectory adherence, with EPE reductions ranging from 11\% to over 10$\times$ depending on the dataset and baseline. The same gains persist on MOT17 and DAVIS, which are never seen during training, confirming that our method generalizes beyond the synthetic training distribution. This is particularly notable given that all baselines are trained on large-scale real-world video data, while our method is trained exclusively on synthetic scenes. We acknowledge that the baselines were not trained on our synthetic evaluation datasets and could benefit from fine-tuning on them. However, on MOT17 and DAVIS, which are real-world scenes closer to the baselines' training distribution, our method still leads, demonstrating generalization beyond our synthetic training data. Appendix~\ref{app:sav} complements this by training our method on real moving-camera video. On MagicBench~\citep{li2025magicmotion} and MoveBench-Multi~\citep{wanmove}, it also leads in PSNR, LPIPS, and FVD.

\subsection{Object count and trajectory crossings}
\label{sec:scaling}

\noindent \textbf{MOTSynth-Multi} \quad To measure how performance changes as scenes become crowded, we construct MOTSynth-Multi from held-out MOTSynth test scenes: four bins of 20 clips with 5, 10, 15, and 17--20 controlled pedestrians (Appendix~\ref{app:msm}). As shown in Table~\ref{tab:scaling}, our method with WaN leads on every metric at every object count. With CogVideoX, it leads on PSNR and LPIPS at every count (tied on LPIPS in the densest bin) and on FVD and EPE in three of the four bins.

\noindent \textbf{Trajectory crossings} \quad We also bin the MoVi-Extended test clips by the number of object pairs whose paths cross (Table~\ref{tab:cross_movi}). All methods degrade as crossings increase, yet ours leads on every metric in every bin and, even with two or more crossings, beats every baseline on crossing-free clips.

\begin{figure*}[t]
\vspace{-8pt}
\centering
\setlength{\tabcolsep}{0.5pt}
\renewcommand{\arraystretch}{0.3}
\newcommand{\qimg}[1]{\includegraphics[width=0.155\textwidth]{#1}}
 \newcommand{\framesarrow}[1]{%
    \begin{tikzpicture}[baseline=0pt]
      \draw[->, line width=0.5pt] (0,0) -- (0.465\textwidth, 0)
        node[midway, fill=white, inner sep=1pt] {\scriptsize #1};
    \end{tikzpicture}%
  }
\begin{tabular}{@{}c ccc @{\hspace{6pt}} c ccc@{}}
& \multicolumn{3}{c}{\framesarrow{DAVIS frames}} & & \multicolumn{3}{c}{\framesarrow{MOT17 frames}} \\[1pt]
\rotatebox{90}{\scriptsize\hspace{8pt}GT} &
\qimg{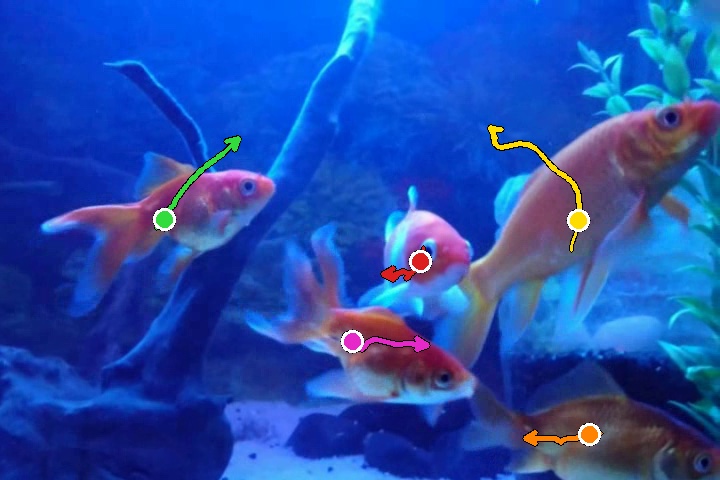} &
\qimg{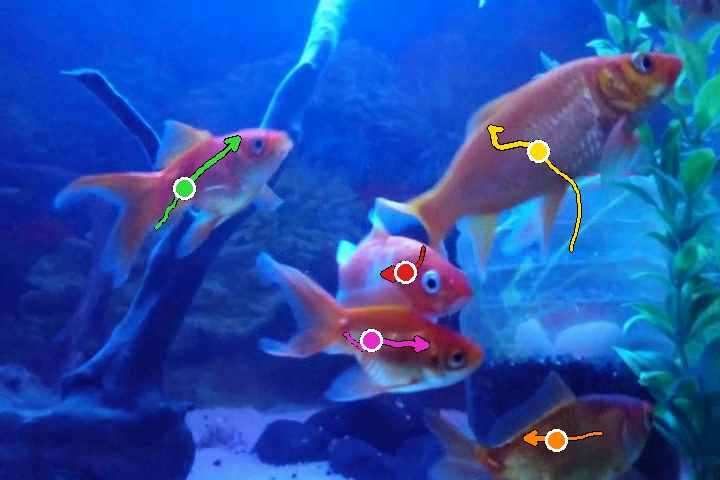} &
\qimg{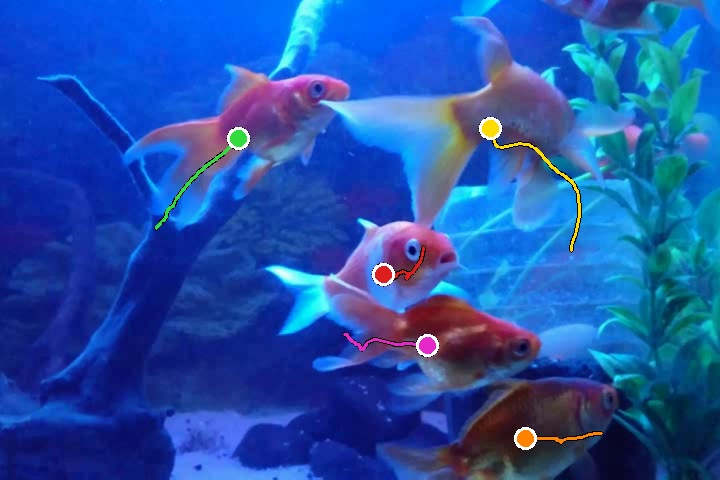} &
\rotatebox{90}{\scriptsize\hspace{8pt}GT} &
\qimg{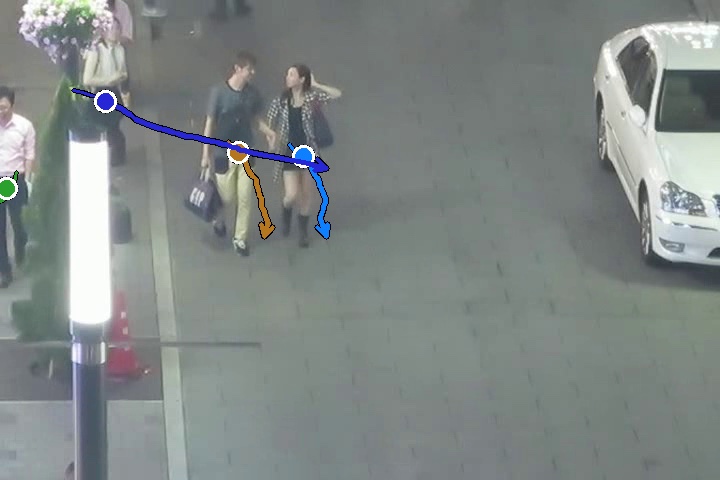} &
\qimg{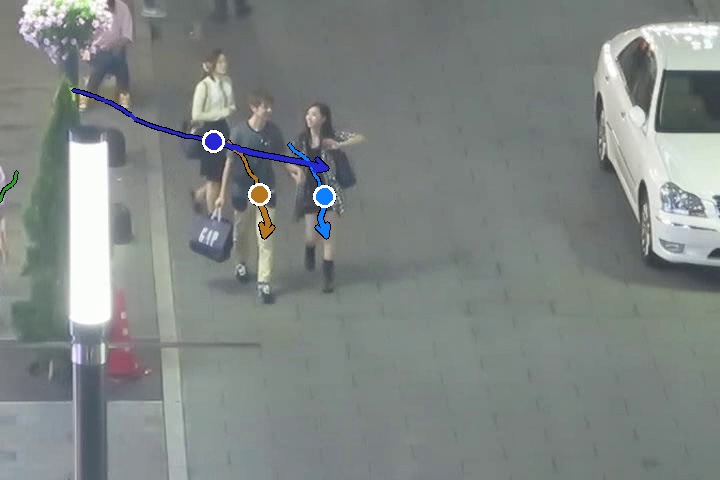} &
\qimg{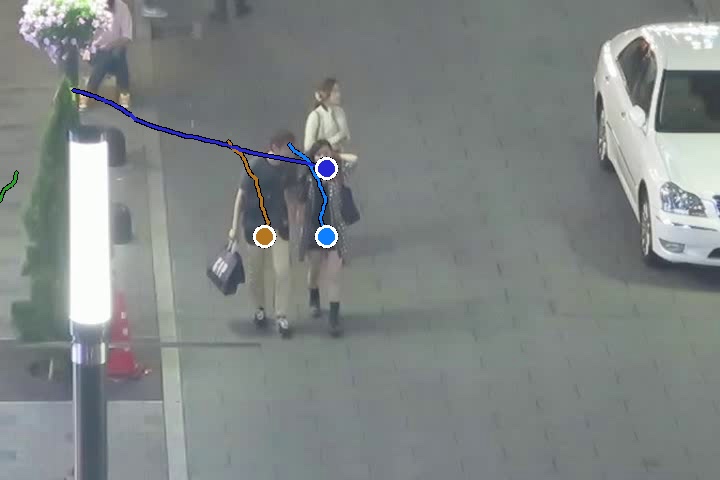} \\[0.5pt]
\rotatebox{90}{\scriptsize\hspace{4pt}Ours-CogV} &
\qimg{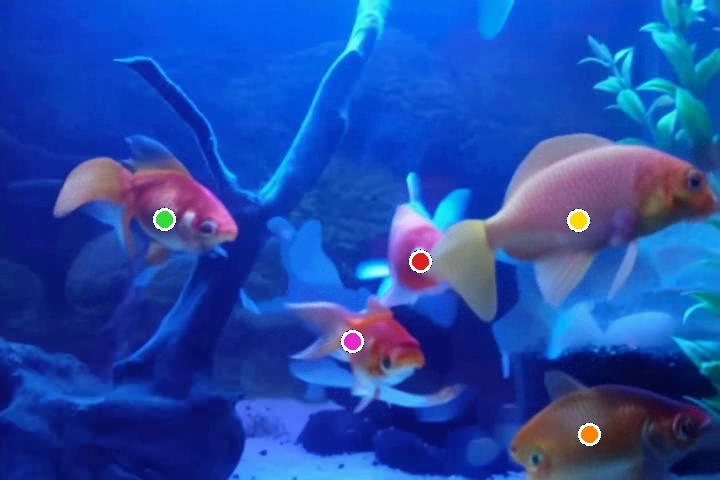} &
\qimg{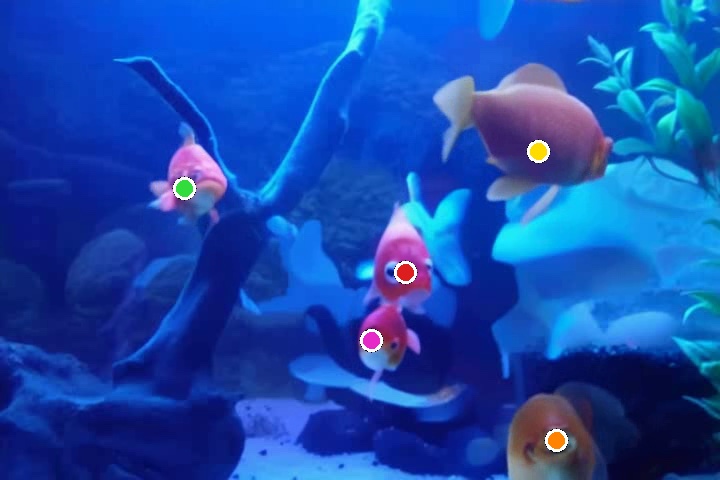} &
\qimg{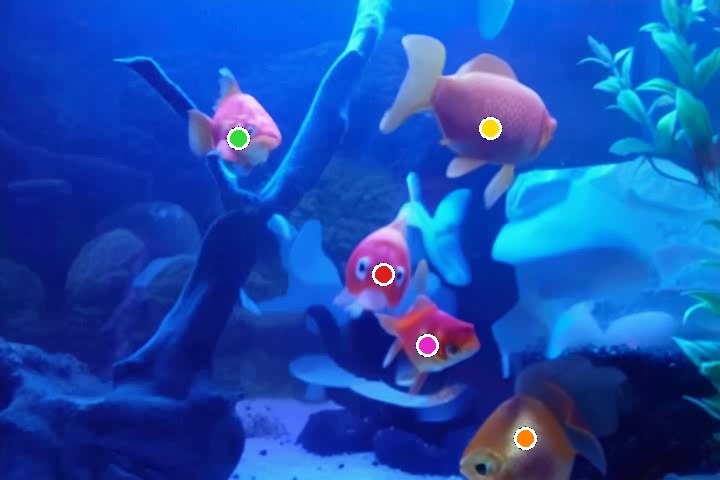} &
\rotatebox{90}{\scriptsize\hspace{4pt}Ours-WaN} &
\qimg{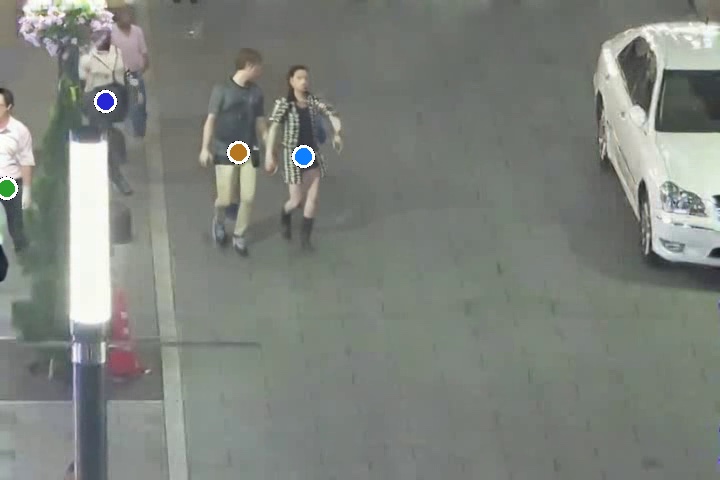} &
\qimg{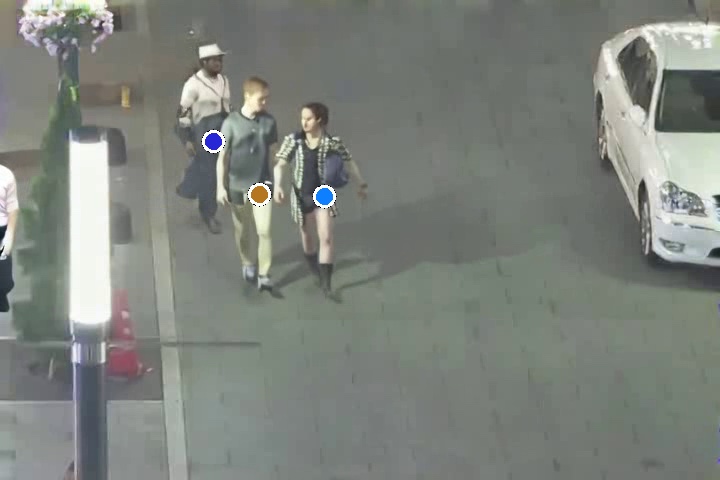} &
\qimg{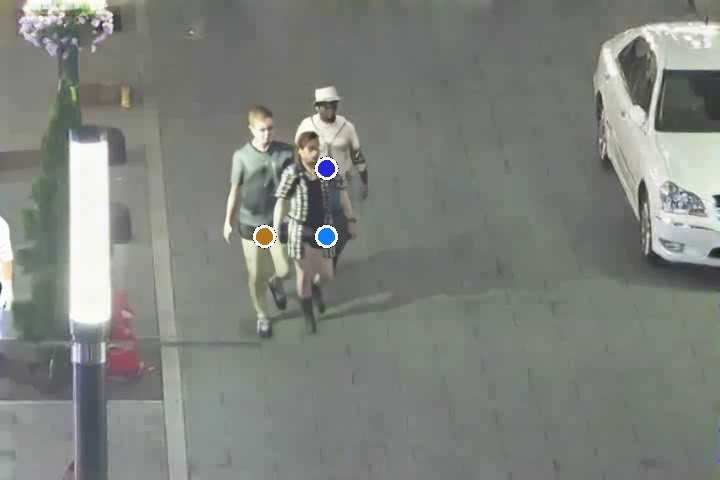} \\[0.5pt]
\rotatebox{90}{\scriptsize\hspace{4pt}MagicMotion} &
\qimg{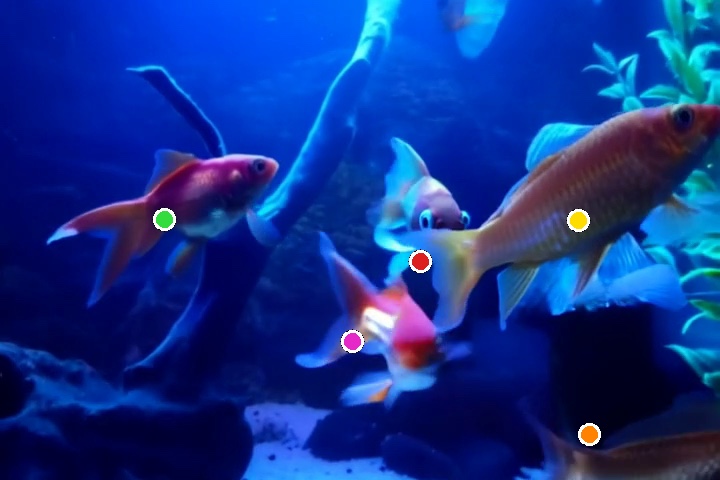} &
\qimg{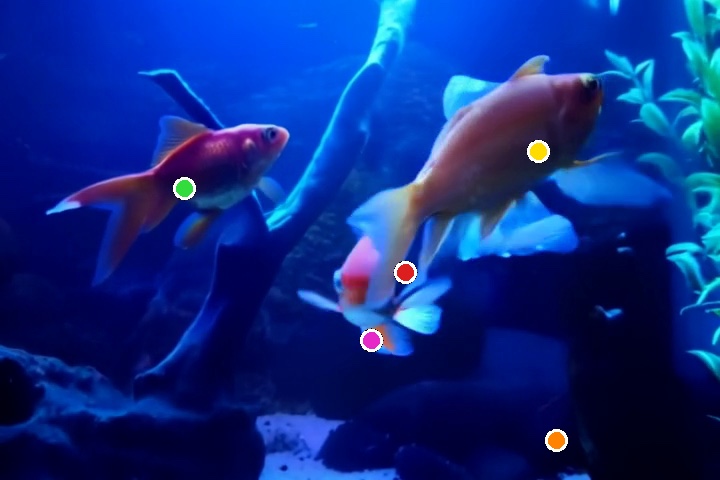} &
\qimg{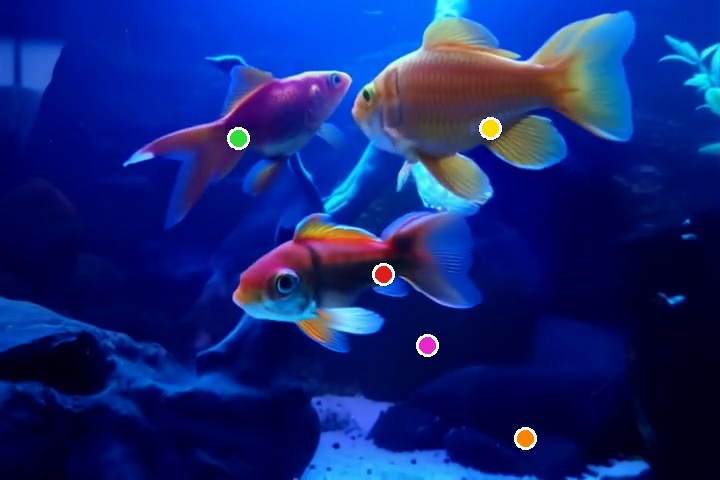} &
\rotatebox{90}{\scriptsize\hspace{8pt}ATI} &
\qimg{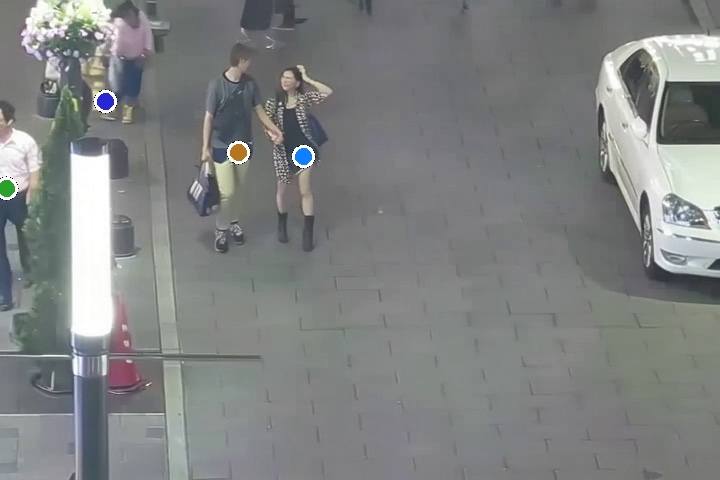} &
\qimg{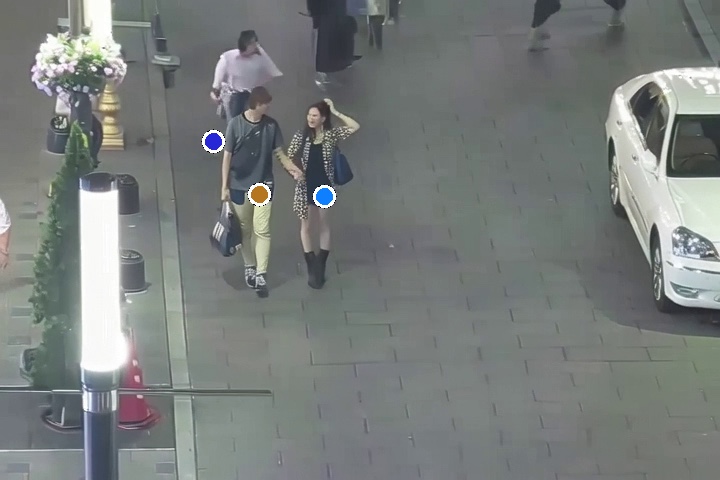} &
\qimg{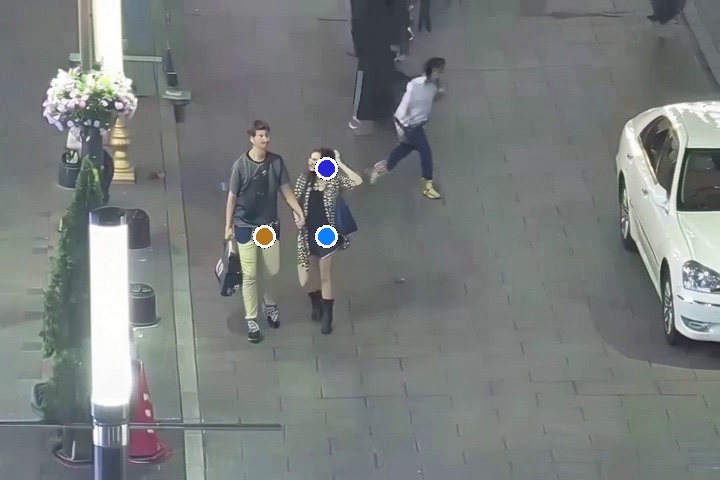} \\[0.5pt]
\rotatebox{90}{\scriptsize\hspace{8pt}Tora} &
\qimg{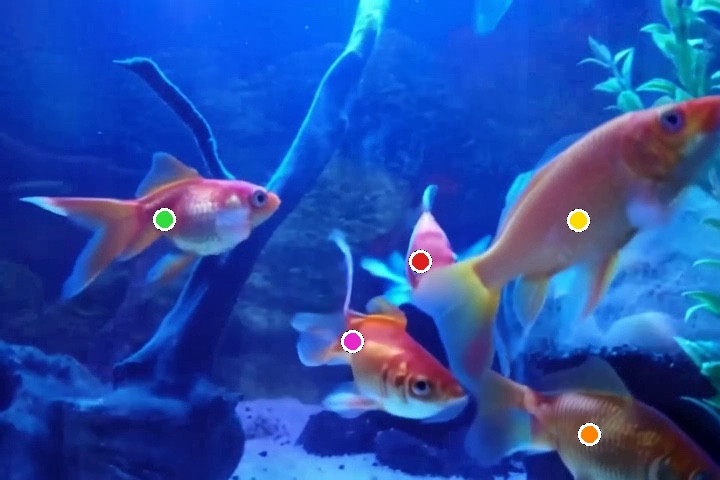} &
\qimg{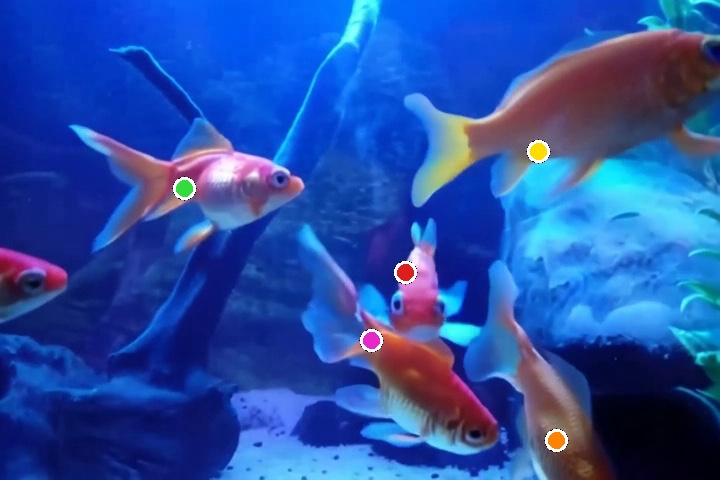} &
\qimg{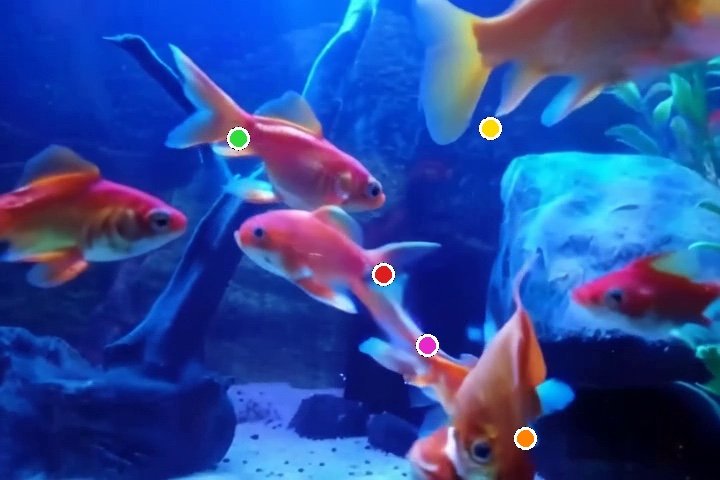} &
\rotatebox{90}{\scriptsize\hspace{4pt}Wan-Move} &
\qimg{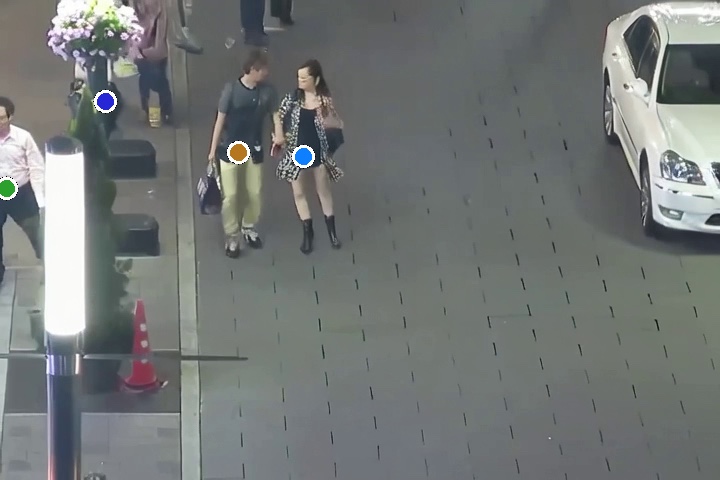} &
\qimg{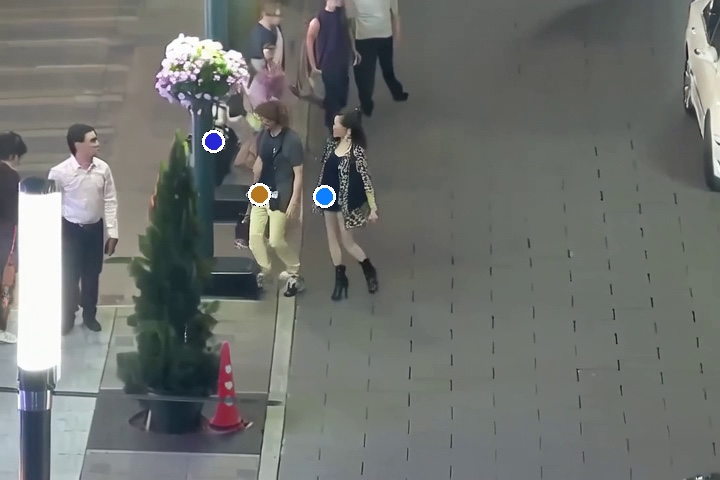} &
\qimg{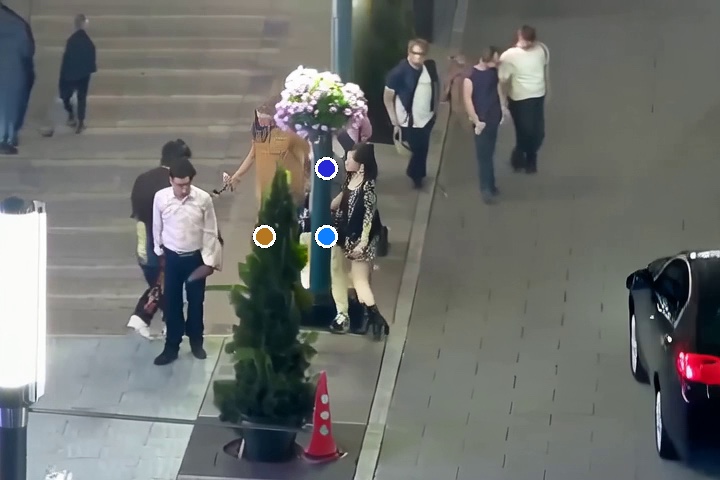} \\[3pt]
& \multicolumn{3}{c}{\framesarrow{MoVi frames}} & & \multicolumn{3}{c}{\framesarrow{MOTSynth frames}} \\[1pt]
\rotatebox{90}{\scriptsize\hspace{8pt}GT} &
\qimg{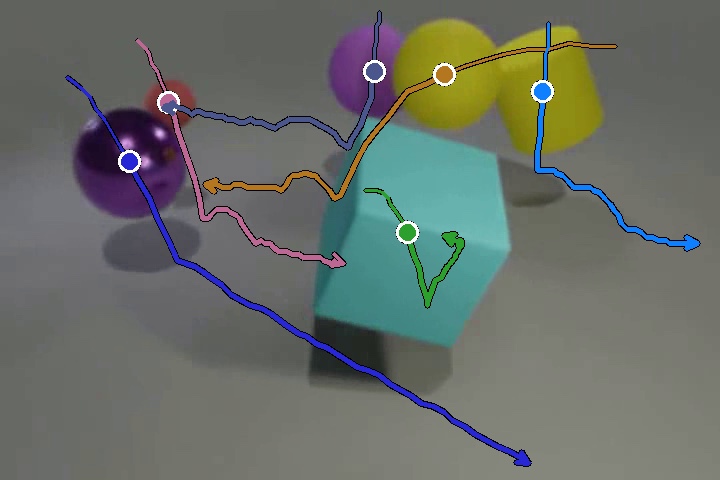} &
\qimg{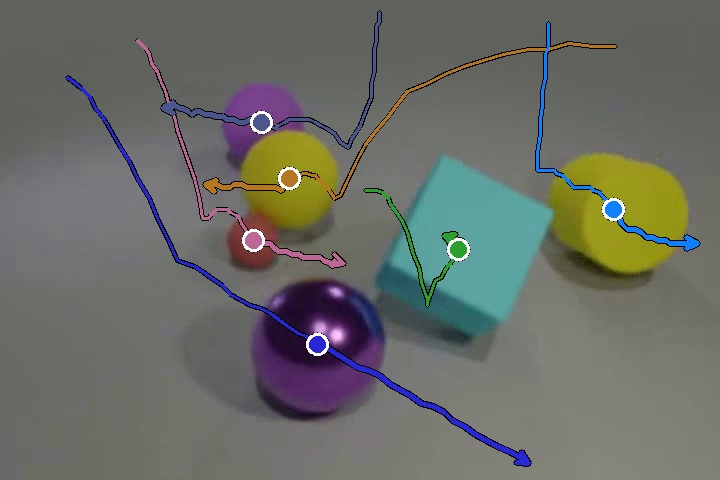} &
\qimg{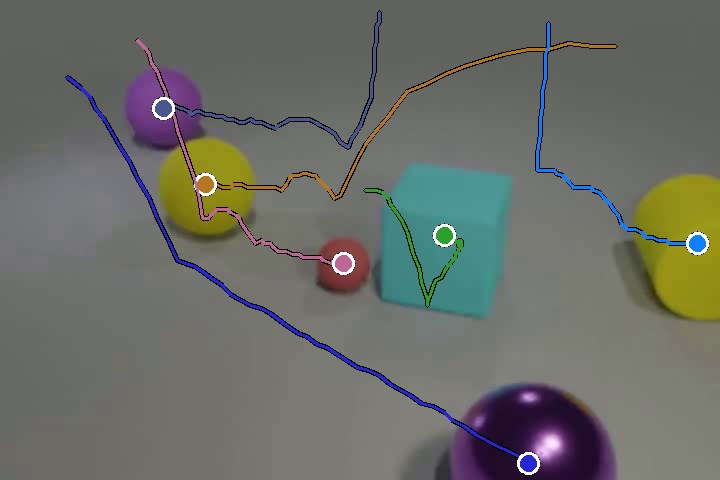} &
\rotatebox{90}{\scriptsize\hspace{8pt}GT} &
\qimg{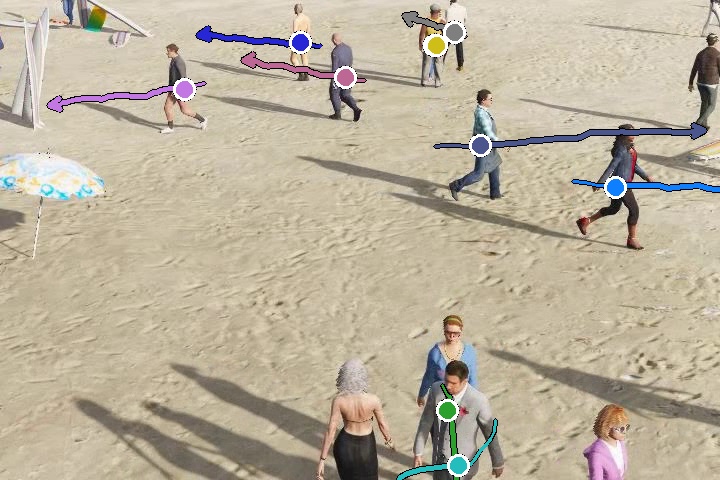} &
\qimg{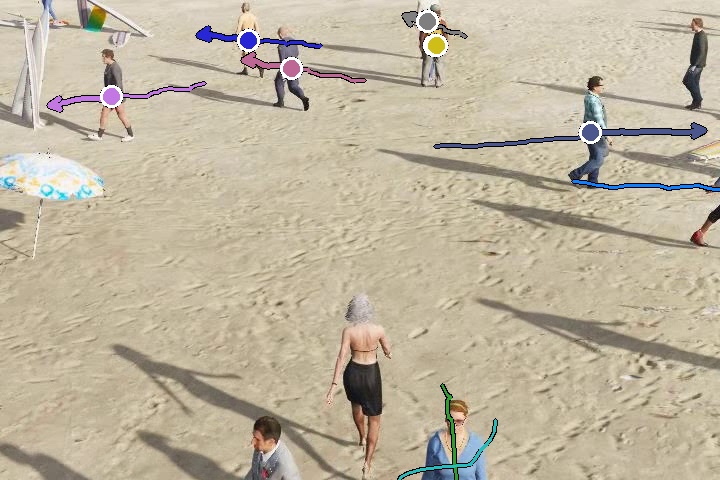} &
\qimg{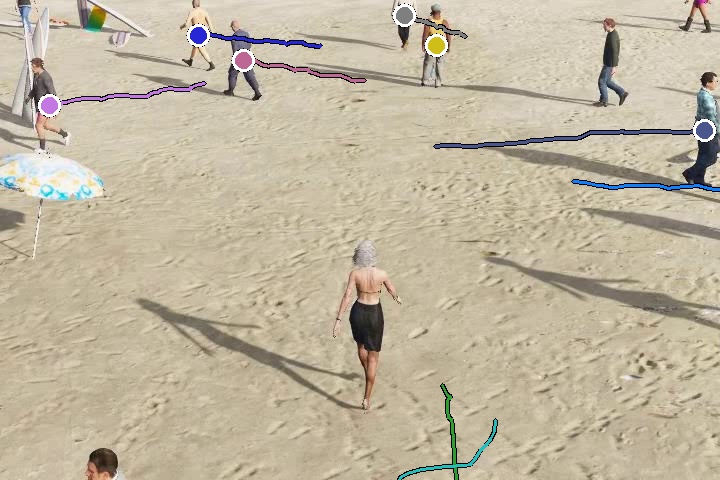} \\[0.5pt]
\rotatebox{90}{\scriptsize\hspace{4pt}Ours-CogV} &
\qimg{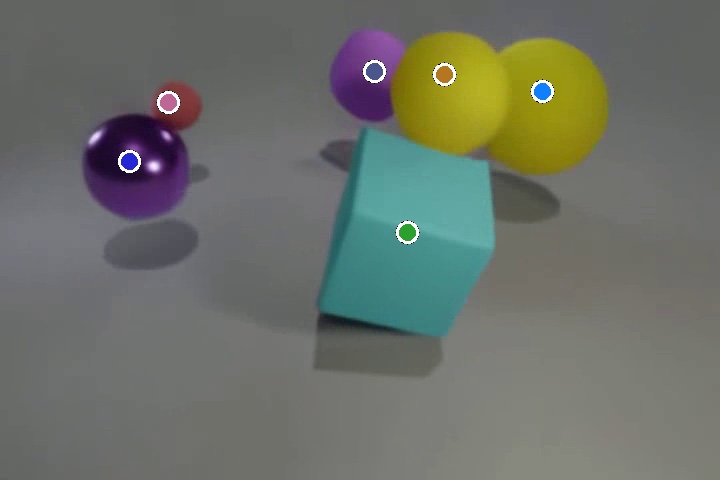} &
\qimg{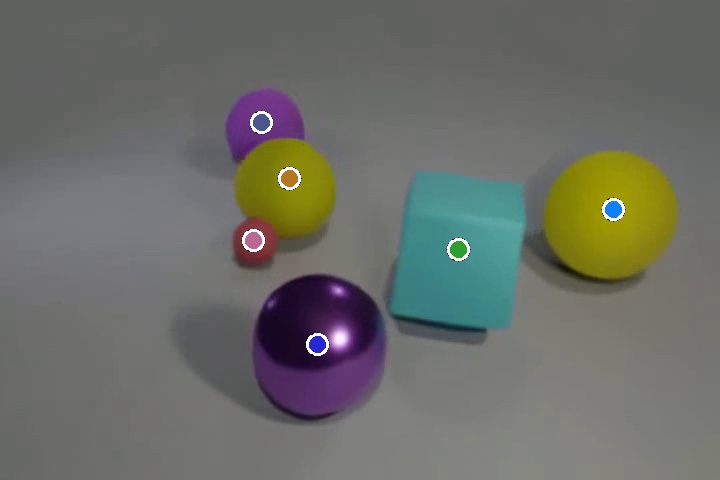} &
\qimg{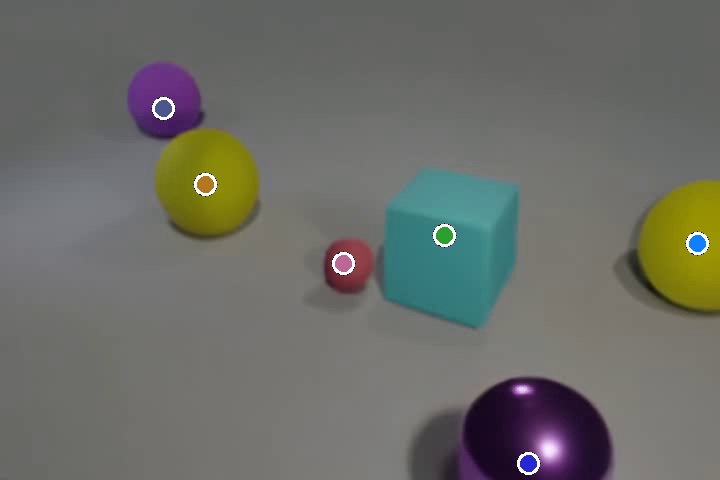} &
\rotatebox{90}{\scriptsize\hspace{4pt}Ours-WaN} &
\qimg{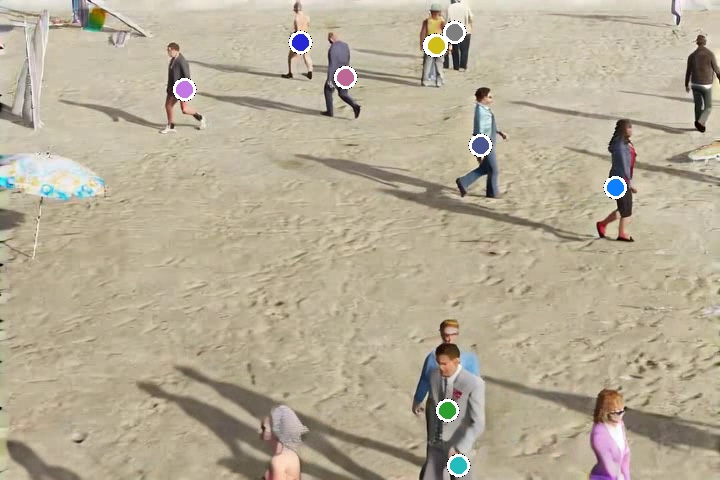} &
\qimg{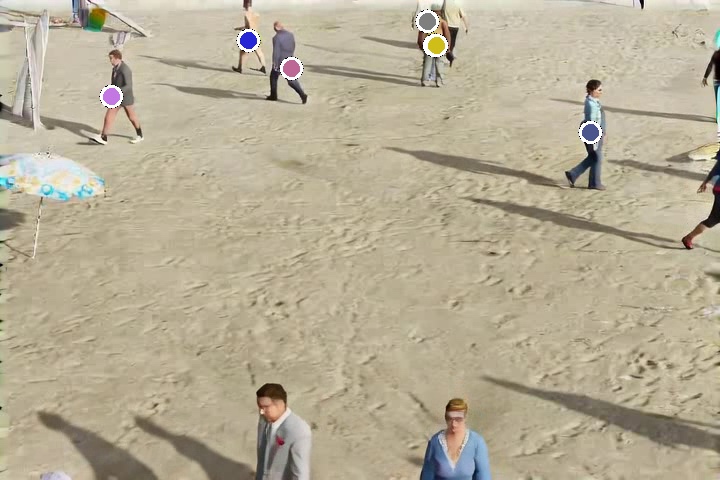} &
\qimg{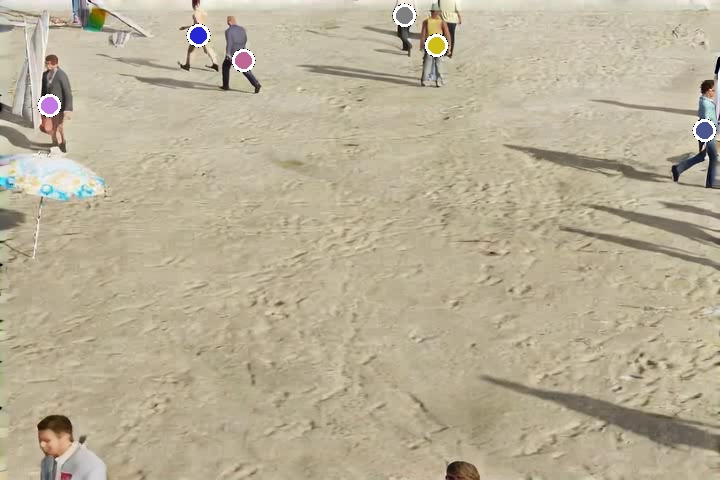} \\[0.5pt]
\rotatebox{90}{\scriptsize\hspace{4pt}MagicMotion} &
\qimg{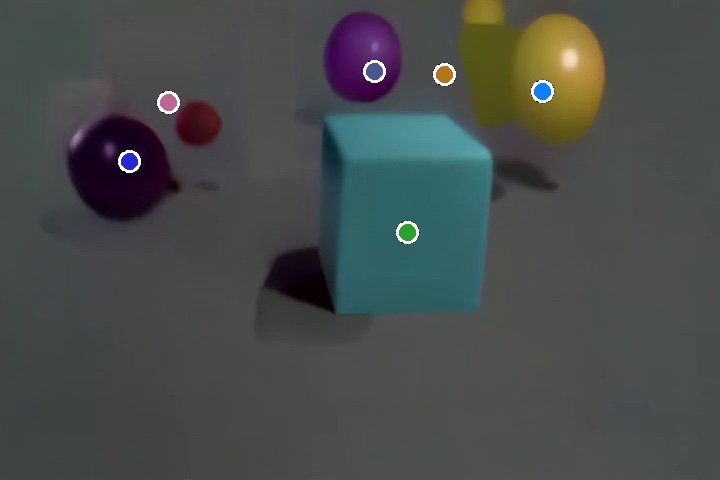} &
\qimg{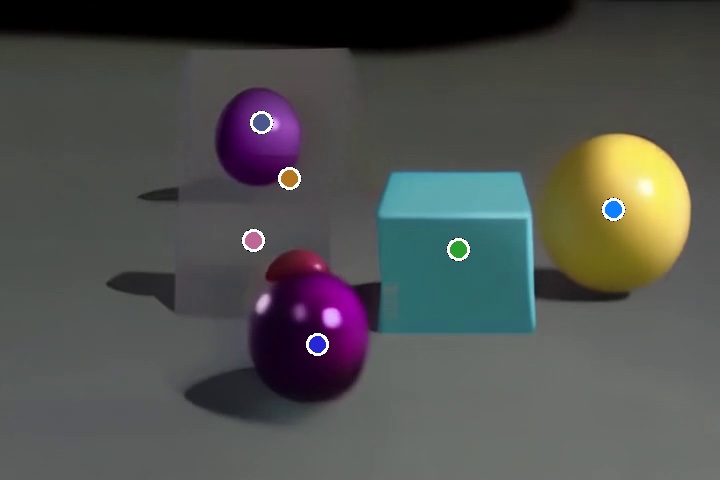} &
\qimg{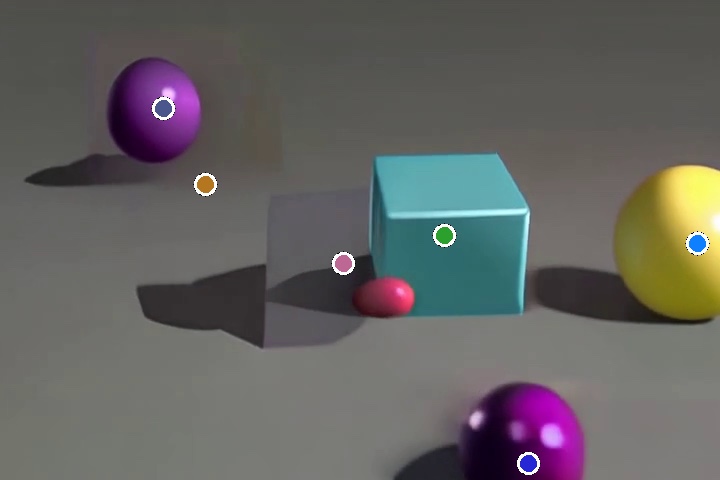} &
\rotatebox{90}{\scriptsize\hspace{8pt}ATI} &
\qimg{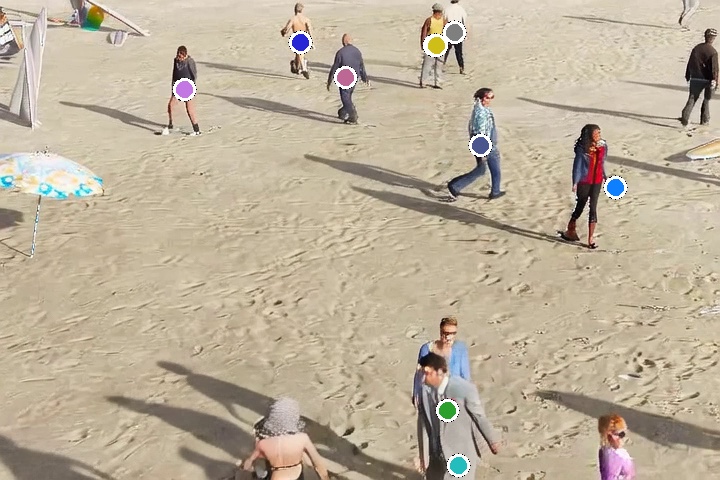} &
\qimg{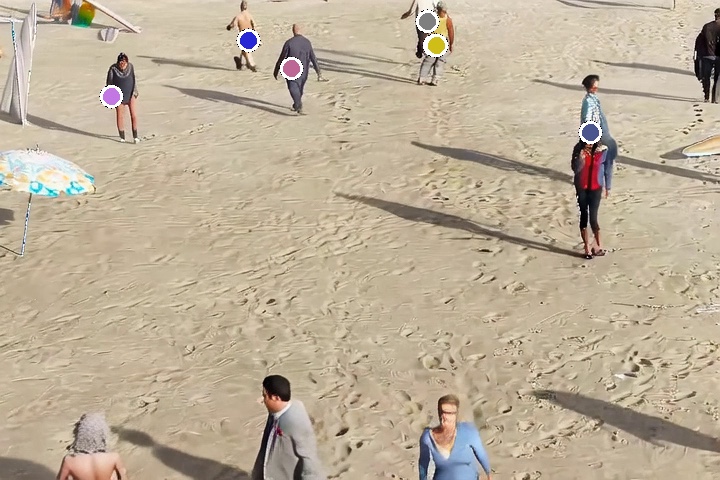} &
\qimg{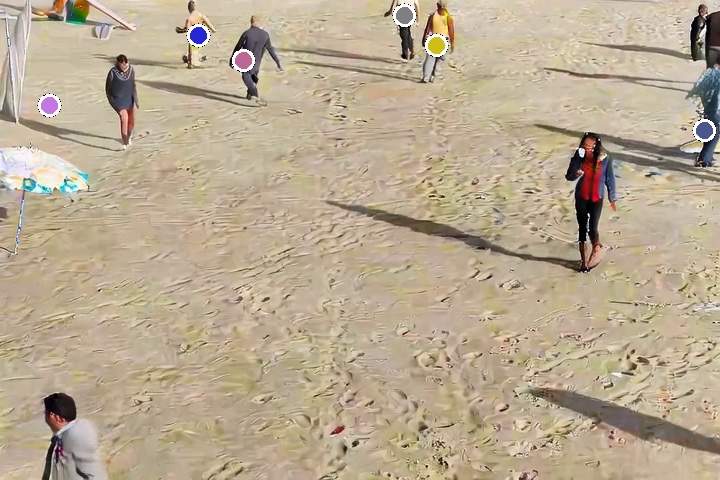} \\[0.5pt]
\rotatebox{90}{\scriptsize\hspace{8pt}Tora} &
\qimg{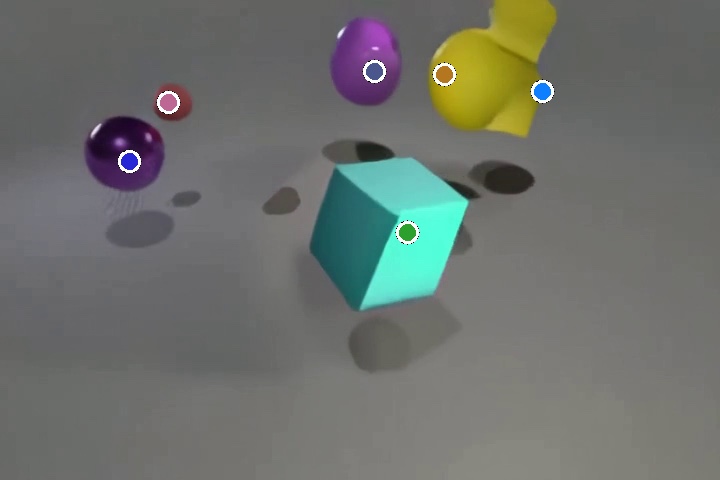} &
\qimg{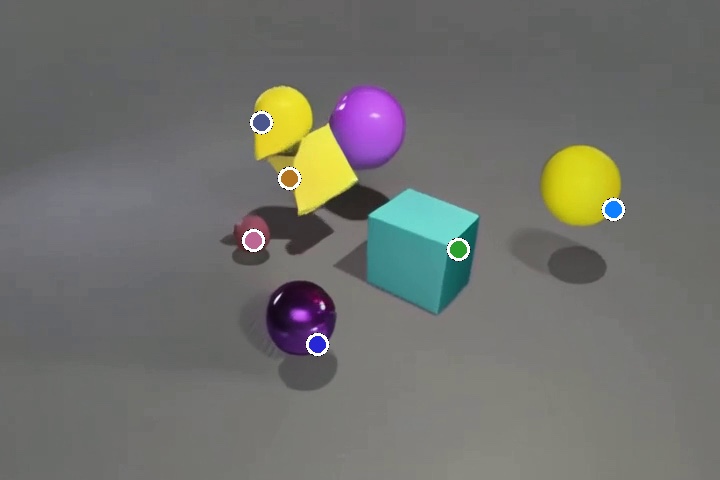} &
\qimg{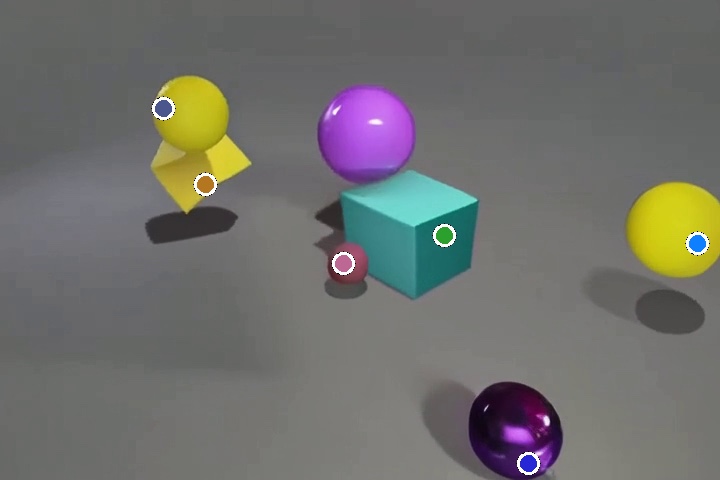} &
\rotatebox{90}{\scriptsize\hspace{4pt}Wan-Move} &
\qimg{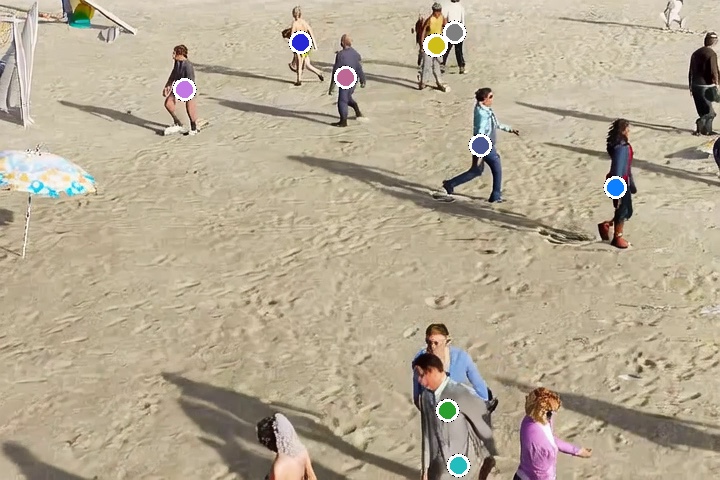} &
\qimg{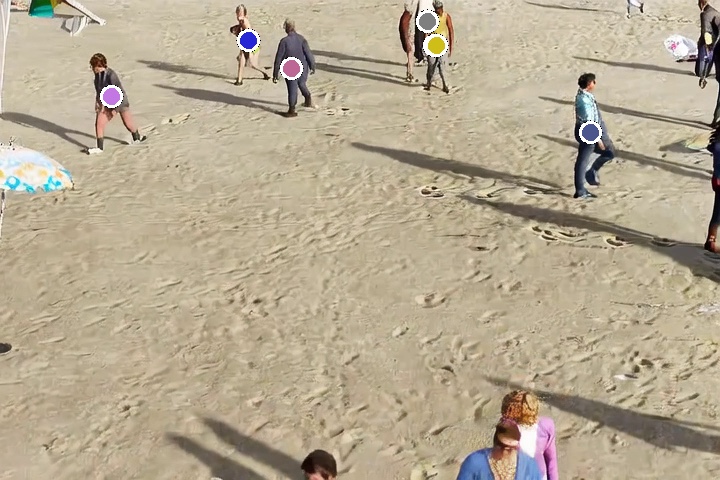} &
\qimg{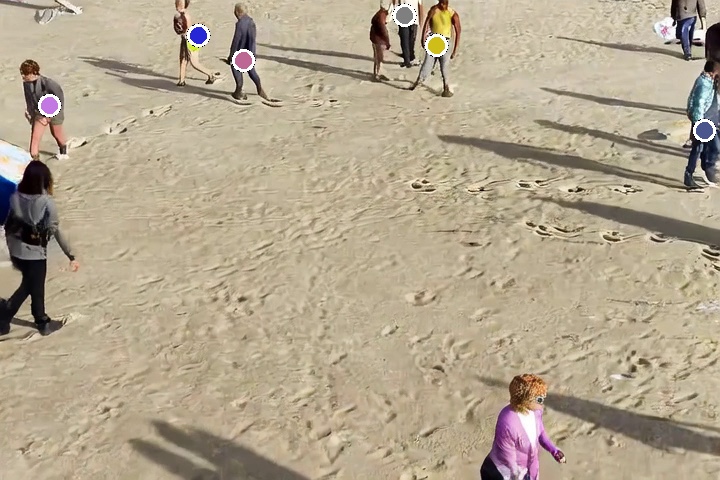} \\
\end{tabular}
\caption{\textbf{Qualitative comparison.} Top-left: CogVideoX-5B results on DAVIS fish (5 objects, real-world). Top-right: WaN~2.1-14B results on MOT17 (4 pedestrians, real-world). Bottom-left: CogVideoX-5B results on MoVi-Extended (6 objects, synthetic). Bottom-right: WaN~2.1-14B results on MOTSynth (10 pedestrians, synthetic). Each row shows three generated frames from a different method, with ground-truth object positions overlaid as colored dots. Our method places all objects at their target locations throughout the sequence, while baselines exhibit missing objects, positional drift, or hallucinated entities. Best viewed in color.}
\label{fig:qual_unified}
\vspace{-14pt}
\end{figure*}

\subsection{Qualitative Evaluation}
\label{sec:qual_results}

To qualitatively assess our method, we compare against all baselines on representative scenes in Figure~\ref{fig:qual_unified}. Each row shows three generated frames spanning the video (frames 9, 29, 49) where colored dots mark the ground-truth position of each object at that frame. The top row shows the original video with prescribed trajectory paths overlaid.

In the top row of panels, CogVideoX-5B is compared against MagicMotion and Tora on the real-scene DAVIS fish scene (5 fish). MagicMotion fails to place objects at several target locations, for example the orange trajectory dot has no corresponding fish in the generated frame. Tora preserves all five entities but places them at incorrect positions, such as the large fish at the top right of the frame. Our method maintains all five fish at their target locations with consistent identity throughout the sequence. WaN~2.1-14B is compared against ATI and Wan-Move on the real-scene MOT17 (4 pedestrians). ATI loses tracking on the dark-blue-trajectory pedestrian. Wan-Move makes the light-blue-trajectory and orange-trajectory pedestrians hover unnaturally and hallucinates additional details. Our method keeps all four pedestrians aligned with their prescribed trajectories. The bottom-left panel compares CogVideoX-5B against MagicMotion and Tora on a MoVi-Extended scene (6 geometric objects). Both baselines change object appearance and lose tracking on several trajectories, while our method accurately follows all six prescribed paths. The bottom-right panel shows WaN~2.1-14B against ATI and Wan-Move on a MOTSynth scene (10 pedestrians). ATI fails to follow individual trajectories such as the blue one on the right, which stops tracking its prescribed path, and Wan-Move duplicates the gray-trajectory pedestrian at the top of the final frame. Our method keeps every pedestrian aligned with its prescribed trajectory. Additional qualitative comparisons on MoVi, MOTSynth, MOT17, and DAVIS are provided in Appendix~\ref{app:qual_extra}. Full video results, which better convey object motion along the prescribed trajectories and identity preservation across frames, are available on our project page (\url{https://sela-omer.github.io/traj-loc/}).

\subsection{Ablation study}
\label{sec:ablations}

We examine the contribution of each component on WaN~2.1-14B (Table~\ref{tab:ablation_combined}), both by removing one component at a time and by building the model up from the bare backbone, on MoVi and MOTSynth, the in-distribution datasets with the most objects per scene, so that the measured effect of each component reflects the component itself rather than domain shift. Attention localization alone recovers most of the gap between the backbone and the full model (row 2: +3.8\,dB PSNR on MoVi and +1.7\,dB on MOTSynth), whereas the tokens without it barely change the backbone (rows 4--5). The tokens carry per-object motion, depth, and appearance, but only attention localization binds each token to its object in the video; without it, they act as diffuse, global conditioning. Removing the attention localization from the full model causes the largest degradation across all metrics, with PSNR dropping by 6.7\,dB and EPE increasing over 3$\times$ on MoVi, confirming that the hard spatial constraint is the most critical component. Removing the trajectory tokens primarily affects depth-sensitive scenes, with EPE on MOTSynth increasing from 2.11 to 2.68. Notably, removing the depth channel alone produces degradation comparable to removing the full trajectory token on MoVi, suggesting that depth encoding is a key contribution of the token. On Pool, however, the full token provides substantially more than depth alone (PSNR drops by 4~dB vs. less than 1~dB; Appendix~\ref{app:lattice}), indicating it also captures spatial motion information. Removing the appearance token costs 0.6\,dB PSNR on MOTSynth and increases FVD on both datasets. The full model achieves the best result on every metric, demonstrating that all components are complementary. The bounding-box loss and inserting the trajectory token before the text encoder both matter (Appendix~\ref{app:design}), while depth is optional (Appendix~\ref{app:depth}). An ablation of the attention localization approximation used for CogVideoX-5B is provided in Appendix~\ref{app:cogv_ablation}.

\begin{table*}[t]

\vspace{-4pt}
\caption{\textbf{Ablation study on WaN~2.1-14B} with additive and leave-one-out configurations. A/T/P/d: attention localization, trajectory token, appearance token, depth. Leave-one-out PSNR varies by at most 0.10\,dB over nine seeds (Appendix~\ref{app:seeds}); Pool and Football are in Appendix~\ref{app:lattice}. Best in bold.}
\label{tab:ablation_combined}
\centering
\footnotesize
\setlength{\tabcolsep}{3.2pt}
\begin{tabular}{rl cccc cccc cccc}
\toprule
& & & & & & \multicolumn{4}{c}{\textbf{MoVi}} & \multicolumn{4}{c}{\textbf{MOTSynth}} \\
\cmidrule(lr){7-10} \cmidrule(lr){11-14}
\# & Configuration & A & T & P & d & PSNR$\uparrow$ & LPIPS$\downarrow$ & FVD$\downarrow$ & EPE$\downarrow$ & PSNR$\uparrow$ & LPIPS$\downarrow$ & FVD$\downarrow$ & EPE$\downarrow$ \\
\midrule
1 & Backbone only &  &  &  &  & 22.95 & .284 & 34.27 & 1.46 & 19.02 & .337 & 62.60 & 4.05 \\
2 & Attention only & \cmark &  &  &  & 26.79 & .159 & 28.63 & 0.74 & 20.69 & .266 & 46.44 & 2.26 \\
\midrule
3 & w/o trajectory token & \cmark &  & \cmark &  & 27.87 & .146 & 27.15 & 0.65 & 21.30 & .256 & 44.01 & 2.68 \\
4 & Trajectory token only &  & \cmark &  & \cmark & 23.53 & .272 & 37.03 & 1.32 & 19.35 & .314 & 53.79 & 3.34 \\
\midrule
5 & w/o attention localization &  & \cmark & \cmark & \cmark & 22.58 & .313 & 40.86 & 1.87 & 19.31 & .316 & 55.96 & 3.22 \\
6 & w/o appearance token & \cmark & \cmark &  & \cmark & 28.40 & .146 & 28.09 & 0.67 & 20.78 & .258 & 44.95 & 2.21 \\
7 & w/o depth channel & \cmark & \cmark & \cmark &  & 27.56 & .145 & 27.66 & 0.62 & 21.32 & .250 & 43.11 & 2.28 \\
\midrule
8 & \textbf{Ours} & \cmark & \cmark & \cmark & \cmark & \textbf{29.24} & \textbf{.131} & \textbf{23.39} & \textbf{0.57} & \textbf{21.39} & \textbf{.244} & \textbf{41.43} & \textbf{2.11} \\
\bottomrule
\end{tabular}

\vspace{-10pt}
\end{table*}

\section{Limitations and Broader Impacts}
\label{sec:impacts}
Our main experiments use static-camera scenes at $720 \!\times\! 480$ resolution, and the training data is predominantly synthetic, which introduces a domain gap on natural scenes. While the method generalizes to real-world data such as DAVIS and MOT17, we occasionally observe the model falling back on synthetic appearance priors, e.g., generating GTA-V pedestrian sprites in place of out-of-distribution objects. Failure cases are shown in  Appendix~\ref{app:failure_cases}. Our real-data models (Appendix~\ref{app:sav}) handle camera motion only implicitly, through background trajectories. Explicit camera control, higher resolutions, and finer-grained, 3D-aware motion control remain future work. As with other generative models, our method carries dual-use potential. Its ability to produce realistic, controllable videos can benefit synthetic data generation, simulation, and professional video editing, but also risks misuse for generating misleading content. We believe it is crucial to develop detection tools for AI-generated videos and to adopt safeguards such as watermarking for safer generative AI usage.

\section{Conclusion}
\label{sec:conclusion}

We presented \method{}, a method for multi-object trajectory-guided video generation that challenges the prevailing assumption that trajectory conditioning requires dense video-sized representations. Instead, we observe that the model's existing conditioning spaces, namely the attention mechanism and the text tokens representations, are sufficient to carry per-object trajectory and identity control. Accordingly, we enforce spatial adherence by replacing attention weight columns with Gaussian heatmaps at the object's associated token and encode each object's trajectory and appearance as dedicated tokens in the text prompt. Evaluated on six datasets spanning synthetic objects, billiard scenes, pedestrian crowds with up to 20 interacting objects, and diverse real-world scenes never seen during training, our method leads on PSNR, LPIPS, and EPE within each backbone group across all six datasets.

\begin{ack}
This work was funded by Amazon. The authors declare no competing interests.
\end{ack}

\bibliography{egbib}
\bibliographystyle{plainnat}

\appendix
\newpage
\addtocontents{toc}{\protect\setcounter{tocdepth}{2}}
\section*{Appendix Contents}
\makeatletter
\@starttoc{toc}
\makeatother
\setcounter{figure}{0}
\setcounter{table}{0}
\renewcommand{\thefigure}{S\arabic{figure}}
\renewcommand{\thetable}{S\arabic{table}}

\section{Real-Data Models for Moving-Camera Footage}
\label{app:sav}

The models of Sec.~\ref{sec:experiments} are trained on synthetic, static-camera clips, while all four baselines are trained on large-scale real-world video. To evaluate our method under training conditions closer to those of the baselines, we retrain it on real, moving-camera footage and evaluate it on MagicBench~\citep{li2025magicmotion} and MoveBench-Multi~\citep{wanmove}, benchmarks introduced with MagicMotion and Wan-Move, respectively. The method itself is unchanged: attention localization, the trajectory and appearance tokens, the training objective, and all hyperparameters are those of Sec.~\ref{sec:method} and Sec.~\ref{sec:setup}. Only the training data and the inputs derived from it differ, as described below. The checkpoints released with our code are the real-data models of this section.

\subsection{Training Data and Inputs}
\label{app:sav_data}

\noindent \textbf{Clips and objects} \quad We build the training set from SA-V~\citep{sam2}, a large and diverse corpus of real videos with frequent camera motion, using the SVG2~\citep{gao2026svg2} panoptic-mask and scene-graph annotations. From about 41K source videos we cut 120,643 clips of 49 frames at $720 \!\times\! 480$, taking up to three non-overlapping clips per video and keeping clips with at least two annotated objects. In each clip we keep the most prominent objects by image area, merge parts such as clothing into their parent person or vehicle, and discard abstract or atmospheric labels.

\noindent \textbf{Trajectories} \quad Under camera motion, an object's image position can change abruptly. We track each object with CoTracker3~\citep{karaev2025cotracker3}, seeding a small set of candidate points at the frame where the object is most visible and keeping the candidate whose track stays inside the object's mask the longest. At every frame, the tracked position is accepted only if it falls inside the object's mask and is marked as absent otherwise, which prevents the track from jumping to a similar-looking neighbor under occlusion or fast camera motion.

\noindent \textbf{Camera motion through background trajectories} \quad Our method has no dedicated camera input. Because the SA-V annotations are panoptic, static scene elements such as walls, ground, or vegetation are tracked in the same way and given to the model as ordinary trajectory-controlled objects. Camera and scene motion are therefore conveyed by the same per-object mechanism as the motion of the foreground objects.

\noindent \textbf{Prompts and depth} \quad The prompt follows the template of Eq.~\ref{eq:prompt}, naming every tracked object with its trajectory token, and is followed by a scene caption generated by Qwen2.5-VL~\citep{Qwen2.5-VL} with the protocol of Appendix~\ref{app:captions}. The depth channel is set to zero, so the trajectory token encodes only image-plane motion.

\noindent \textbf{Training} \quad We fine-tune from the released CogVideoX-5B and WaN~2.1-14B I2V weights with the configuration of Appendix~\ref{app:additional_impl} on a single H100 GPU per backbone, with batch size 3 for CogVideoX and 2 for WaN. We release and report the checkpoints at step 28,500 for CogVideoX (${\sim}$8 days) and step 49,000 for WaN (${\sim}$8.7 days).

\subsection{Benchmarks and Protocol}
\label{app:sav_protocol}

\noindent \textbf{Benchmarks} \quad MagicBench~\citep{li2025magicmotion} is the benchmark released with MagicMotion. It contains 600 real clips with per-frame instance masks, organized into six bins of 100 clips with 1 to 6 or more annotated objects, and ships neither trajectories nor captions. MoveBench-Multi is the multi-object subset of MoveBench~\citep{wanmove}, the benchmark released with Wan-Move. It contains 192 clips with two to four annotated objects and ships ground-truth point tracks, first-frame masks, and a caption per clip. Both benchmarks include many clips with camera motion. Every clip is converted to our format of 49 frames at $720 \!\times\! 480$, taking the first 49 frames of each MoveBench-Multi clip.

\noindent \textbf{Inputs} \quad On MoveBench-Multi, foreground trajectories are the benchmark's ground-truth tracks; on MagicBench, they are obtained from the instance masks with the tracking procedure above. Since neither benchmark annotates the background, we discover up to four background regions per clip automatically by running SAM2~\citep{sam2} in automatic mask-generation mode on the first frame, discarding regions that overlap the foreground, keeping the largest remaining regions, and tracking them with CoTracker3. MoveBench-Multi clips use their shipped caption, and MagicBench clips receive a Qwen2.5-VL caption. All methods, including ours, receive the same first frame, foreground and background trajectories, and caption. All methods are scored with the metric code of Sec.~\ref{sec:setup}.

\subsection{Results}
\label{app:sav_results}

Table~\ref{tab:sav} reports both benchmarks. With both backbones, our method achieves the best PSNR, LPIPS, and FVD on both benchmarks, and the best EPE in three of the four backbone and benchmark combinations; on MagicBench with WaN, Wan-Move attains a lower EPE (2.28 vs.\ 2.39). The margins, 0.2--1.1\,dB PSNR over the strongest same-backbone baseline, are smaller than in Tables~\ref{tab:vin_results}--\ref{tab:ped_results}, where four of the six datasets lie within our training distribution, but consistent across both benchmarks and backbones. Figures~\ref{fig:qual_sav_movebench} and~\ref{fig:qual_sav_magicbench} compare all methods on scenes from both benchmarks, and Figure~\ref{fig:qual_sav_solo} shows further results of our models alone.

\begin{table}[h]
\caption{\textbf{Real-data models on MagicBench and MoveBench-Multi.} MagicBench (600 clips) and MoveBench-Multi (192 clips) both include moving-camera footage. All methods receive the identical first frame, foreground and background trajectories, and caption; no depth is used. Best within each backbone group in bold.}
\label{tab:sav}
\centering
\footnotesize
\setlength{\tabcolsep}{3.5pt}
\begin{tabular}{ll cccc cccc}
\toprule
& & \multicolumn{4}{c}{\textbf{MagicBench}} & \multicolumn{4}{c}{\textbf{MoveBench-Multi}} \\
\cmidrule(lr){3-6} \cmidrule(lr){7-10}
& Method & PSNR$\uparrow$ & LPIPS$\downarrow$ & FVD$\downarrow$ & EPE$\downarrow$ & PSNR$\uparrow$ & LPIPS$\downarrow$ & FVD$\downarrow$ & EPE$\downarrow$ \\
\midrule
\multirow{3}{*}{\rotatebox{90}{\scriptsize Wan 14B}}
& \textbf{Ours} & \textbf{18.65} & \textbf{.276} & \textbf{11.51} & 2.39 & \textbf{18.95} & \textbf{.244} & \textbf{26.91} & \textbf{1.74} \\
& ATI & 17.18 & .336 & 12.82 & 2.61 & 18.57 & .274 & 28.62 & 1.98 \\
& Wan-Move & 17.57 & .293 & 11.67 & \textbf{2.28} & 17.83 & .259 & 27.32 & 1.79 \\
\midrule
\multirow{3}{*}{\rotatebox{90}{\scriptsize CogV}}
& \textbf{Ours} & \textbf{18.83} & \textbf{.267} & \textbf{10.94} & \textbf{1.97} & \textbf{19.35} & \textbf{.228} & \textbf{26.03} & \textbf{1.62} \\
& MagicMotion & 18.19 & .300 & 13.37 & 2.51 & 19.17 & .242 & 27.41 & 1.71 \\
& Tora & 15.88 & .431 & 44.75 & 3.03 & 17.51 & .319 & 42.23 & 1.96 \\
\bottomrule
\end{tabular}
\end{table}

\begin{figure*}[p]

\centering
\setlength{\tabcolsep}{0.5pt}
\renewcommand{\arraystretch}{0.3}
\newcommand{\qimg}[1]{\includegraphics[width=0.155\textwidth]{#1}}
\newcommand{\framesarrow}[1]{%
  \begin{tikzpicture}[baseline=0pt]
    \draw[->, line width=0.5pt] (0,0) -- (0.465\textwidth, 0)
      node[midway, fill=white, inner sep=1pt] {\scriptsize #1};
  \end{tikzpicture}%
}
\begin{tabular}{@{}c ccc @{\hspace{6pt}} c ccc@{}}
& \multicolumn{3}{c}{\framesarrow{MoveBench-Multi frames}} & & \multicolumn{3}{c}{\framesarrow{MoveBench-Multi frames}} \\[1pt]
\rotatebox{90}{\scriptsize\hspace{8pt}GT} &
\qimg{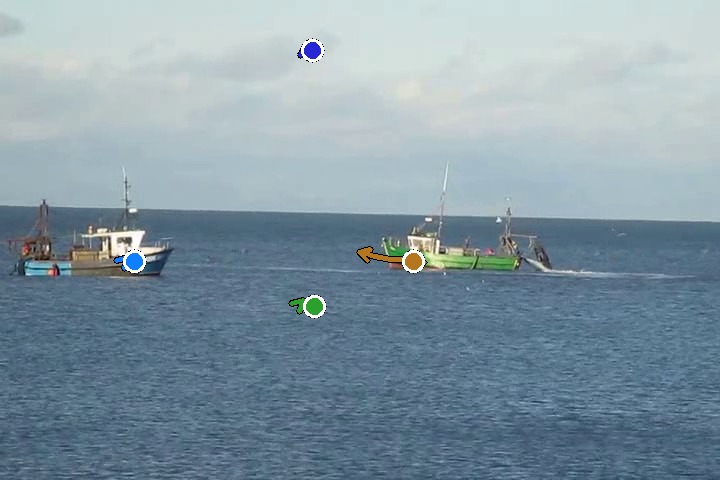} &
\qimg{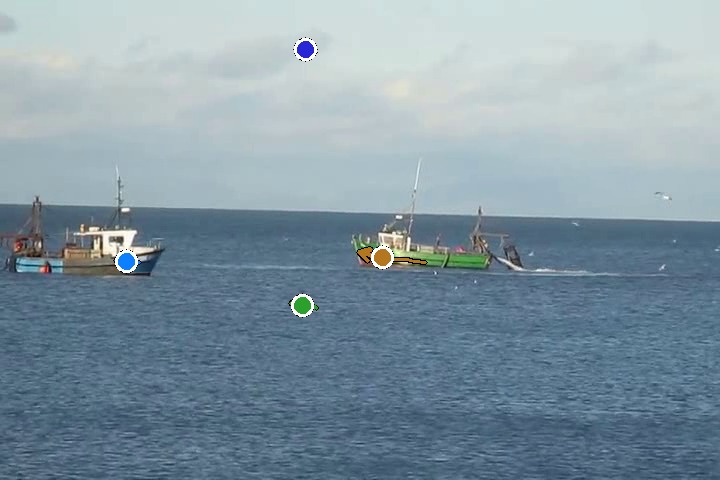} &
\qimg{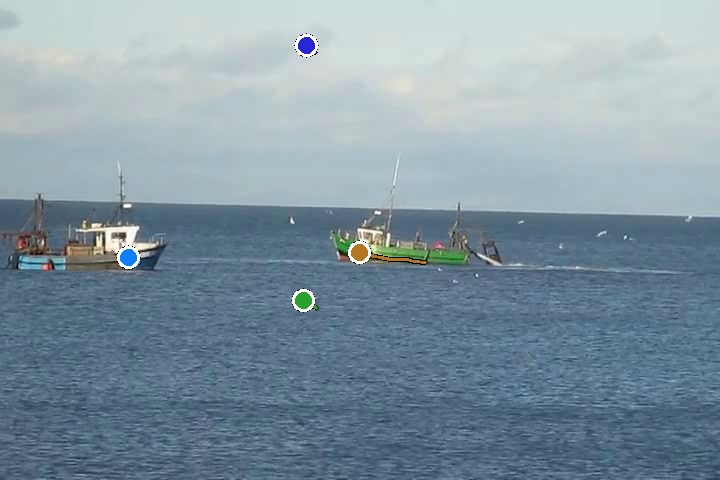} &
\rotatebox{90}{\scriptsize\hspace{8pt}GT} &
\qimg{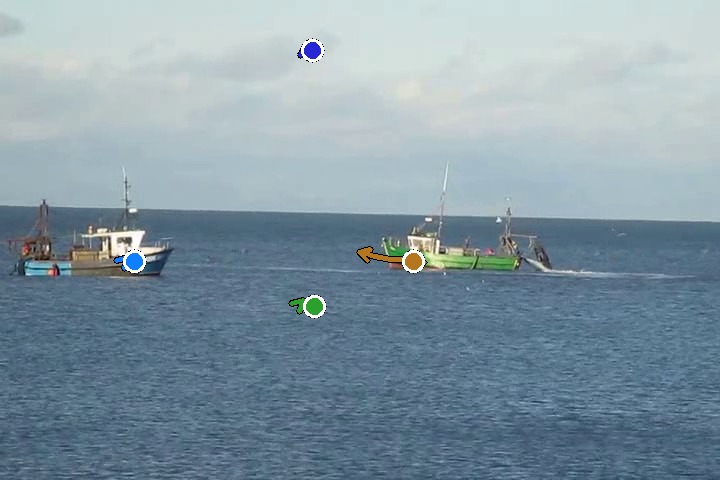} &
\qimg{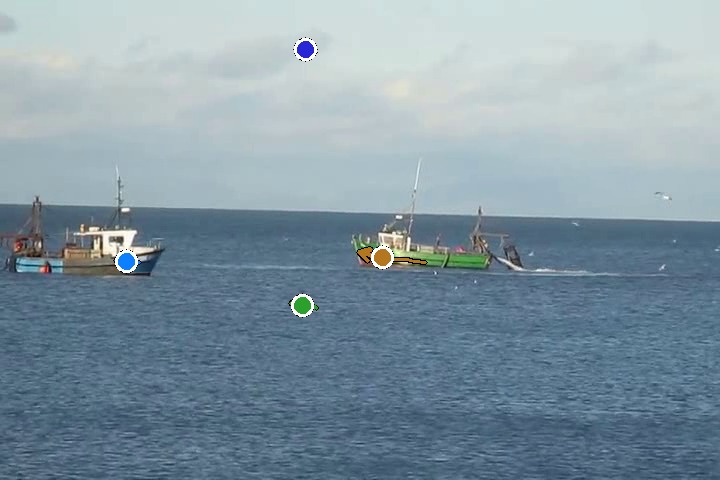} &
\qimg{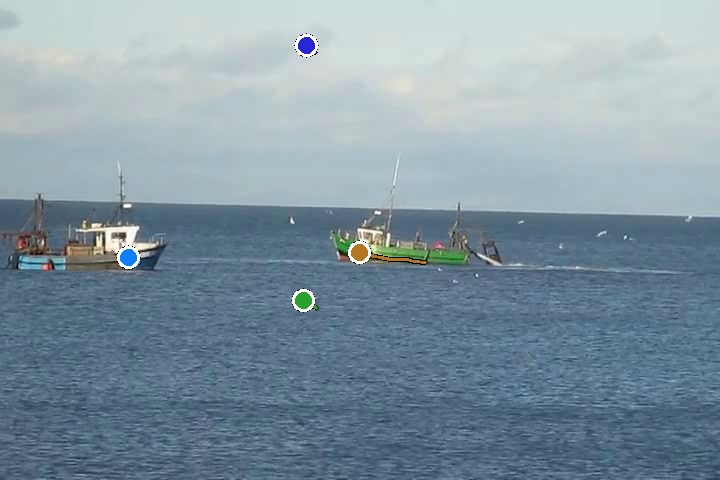} \\[0.5pt]
\rotatebox{90}{\scriptsize\hspace{4pt}Ours-CogV} &
\qimg{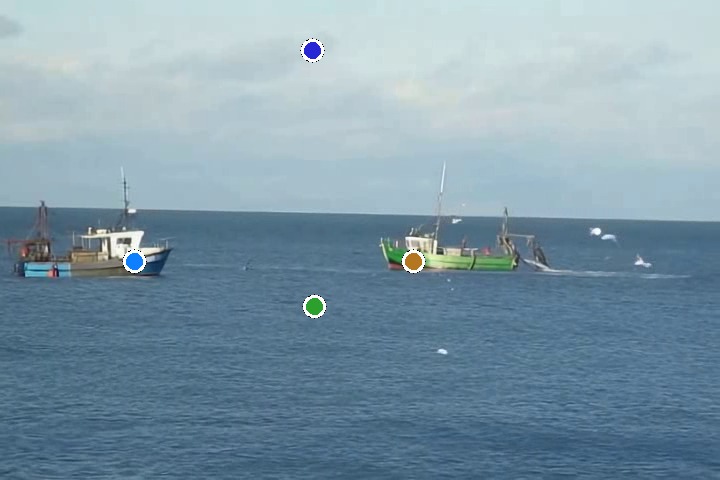} &
\qimg{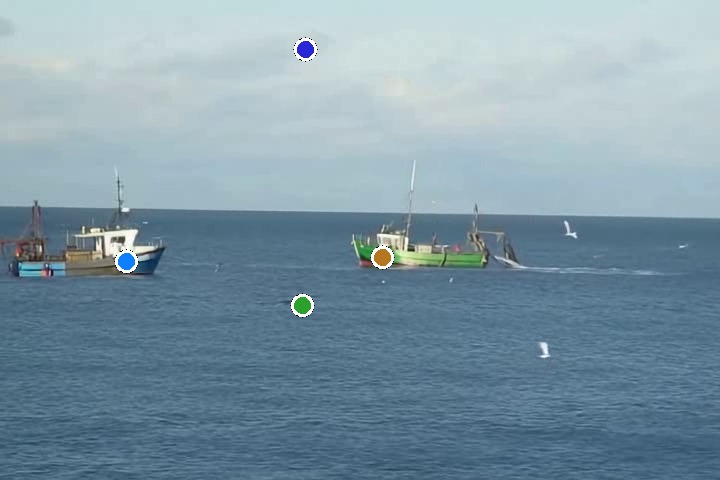} &
\qimg{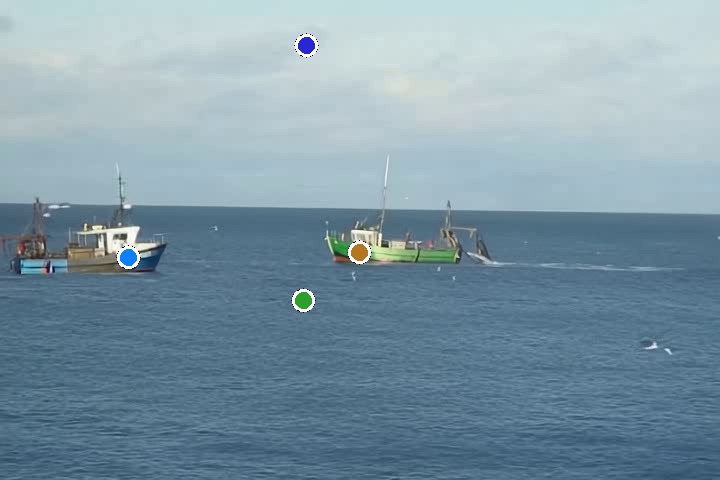} &
\rotatebox{90}{\scriptsize\hspace{4pt}Ours-WaN} &
\qimg{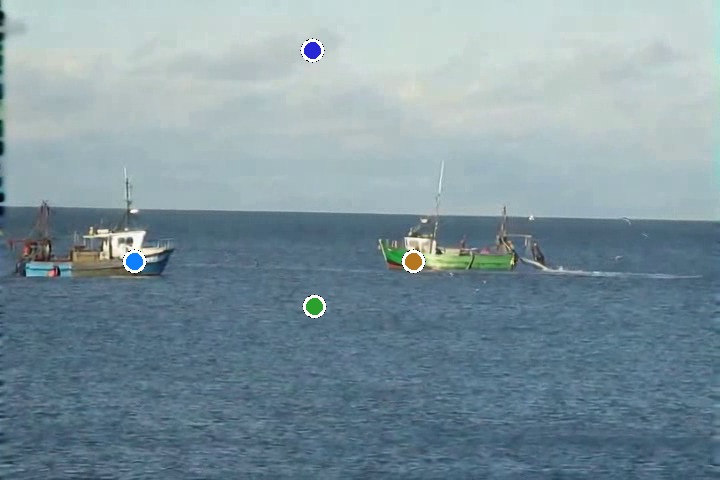} &
\qimg{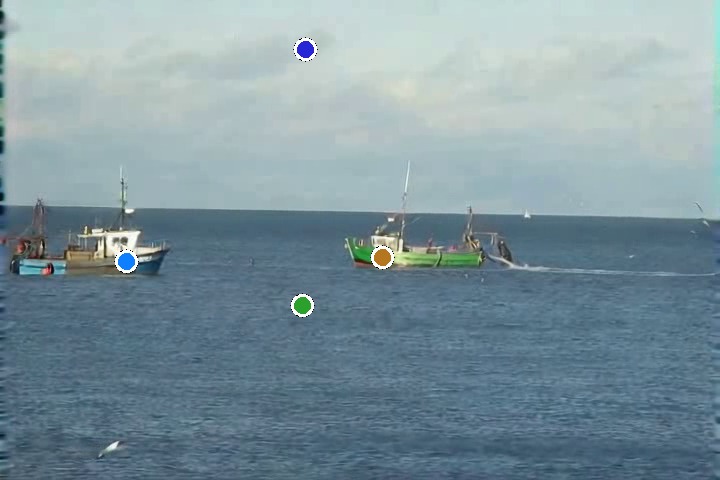} &
\qimg{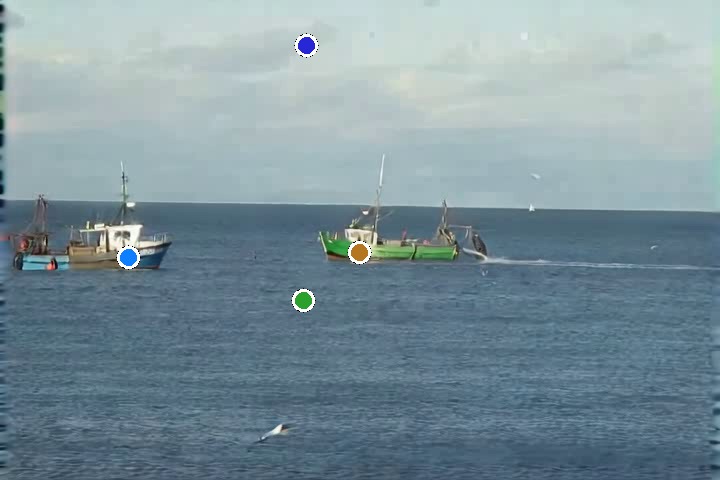} \\[0.5pt]
\rotatebox{90}{\scriptsize\hspace{4pt}MagicMotion} &
\qimg{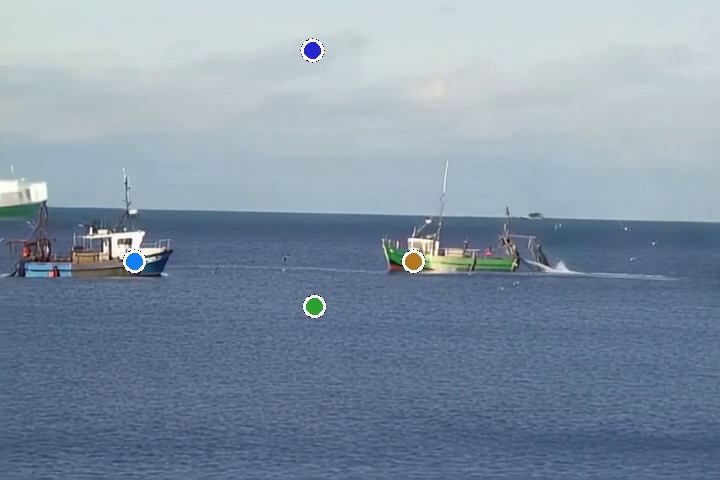} &
\qimg{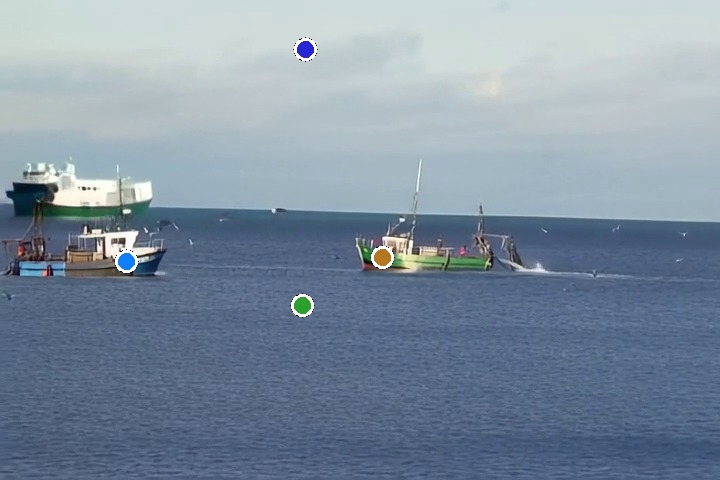} &
\qimg{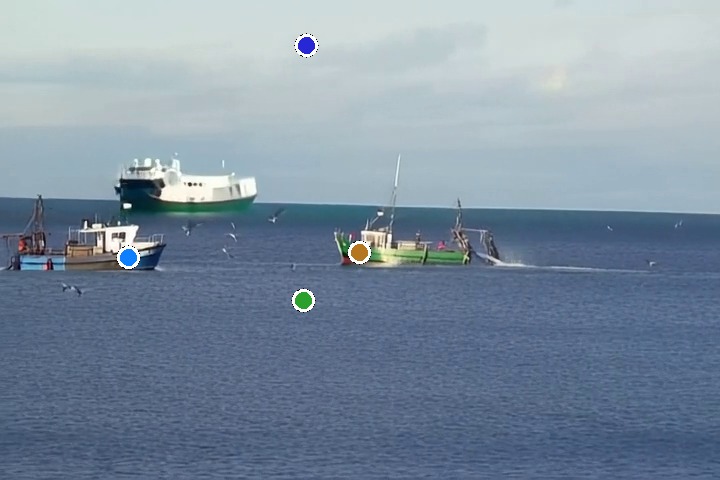} &
\rotatebox{90}{\scriptsize\hspace{8pt}ATI} &
\qimg{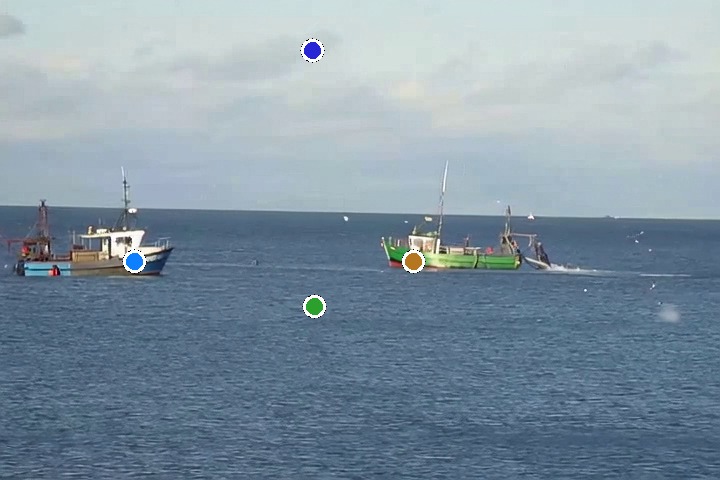} &
\qimg{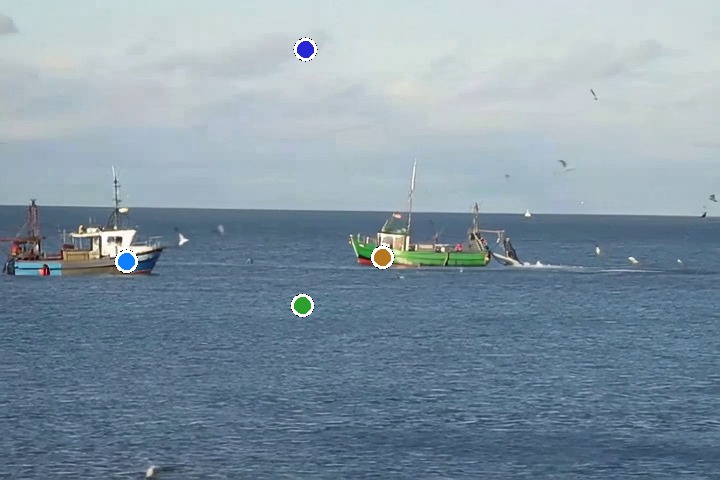} &
\qimg{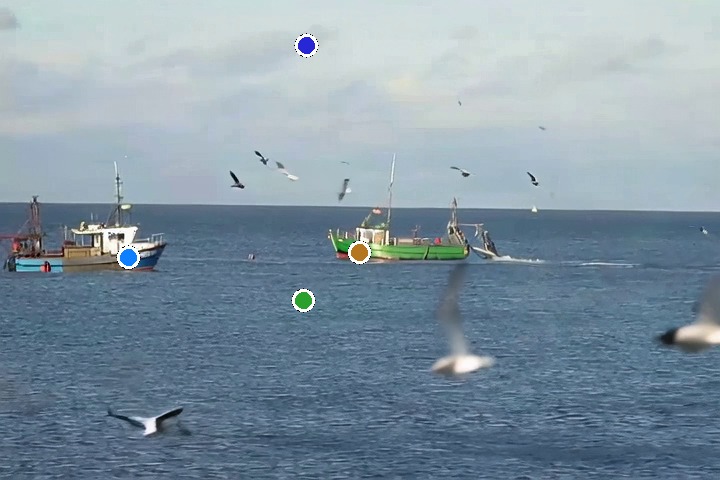} \\[0.5pt]
\rotatebox{90}{\scriptsize\hspace{8pt}Tora} &
\qimg{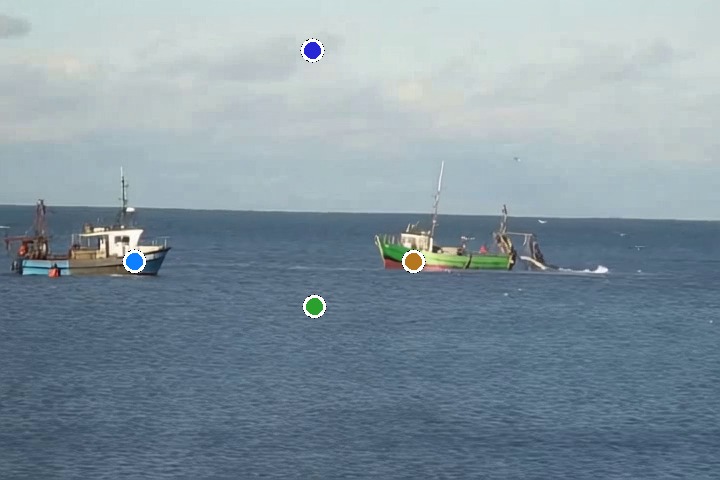} &
\qimg{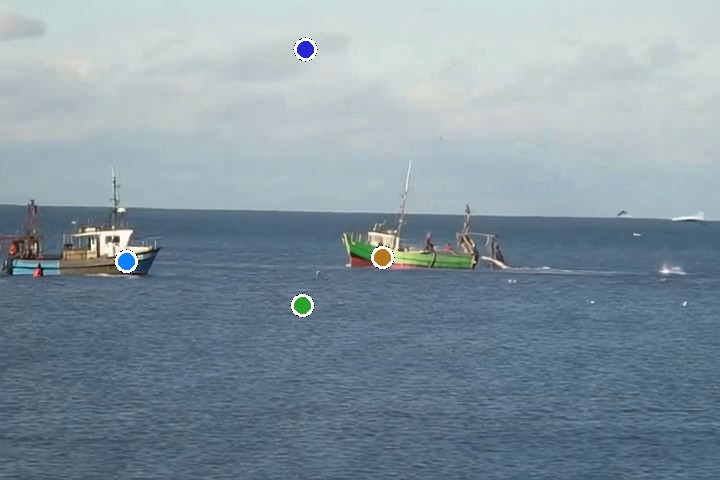} &
\qimg{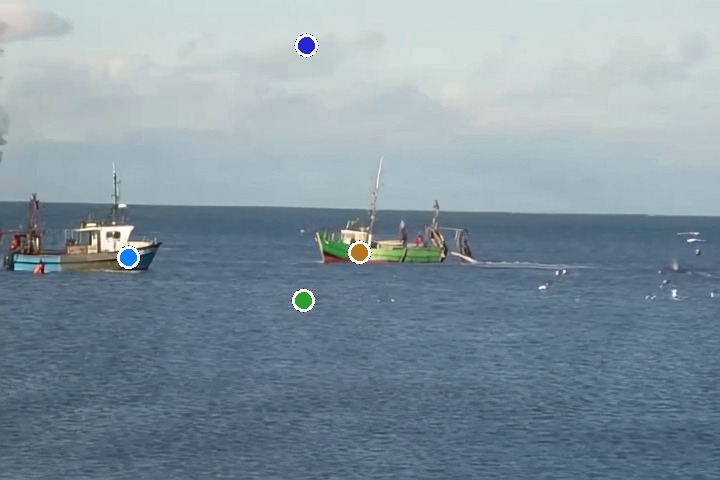} &
\rotatebox{90}{\scriptsize\hspace{4pt}Wan-Move} &
\qimg{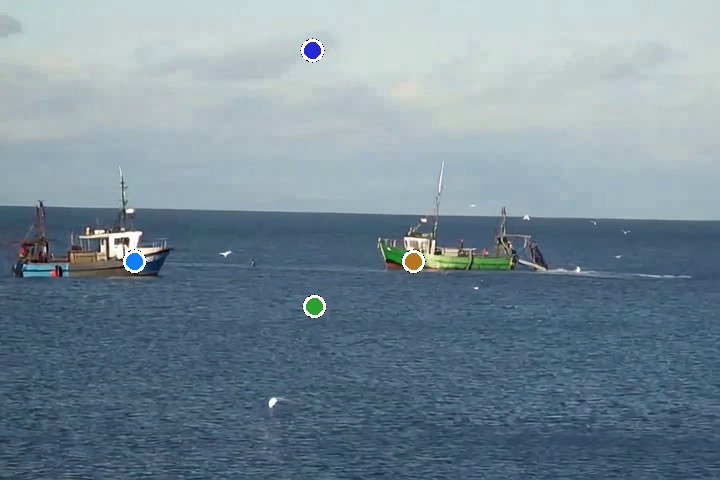} &
\qimg{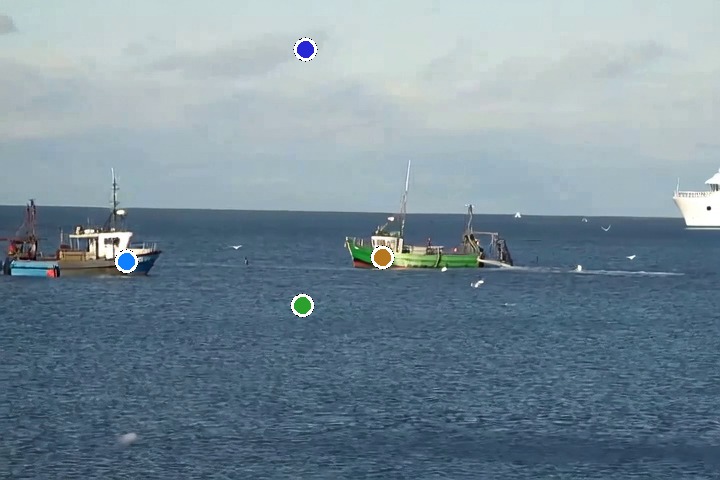} &
\qimg{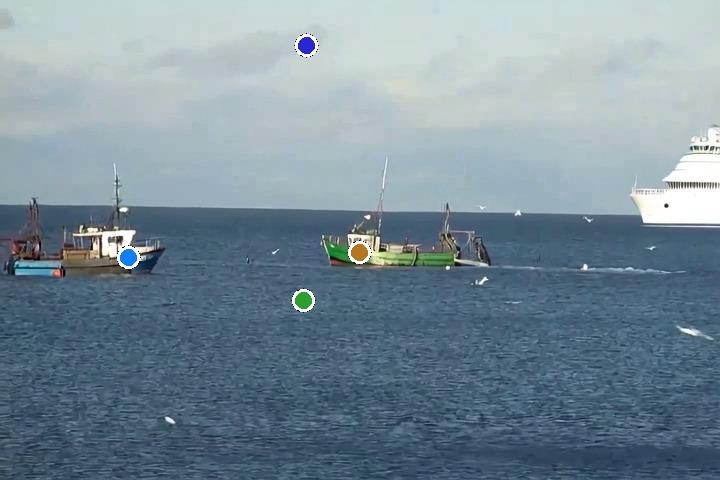} \\
\end{tabular}
\caption{\textbf{Qualitative comparison on MoveBench-Multi with our real-data models.} The same scene is shown for both backbones: CogVideoX-5B against MagicMotion and Tora (left) and WaN~2.1-14B against ATI and Wan-Move (right). Dots include the background regions given to every method as additional trajectories. MagicMotion and Wan-Move insert a large white ship that is absent from the ground truth, and ATI fills the foreground with seagulls, while our method keeps both fishing boats on their trajectories without inserting new vessels.}
\label{fig:qual_sav_movebench}

\vspace{14pt}
\begin{tabular}{@{}c ccc @{\hspace{6pt}} c ccc@{}}
& \multicolumn{3}{c}{\framesarrow{MagicBench frames}} & & \multicolumn{3}{c}{\framesarrow{MagicBench frames}} \\[1pt]
\rotatebox{90}{\scriptsize\hspace{8pt}GT} &
\qimg{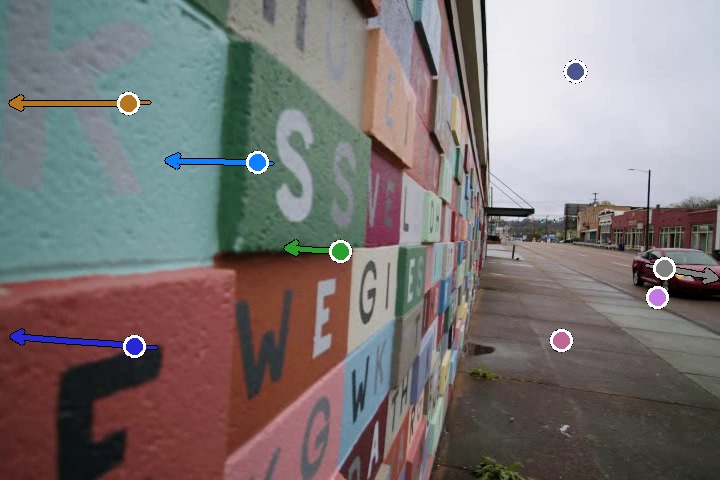} &
\qimg{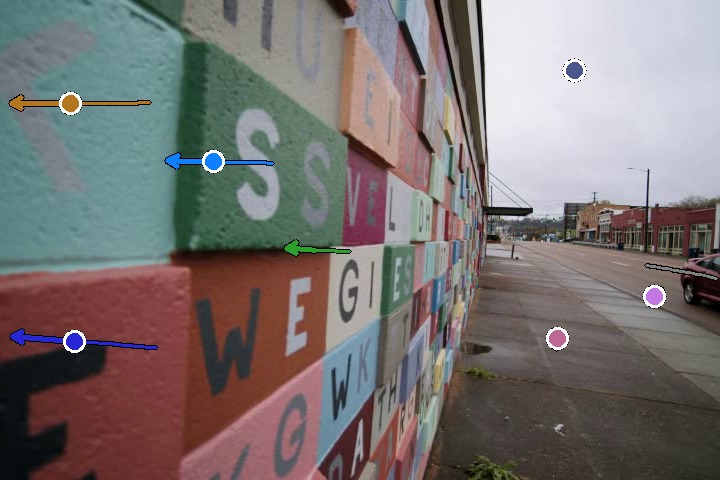} &
\qimg{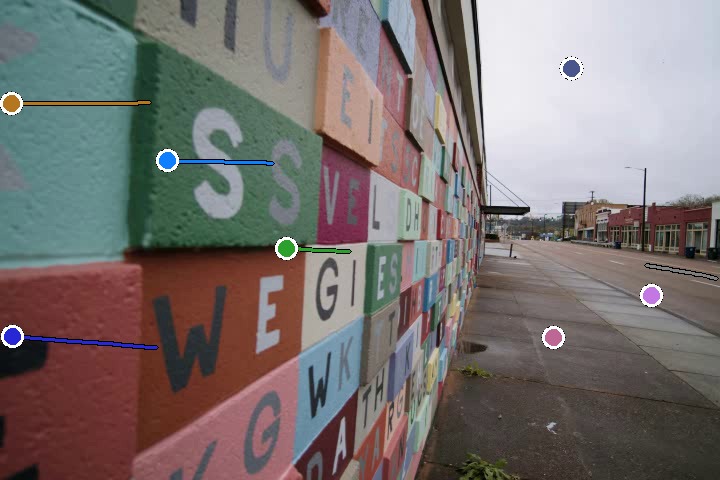} &
\rotatebox{90}{\scriptsize\hspace{8pt}GT} &
\qimg{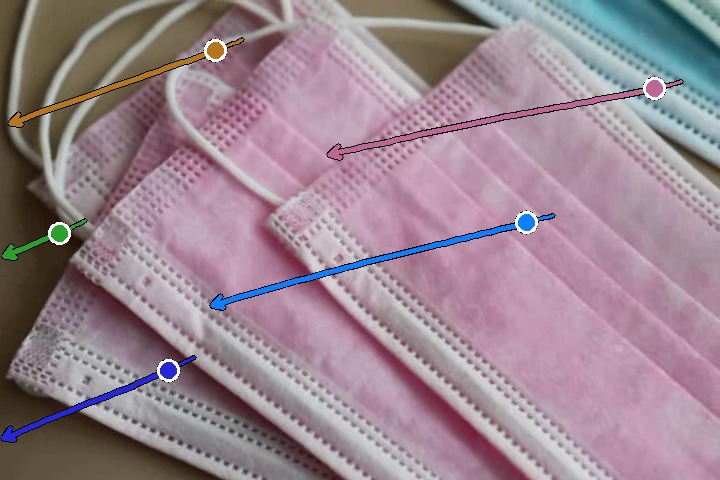} &
\qimg{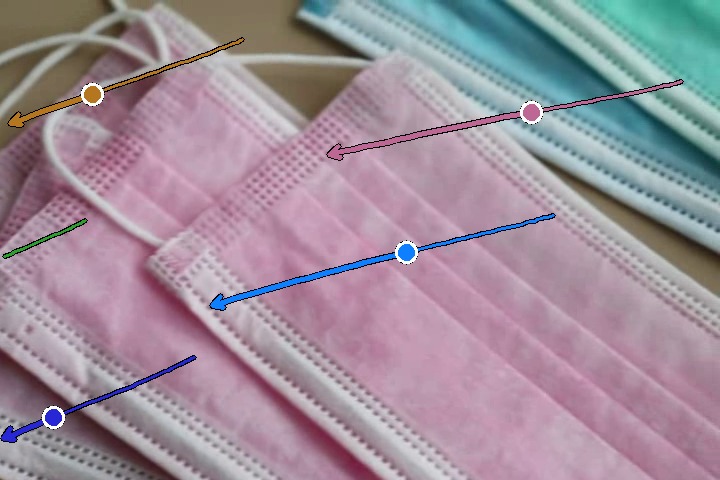} &
\qimg{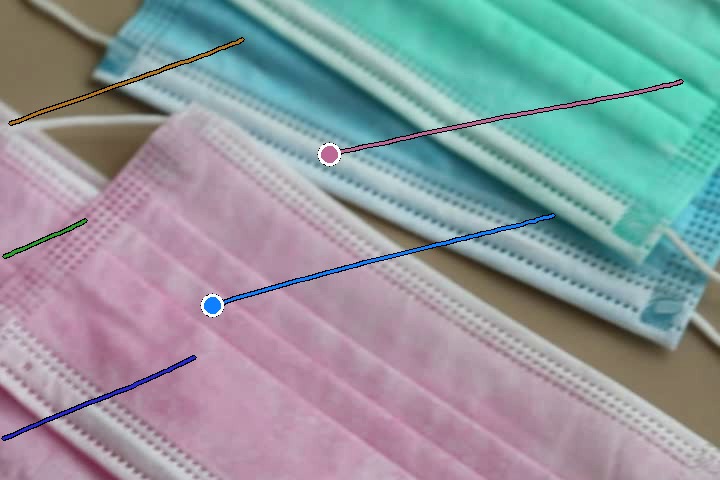} \\[0.5pt]
\rotatebox{90}{\scriptsize\hspace{4pt}Ours-CogV} &
\qimg{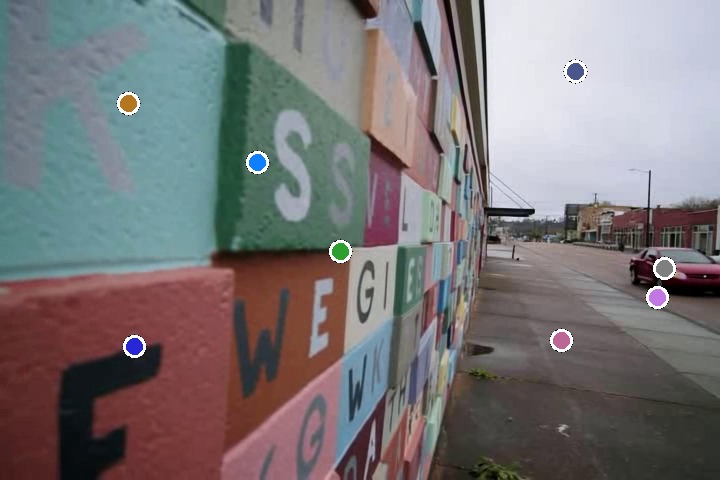} &
\qimg{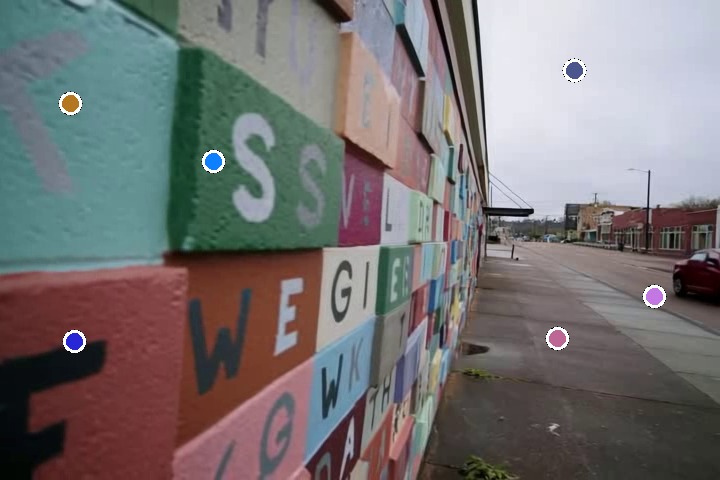} &
\qimg{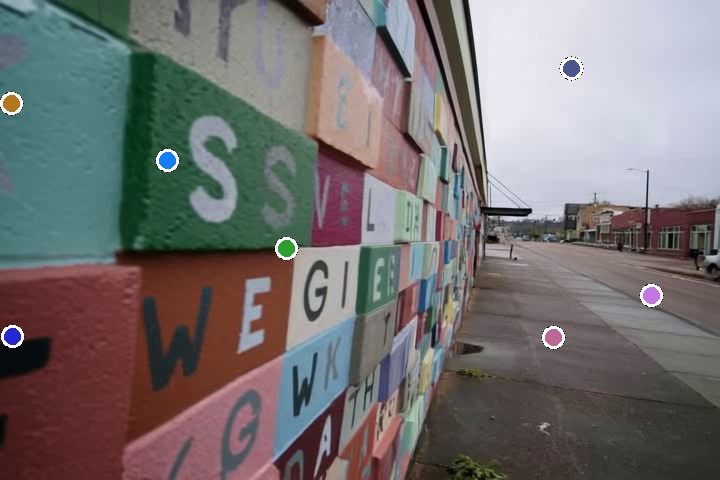} &
\rotatebox{90}{\scriptsize\hspace{4pt}Ours-WaN} &
\qimg{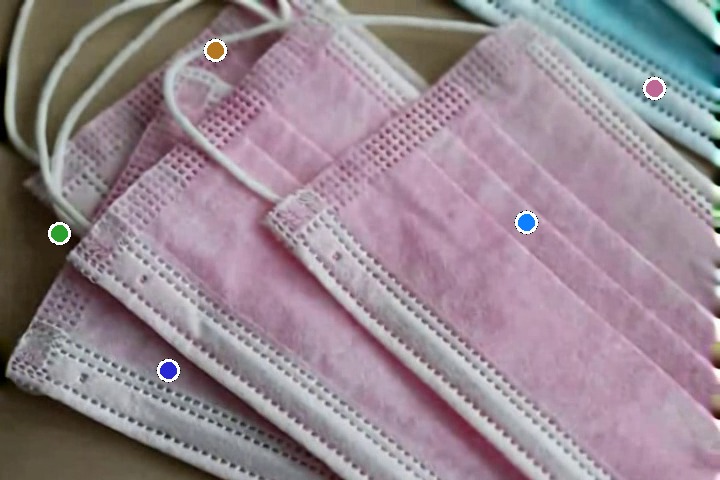} &
\qimg{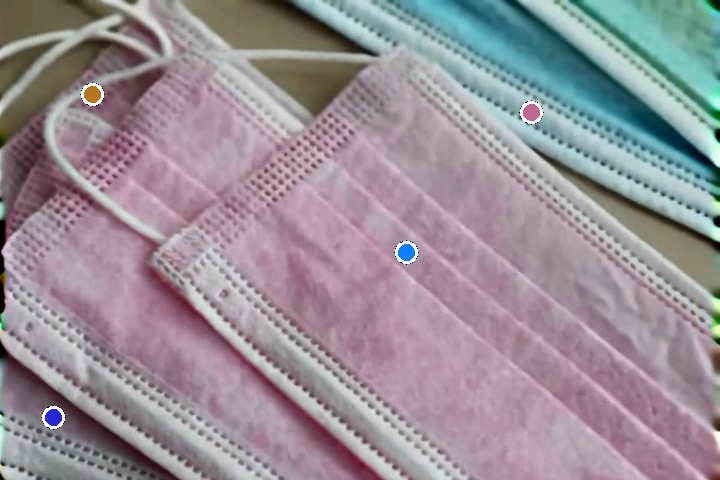} &
\qimg{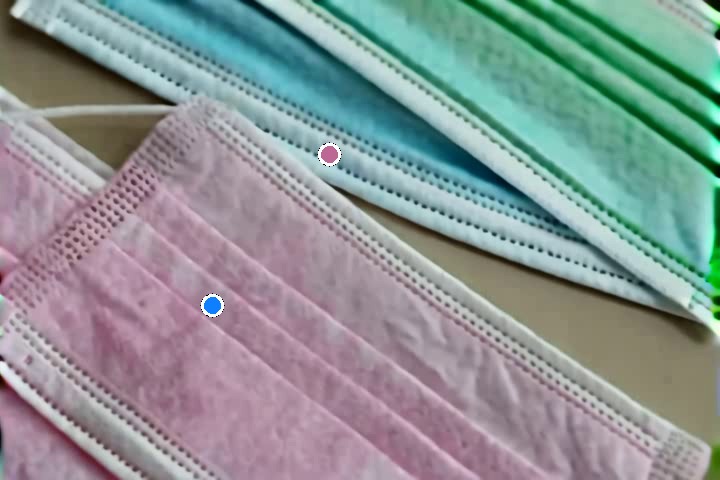} \\[0.5pt]
\rotatebox{90}{\scriptsize\hspace{4pt}MagicMotion} &
\qimg{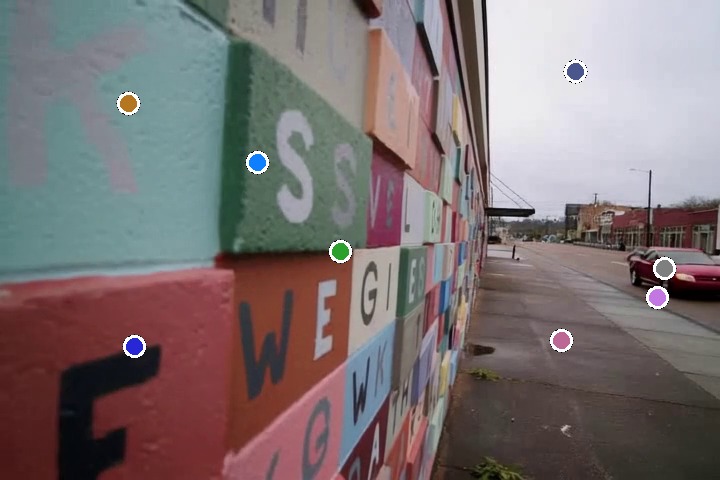} &
\qimg{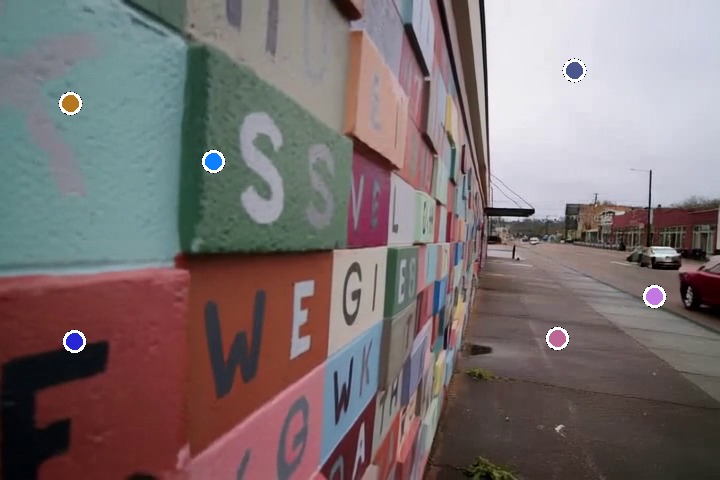} &
\qimg{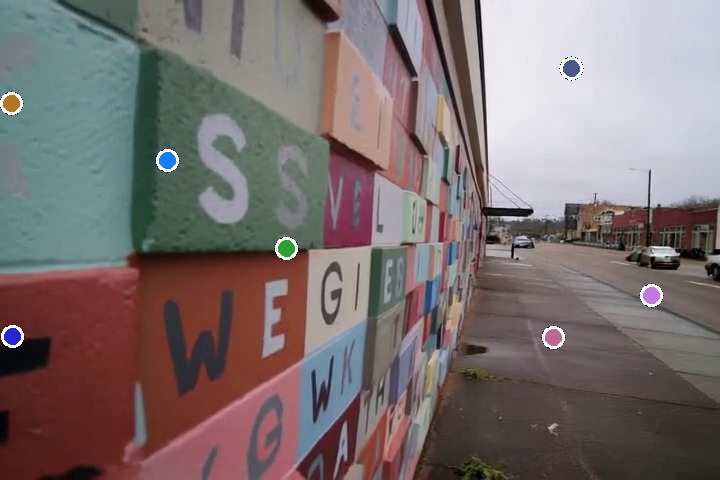} &
\rotatebox{90}{\scriptsize\hspace{8pt}ATI} &
\qimg{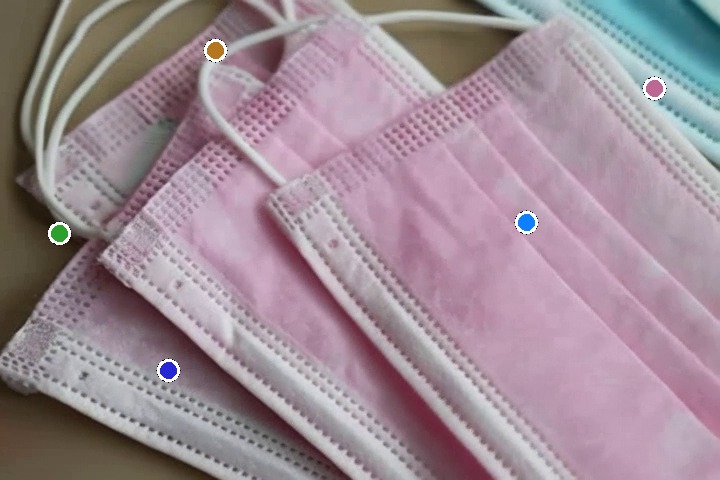} &
\qimg{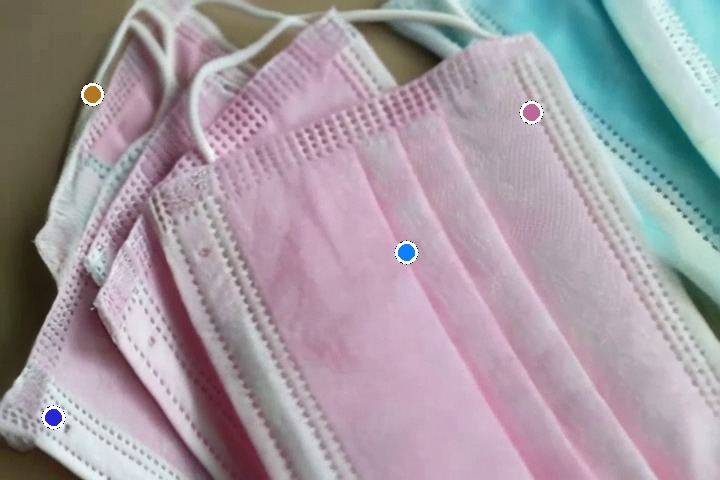} &
\qimg{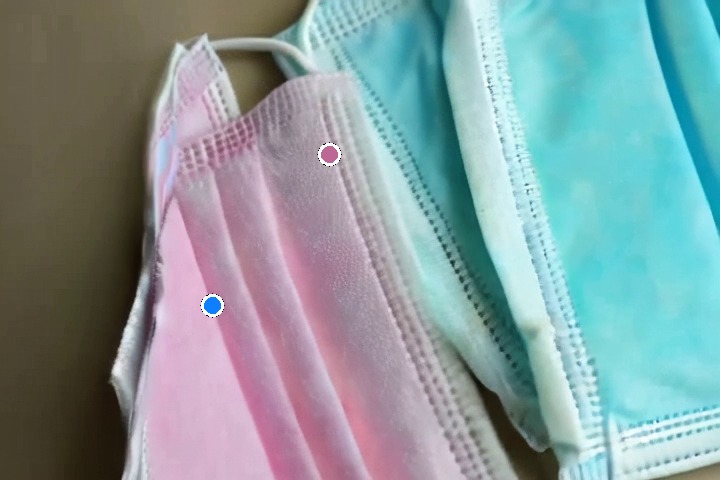} \\[0.5pt]
\rotatebox{90}{\scriptsize\hspace{8pt}Tora} &
\qimg{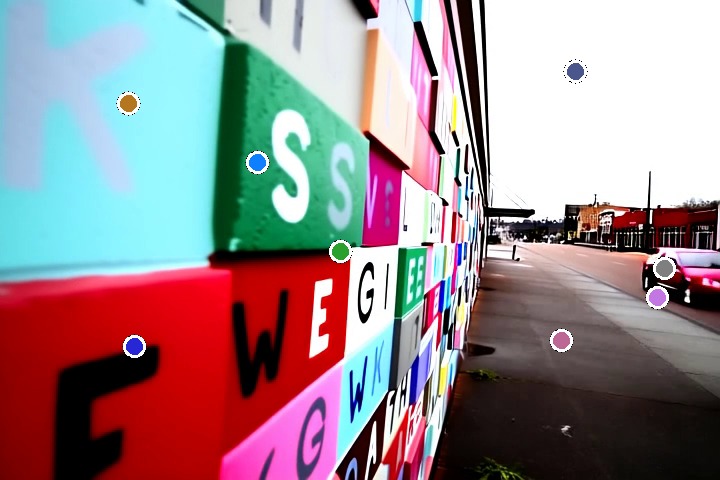} &
\qimg{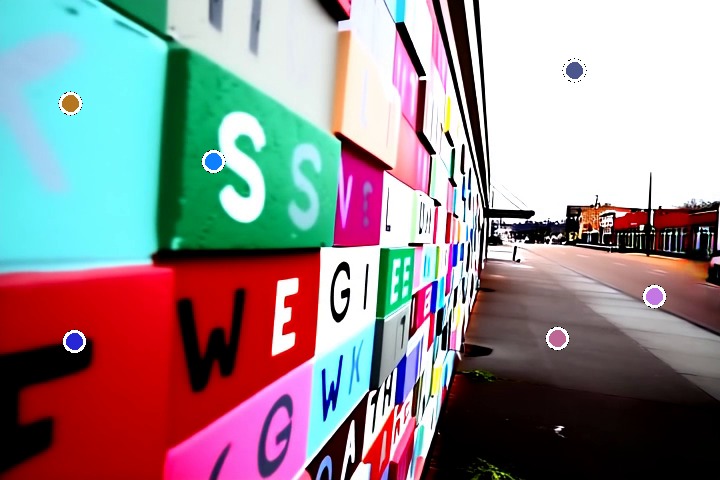} &
\qimg{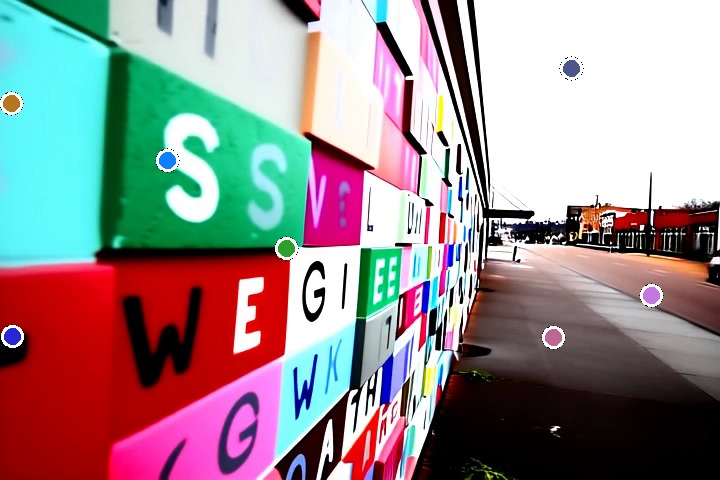} &
\rotatebox{90}{\scriptsize\hspace{4pt}Wan-Move} &
\qimg{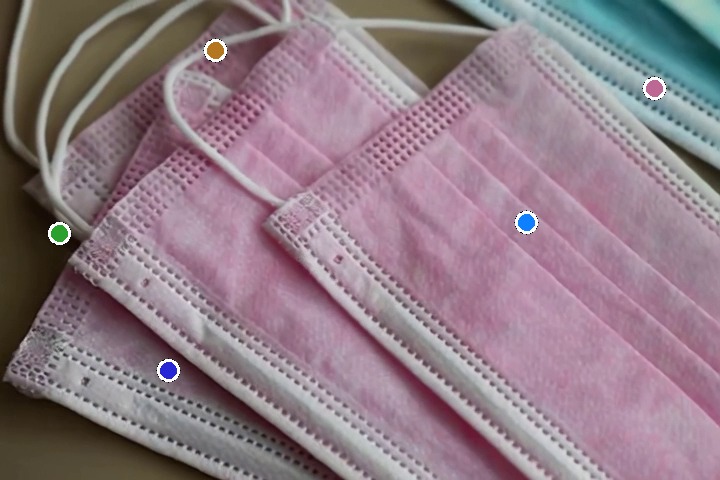} &
\qimg{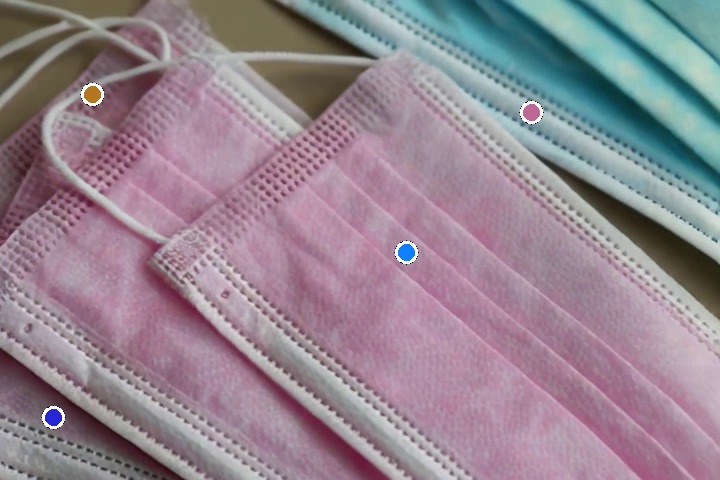} &
\qimg{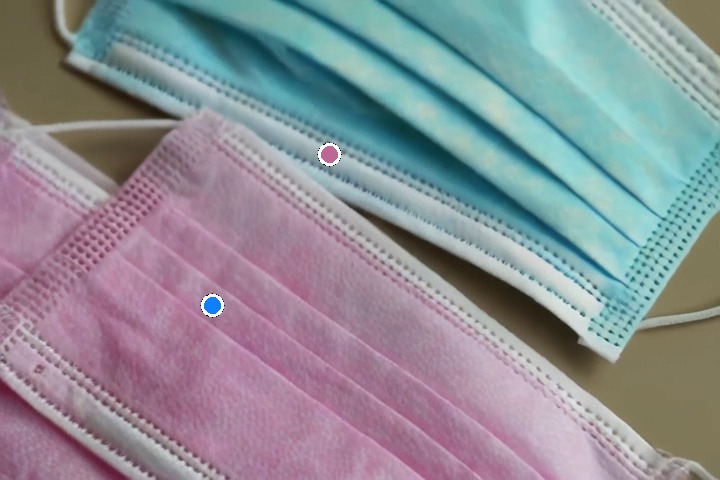} \\
\end{tabular}
\caption{\textbf{Qualitative comparison on MagicBench with our real-data models.} Both scenes are filmed by a moving camera, whose motion is conveyed to every method through the trajectories of background regions. Left: CogVideoX-5B against MagicMotion and Tora; Tora oversaturates the painted wall and washes out the sky. Right: WaN~2.1-14B against ATI and Wan-Move; ATI folds and distorts the masks as the camera pans. Our method follows the camera motion of the ground truth, revealing the blue masks at the top right, while preserving the appearance of the scene.}
\label{fig:qual_sav_magicbench}

\end{figure*}

\begin{figure*}[p]

\centering
\setlength{\tabcolsep}{0.8pt}
\renewcommand{\arraystretch}{0.2}
\newcommand{\soloimg}[1]{\includegraphics[width=0.236\textwidth]{#1}}
\newcommand{\solorow}[3]{%
\rotatebox{90}{\scriptsize\hspace{6pt}#3} &
\soloimg{figures/qual/sav_noeq/#1_gt_f0.jpg} &
\soloimg{figures/qual/sav_noeq/#1_ours_#2_f8.jpg} &
\soloimg{figures/qual/sav_noeq/#1_ours_#2_f28.jpg} &
\soloimg{figures/qual/sav_noeq/#1_ours_#2_f48.jpg}}
\newcommand{\soloframesarrow}{%
  \begin{tikzpicture}[baseline=(mid.base)]
    \draw[->, line width=0.5pt] (0,0) -- (0.712\textwidth, 0);
    \node[fill=white, inner sep=1pt] (mid) at (0.356\textwidth, 0) {\small frames};
  \end{tikzpicture}%
}
\begin{tabular}{@{}c c@{\hspace{5pt}}ccc@{}}
& \makebox[0.236\textwidth]{\small condition} & \multicolumn{3}{c}{\soloframesarrow} \\[1pt]
\solorow{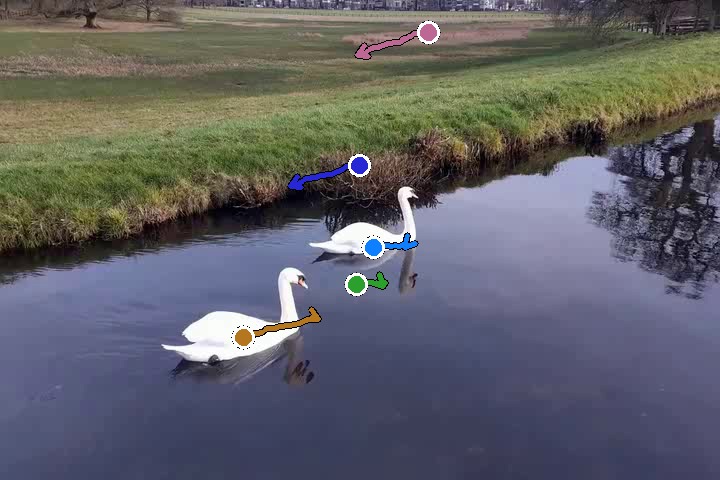}{wan}{WaN} \\[0.8pt]
\solorow{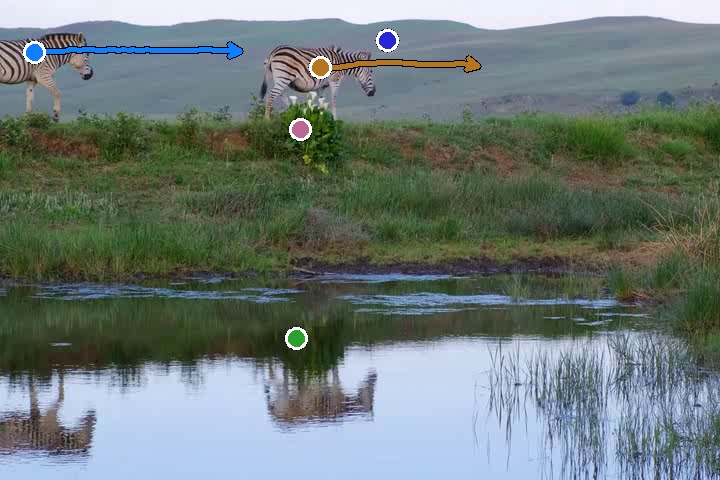}{wan}{WaN} \\[0.8pt]
\solorow{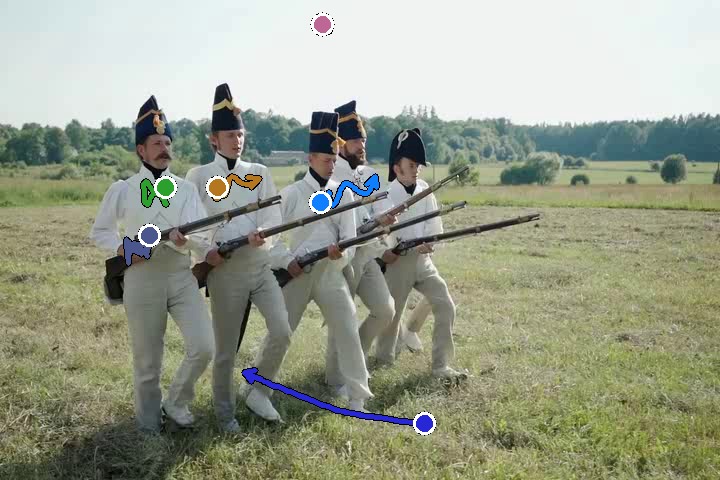}{cogv}{CogV} \\[0.8pt]
\solorow{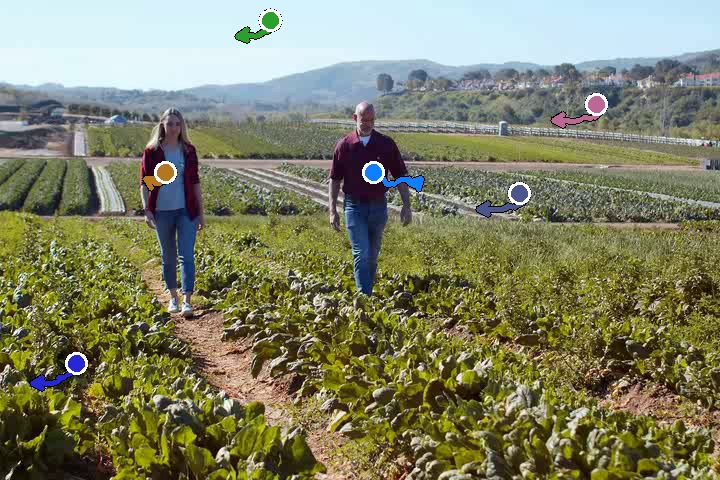}{wan}{WaN} \\[0.8pt]
\solorow{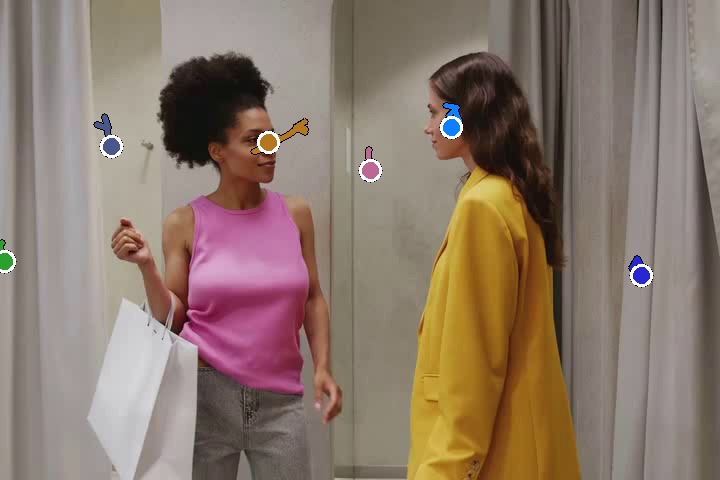}{cogv}{CogV} \\[0.8pt]
\solorow{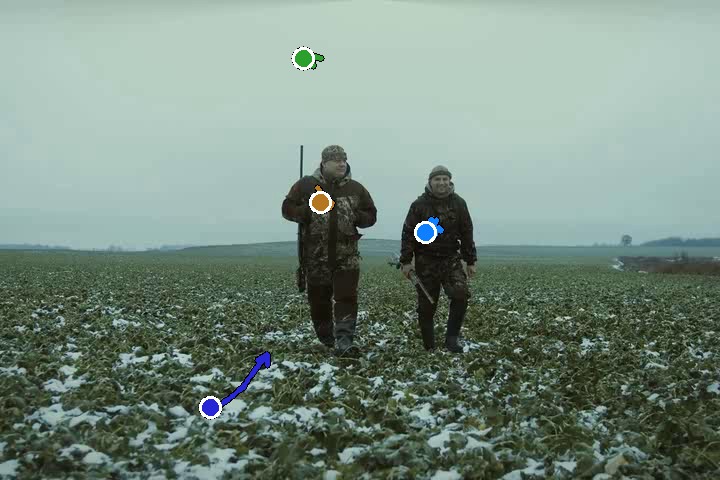}{wan}{WaN} \\[0.8pt]
\solorow{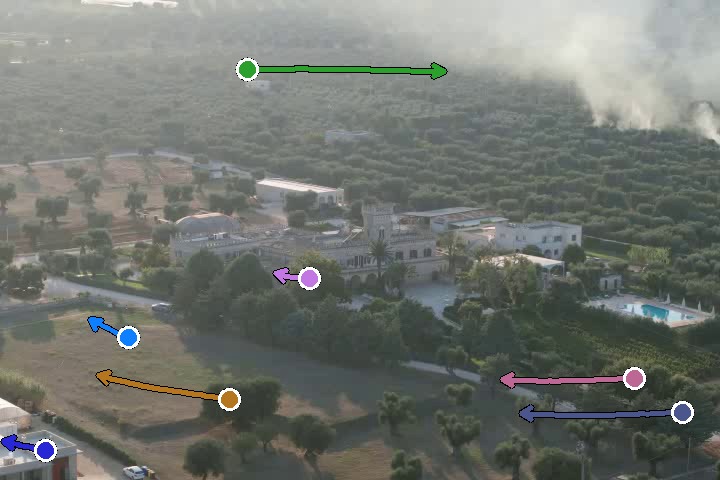}{cogv}{CogV} \\[0.8pt]
\solorow{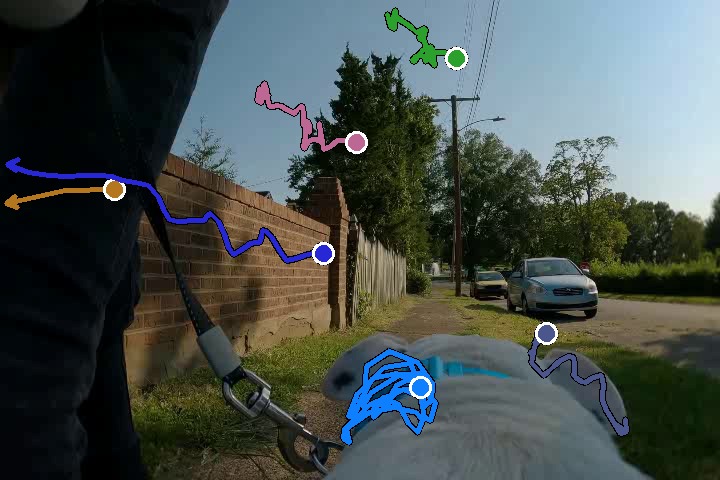}{wan}{WaN} \\
\end{tabular}
\caption{\textbf{Results of our real-data models on real, moving-camera footage.} Rows 1--6 are from MoveBench-Multi and rows 7--8 from MagicBench; the backbone is indicated on the left. The first column shows the input first frame with every prescribed trajectory, including the background regions that convey camera motion; the remaining columns show generated frames 9, 29 and 49 with the target position of each trajectory as a colored dot. Full videos are on the project page.}
\label{fig:qual_sav_solo}

\end{figure*}

\clearpage

\section{Additional Qualitative Comparisons}
\label{app:qual_extra}
\suppressfloats[t]

\begin{figure*}[!h]
\centering
\setlength{\tabcolsep}{0.5pt}
\renewcommand{\arraystretch}{0.3}
\newcommand{\qimg}[1]{\includegraphics[width=0.155\textwidth]{#1}}
\newcommand{\framesarrow}{%
  \begin{tikzpicture}[baseline=0pt]
    \draw[->, line width=0.5pt] (0,0) -- (0.465\textwidth, 0)
      node[midway, fill=white, inner sep=1pt] {\scriptsize frames};
  \end{tikzpicture}%
}
\begin{tabular}{@{}c ccc @{\hspace{6pt}} c ccc@{}}
& \multicolumn{3}{c}{\framesarrow} & & \multicolumn{3}{c}{\framesarrow} \\[1pt]
\rotatebox{90}{\scriptsize\hspace{8pt}GT} &
\qimg{figures/qual/movi_noeq/movi_cogv_gt_f8.jpg} &
\qimg{figures/qual/movi_noeq/movi_cogv_gt_f28.jpg} &
\qimg{figures/qual/movi_noeq/movi_cogv_gt_f48.jpg} &
\rotatebox{90}{\scriptsize\hspace{8pt}GT} &
\qimg{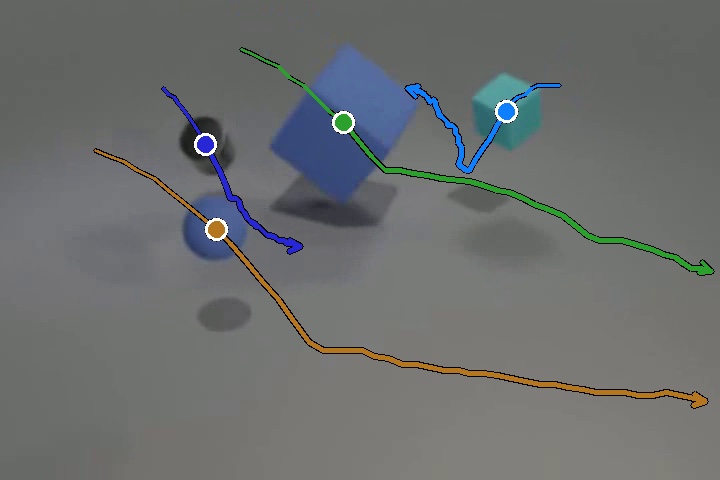} &
\qimg{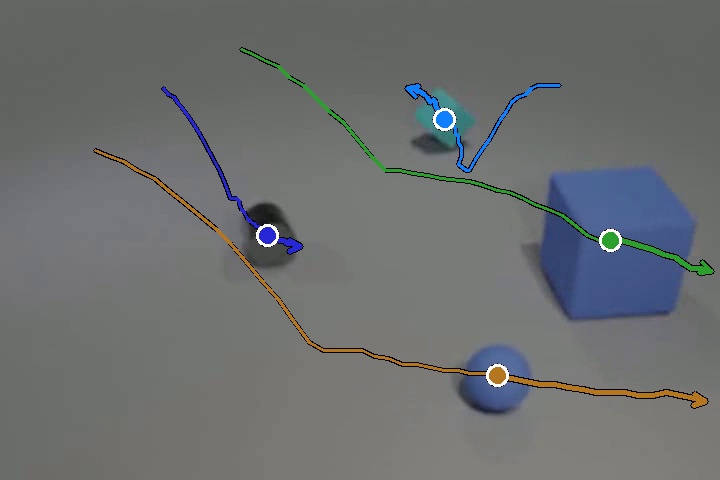} &
\qimg{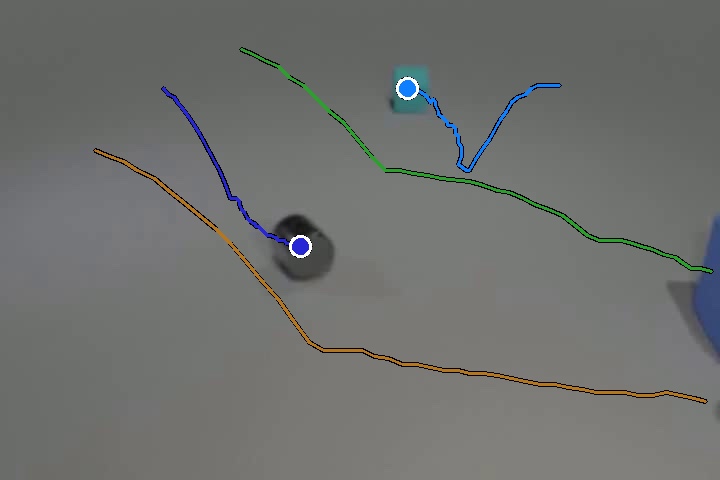} \\[0.5pt]
\rotatebox{90}{\scriptsize\hspace{4pt}Ours-CogV} &
\qimg{figures/qual/movi_noeq/movi_cogv_ours_cogv_f8.jpg} &
\qimg{figures/qual/movi_noeq/movi_cogv_ours_cogv_f28.jpg} &
\qimg{figures/qual/movi_noeq/movi_cogv_ours_cogv_f48.jpg} &
\rotatebox{90}{\scriptsize\hspace{4pt}Ours-WaN} &
\qimg{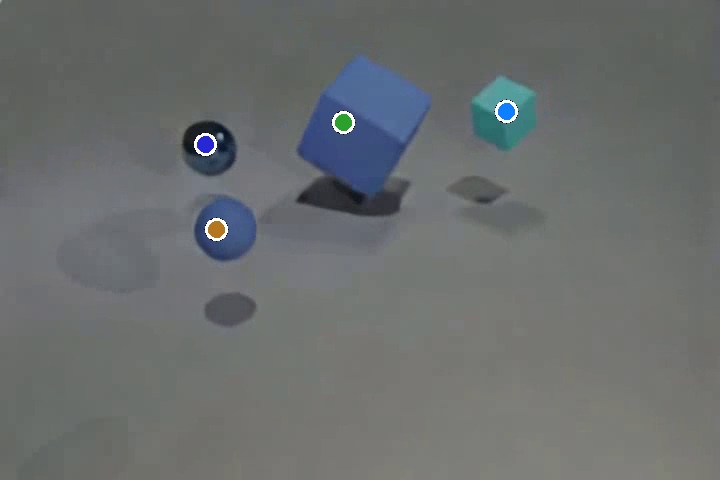} &
\qimg{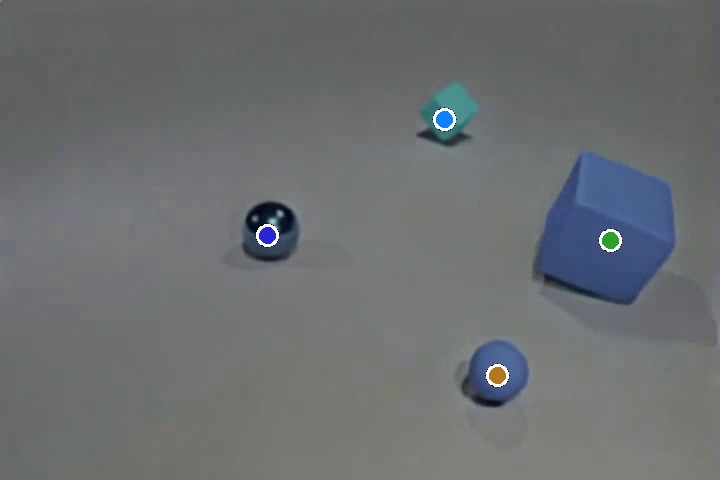} &
\qimg{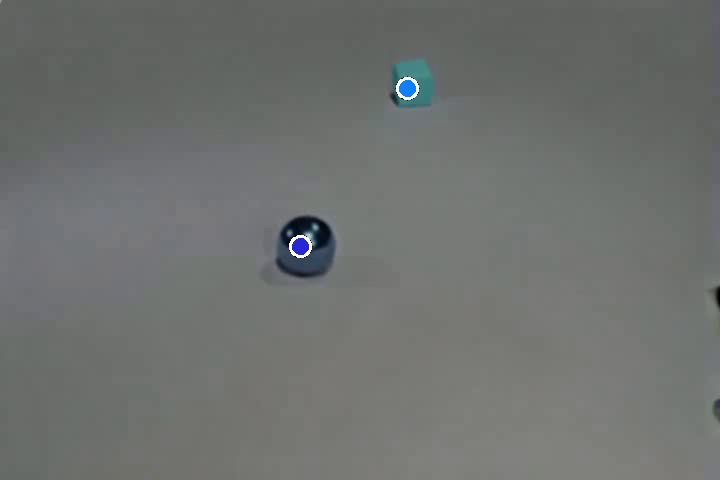} \\[0.5pt]
\rotatebox{90}{\scriptsize\hspace{4pt}MagicMotion} &
\qimg{figures/qual/movi_noeq/movi_cogv_magicmotion_f8.jpg} &
\qimg{figures/qual/movi_noeq/movi_cogv_magicmotion_f28.jpg} &
\qimg{figures/qual/movi_noeq/movi_cogv_magicmotion_f48.jpg} &
\rotatebox{90}{\scriptsize\hspace{8pt}ATI} &
\qimg{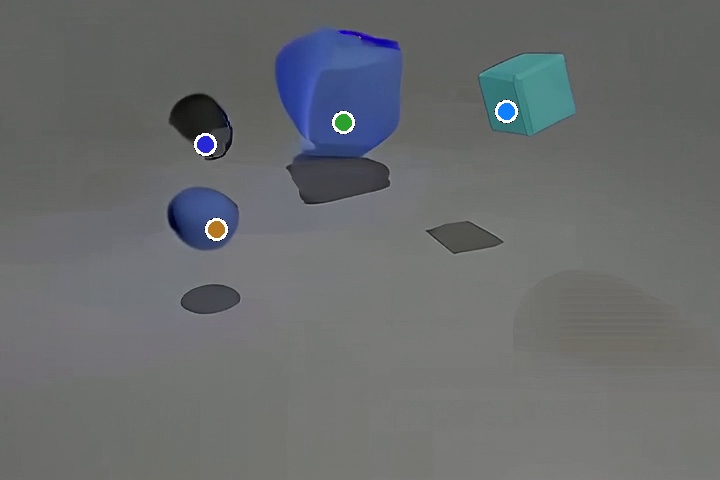} &
\qimg{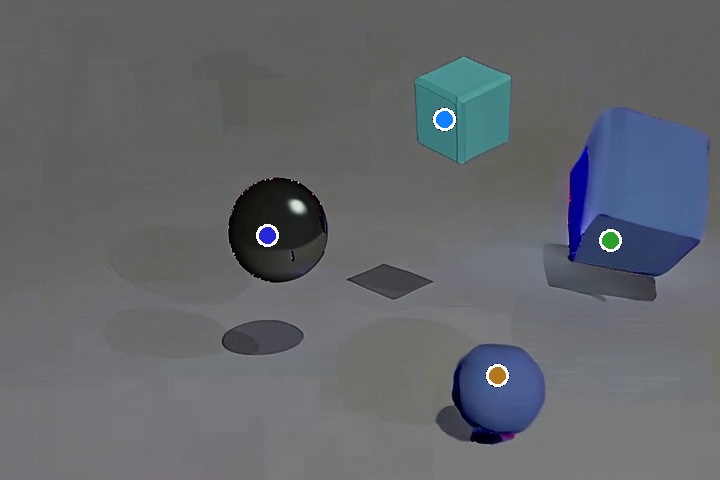} &
\qimg{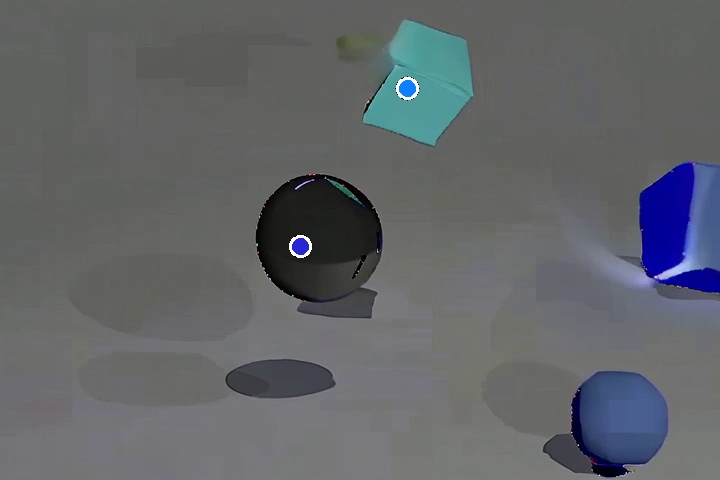} \\[0.5pt]
\rotatebox{90}{\scriptsize\hspace{8pt}Tora} &
\qimg{figures/qual/movi_noeq/movi_cogv_tora_f8.jpg} &
\qimg{figures/qual/movi_noeq/movi_cogv_tora_f28.jpg} &
\qimg{figures/qual/movi_noeq/movi_cogv_tora_f48.jpg} &
\rotatebox{90}{\scriptsize\hspace{4pt}Wan-Move} &
\qimg{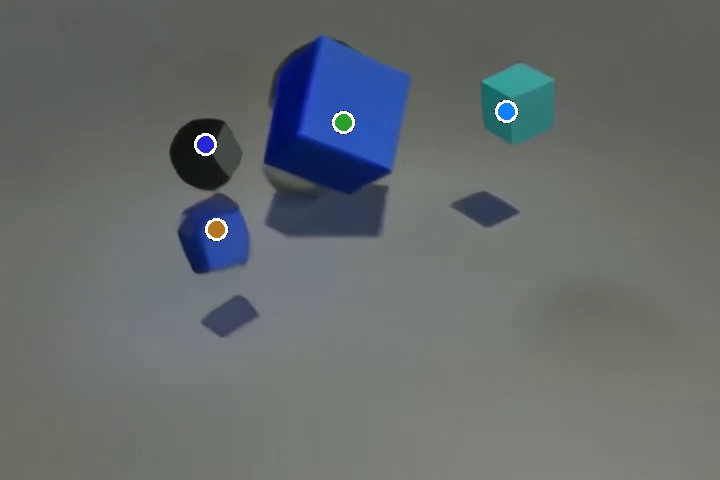} &
\qimg{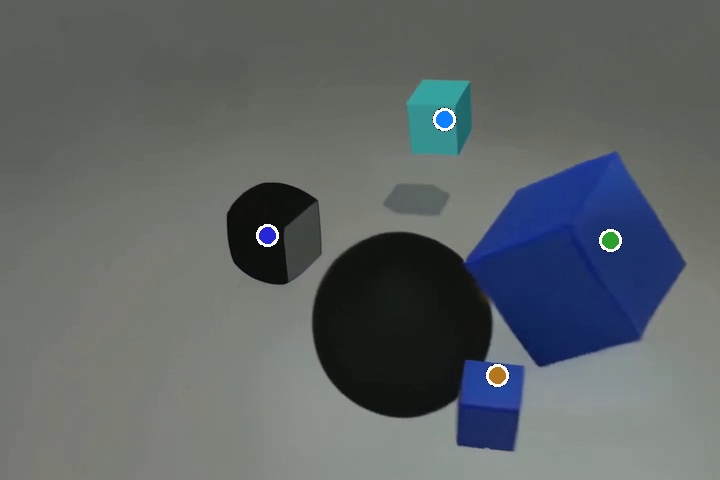} &
\qimg{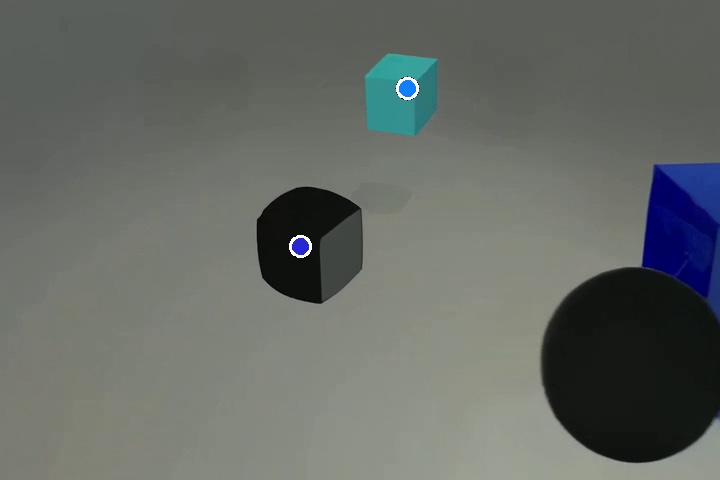} \\
\end{tabular}
\caption{\textbf{Additional qualitative comparison on MoVi.} Left: CogVideoX-5B against MagicMotion and Tora. Right: Wan2.1-14B against ATI and Wan-Move. MagicMotion fails to move the object at the orange dot, and Tora removes the blue-dot object from the scene entirely. ATI and Wan-Move retain the green and orange objects even after their trajectories have left the frame. Our method follows all trajectories with consistent identity in both backbones.}
\label{fig:qual_extra_movi}
\end{figure*}

\begin{figure*}[t]
\centering
\setlength{\tabcolsep}{0.5pt}
\renewcommand{\arraystretch}{0.3}
\newcommand{\qimg}[1]{\includegraphics[width=0.155\textwidth]{#1}}
\newcommand{\framesarrow}{%
  \begin{tikzpicture}[baseline=0pt]
    \draw[->, line width=0.5pt] (0,0) -- (0.465\textwidth, 0)
      node[midway, fill=white, inner sep=1pt] {\scriptsize frames};
  \end{tikzpicture}%
}
\begin{tabular}{@{}c ccc @{\hspace{6pt}} c ccc@{}}
& \multicolumn{3}{c}{\framesarrow} & & \multicolumn{3}{c}{\framesarrow} \\[1pt]
\rotatebox{90}{\scriptsize\hspace{8pt}GT} &
\qimg{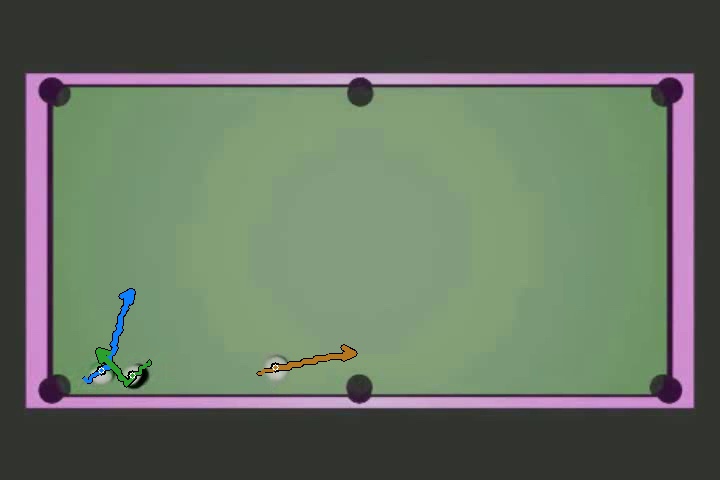} &
\qimg{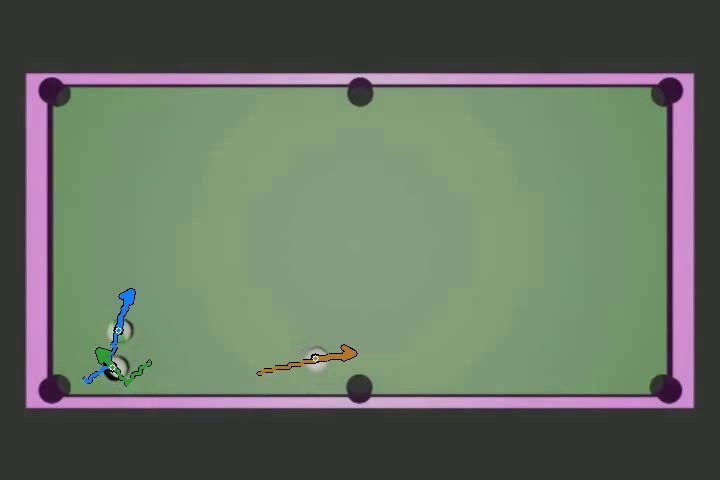} &
\qimg{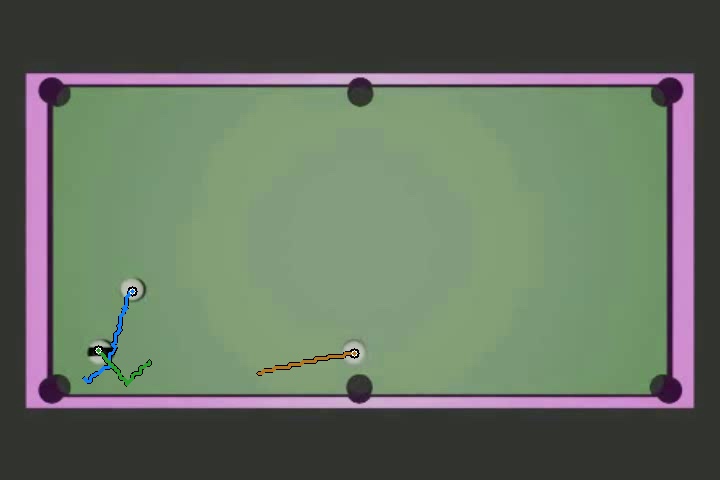} &
\rotatebox{90}{\scriptsize\hspace{8pt}GT} &
\qimg{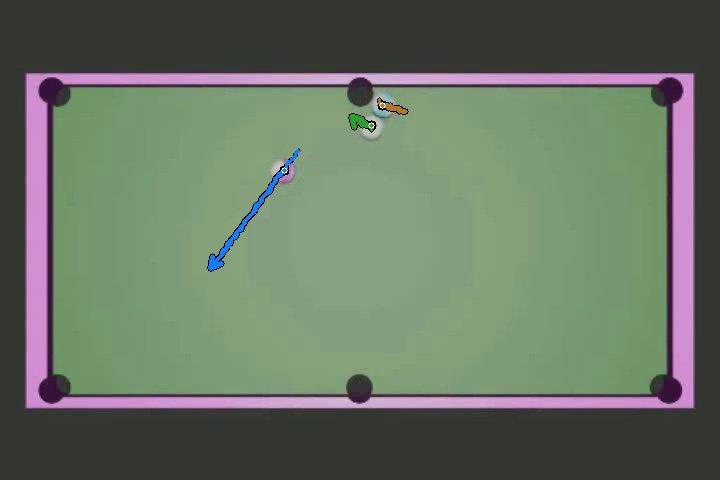} &
\qimg{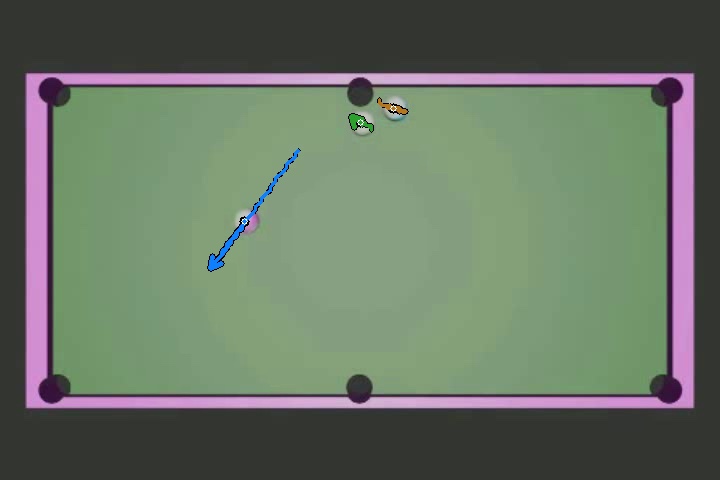} &
\qimg{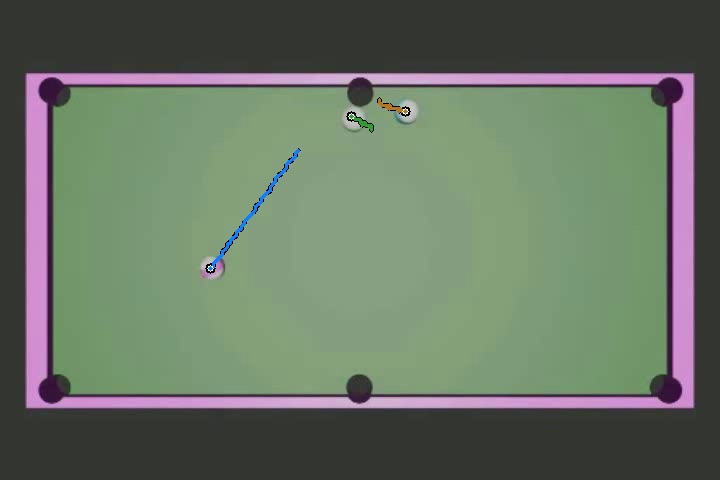} \\[0.5pt]
\rotatebox{90}{\scriptsize\hspace{4pt}Ours-CogV} &
\qimg{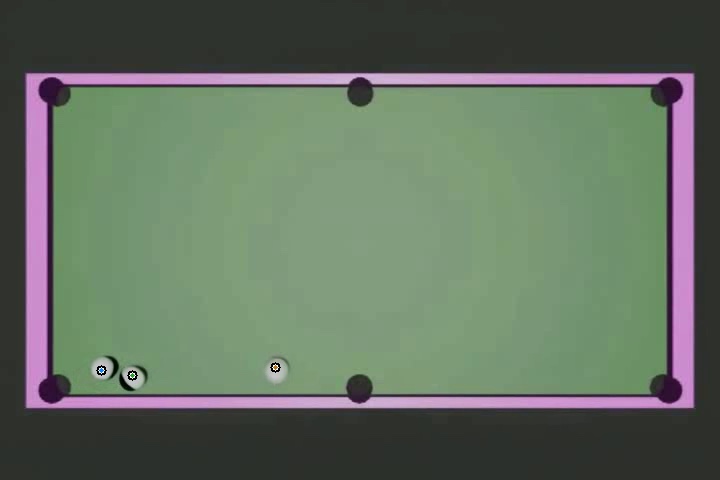} &
\qimg{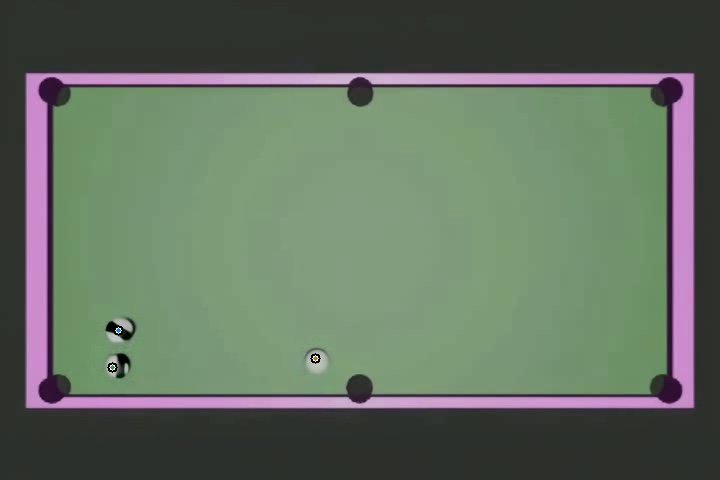} &
\qimg{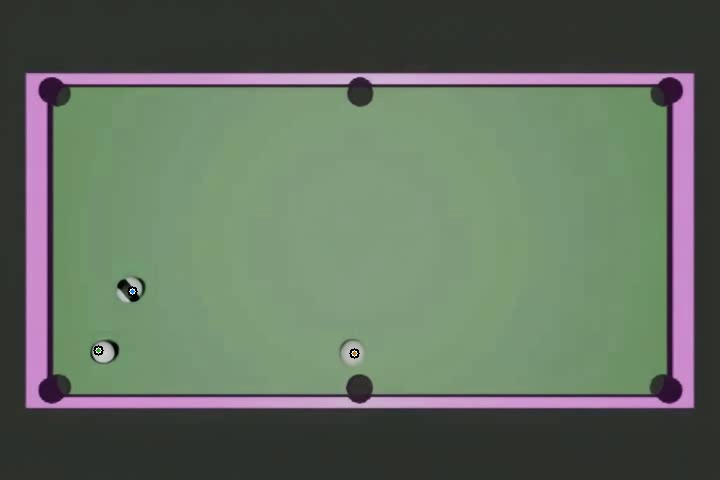} &
\rotatebox{90}{\scriptsize\hspace{4pt}Ours-WaN} &
\qimg{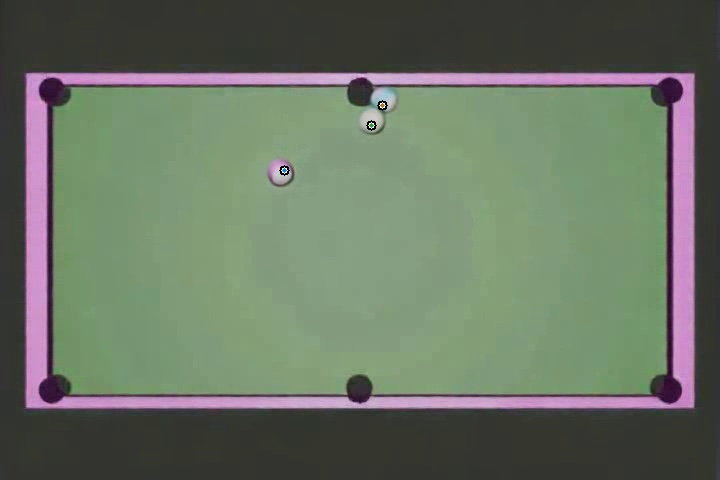} &
\qimg{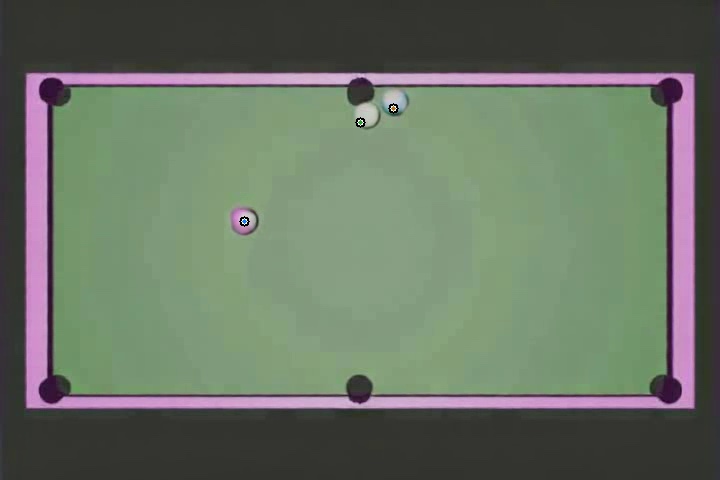} &
\qimg{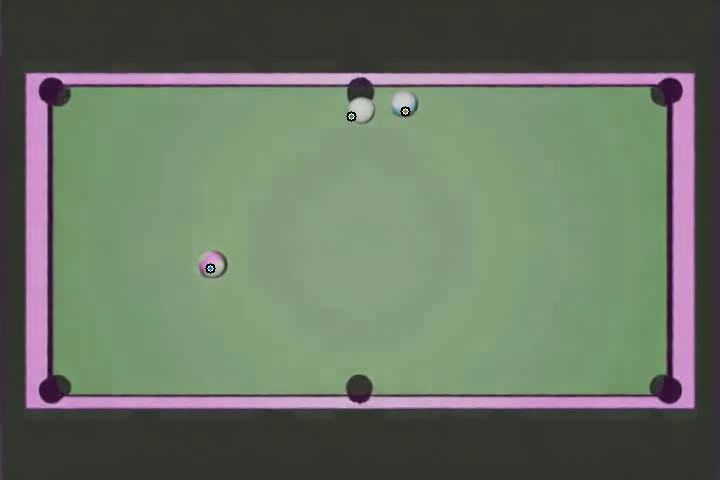} \\[0.5pt]
\rotatebox{90}{\scriptsize\hspace{4pt}MagicMotion} &
\qimg{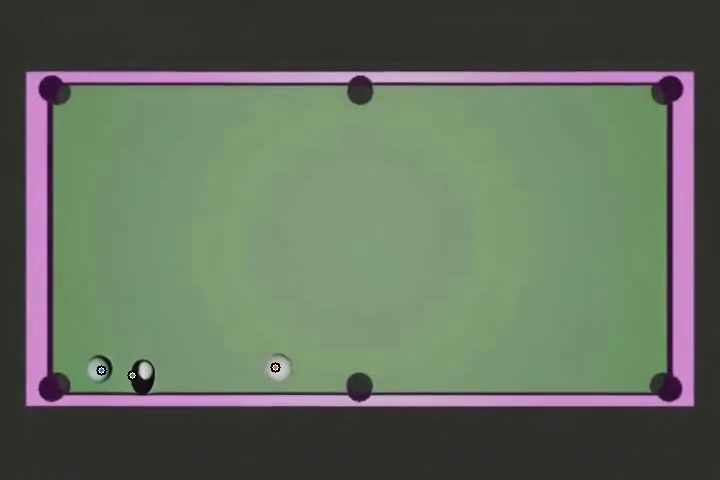} &
\qimg{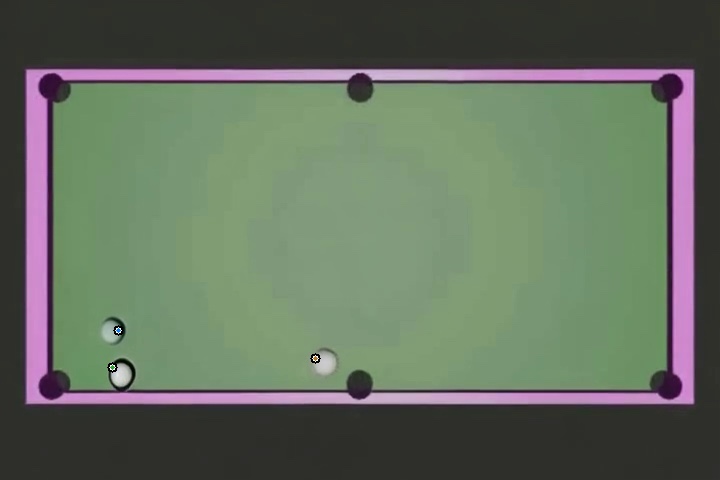} &
\qimg{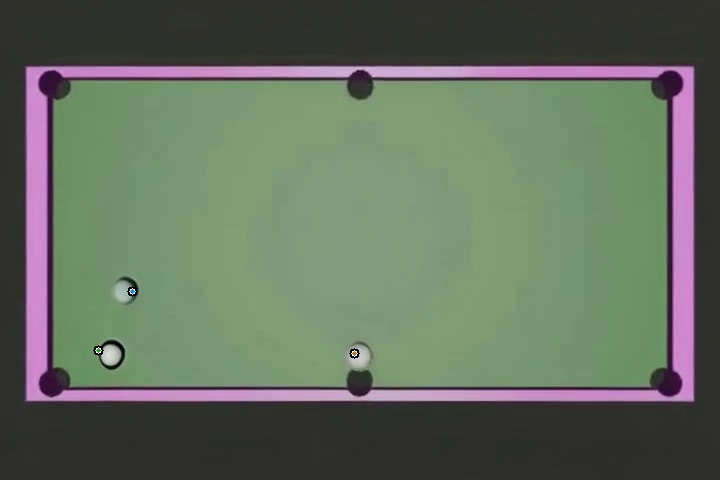} &
\rotatebox{90}{\scriptsize\hspace{8pt}ATI} &
\qimg{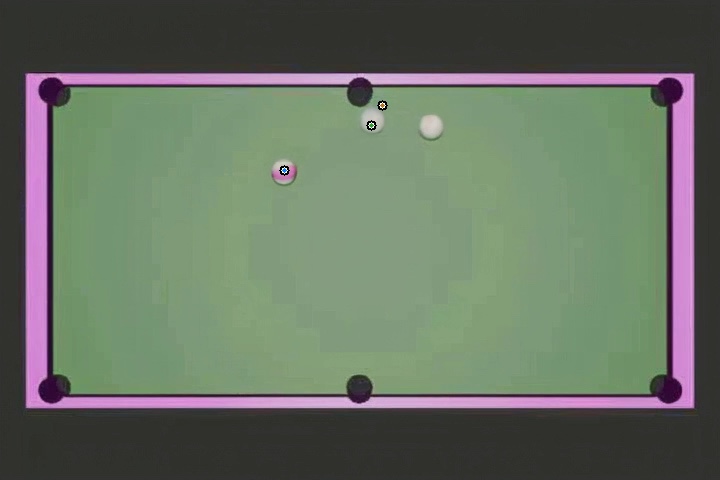} &
\qimg{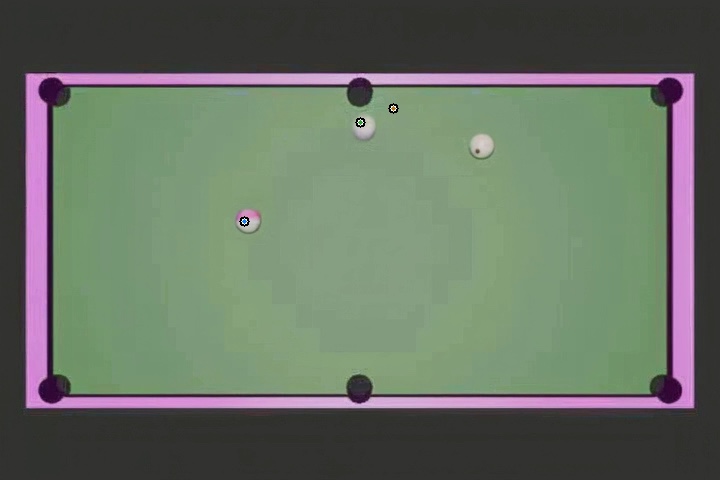} &
\qimg{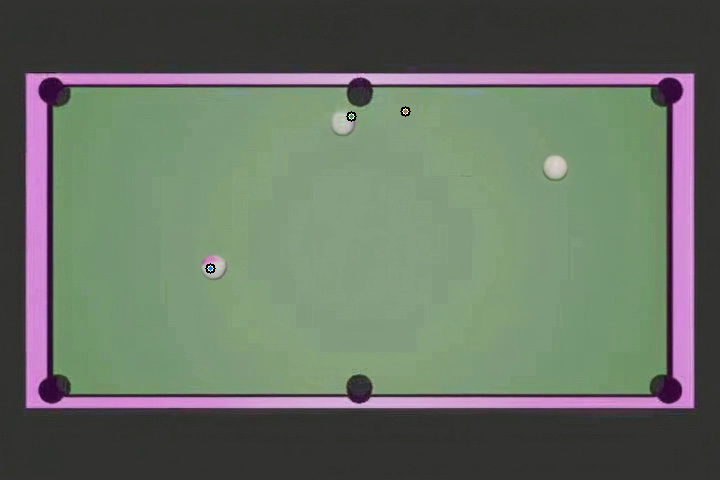} \\[0.5pt]
\rotatebox{90}{\scriptsize\hspace{8pt}Tora} &
\qimg{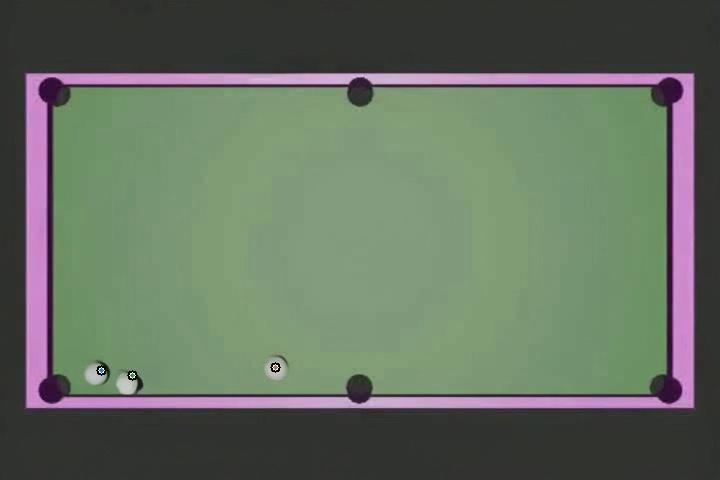} &
\qimg{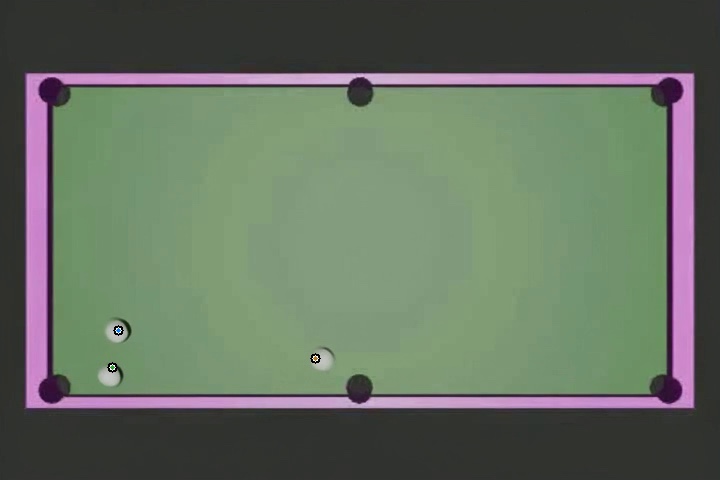} &
\qimg{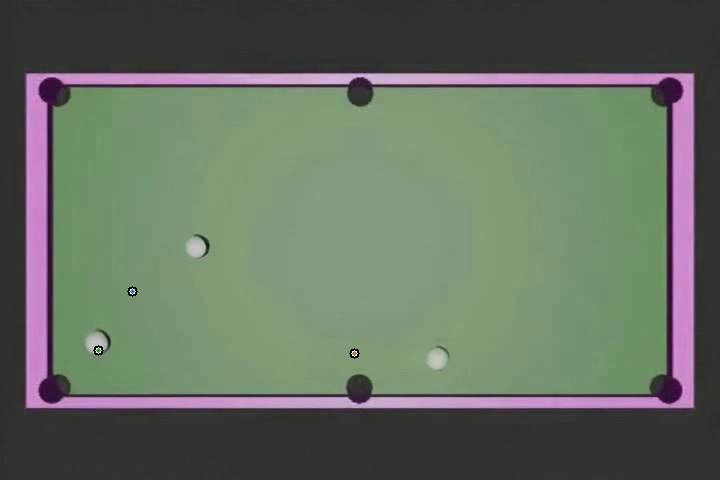} &
\rotatebox{90}{\scriptsize\hspace{4pt}Wan-Move} &
\qimg{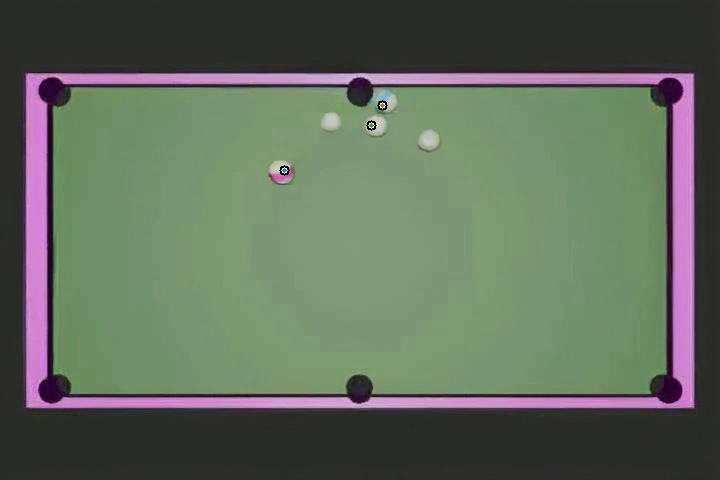} &
\qimg{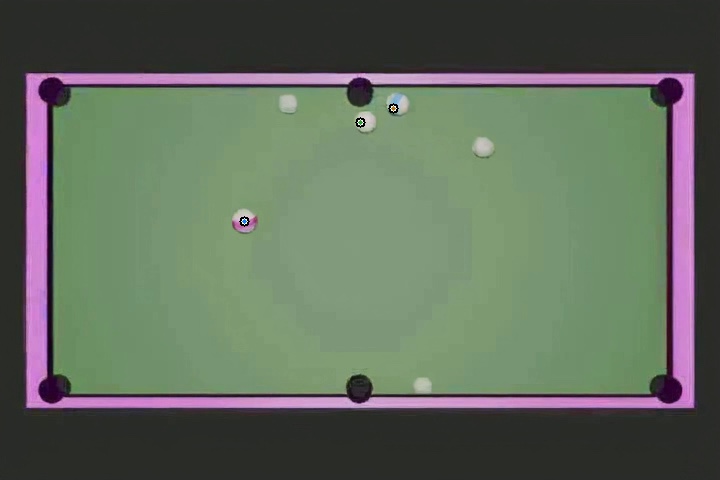} &
\qimg{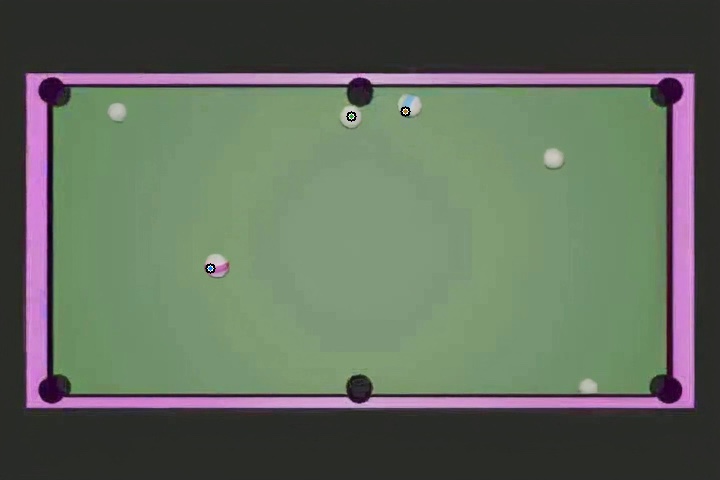} \\
\end{tabular}
\caption{\textbf{Additional qualitative comparison on Pool.} Left: CogVideoX-5B against MagicMotion and Tora. Right: Wan2.1-14B against ATI and Wan-Move. Our method follows the ball locations exactly throughout the sequence. Tora and ATI maintain the correct number of balls but drift from the prescribed positions, as visible in the final frame. Wan-Move follows the trajectories but hallucinates additional objects, and MagicMotion shows slight positional errors.}
\label{fig:qual_extra_pool}
\end{figure*}

\begin{figure*}[t]
\centering
\setlength{\tabcolsep}{0.5pt}
\renewcommand{\arraystretch}{0.3}
\newcommand{\qimg}[1]{\includegraphics[width=0.155\textwidth]{#1}}
\newcommand{\framesarrow}{%
  \begin{tikzpicture}[baseline=0pt]
    \draw[->, line width=0.5pt] (0,0) -- (0.465\textwidth, 0)
      node[midway, fill=white, inner sep=1pt] {\scriptsize frames};
  \end{tikzpicture}%
}
\begin{tabular}{@{}c ccc @{\hspace{6pt}} c ccc@{}}
& \multicolumn{3}{c}{\framesarrow} & & \multicolumn{3}{c}{\framesarrow} \\[1pt]
\rotatebox{90}{\scriptsize\hspace{8pt}GT} &
\qimg{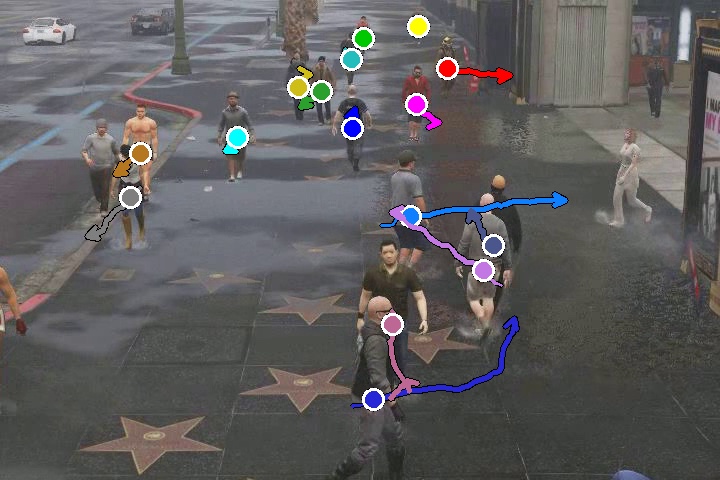} &
\qimg{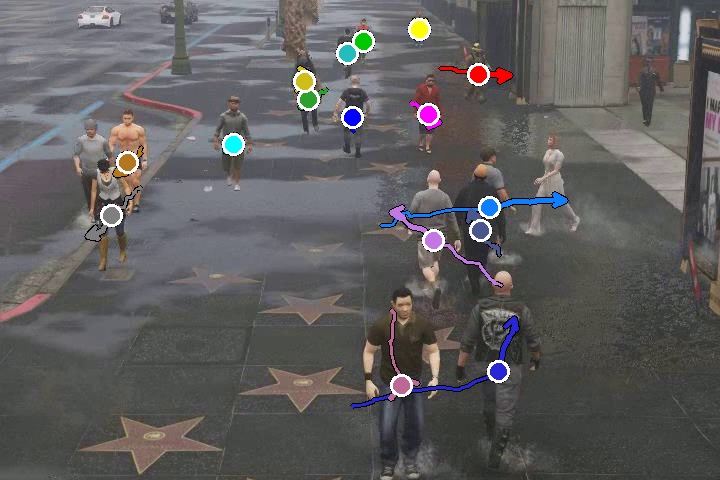} &
\qimg{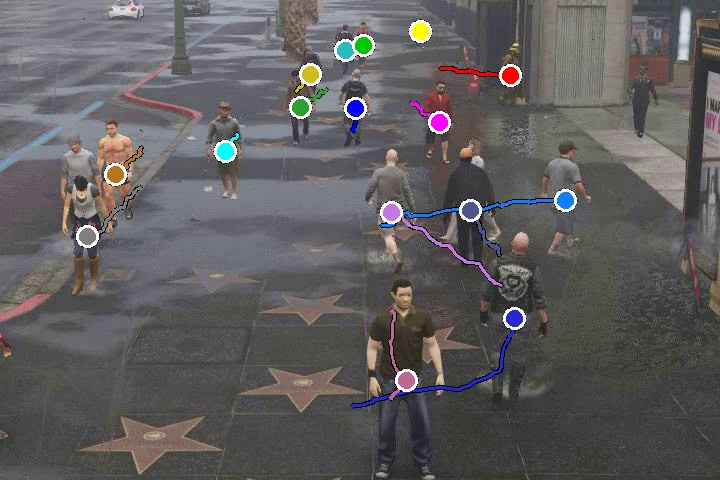} &
\rotatebox{90}{\scriptsize\hspace{8pt}GT} &
\qimg{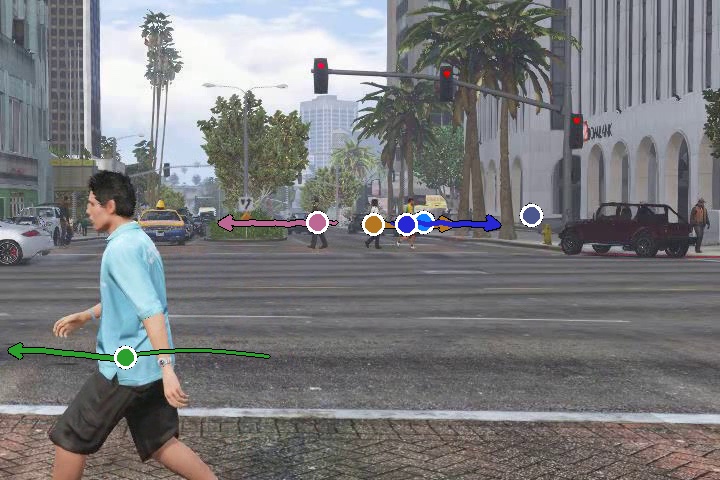} &
\qimg{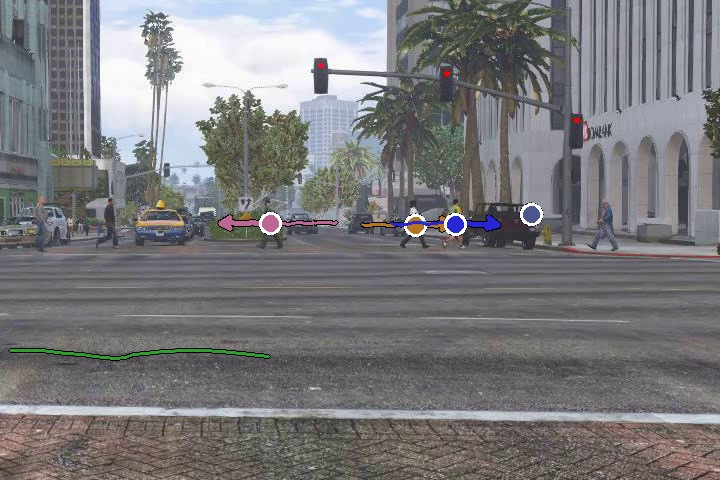} &
\qimg{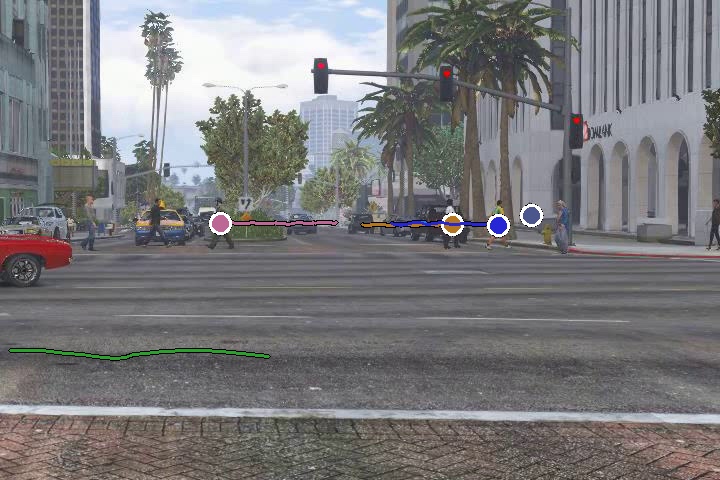} \\[0.5pt]
\rotatebox{90}{\scriptsize\hspace{4pt}Ours-CogV} &
\qimg{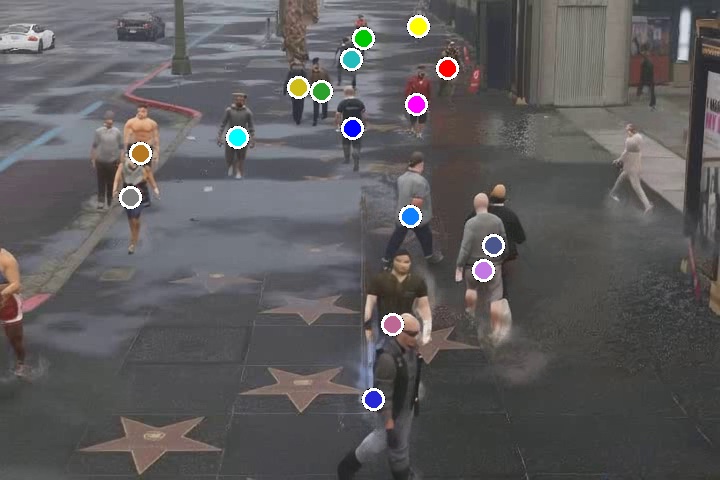} &
\qimg{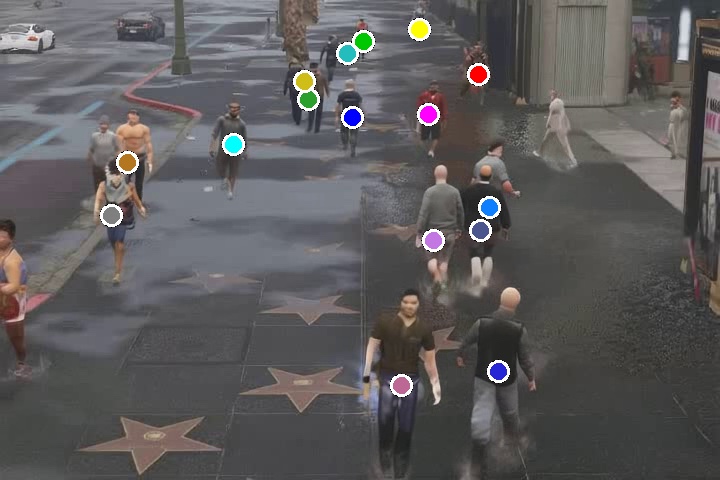} &
\qimg{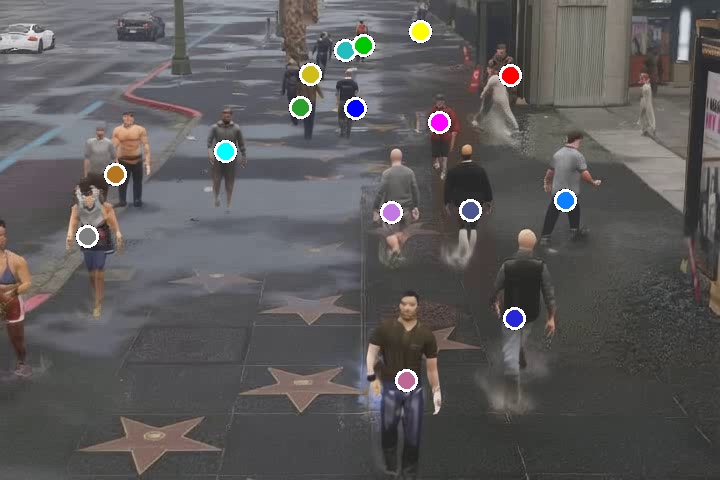} &
\rotatebox{90}{\scriptsize\hspace{4pt}Ours-WaN} &
\qimg{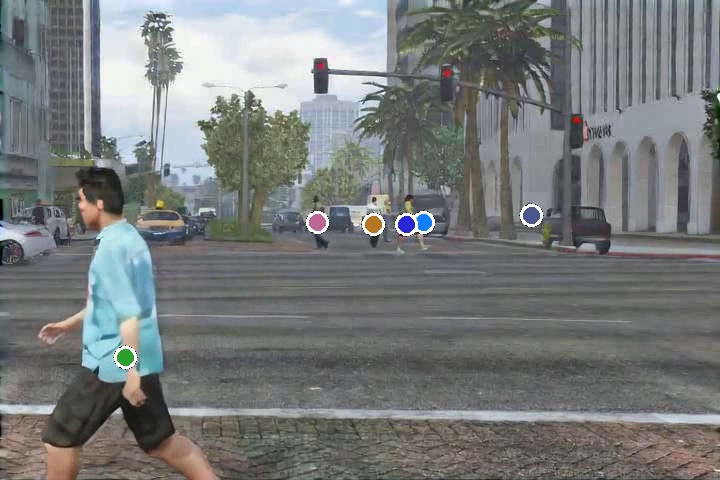} &
\qimg{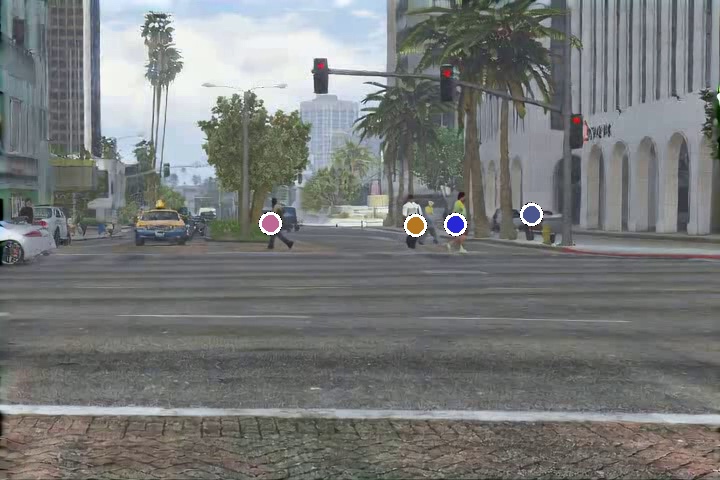} &
\qimg{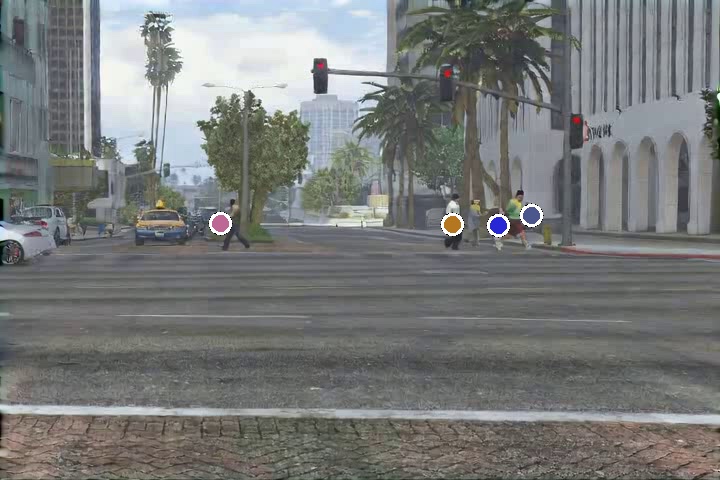} \\[0.5pt]
\rotatebox{90}{\scriptsize\hspace{4pt}MagicMotion} &
\qimg{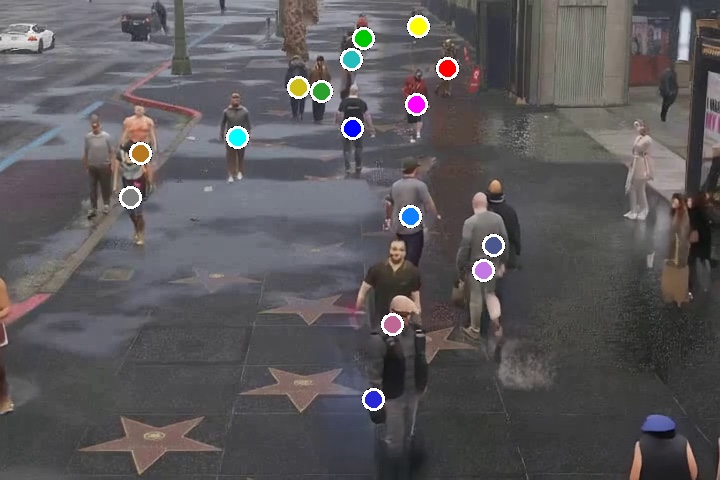} &
\qimg{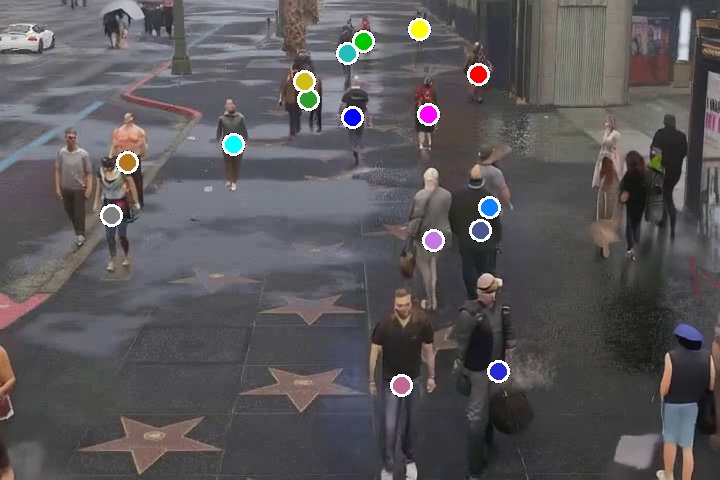} &
\qimg{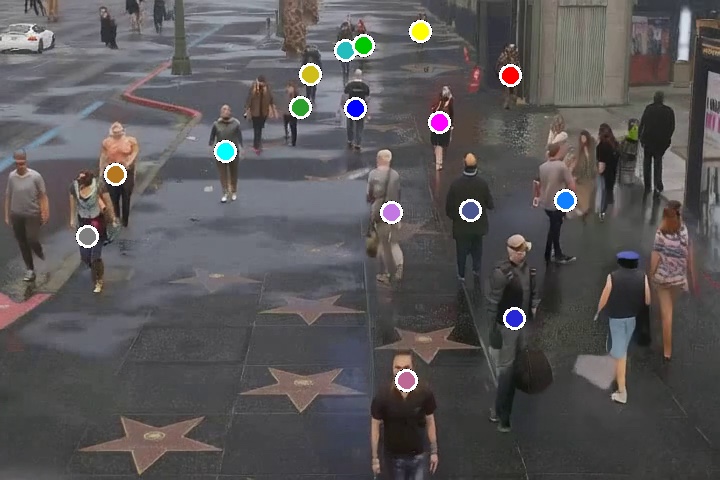} &
\rotatebox{90}{\scriptsize\hspace{8pt}ATI} &
\qimg{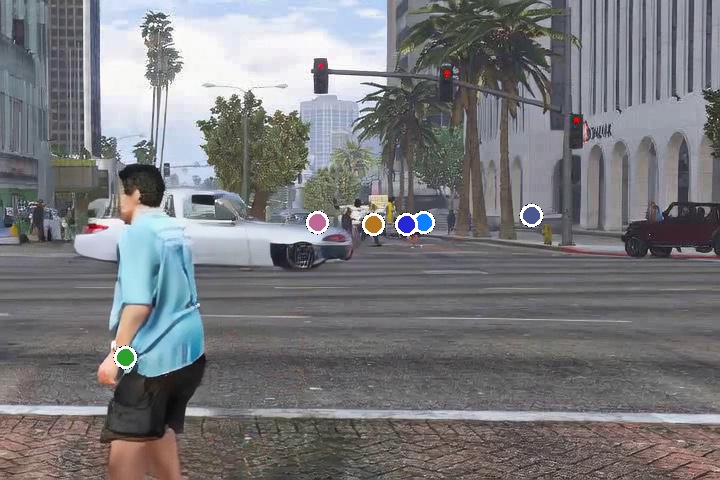} &
\qimg{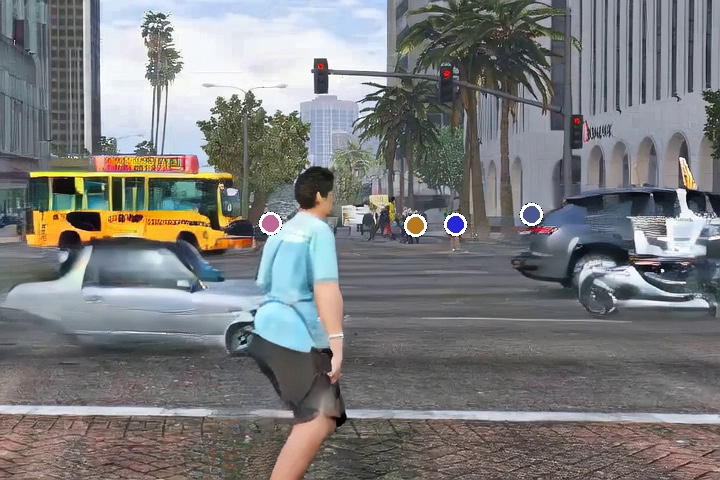} &
\qimg{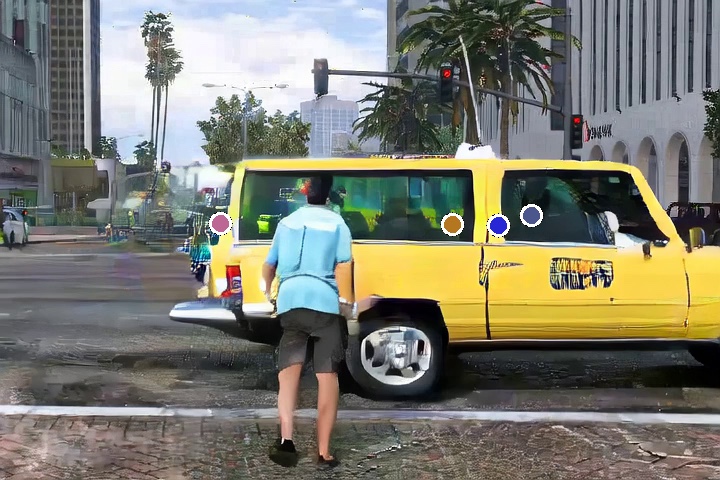} \\[0.5pt]
\rotatebox{90}{\scriptsize\hspace{8pt}Tora} &
\qimg{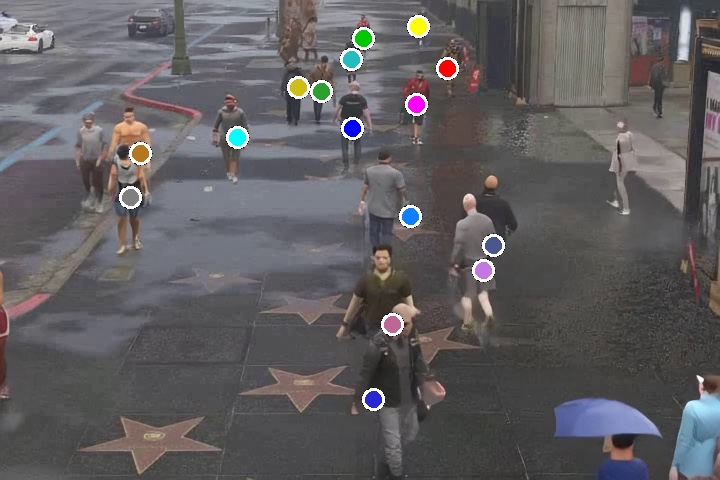} &
\qimg{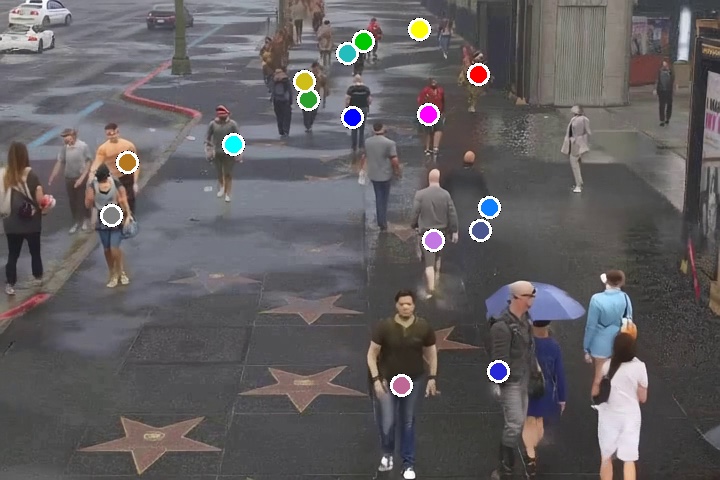} &
\qimg{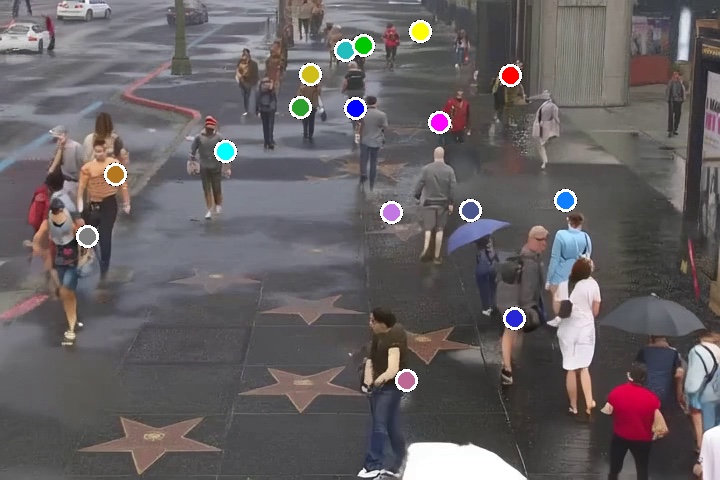} &
\rotatebox{90}{\scriptsize\hspace{4pt}Wan-Move} &
\qimg{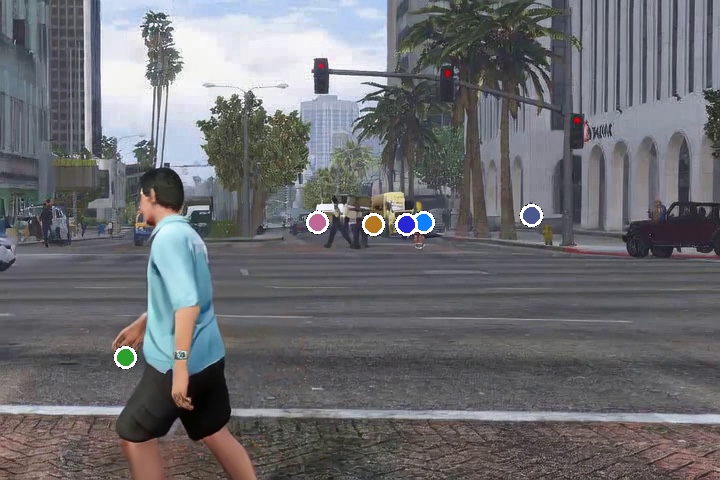} &
\qimg{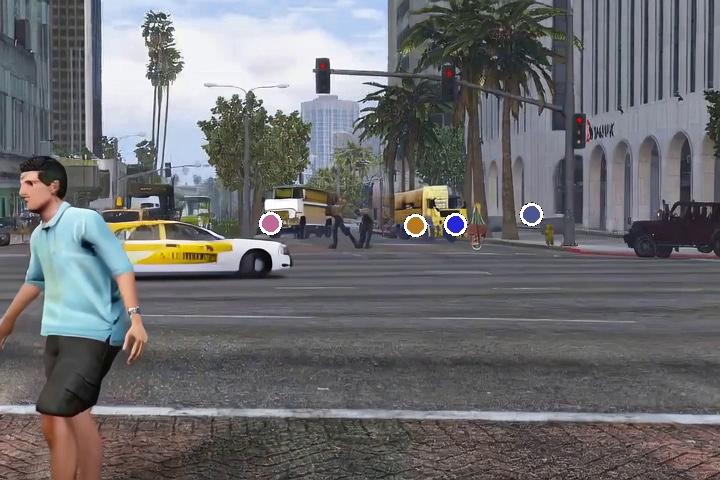} &
\qimg{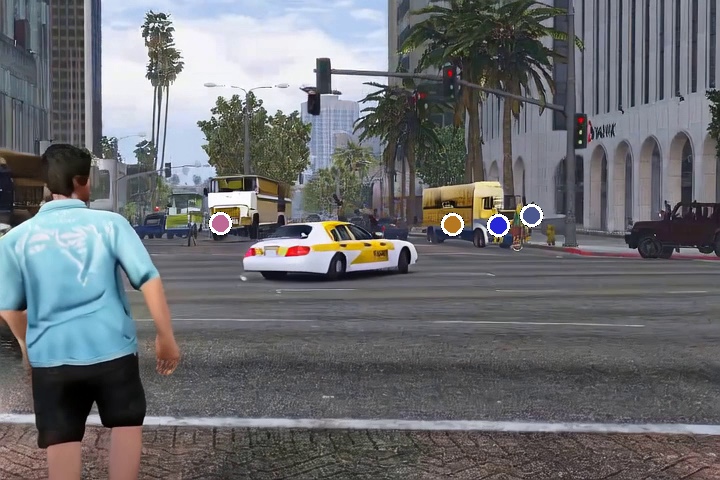} \\
\end{tabular}
\caption{\textbf{Additional qualitative comparison on MOTSynth.} Left: CogVideoX-5B against MagicMotion and Tora. Right: Wan2.1-14B against ATI and Wan-Move. MagicMotion fails to follow several pedestrians after trajectory intersections, as seen at the brown and pink dots where objects drift from their prescribed positions. Tora  also struggles with occlusion, notably the gray-dot pedestrian on the left side of the scene. ATI and Wan-Move also fail, for example the green-dot pedestrian should exit the scene but remains visible.}
\label{fig:qual_extra_motsynth}
\end{figure*}

\begin{figure*}[t]
\centering
\setlength{\tabcolsep}{0.5pt}
\renewcommand{\arraystretch}{0.3}
\newcommand{\qimg}[1]{\includegraphics[width=0.155\textwidth]{#1}}
\newcommand{\framesarrow}{%
  \begin{tikzpicture}[baseline=0pt]
    \draw[->, line width=0.5pt] (0,0) -- (0.465\textwidth, 0)
      node[midway, fill=white, inner sep=1pt] {\scriptsize frames};
  \end{tikzpicture}%
}
\begin{tabular}{@{}c ccc @{\hspace{6pt}} c ccc@{}}
& \multicolumn{3}{c}{\framesarrow} & & \multicolumn{3}{c}{\framesarrow} \\[1pt]
\rotatebox{90}{\scriptsize\hspace{8pt}GT} &
\qimg{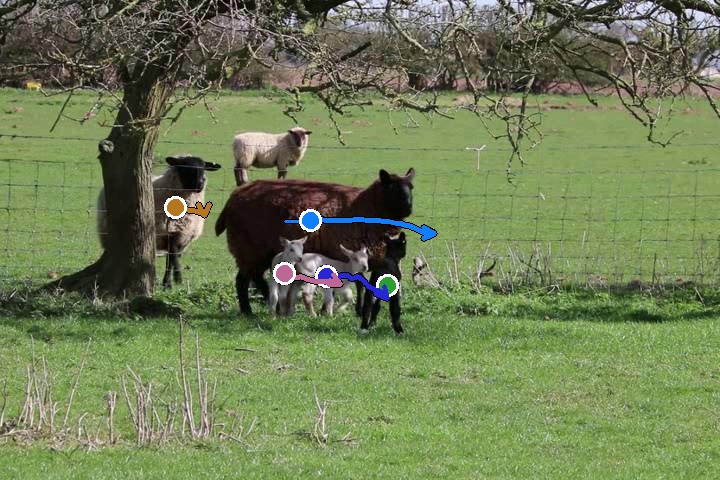} &
\qimg{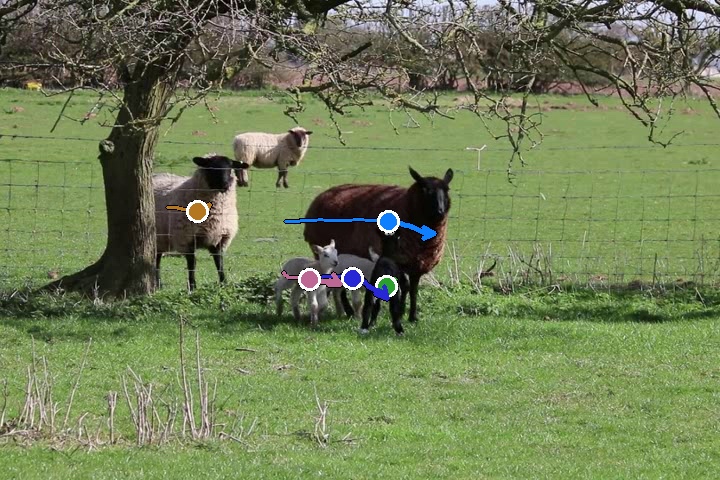} &
\qimg{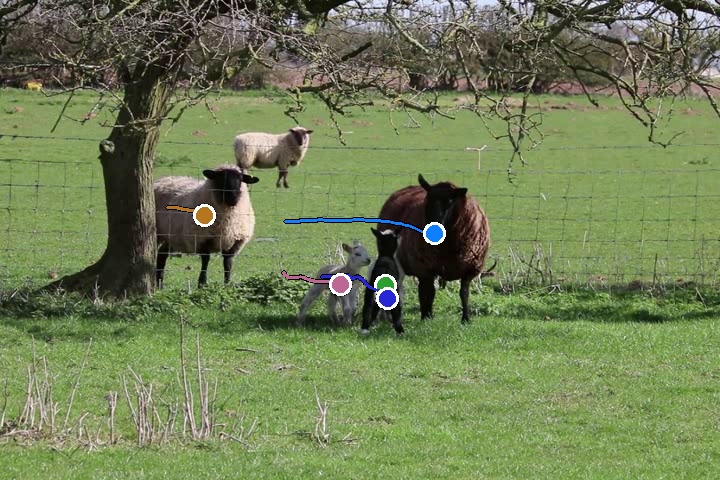} &
\rotatebox{90}{\scriptsize\hspace{8pt}GT} &
\qimg{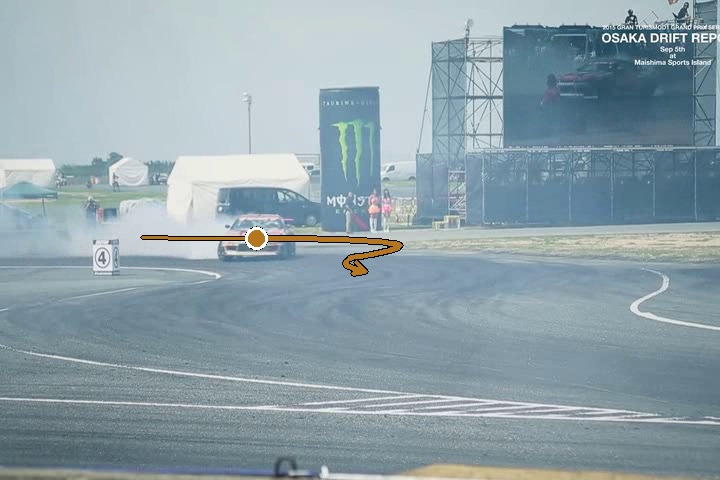} &
\qimg{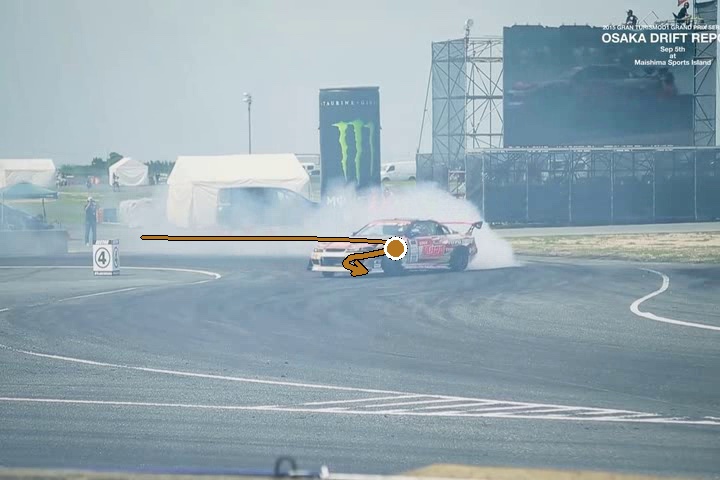} &
\qimg{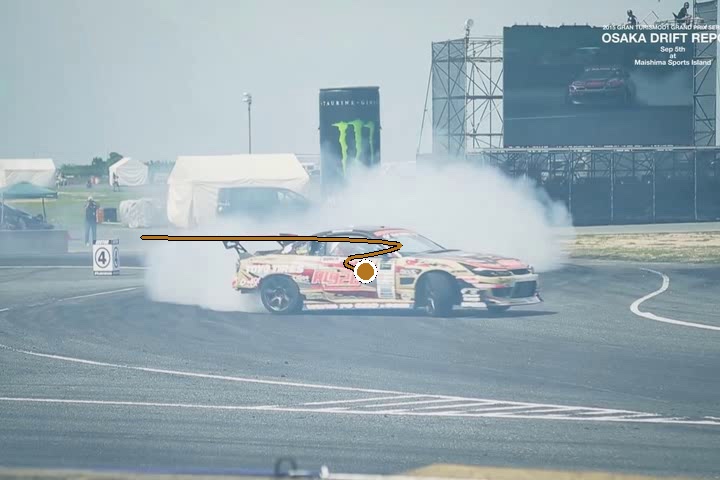} \\[0.5pt]
\rotatebox{90}{\scriptsize\hspace{4pt}Ours-CogV} &
\qimg{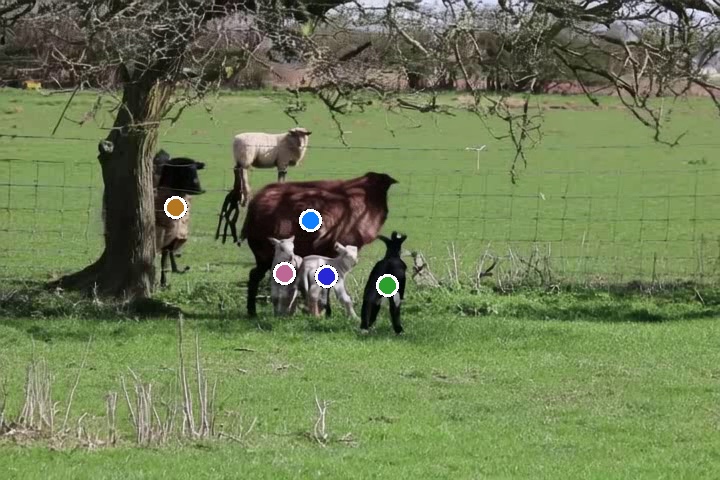} &
\qimg{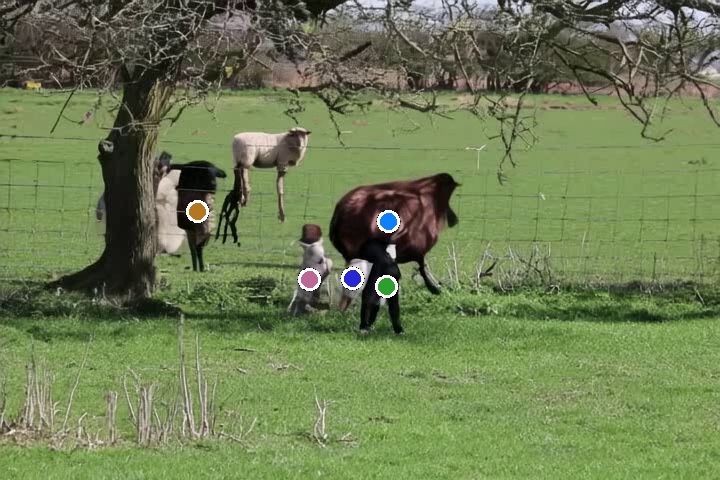} &
\qimg{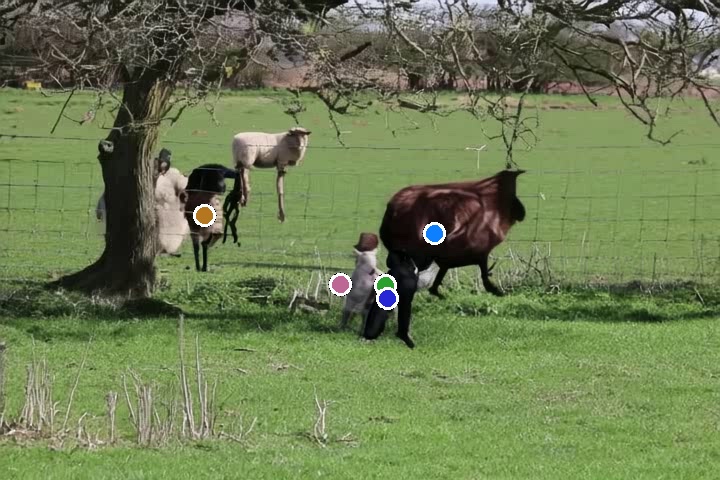} &
\rotatebox{90}{\scriptsize\hspace{4pt}Ours-WaN} &
\qimg{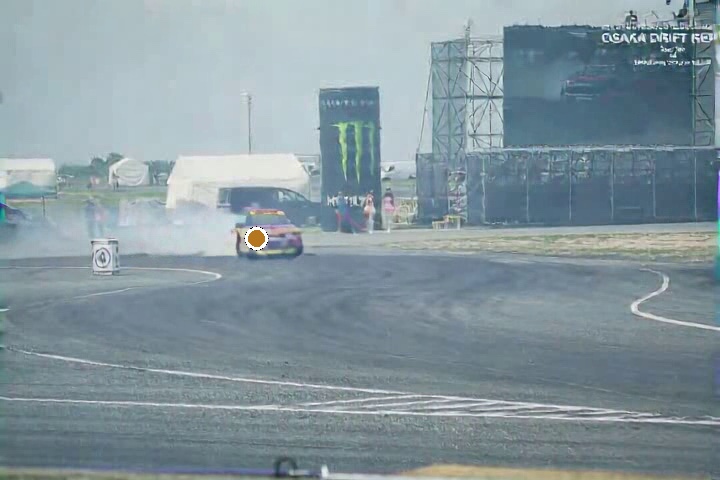} &
\qimg{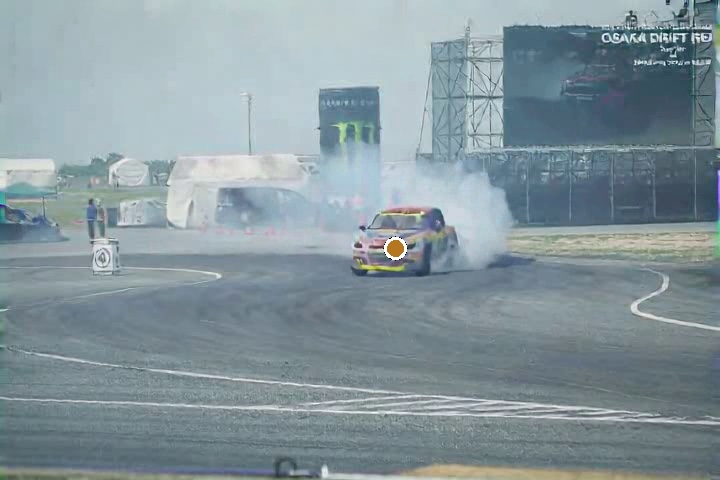} &
\qimg{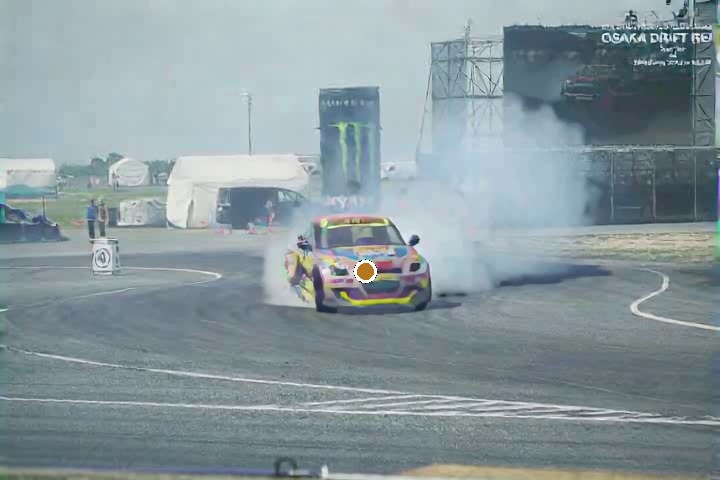} \\[0.5pt]
\rotatebox{90}{\scriptsize\hspace{4pt}MagicMotion} &
\qimg{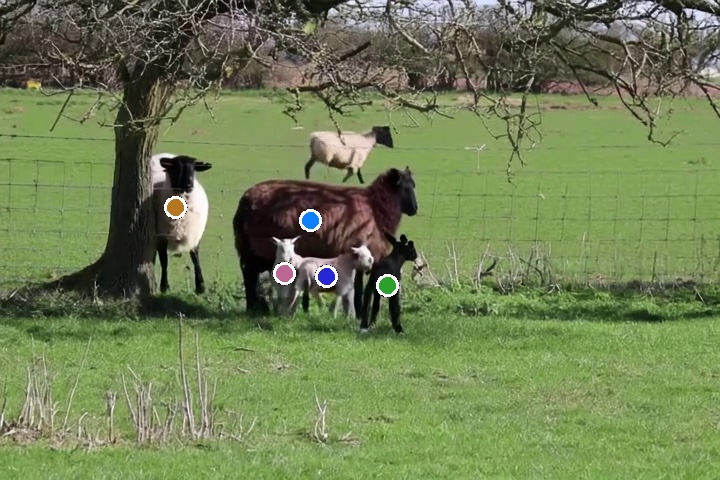} &
\qimg{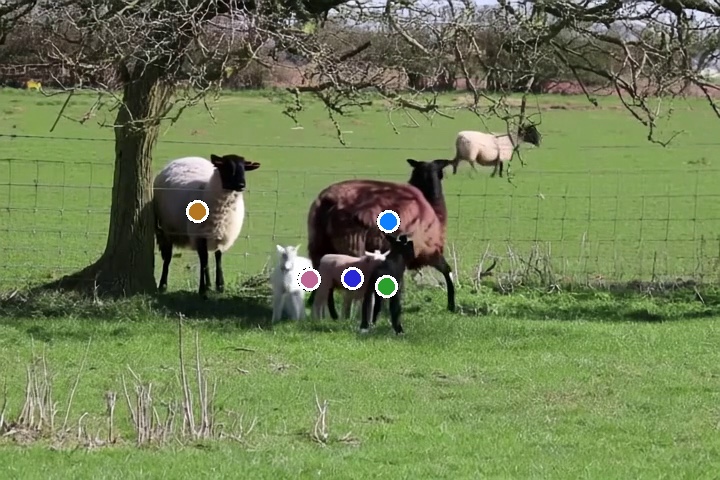} &
\qimg{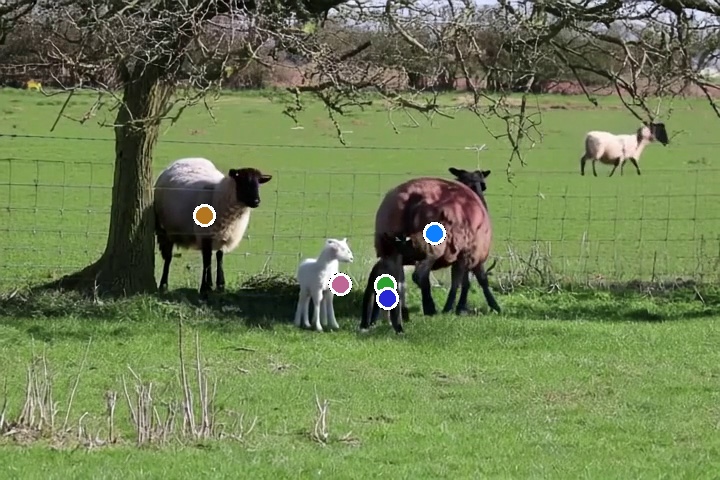} &
\rotatebox{90}{\scriptsize\hspace{8pt}ATI} &
\qimg{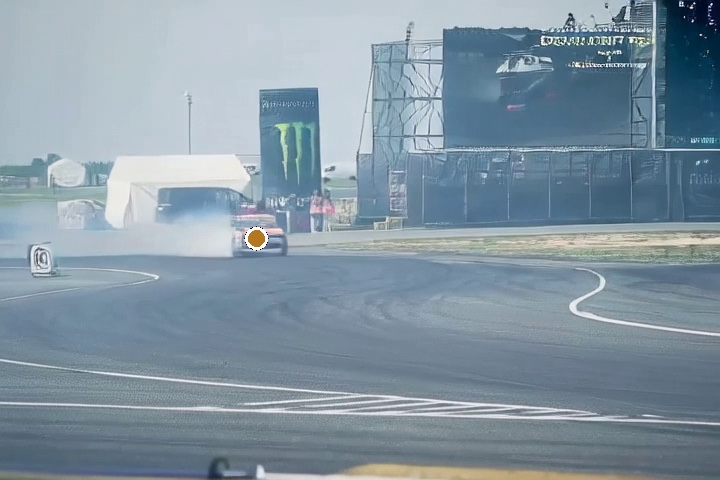} &
\qimg{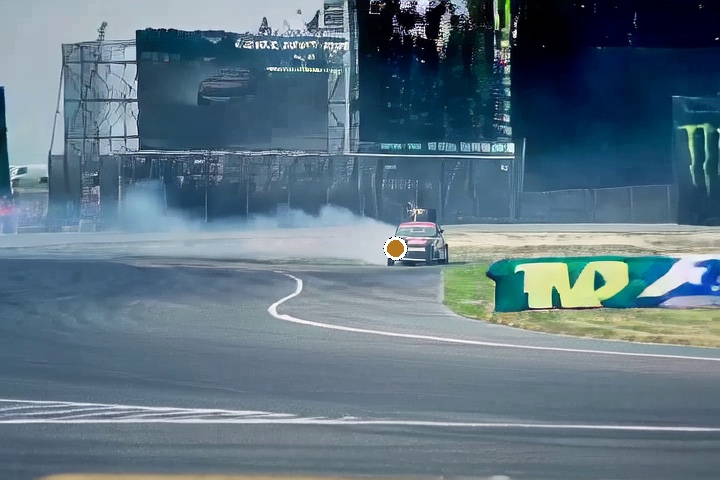} &
\qimg{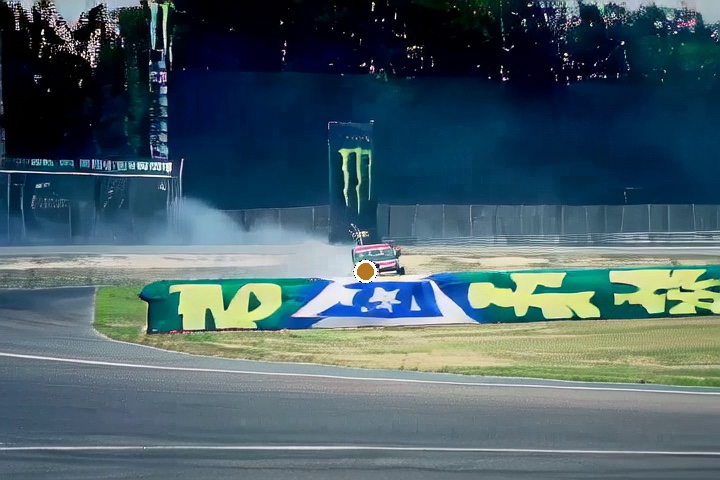} \\[0.5pt]
\rotatebox{90}{\scriptsize\hspace{8pt}Tora} &
\qimg{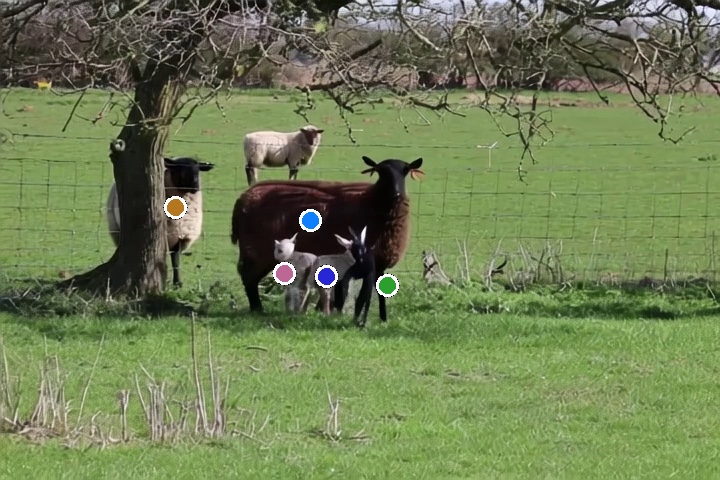} &
\qimg{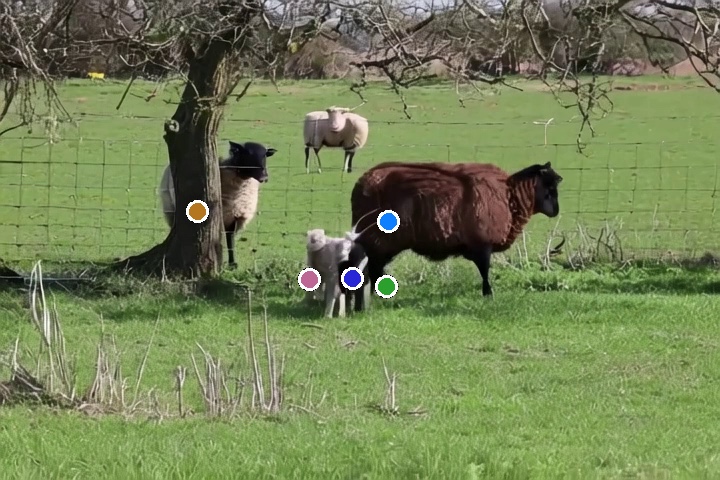} &
\qimg{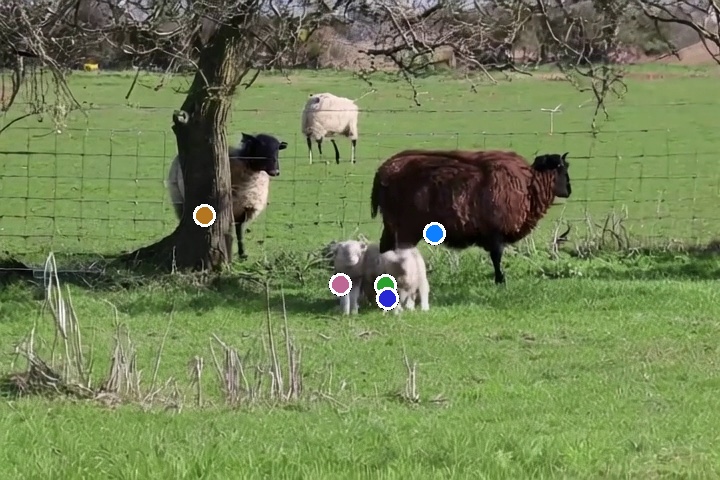} &
\rotatebox{90}{\scriptsize\hspace{4pt}Wan-Move} &
\qimg{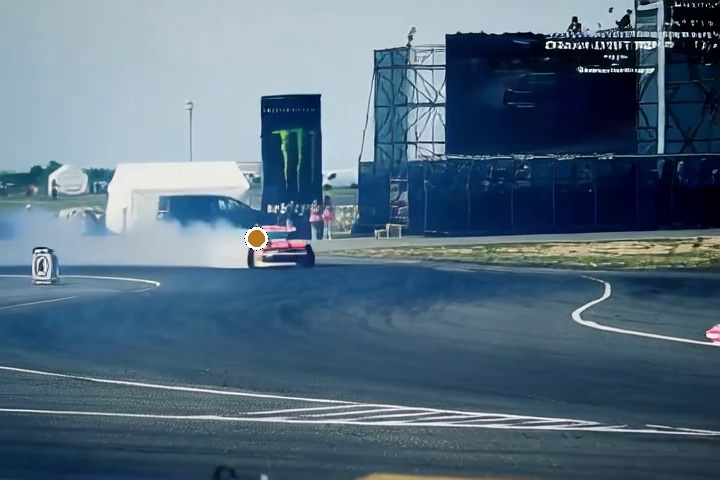} &
\qimg{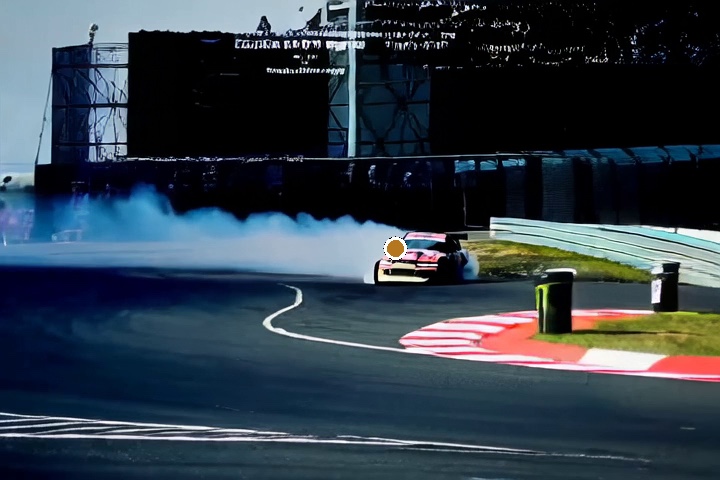} &
\qimg{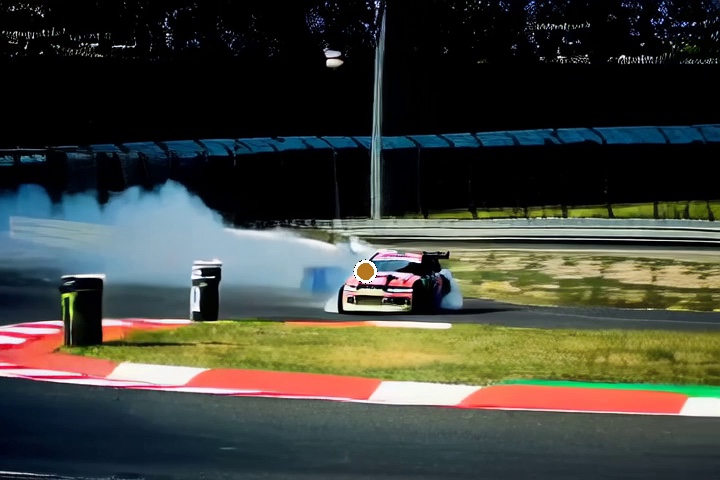} \\
\end{tabular}
\caption{\textbf{Additional qualitative comparison on DAVIS.} Left: CogVideoX-5B against MagicMotion and Tora on \texttt{sheep}. Right: Wan2.1-14B against ATI and Wan-Move on \texttt{drift-chicane}. On the \texttt{sheep} scene, Tora struggles with the blue-dot object while MagicMotion performs reasonably well. On \texttt{drift-chicane}, a single-object scene with no interactions, ATI and Wan-Move move the camera rather than the car to follow the prescribed trajectory. Our method correctly moves the object itself in both cases.}
\label{fig:qual_extra_davis}
\end{figure*}

\begin{figure*}[t]
\centering
\setlength{\tabcolsep}{0.5pt}
\renewcommand{\arraystretch}{0.3}
\newcommand{\qimg}[1]{\includegraphics[width=0.155\textwidth]{#1}}
\newcommand{\framesarrow}{%
  \begin{tikzpicture}[baseline=0pt]
    \draw[->, line width=0.5pt] (0,0) -- (0.465\textwidth, 0)
      node[midway, fill=white, inner sep=1pt] {\scriptsize frames};
  \end{tikzpicture}%
}
\begin{tabular}{@{}c ccc @{\hspace{6pt}} c ccc@{}}
& \multicolumn{3}{c}{\framesarrow} & & \multicolumn{3}{c}{\framesarrow} \\[1pt]
\rotatebox{90}{\scriptsize\hspace{8pt}GT} &
\qimg{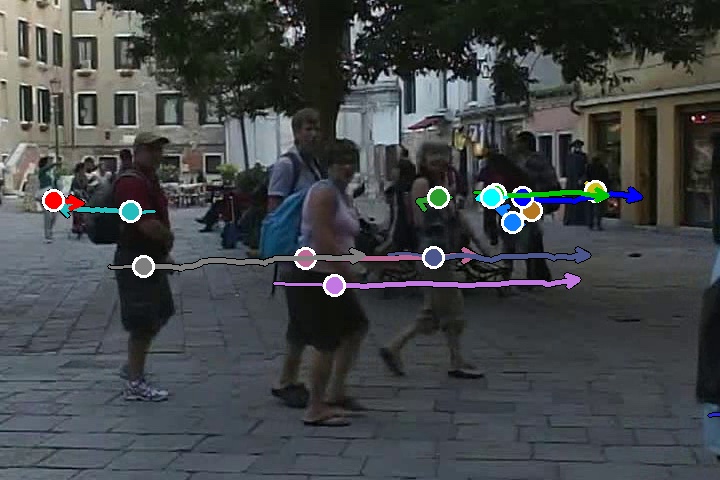} &
\qimg{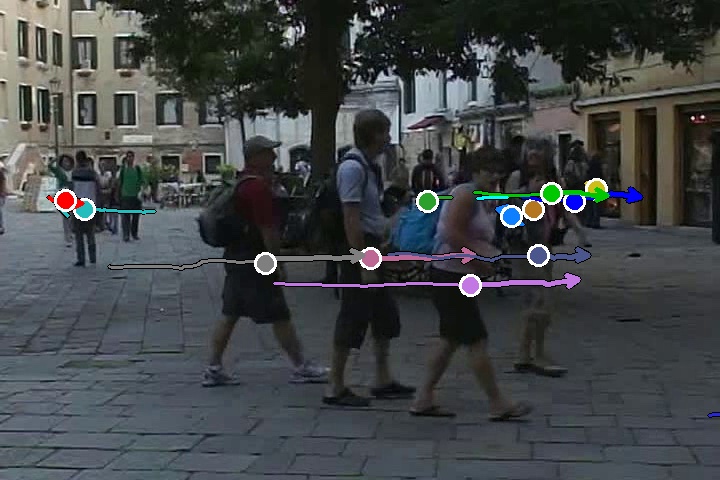} &
\qimg{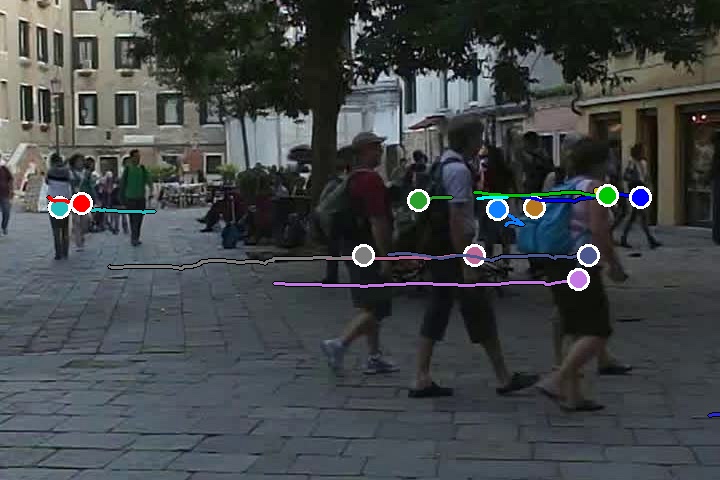} &
\rotatebox{90}{\scriptsize\hspace{8pt}GT} &
\qimg{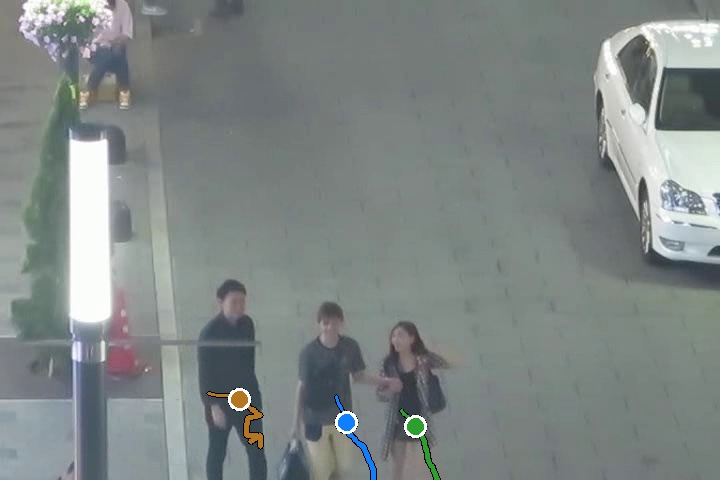} &
\qimg{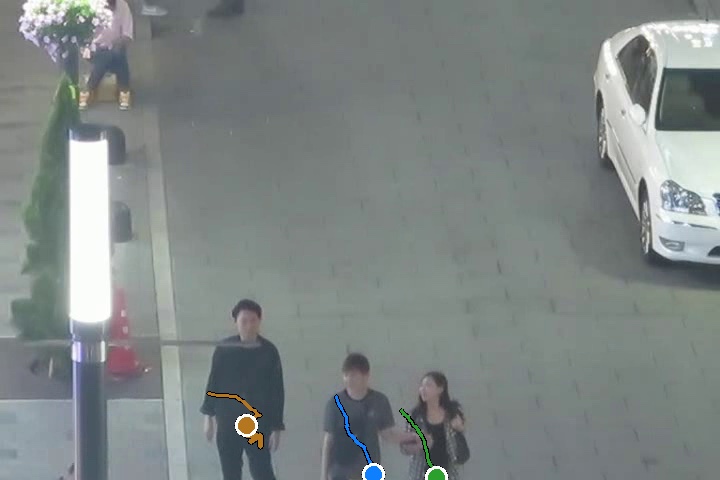} &
\qimg{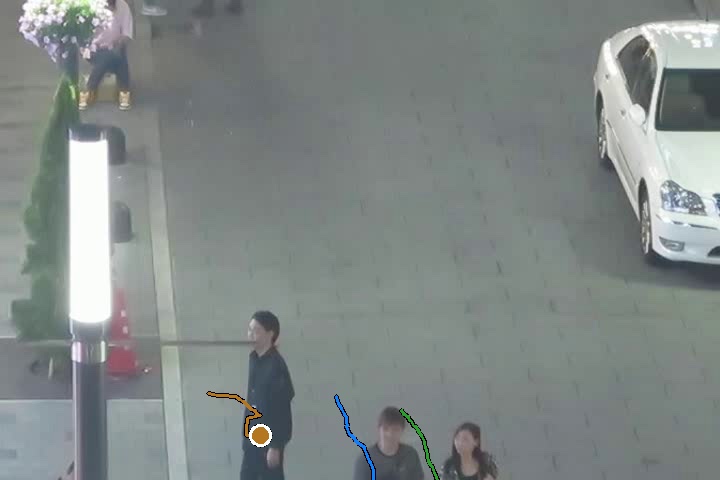} \\[0.5pt]
\rotatebox{90}{\scriptsize\hspace{4pt}Ours-CogV} &
\qimg{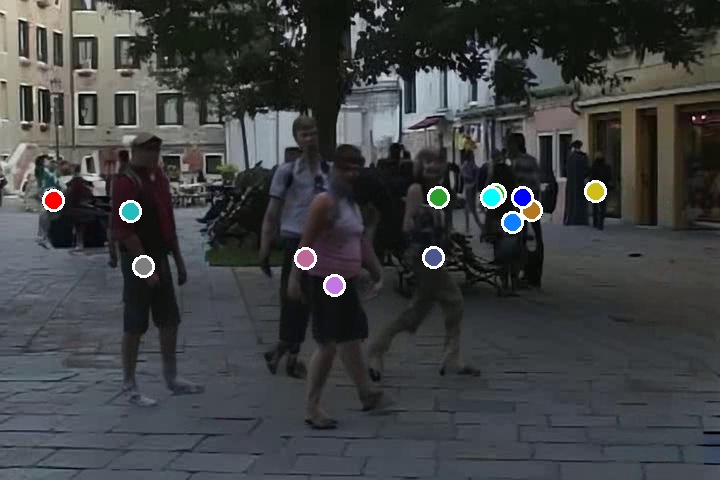} &
\qimg{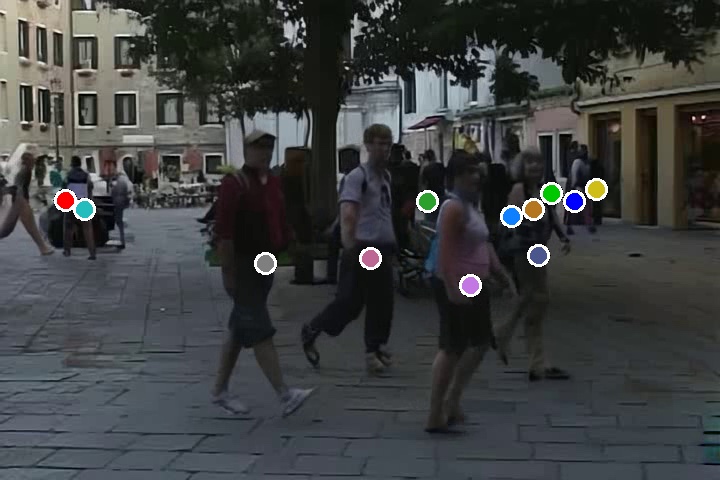} &
\qimg{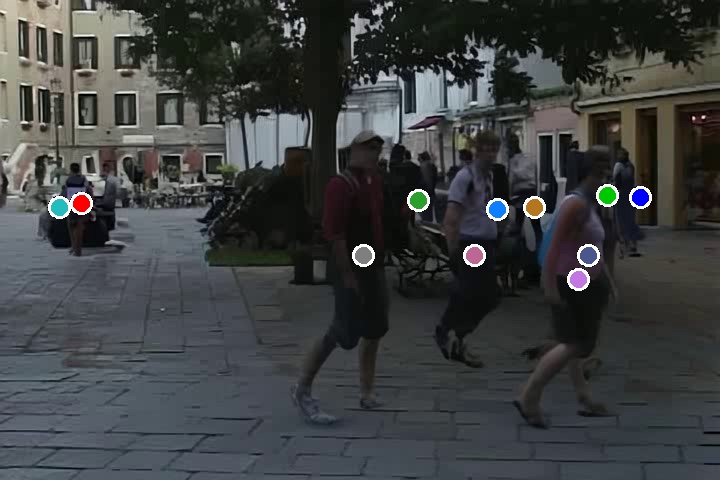} &
\rotatebox{90}{\scriptsize\hspace{4pt}Ours-WaN} &
\qimg{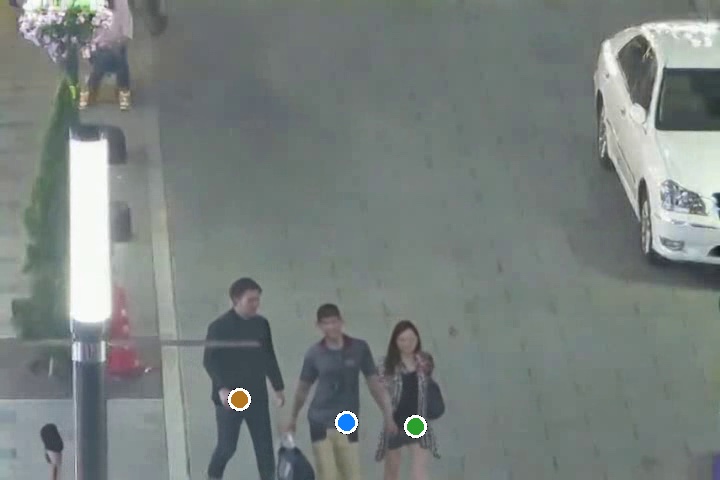} &
\qimg{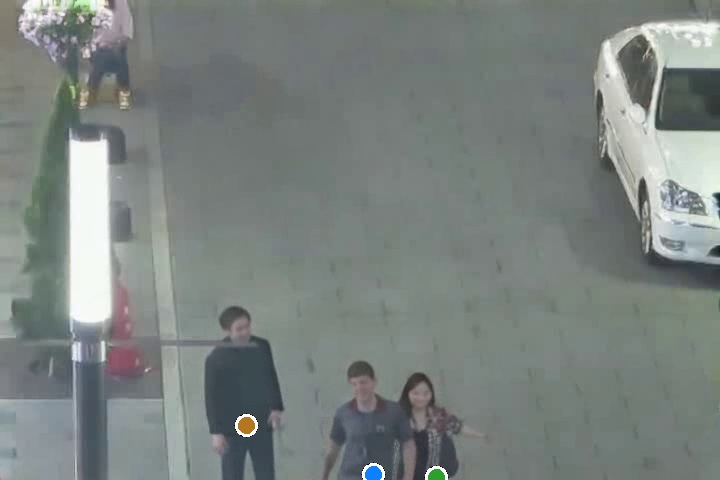} &
\qimg{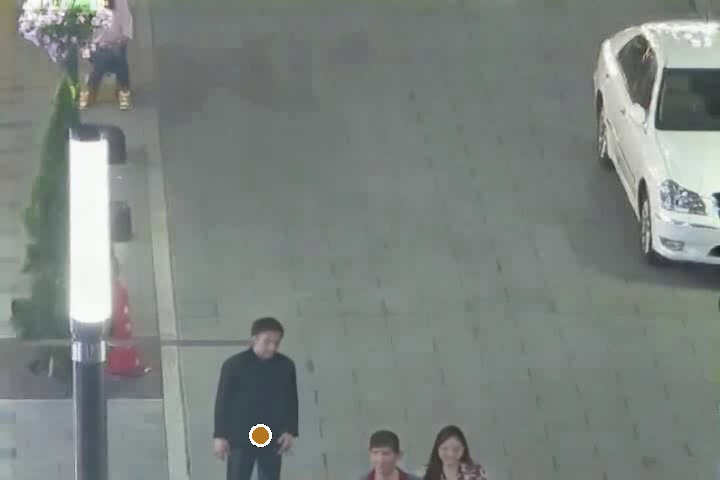} \\[0.5pt]
\rotatebox{90}{\scriptsize\hspace{4pt}MagicMotion} &
\qimg{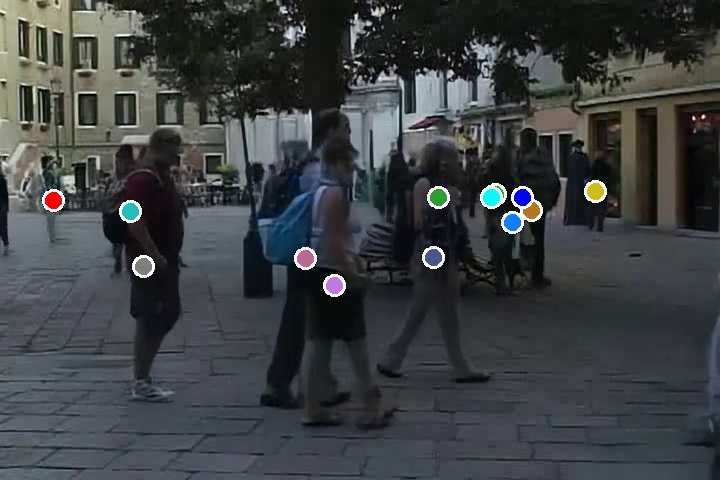} &
\qimg{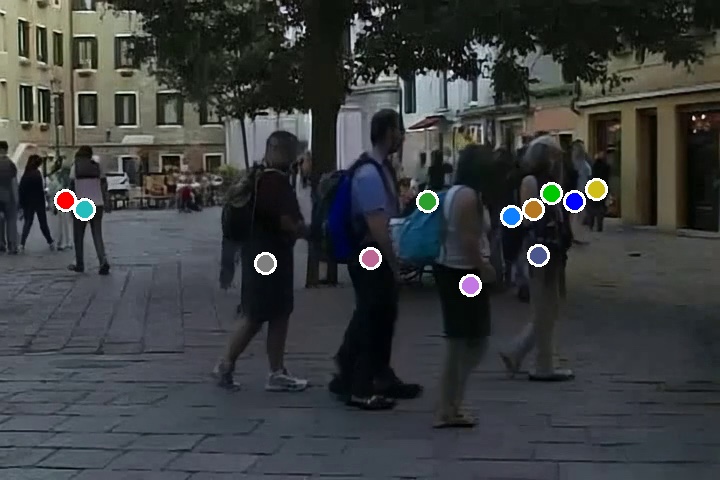} &
\qimg{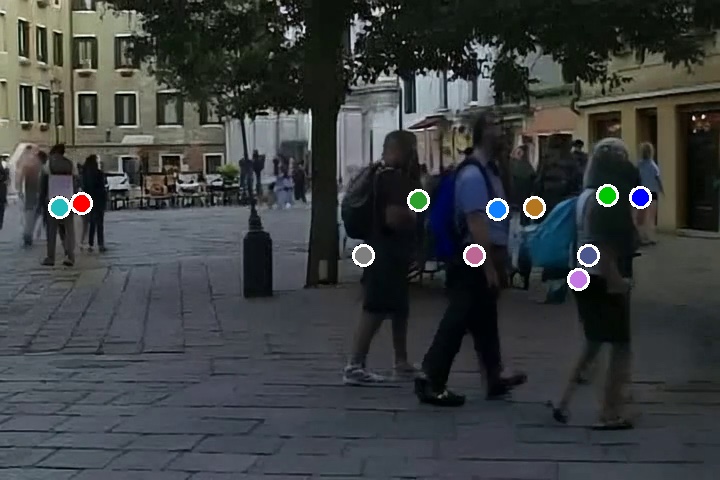} &
\rotatebox{90}{\scriptsize\hspace{8pt}ATI} &
\qimg{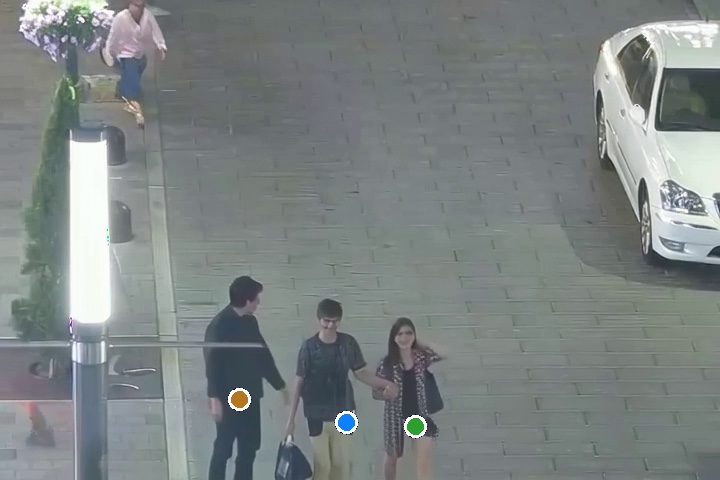} &
\qimg{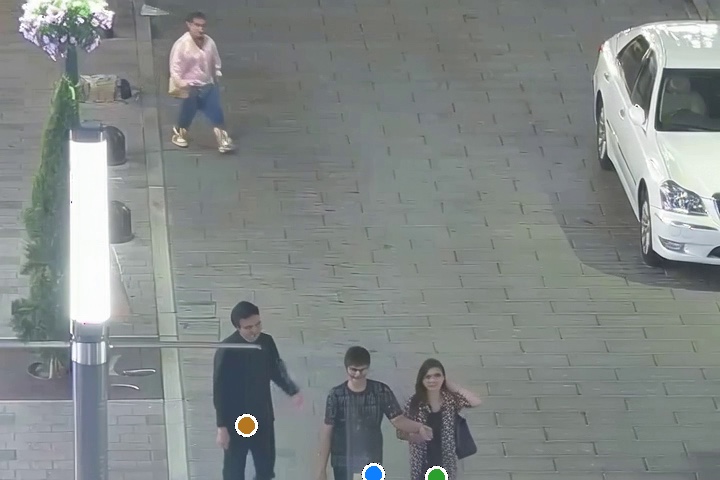} &
\qimg{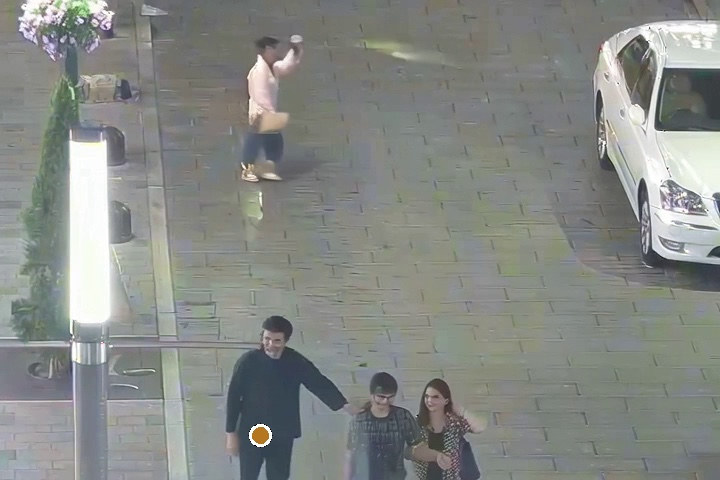} \\[0.5pt]
\rotatebox{90}{\scriptsize\hspace{8pt}Tora} &
\qimg{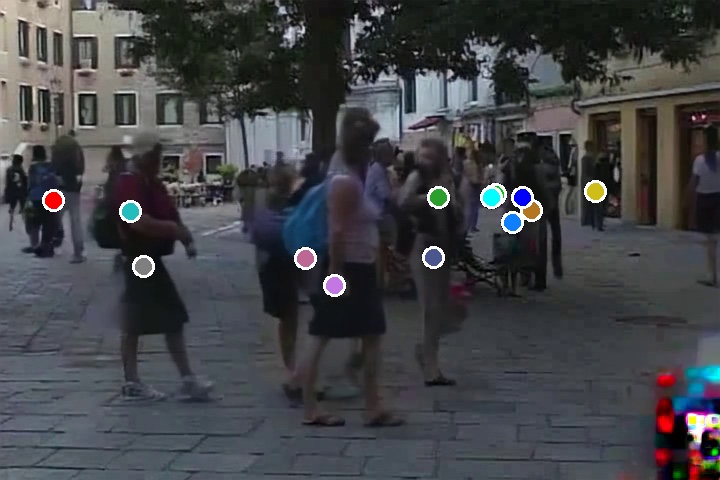} &
\qimg{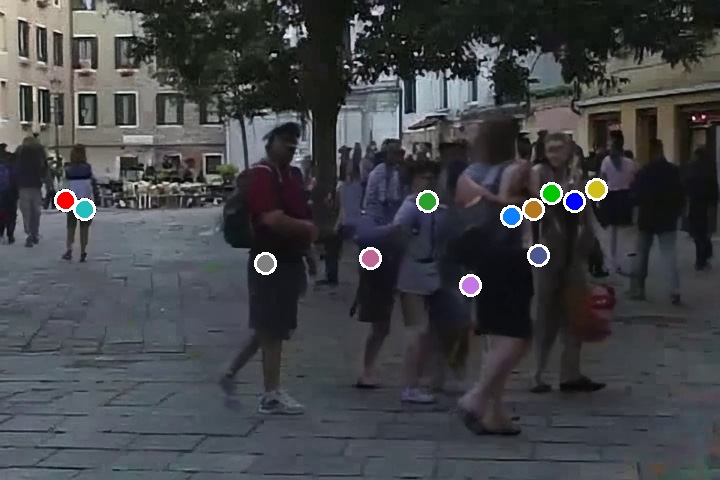} &
\qimg{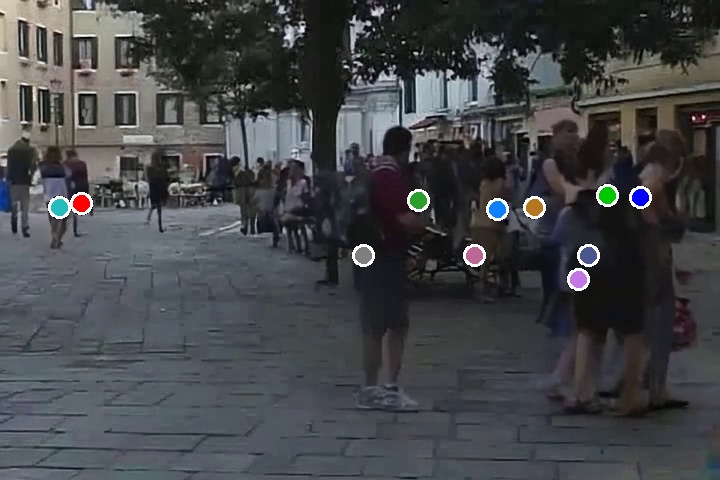} &
\rotatebox{90}{\scriptsize\hspace{4pt}Wan-Move} &
\qimg{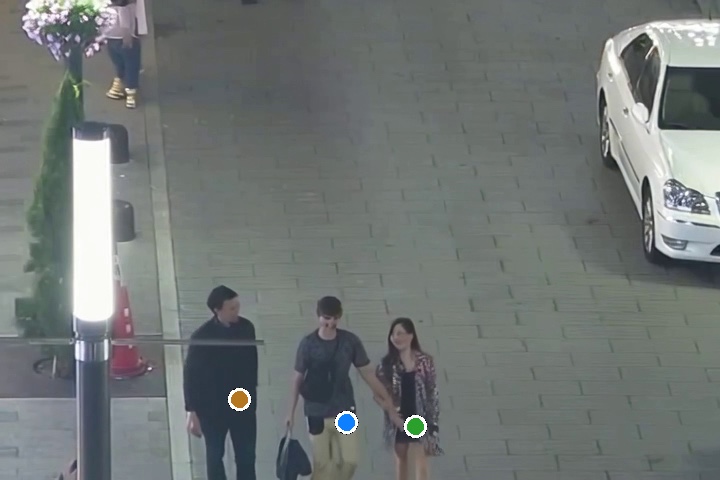} &
\qimg{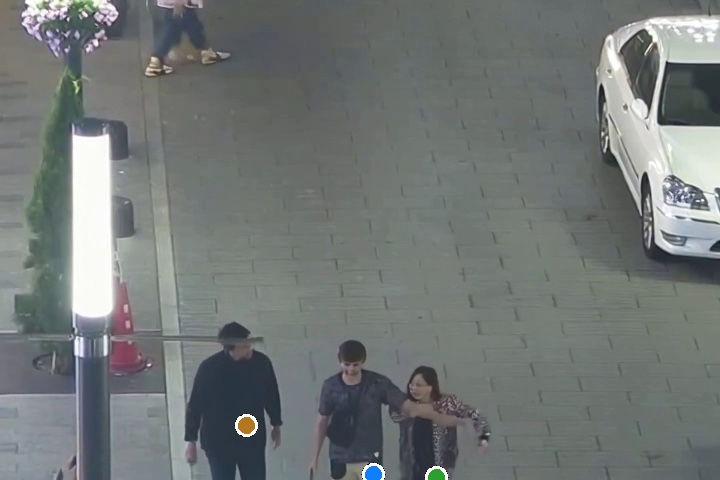} &
\qimg{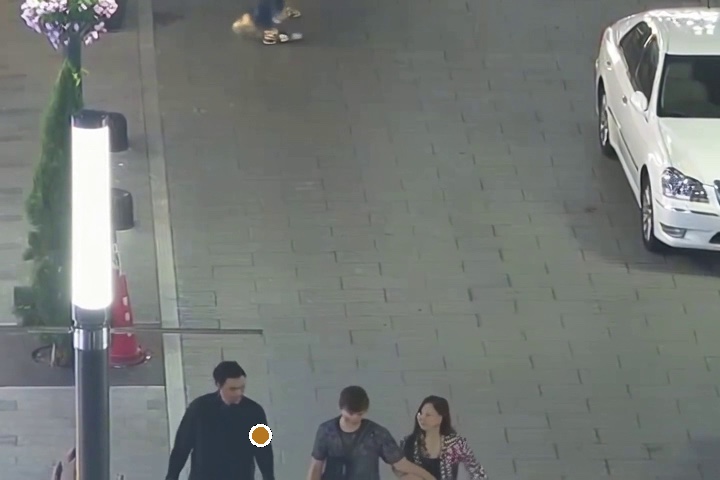} \\
\end{tabular}
\caption{\textbf{Additional qualitative comparison on MOT17.} Left: CogVideoX-5B against MagicMotion and Tora. Right: WaN~2.1-14B against ATI and Wan-Move. This real-world scene features many pedestrians with frequent occlusions (video results on our project page provide a clearer view). MagicMotion changes the appearance of the pink-dot pedestrian (white shirt), and Tora causes him to disappear entirely, in addition to inaccurate motion on other trajectories. ATI and Wan-Move succeed in moving objects but with less precision, while our method places each pedestrian's center exactly on its target dot.}
\label{fig:qual_extra_mot17}
\end{figure*}

We provide additional qualitative comparisons extending the main paper (Figure~\ref{fig:qual_unified}), with one scene per backbone on every evaluation dataset. In all figures below, each row shows three uniformly spaced frames (9, 29, 49). GT rows show full trajectory paths, and other rows show only the per-object ground-truth location as a dot. Across the synthetic ViN datasets (MoVi,
Figure~\ref{fig:qual_extra_movi}; Pool, Figure~\ref{fig:qual_extra_pool}), the synthetic pedestrian dataset (MOTSynth, Figure~\ref{fig:qual_extra_motsynth}), and the two real-world scene datasets (MOT17, Figure~\ref{fig:qual_extra_mot17}; DAVIS, Figure~\ref{fig:qual_extra_davis}), our method keeps each object aligned with its prescribed trajectory while the baselines drift, merge trajectories, or fail to move objects at all. The real-world scenes are out-of-distribution for our synthetically trained model but within the training domain of the baselines, which are all trained on large-scale real-world video data.

\newpage

\clearpage

\section{Failure Cases}
\label{app:failure_cases}

\subsection{GTA-V Training-Distribution Leakage on Out-of-Distribution Scenes}
\label{app:gta_hallucination}

Our CogVideoX model exhibits a failure mode on out-of-distribution scenes. When trajectory guidance forces an object through large displacements, the model occasionally morphs the object's appearance into a GTA-V pedestrian sprite mid-generation. We attribute this to the dominance of MOTSynth~\citep{motsynth} in the training mixture (${\sim}$44\% of clips, all rendered from GTA~V). When the object category is underrepresented in training, the appearance encoder's signal becomes insufficient to override the strong GTA-V prior learned by the backbone, causing the model to default to the most frequently seen appearance. The same effect occurs on MOT17, though it is less visually pronounced since the scenes already contain pedestrians and the GTA-V style is closer to the target domain. 

\begin{figure}[t]
\centering
\setlength{\tabcolsep}{0.5pt}
\renewcommand{\arraystretch}{0.3}
\newcommand{\qimgr}[1]{\includegraphics[width=0.24\textwidth]{figures/qual/rallye_eq/#1}}
\begin{tabular}{@{}c cccc@{}}
& \scriptsize Frame 1 & \scriptsize Frame 17 & \scriptsize Frame 33 & \scriptsize Frame 49 \\[1pt]
\rotatebox{90}{\scriptsize\hspace{10pt}GT} &
\qimgr{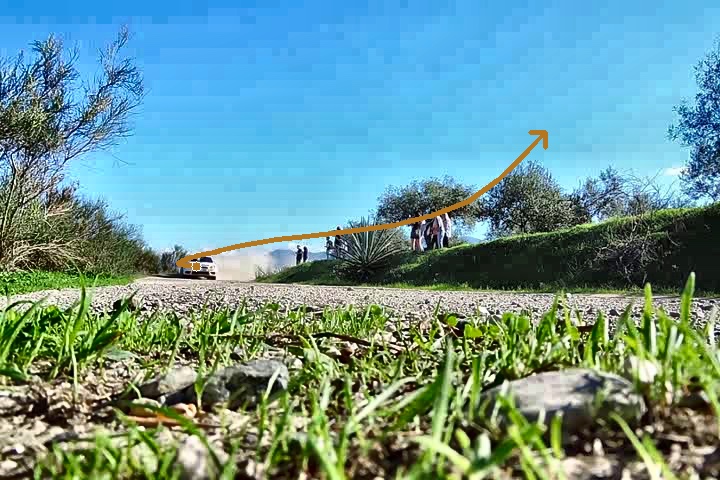} & \qimgr{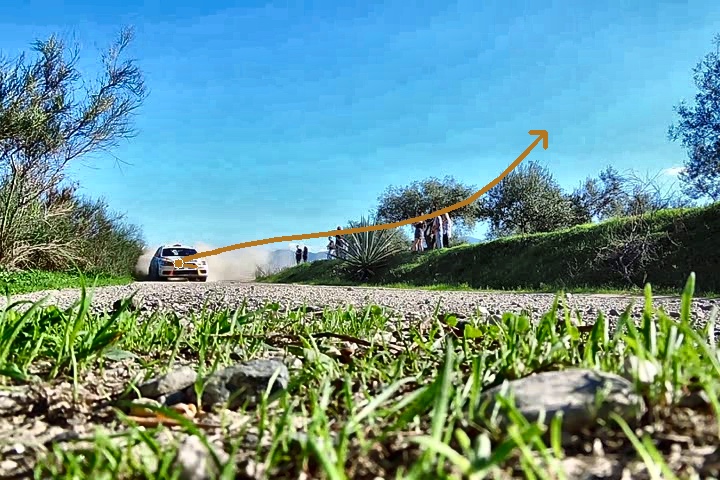} & \qimgr{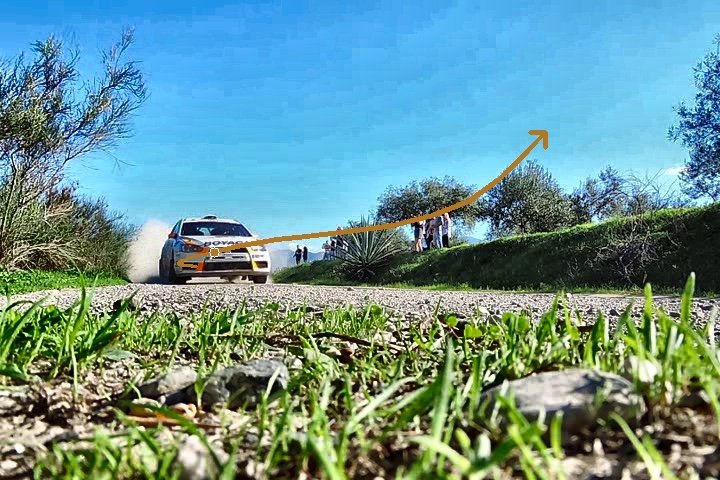} & \qimgr{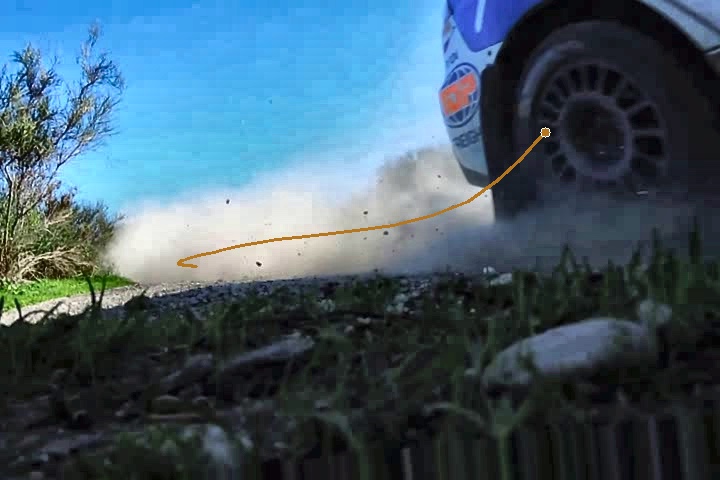} \\[0.5pt]
\rotatebox{90}{\scriptsize\hspace{3pt}Ckpt 8{,}500} &
\qimgr{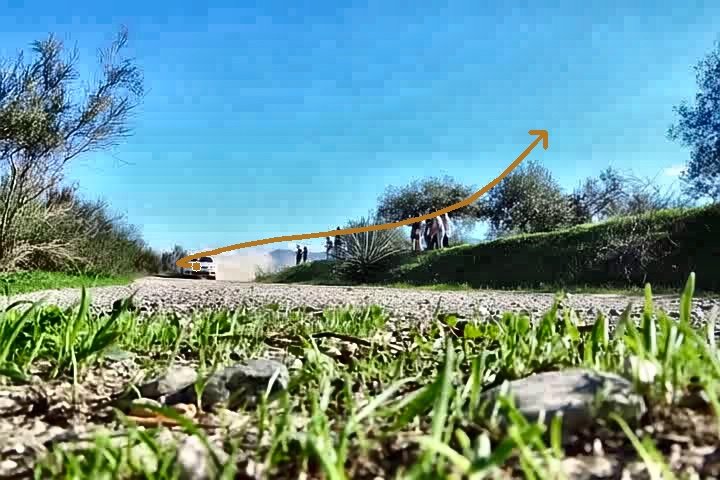} & \qimgr{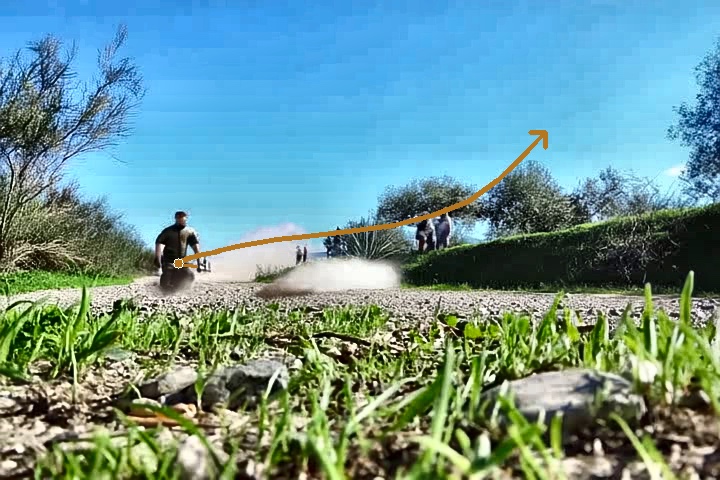} & \qimgr{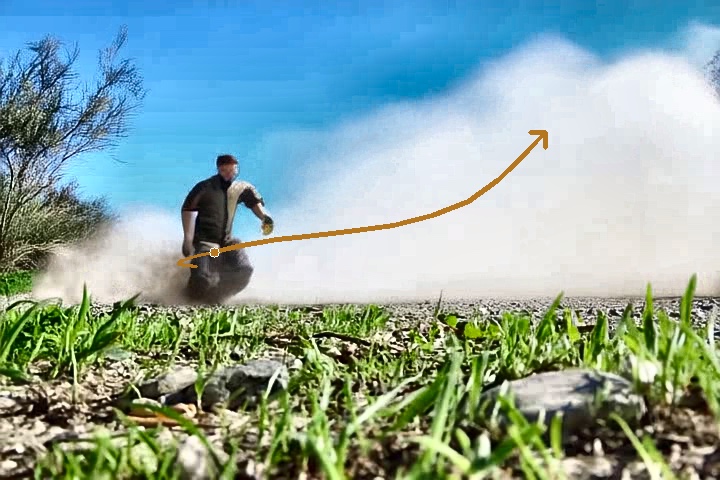} & \qimgr{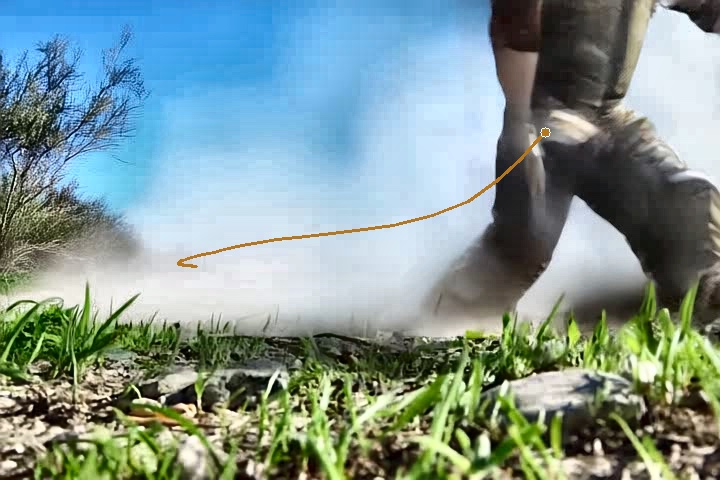} \\[0.5pt]
\rotatebox{90}{\scriptsize\hspace{3pt}Ckpt 5{,}500} &
\qimgr{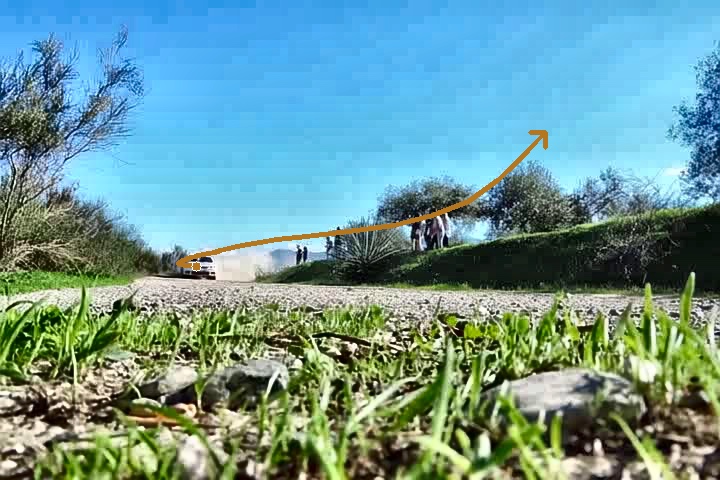} & \qimgr{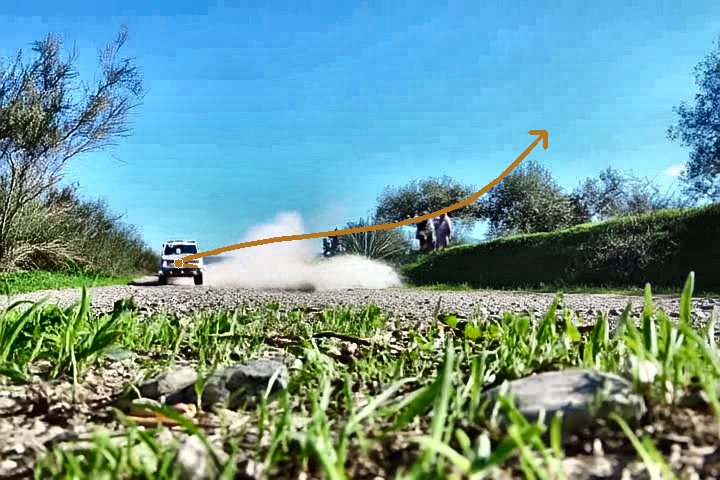} & \qimgr{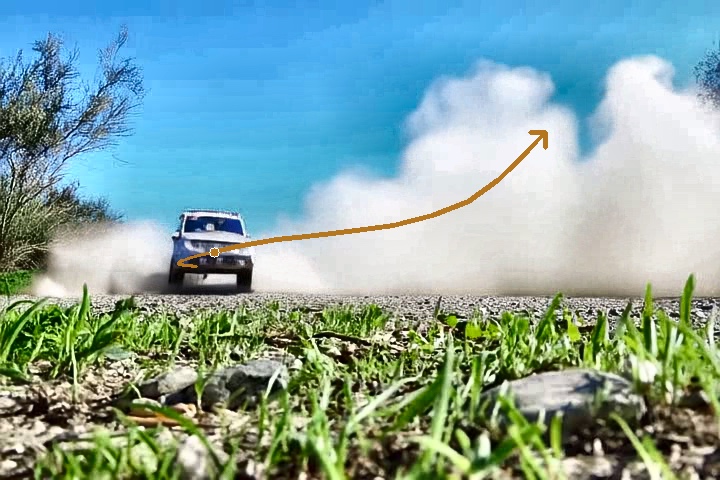} & \qimgr{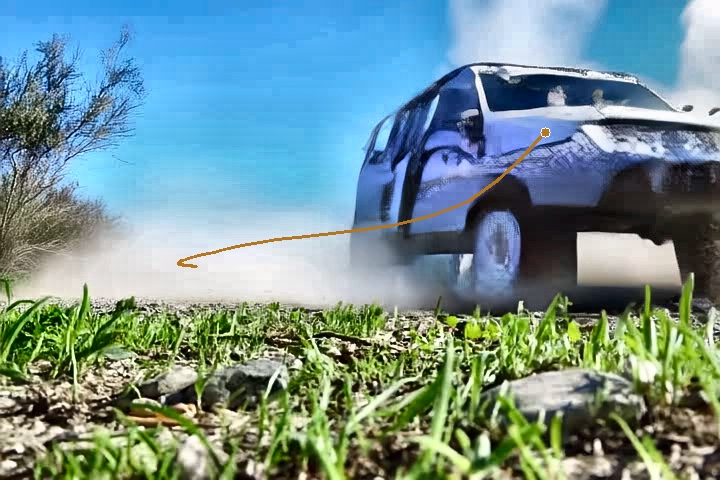} \\
\end{tabular}
\caption{\textbf{GTA-V training-distribution leakage on the DAVIS \texttt{rallye} scene.} Checkpoint 8{,}500 (middle row) generates the rally car correctly at the start but morphs it into a GTA-V pedestrian sprite mid-clip. Checkpoint 5{,}500 (bottom row), from earlier in training, retains stable car appearance throughout.}
\label{fig:qual_rallye_failure}
\end{figure}

Figure~\ref{fig:qual_rallye_failure} illustrates this on the DAVIS \texttt{rallye} scene. The rally car is generated correctly at the start but visibly morphs into a GTA-V pedestrian sprite mid-clip. To support our hypothesis that this stems from overfitting to the synthetic data, we generate with an earlier checkpoint (5,500 steps) of the same model, which produces a rally car with stable appearance throughout. Quantitatively, the earlier checkpoint improves PSNR by 3\%, EPE by 5\%, and FVD by 12\% on DAVIS. We report the later checkpoint in the main tables because it yields the best performance across all six datasets.

This observation reflects a broader limitation of the current training setup. There is no publicly available large-scale static-camera dataset of real-world multi-object trajectories at the scale required for video diffusion fine-tuning, so our training mixture is heavily weighted toward synthetic sources. When the evaluation distribution drifts far from those sources, style priors from the dominant source leak into the output. Scaling this line of work to broader real-world scenes will require collecting or curating a comparably large multi-object trajectory dataset outside the synthetic domain.

\begin{figure}[h]
\centering
\setlength{\tabcolsep}{0.5pt}
\renewcommand{\arraystretch}{0.3}
\newcommand{\qimgw}[1]{\includegraphics[width=0.30\textwidth]{figures/qual/davis_noeq/#1}}
\begin{tabular}{@{}c ccc@{}}
& \scriptsize GT & \scriptsize gs=5.0 (reported) & \scriptsize gs=3.0 \\[1pt]
\rotatebox{90}{\scriptsize\hspace{3pt}gold-fish} &
\qimgw{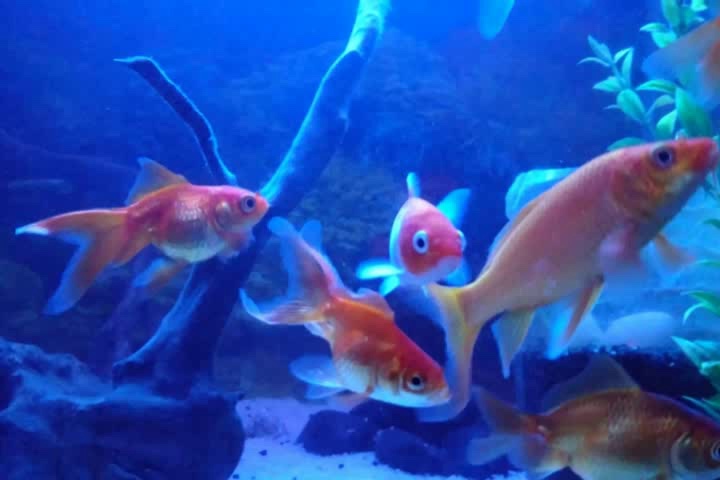} & \qimgw{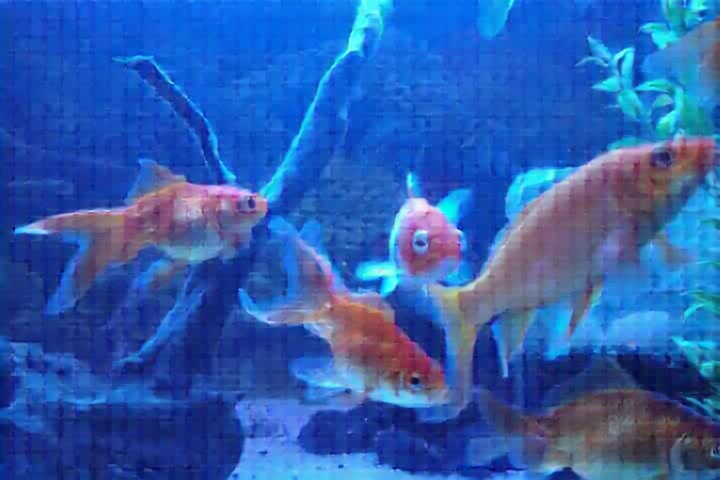} & \qimgw{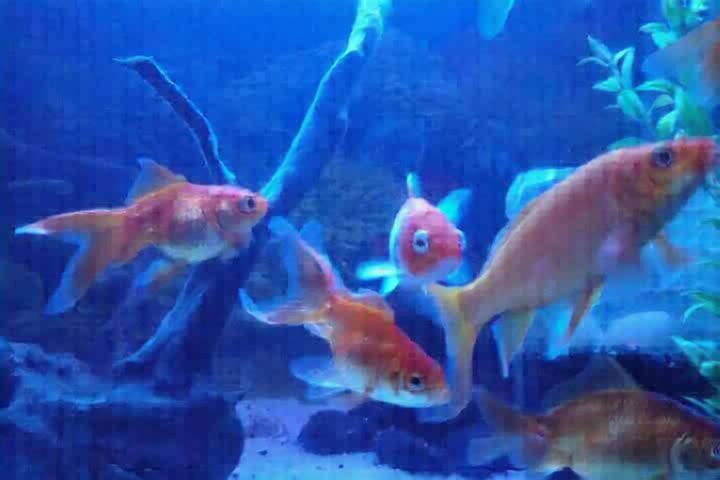} \\[0.5pt]
\rotatebox{90}{\scriptsize\hspace{2pt}schoolgirls} &
\qimgw{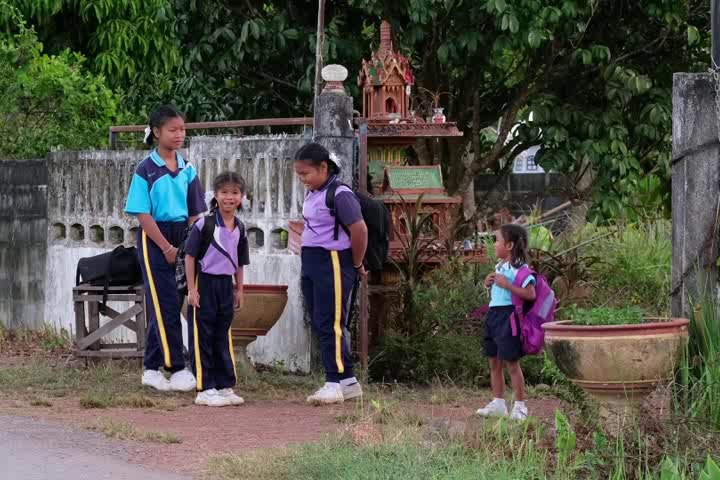} & \qimgw{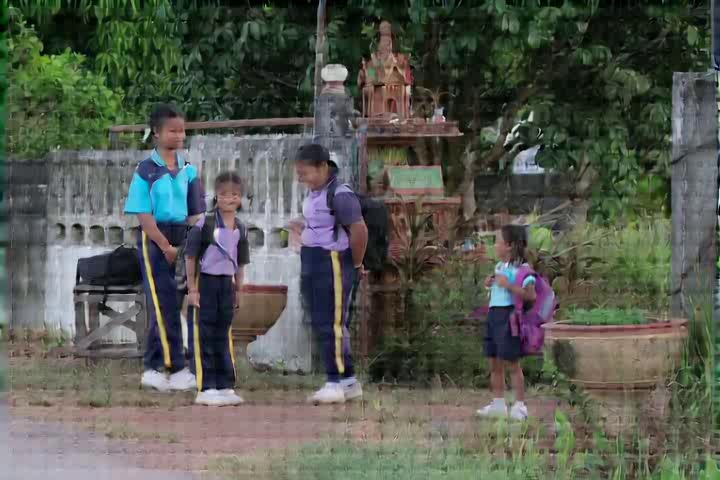} & \qimgw{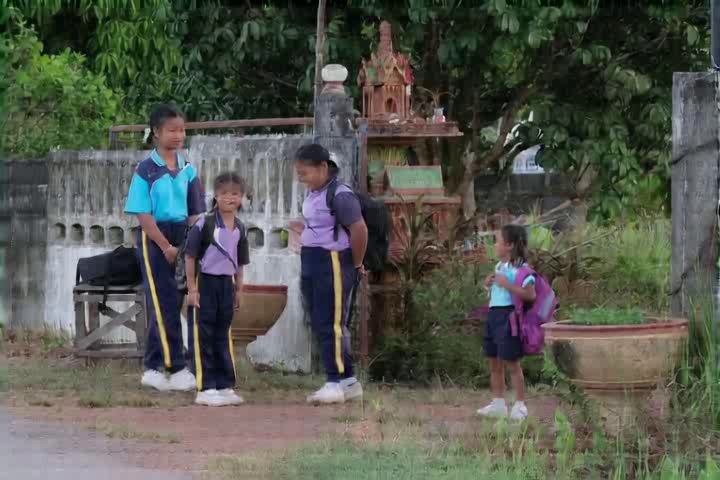} \\[0.5pt]
\rotatebox{90}{\scriptsize\hspace{2pt}soccerball} &
\qimgw{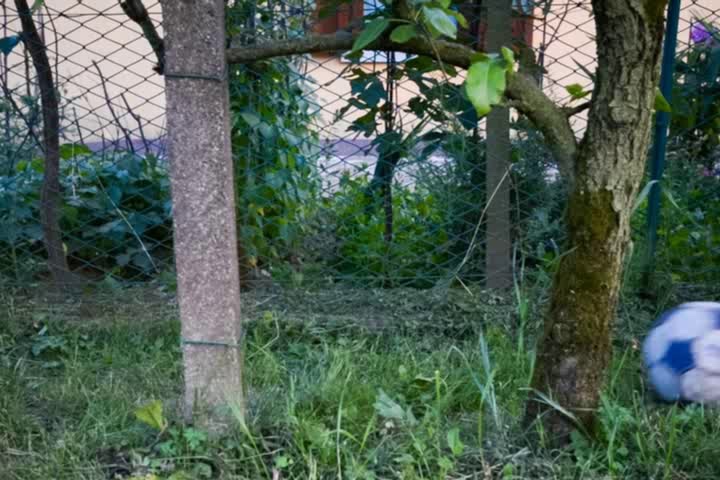} & \qimgw{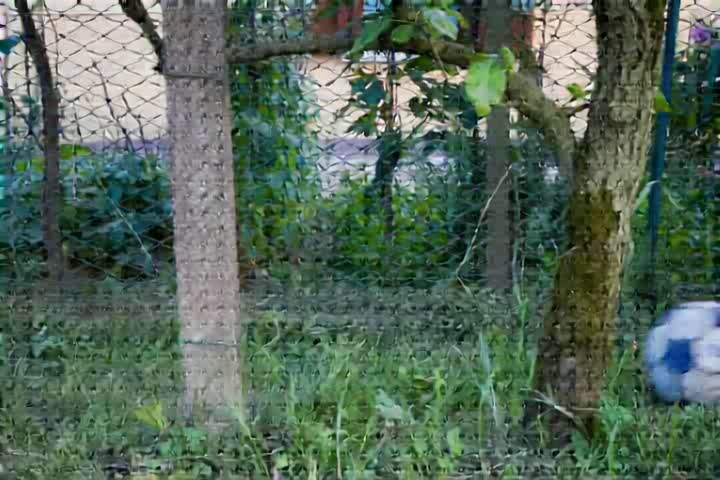} & \qimgw{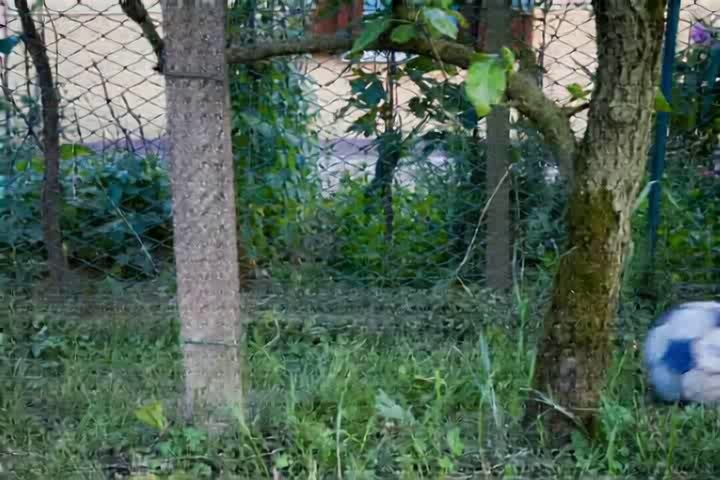} \\
\end{tabular}
\caption{\textbf{First-frame rendering artifacts on the WaN backbone.} Frame~1 of three DAVIS scenes. With the reported guidance scale gs{=}5.0 (middle column), the first few frames of some scenes show visible rendering artifacts that do not appear in the GT. Lowering to gs{=}3.0 (right column) mitigates the effect while leaving object appearance intact.}
\label{fig:qual_wan_cfg_artifacts}
\end{figure}

\clearpage
\begin{figure}[h]
\centering
\setlength{\tabcolsep}{0.5pt}
\renewcommand{\arraystretch}{0.3}
\newcommand{\qimgm}[1]{\includegraphics[width=0.225\textwidth]{figures/qual/motsynth_noeq/#1}}
\newcommand{\framesarrowm}{%
  \begin{tikzpicture}[baseline=0pt]
    \draw[->, line width=0.5pt] (0,0) -- (0.9\textwidth, 0)
      node[midway, fill=white, inner sep=1pt] {\scriptsize frames};
  \end{tikzpicture}%
}
\begin{tabular}{@{}c cccc@{}}
& \multicolumn{4}{c}{\framesarrowm} \\[1pt]
\rotatebox{90}{\scriptsize\hspace{8pt}GT} &
\qimgm{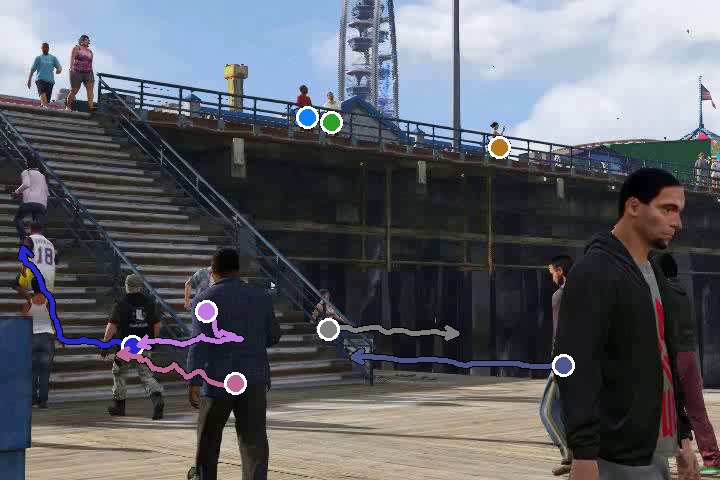} & \qimgm{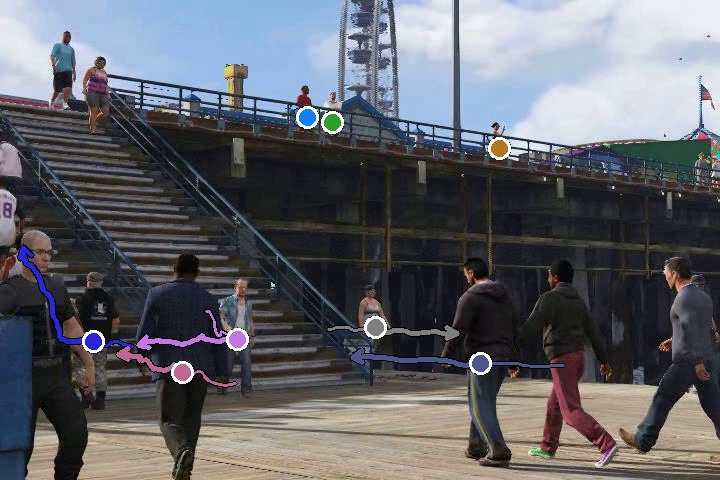} &
\qimgm{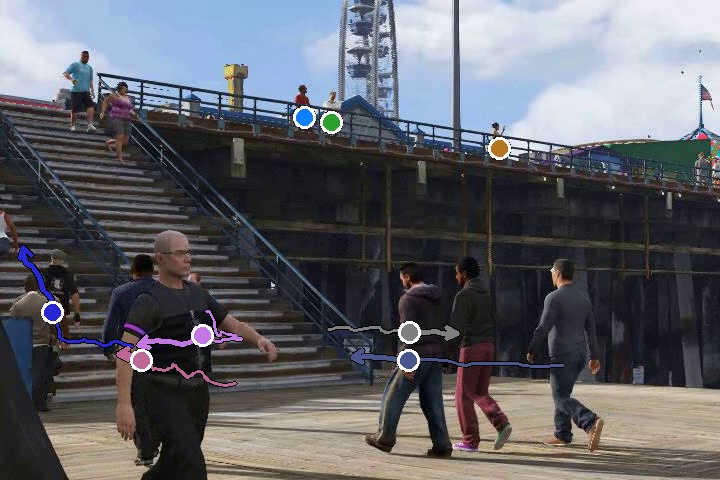} & \qimgm{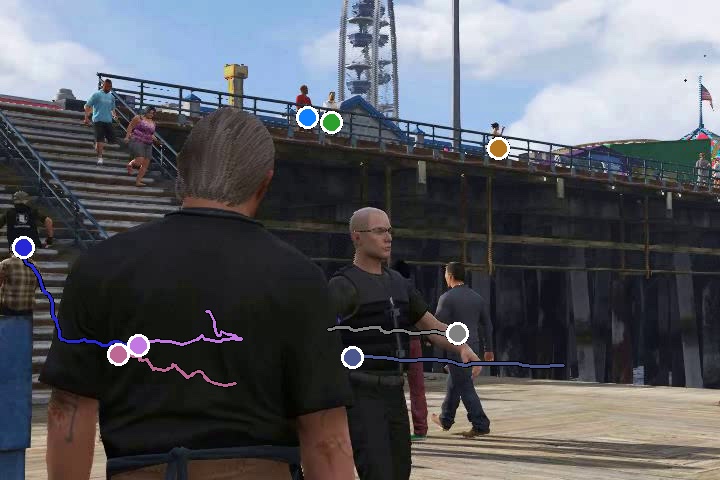} \\[0.5pt]
\rotatebox{90}{\scriptsize\hspace{2pt}Ours-WaN} &
\qimgm{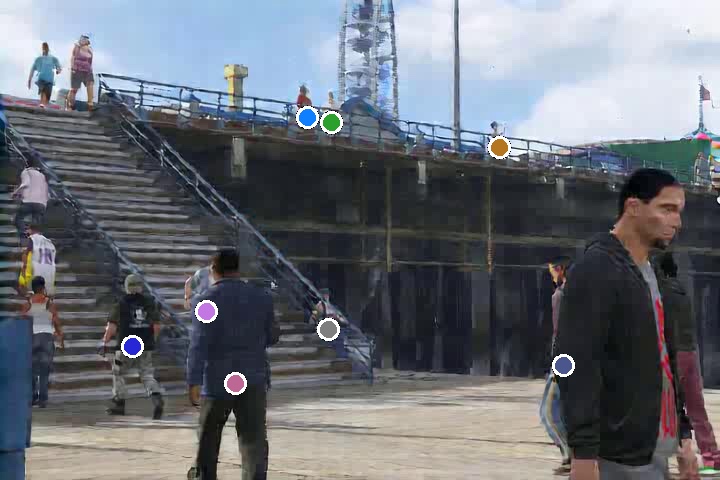} & \qimgm{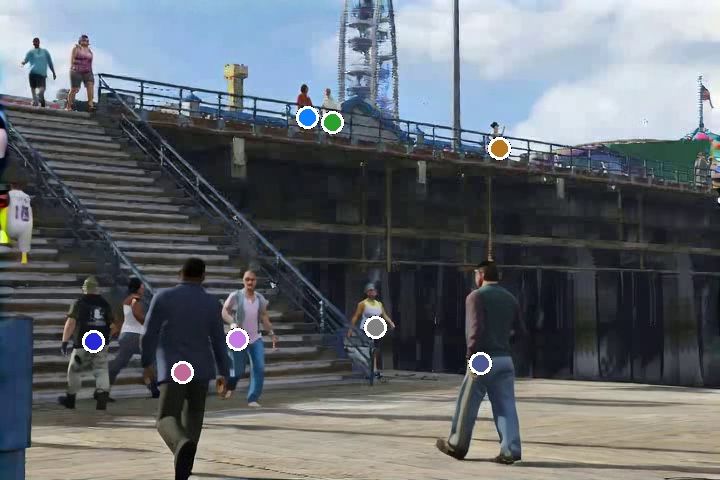} &
\qimgm{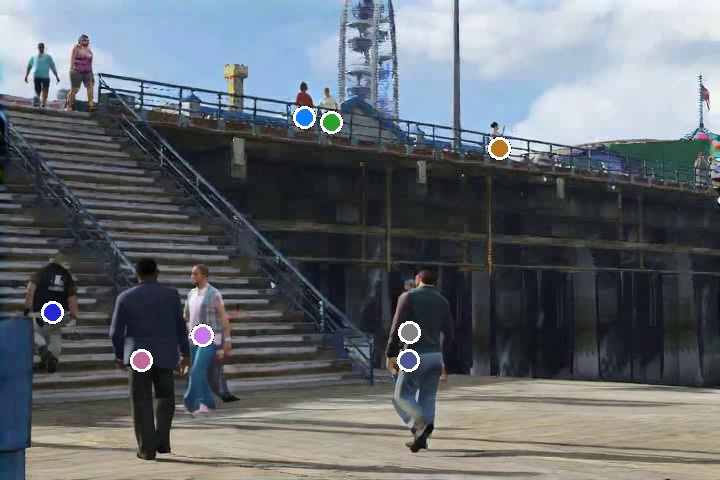} & \qimgm{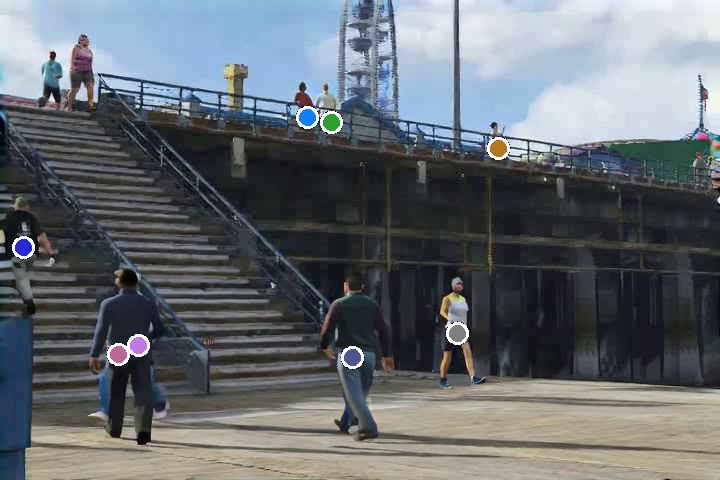} \\[0.5pt]
\rotatebox{90}{\scriptsize\hspace{1pt}Ours-CogV} &
\qimgm{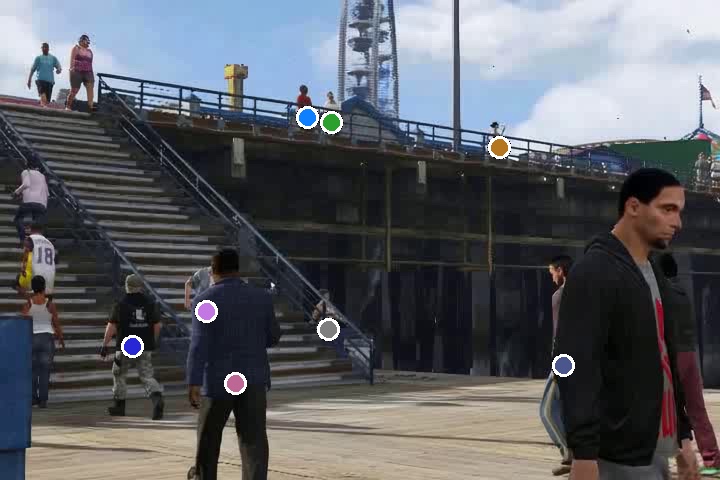} & \qimgm{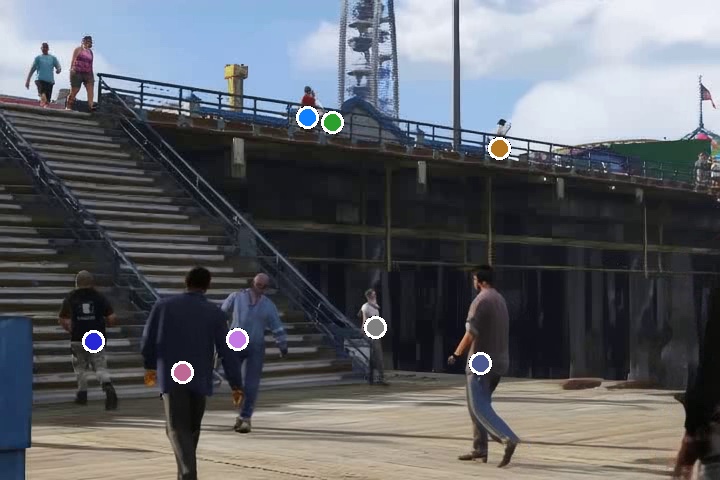} &
\qimgm{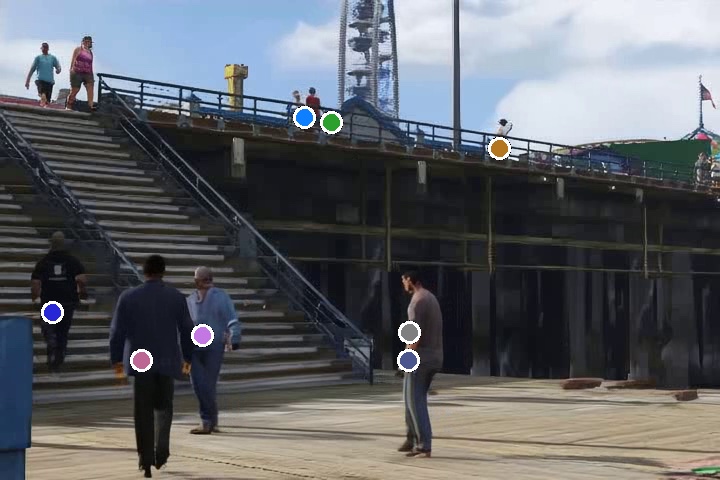} & \qimgm{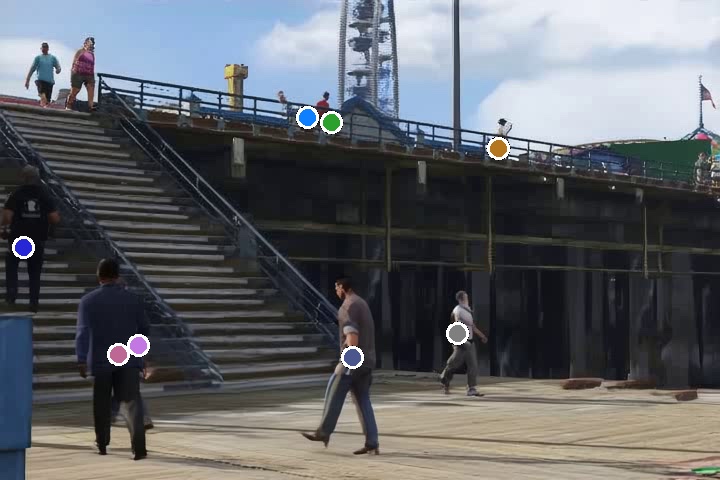} \\
\end{tabular}
\caption{\textbf{Unnatural motion of untracked objects.} A MOTSynth scene shown at four uniformly spaced frames (1, 17, 33, 49). The GT row shows the full trajectory paths provided as input. The Ours rows show only the dots at the current frame, so any residual motion visible in the scene comes from the model rather than the conditioning. Both backbones follow the tracked trajectories but move \emph{untracked} objects implausibly. The woman in a white tank top in the bottom-left of frame~1. On Ours-WaN she merges with a tracked pedestrian under occlusion. On Ours-CogV she drifts backwards out of frame.}
\label{fig:qual_untracked_motion}
\end{figure}

\subsection{First-Frame Rendering Artifacts on the WaN Backbone}
\label{app:wan_cfg_artifacts}

On some scenes, the first few frames of our WaN model show visible rendering artifacts that are absent from the ground truth and do not persist beyond those opening frames. We hypothesize that this artifact stems from the interaction between our method and classifier-free guidance (CFG). During CFG, the model performs both a conditional and an unconditional forward pass, and our method is applied identically to both branches. Since the conditional branch contains the learned subject tokens while the unconditional branch does not, applying the same intervention to both may introduce inconsistencies. We find that simply lowering the CFG scale from gs{=}5.0 to gs{=}3.0, while keeping our method unchanged, visibly reduces the artifacts. This supports our hypothesis, as a lower scale reduces the influence of the unconditional branch. We note that all metrics reported in the paper are computed with the default parameters (gs{=}5.0). Figure~\ref{fig:qual_wan_cfg_artifacts} shows the first frame of three DAVIS examples. First-frame PSNR (gs{=}5.0 $\rightarrow$ gs{=}3.0) improves from 24.82 to 27.94 on gold-fish, 22.71 to 24.17 on schoolgirls, and 22.61 to 25.15 on soccerball. Designing branch-specific interventions that handle the conditional and unconditional passes differently is left for future work.

\subsection{Unnatural Motion of Untracked Objects}
\label{app:untracked_motion}

Our trajectory inputs specify the motion of a subset of the objects in the scene, and we rely on the base I2V model's generative prior to animate everything else (background, crowds, occluders). On crowded MOTSynth scenes, we occasionally observe implausible motion of these untracked objects. Figure~\ref{fig:qual_untracked_motion} shows a representative example: the woman in a white tank top in the bottom-left of the first frame is not part of the trajectory inputs. On Ours-WaN she merges with a tracked pedestrian under occlusion, and on Ours-CogV she drifts backwards out of frame in a way inconsistent with natural walking. The tracked objects follow their prescribed trajectories correctly in both cases. The failure reflects the limits of what trajectory conditioning alone can enforce: the model has no signal for what untracked objects should do, and nothing in our attention replacement constrains them.

\section{Joint Self-Attention: Derivations and Feasibility}
\label{app:joint_attention}

\subsection{Attention-Weight Matrix Sizes: Cross-Attention vs.\ Joint Self-Attention}
\label{app:SA_replacement}

Our column replacement (Eq.~\ref{eq:col_replace}) overwrites entries of the attention-weight matrix $\bA$ before it is multiplied by $\bV$. Whether this can be applied exactly depends on whether $\bA$ is small enough to store in memory at the scale of the target backbone. The two backbones we consider sit on opposite sides of this boundary, and this appendix makes the distinction quantitative.

We instantiate both cases at $720{\times}480$ resolution with $49$ video frames, which yields a latent spatiotemporal grid of $F{\times}h{\times}w = 13{\times}30{\times}45$ and therefore $M = F\!\cdot\! h\!\cdot\! w = 17{,}550$ video tokens.

\noindent \textbf{Explicit cross-attention (WaN~2.1-14B)} \quad
WaN~2.1 uses separate cross-attention: video tokens $\bQ_{\mathrm{vid}}\in\RR^{M\times d}$ attend to text tokens $\bK_{\mathrm{txt}},\bV_{\mathrm{txt}}\in\RR^{L\times d}$, with $L = 512$ and $d = 128$ over $40$ heads. The per-head attention matrix is therefore
\begin{equation}
    \bA \in \RR^{M \times L} = \RR^{17{,}550 \times 512},
\end{equation}
which is approximately $17{,}550 \times 512 \times 2~\text{bytes} \approx 18~\text{MB}$ per head in bfloat16. At $40$ heads, the full per-layer footprint is ${\sim}720$~MB, which fits comfortably alongside the model activations. $\bA$ can therefore be stored in memory, the supervised columns at $\{q_i\}$ overwritten, and the row-wise renormalization of Eq.~\ref{eq:col_replace} applied directly.

\noindent \textbf{Joint self-attention (CogVideoX-5B)} \quad
CogVideoX concatenates text and video tokens into a single sequence of length $L_{\mathrm{tot}} = L + M = 226 + 17{,}550 = 17{,}776$ and performs a single self-attention pass over the joint sequence. The per-head attention matrix is
\begin{equation}
    \bA \in \RR^{L_{\mathrm{tot}} \times L_{\mathrm{tot}}} = \RR^{17{,}776 \times 17{,}776},
\end{equation}
approximately $17{,}776^2 \times 2~\text{bytes} \approx 0.6~\text{GB}$ per head in bfloat16. With $48$ heads the per-layer footprint is ${\sim}30$~GB, and the transformer has tens of such layers. Storing $\bA$ in memory is therefore infeasible in practice, which is precisely why production implementations of joint self-attention rely on FlashAttention-style kernels that never store $\bA$ explicitly.

\subsection{Weight Replacement Approximation in Self-Attention}
\label{app:two_sdpa}
\label{sec:two_sdpa}

In joint self-attention architectures such as CogVideoX~\citep{cogvideox}, text and video tokens form a single sequence. The full attention matrix is then too large to store in memory, reaching roughly $30$~GB per layer for CogVideoX (Appendix~\ref{app:SA_replacement}). Standard efficient kernels (e.g., FlashAttention) therefore never build $\bA$ explicitly, and applying Eq.~\ref{eq:col_replace} exactly would require a custom kernel that performs the column overwrite inside the tiled attention loop. Instead, we derive two alternatives that operate entirely with standard scaled dot-product attention (SDPA) calls.

Below we first state the exact column-replacement output for joint self-attention, then derive two relaxations of it that use only standard SDPA calls.

\noindent \textbf{Exact formulation} \quad
We derive the counterpart of the cross-attention output $\bO' = \bA' \bV_{\mathrm{txt}}$ (\S\ref{sec:xattn_replace}) for the joint self-attention setting. Let $\bQ, \bK, \bV \in \RR^{L_{\mathrm{tot}} \times d}$ denote the joint query, key, and value matrices, with $L_{\mathrm{tot}} = L + M$ and per-head dimension $d$. We order joint tokens as $[\text{text};\text{video}]$, so video position $m\in\{1,\dots,M\}$ corresponds to joint index $L+m$, and we write $\bQ_{\mathrm{vid}}$ for the video-row submatrix of $\bQ$. We write $\mathrm{SDPA}(\bQ,\bK,\bV)$ for the scaled dot-product attention output and $\mathrm{SDPA}(\bQ,\bK,\bV)[m]$ for its $m$-th row. Throughout this section and Appendices~\ref{app:joint_exact} and~\ref{app:additive_derivation}, we write $\bA[m,\cdot]$ as shorthand for the $m$-th video-query row of the joint attention matrix (i.e., joint row $L+m$), for $m\in\{1,\dots,M\}$.

Let $\{q_i\}_{i=1}^{N}$ denote the object-token columns introduced in \S\ref{sec:xattn_replace} (the columns of $\bA$ where attention weights are replaced by trajectory heatmaps); we refer to these as \emph{supervised positions}. Let $\mathcal U = \{1,\dots,L_{\mathrm{tot}}\}\setminus\{q_i\}_{i=1}^{N}$ be the complementary set of non-supervised positions, and write $\bK_{\mathcal U}, \bV_{\mathcal U}$ for the submatrices restricted to those rows. Define
\begin{equation}
    H[m]=\sum_{i=1}^{N}\bh_i[m],
    \qquad
    S_{\mathcal U}[m]=\sum_{j\in\mathcal U}\bA[m,j],
    \label{eq:su_h_def}
\end{equation}
where $H[m]$ is the total trajectory heatmap mass at video position $m$, and $S_{\mathcal U}[m]$ is the native attention mass assigned to the non-supervised positions. The exact counterpart of $\bO' = \bA' \bV_{\mathrm{txt}}$ for joint self-attention is
\begin{equation}
    \bO_{\mathrm{vid}}^{\mathrm{exact}}[m]
    =
    \frac{S_{\mathcal U}[m]}{S_{\mathcal U}[m]+H[m]}
    \,\mathrm{SDPA}(\bQ_{\mathrm{vid}},\bK_{\mathcal U},\bV_{\mathcal U})[m]
    +
    \frac{1}{S_{\mathcal U}[m]+H[m]}
    \sum_{i=1}^{N}\bh_i[m]\bV[q_i].
    \label{eq:exact_joint}
\end{equation}
The derivation of Eq.~\ref{eq:exact_joint} is provided in Appendix~\ref{app:joint_exact}.

\noindent \textbf{First relaxation: additive correction} \quad
A simple relaxation preserves the native joint self-attention output at the video-token positions and adds the supervised contribution:
\begin{equation}
\begin{aligned}
    \bO_{\mathrm{vid}}^{\mathrm{add}}[m]
    &= \mathrm{SDPA}(\bQ,\bK,\bV)[L+m]
      + \sum_{i=1}^{N}\bh_i[m]\,\bV[q_i] \\
    &= S_{\mathcal U}[m]\,\mathrm{SDPA}(\bQ_{\mathrm{vid}},\bK_{\mathcal U},\bV_{\mathcal U})[m]
      + \sum_{i=1}^{N}\bigl(\bA[m,q_i] + \bh_i[m]\bigr)\,\bV[q_i].
\end{aligned}
\label{eq:additive}
\end{equation}
This form is close to Eq.~\ref{eq:exact_joint} but weaker: the supervised contribution carries $\bigl(\bA[m,q_i] + \bh_i[m]\bigr)$, so the native attention weight $\bA[m,q_i]$ sits \emph{on top of} the heatmap $\bh_i[m]$ rather than being replaced by it, and the non-supervised coefficient $S_{\mathcal U}[m]$ is not renormalized against the supervised mass. The derivation of the second line is given in Appendix~\ref{app:additive_derivation}. The second line of Eq.~\ref{eq:additive} is an equivalent rewrite of the first, not a computational recipe: the additive correction is implemented using the first line, which avoids forming $\bA$ and therefore avoids accessing the $S_{\mathcal U}[m]$ and $\bA[m,q_i]$ quantities that appear only for comparison with Eq.~\ref{eq:exact_joint}.

\noindent \textbf{Second relaxation: Two-SDPA approximation} \quad
A stronger relaxation follows the same supervised/non-supervised decomposition as Eq.~\ref{eq:exact_joint}, but replaces the full-sequence SDPA call with an SDPA call restricted to $\mathcal U$. The supervised term then drops the native weight $\bA[m,q_i]$ and carries only the heatmap; the non-supervised term is rescaled by $1-H[m]$, the mass complement of the supervised contribution, yielding
\begin{equation}
    \bO_{\mathrm{vid}}^{\mathrm{2sdpa}}[m]
    =
    (1-H[m])\,\mathrm{SDPA}(\bQ_{\mathrm{vid}},\bK_{\mathcal U},\bV_{\mathcal U})[m]
    +
    \sum_{i=1}^{N}\bh_i[m]\bV[q_i].
    \label{eq:two_sdpa}
\end{equation}
The approximation becomes exact whenever the base model's native attention at the supervised columns already matches the target heatmap mass, $\sum_{i=1}^{N}\bA[m,q_i] = H[m]$ (equivalently, $S_{\mathcal U}[m] = 1-H[m]$). Both formulations also coincide when $H[m]=0$, i.e., far from all trajectories, where the supervised contribution vanishes and the non-supervised coefficient evaluates to $1$ in both forms, reducing to $\mathrm{SDPA}(\bQ_{\mathrm{vid}},\bK_{\mathcal U},\bV_{\mathcal U})[m]$. When $H[m]>0$, both formulations combine the same untouched-column term with the same supervised contribution $\sum_i \bh_i[m]\bV[q_i]$, differing only in their relative weighting.
This formulation is implemented with two standard SDPA calls, incurring approximately a $2\times$ attention overhead per layer.

\subsection{Exact Form of Joint-Attention Replacement}
\label{app:joint_exact}

In this section, we derive the exact analogue of the column replacement in Eq.~\ref{eq:col_replace} for joint self-attention architectures (Eq.~\ref{eq:exact_joint}).

We use the notation established in Appendix~\ref{app:two_sdpa}: the joint matrices $\bQ,\bK,\bV\in\RR^{L_{\mathrm{tot}}\times d}$, the supervised positions $\{q_i\}_{i=1}^{N}$ and their complement $\mathcal U$, and the per-position masses $H[m]$ and $S_{\mathcal U}[m]$. Let $\bA=\mathrm{softmax}(\bQ\bK^\top/\sqrt d)\in\RR^{L_{\mathrm{tot}}\times L_{\mathrm{tot}}}$ denote the hypothetical full joint attention matrix (which cannot be stored in practice; Appendix~\ref{app:SA_replacement}).

Apply the column replacement (Eq.~\ref{eq:col_replace}) to $\bA$: the modified matrix $\bA'$ satisfies $\bA'[m,j] = \bA[m,j]$ for $j\in\mathcal U$ (unsupervised columns untouched) and $\bA'[m,q_i] = \bh_i[m]$ (supervised columns overwritten). The row-$m$ sum after replacement is
\begin{equation}
    \sum_{k=1}^{L_{\mathrm{tot}}}\bA'[m,k]
    =
    \underbrace{\sum_{j\in\mathcal U}\bA[m,j]}_{S_{\mathcal U}[m]}
    +
    \underbrace{\sum_{i=1}^{N}\bh_i[m]}_{H[m]}.
\end{equation}
The renormalized output at video position $m$ is then
\begin{equation}
    \bO_{\mathrm{vid}}^{\mathrm{exact}}[m]
    =
    \frac{1}{S_{\mathcal U}[m]+H[m]}
    \Bigl(\sum_{j\in\mathcal U}\bA[m,j]\,\bV[j]
    +
    \sum_{i=1}^{N}\bh_i[m]\,\bV[q_i]\Bigr).
\end{equation}
For the unsupervised sum, factor out $S_{\mathcal U}[m]$:
\begin{equation}
    \sum_{j\in\mathcal U}\bA[m,j]\,\bV[j]
    =
    S_{\mathcal U}[m]\sum_{j\in\mathcal U}\frac{\bA[m,j]}{S_{\mathcal U}[m]}\,\bV[j].
\end{equation}
The ratio $\bA[m,j]/S_{\mathcal U}[m]$ is the softmax over only the unsupervised keys, since normalizing the original softmax weights by $S_{\mathcal U}[m]$ cancels the full-sequence normalization:
\begin{equation}
    \frac{\bA[m,k]}{\sum_{j\in\mathcal U}\bA[m,j]}
    =
    \frac{\exp(\bQ[m]\bK[k]^\top/\sqrt d)}
         {\sum_{j\in\mathcal U}\exp(\bQ[m]\bK[j]^\top/\sqrt d)},
    \qquad k\in\mathcal U.
    \label{eq:appendix_softmax_cancel}
\end{equation}
The unsupervised sum therefore equals $S_{\mathcal U}[m]\,\mathrm{SDPA}(\bQ_{\mathrm{vid}},\bK_{\mathcal U},\bV_{\mathcal U})[m]$. Substituting back yields
\begin{equation}
    \bO_{\mathrm{vid}}^{\mathrm{exact}}[m]
    =
    \frac{S_{\mathcal U}[m]}{S_{\mathcal U}[m]+H[m]}
    \,\mathrm{SDPA}(\bQ_{\mathrm{vid}},\bK_{\mathcal U},\bV_{\mathcal U})[m]
    +
    \frac{1}{S_{\mathcal U}[m]+H[m]}
    \sum_{i=1}^{N}\bh_i[m]\bV[q_i],
    \label{eq:appendix_exact_joint}
\end{equation}
which is Eq.~\ref{eq:exact_joint}. \hfill$\square$

\subsection{Additive Correction: Supervised/Non-Supervised Form}
\label{app:additive_derivation}

We derive the second line of Eq.~\ref{eq:additive}, which rewrites the additive correction $\bO_{\mathrm{vid}}^{\mathrm{add}}[m]$ in the same non-supervised/supervised decomposition used by the exact form in Eq.~\ref{eq:appendix_exact_joint}. Concretely, we show that
\begin{equation}
    \mathrm{SDPA}(\bQ,\bK,\bV)[L+m]
    =
    S_{\mathcal U}[m]\,\mathrm{SDPA}(\bQ_{\mathrm{vid}},\bK_{\mathcal U},\bV_{\mathcal U})[m]
    +
    \sum_{i=1}^{N}\bA[m,q_i]\,\bV[q_i],
    \label{eq:appendix_joint_decomp}
\end{equation}
and then plug this identity into Eq.~\ref{eq:additive} to obtain its reformulated right-hand side.

\paragraph{Step 1: decompose the joint-SDPA row $L+m$.}
By definition of attention, the $m$-th video-query row of the joint SDPA output is
\begin{equation}
    \mathrm{SDPA}(\bQ,\bK,\bV)[L+m]
    =
    \sum_{j=1}^{L_{\mathrm{tot}}}\bA[m,j]\,\bV[j]
    =
    \sum_{j\in\mathcal U}\bA[m,j]\,\bV[j]
    +
    \sum_{i=1}^{N}\bA[m,q_i]\,\bV[q_i].
\end{equation}

\paragraph{Step 2: rewrite the non-supervised sum.}
Factoring out $S_{\mathcal U}[m]$,
\begin{equation}
    \sum_{j\in\mathcal U}\bA[m,j]\,\bV[j]
    =
    S_{\mathcal U}[m]\sum_{j\in\mathcal U}\frac{\bA[m,j]}{S_{\mathcal U}[m]}\,\bV[j].
\end{equation}
The ratio $\bA[m,j]/S_{\mathcal U}[m]$ is exactly the softmax over only the non-supervised keys, by the same cancellation used in Eq.~\ref{eq:appendix_softmax_cancel}:
\begin{equation}
    \frac{\bA[m,j]}{S_{\mathcal U}[m]}
    =
    \mathrm{softmax}\!\bigl(\bQ_{\mathrm{vid}}[m]\bK_{\mathcal U}^\top/\sqrt d\bigr)[j],
    \qquad j\in\mathcal U.
\end{equation}
This is the softmax used by $\mathrm{SDPA}(\bQ_{\mathrm{vid}},\bK_{\mathcal U},\bV_{\mathcal U})$ at query row $m$, so
\begin{equation}
    \sum_{j\in\mathcal U}\bA[m,j]\,\bV[j]
    =
    S_{\mathcal U}[m]\,\mathrm{SDPA}(\bQ_{\mathrm{vid}},\bK_{\mathcal U},\bV_{\mathcal U})[m].
\end{equation}

\paragraph{Step 3: combine.}
Substituting back into Step 1 yields Eq.~\ref{eq:appendix_joint_decomp}.

\paragraph{Step 4: plug into Eq.~\ref{eq:additive} and collect supervised terms.}
Adding $\sum_i \bh_i[m]\,\bV[q_i]$ to Eq.~\ref{eq:appendix_joint_decomp} and grouping the two supervised sums gives
\begin{align}
    \bO_{\mathrm{vid}}^{\mathrm{add}}[m]
    &=
    \mathrm{SDPA}(\bQ,\bK,\bV)[L+m]
    +
    \sum_{i=1}^{N}\bh_i[m]\,\bV[q_i] \\
    &=
    S_{\mathcal U}[m]\,\mathrm{SDPA}(\bQ_{\mathrm{vid}},\bK_{\mathcal U},\bV_{\mathcal U})[m]
    +
    \sum_{i=1}^{N}\bA[m,q_i]\,\bV[q_i]
    +
    \sum_{i=1}^{N}\bh_i[m]\,\bV[q_i] \\
    &=
    S_{\mathcal U}[m]\,\mathrm{SDPA}(\bQ_{\mathrm{vid}},\bK_{\mathcal U},\bV_{\mathcal U})[m]
    +
    \sum_{i=1}^{N}\bigl(\bA[m,q_i]+\bh_i[m]\bigr)\,\bV[q_i],
\end{align}
which is the second line of Eq.~\ref{eq:additive}. \hfill$\square$

Comparing Eq.~\ref{eq:appendix_exact_joint} and this rewritten form of Eq.~\ref{eq:additive} side by side makes the semantic difference explicit. Both formulas combine the same non-supervised SDPA term and a supervised contribution at the same positions $\{q_i\}$, but the exact form uses renormalized coefficients $S_{\mathcal U}[m]/(S_{\mathcal U}[m]+H[m])$ and $1/(S_{\mathcal U}[m]+H[m])$ and relies entirely on the heatmap $\bh_i[m]$ for the supervised contribution. The additive form, in contrast, uses the unnormalized coefficient $S_{\mathcal U}[m]$ and carries the native attention weight $\bA[m,q_i]$ \emph{in addition to} the heatmap, which is why it is a weaker correction.

\subsection{Additive vs.\ Replacement Correction on CogVideoX}
\label{app:cogv_ablation}

Table~\ref{tab:cogv_ablation} compares the additive correction (Eq.~\ref{eq:additive}) with the two-SDPA replacement (Eq.~\ref{eq:two_sdpa}) on CogVideoX-5B across all six datasets. Replacement wins clearly on MOTSynth and MOT17 (all four metrics) and on Pool (three of four, EPE near-tied), while MoVi and Football are mixed. The pattern flips on DAVIS, where additive wins all four metrics. DAVIS is the only dataset whose object categories (animals, vehicles, sports) fall outside the training distribution. We report replacement throughout, as it wins on the most datasets.

\begin{table*}[h]
\caption{Additive vs.\ two-SDPA replacement on CogVideoX-5B across all six evaluation datasets. Both modes are ours; 2-SDPA is the variant used in the main-paper tables. Best per cell in \textbf{bold}.}
\label{tab:cogv_ablation}
\centering
\footnotesize
\setlength{\tabcolsep}{2.5pt}
\begin{tabular}{l cccc cccc cccc}
\toprule
& \multicolumn{4}{c}{\textbf{MoVi}} & \multicolumn{4}{c}{\textbf{Pool}} & \multicolumn{4}{c}{\textbf{Football}} \\
\cmidrule(lr){2-5} \cmidrule(lr){6-9} \cmidrule(lr){10-13}
Mode & PSNR$\uparrow$ & LPIPS$\downarrow$ & FVD$\downarrow$ & EPE$\downarrow$
     & PSNR$\uparrow$ & LPIPS$\downarrow$ & FVD$\downarrow$ & EPE$\downarrow$
     & PSNR$\uparrow$ & LPIPS$\downarrow$ & FVD$\downarrow$ & EPE$\downarrow$ \\
\midrule
2-SDPA (reported) & \textbf{29.77} & \textbf{.108} & 14.98 & 0.53 & \textbf{35.64} & \textbf{.025} & \textbf{8.79} & 0.07 & 24.79 & .090 & \textbf{13.87} & \textbf{0.03} \\
Additive & 29.42 & .116 & \textbf{14.16} & \textbf{0.52} & 33.42 & .029 & 8.97 & \textbf{0.06} & \textbf{24.98} & \textbf{.088} & 14.60 & \textbf{0.03} \\
\bottomrule
\\[4pt]
\toprule
& \multicolumn{4}{c}{\textbf{MOTSynth}} & \multicolumn{4}{c}{\textbf{MOT17}} & \multicolumn{4}{c}{\textbf{DAVIS}} \\
\cmidrule(lr){2-5} \cmidrule(lr){6-9} \cmidrule(lr){10-13}
Mode & PSNR$\uparrow$ & LPIPS$\downarrow$ & FVD$\downarrow$ & EPE$\downarrow$
     & PSNR$\uparrow$ & LPIPS$\downarrow$ & FVD$\downarrow$ & EPE$\downarrow$
     & PSNR$\uparrow$ & LPIPS$\downarrow$ & FVD$\downarrow$ & EPE$\downarrow$ \\
\midrule
2-SDPA (reported) & \textbf{22.34} & \textbf{.232} & \textbf{39.70} & \textbf{2.32} & \textbf{21.59} & \textbf{.265} & \textbf{53.92} & \textbf{2.37} & 18.39 & .267 & 89.39 & 2.43 \\
Additive & 22.16 & .234 & 42.31 & 2.42 & 20.76 & .275 & 60.97 & 3.16 & \textbf{18.84} & \textbf{.259} & \textbf{80.01} & \textbf{2.23} \\
\bottomrule
\end{tabular}
\end{table*}

\section{Encoder Architectures and Training}
\label{app:encoders_training}

\subsection{Trajectory Encoder Details}
\label{app:traj_encoder}

This section provides architectural and training details for the trajectory encoder $\mathcal{E}_\text{traj}$ introduced in Sec.~\ref{sec:traj_encoding}.

\noindent \textbf{Encoder architecture} \quad
The encoder $\mathrm{Enc}_\text{traj}$ consists of two stages. First, three 1D convolutions with stride 2 progressively downsample the temporal dimension while increasing the channel count ($4 \to 64 \to 128 \to 256$), each followed by batch normalization and GELU; the output is flattened rather than pooled and linearly projected to an intermediate representation $\bz_i$ of dimension $d = 32$. The four input channels are the per-frame coordinates $(x_i(t), y_i(t), d_i(t))$ together with a normalized temporal position channel $p(t) = (t-1)/(T-1) \in [0,1]$ for $t \in \{1,\dots,T\}$, concatenated along the channel dimension before the first convolution. Second, a two-layer MLP head maps $\bz_i$ to the token embedding:
\begin{equation}
    \langle traj_i \rangle = W_2 \, \text{GELU}(W_1 \, \bz_i),
    \qquad
    W_1 \in \RR^{512 \times d},\; W_2 \in \RR^{4096 \times 512},
\end{equation}
with the final layer initialized to $\mathcal{N}(0, 0.02^2)$ weights and zero bias. The output is normalized to target $\text{std} \approx 0.15$ for T5-XXL and $\text{std} \approx 0.07$ for UMT5-XXL.

\noindent \textbf{Pretraining pipeline} \quad
Pretraining uses synthetic trajectories generated on-the-fly with statistics matched to real MOTSynth tracks (1{,}793 trajectories). Each trajectory $(x_t, y_t, d_t) \in [0,1]^3$ is generated as a clamped random walk: we sample a start position from a per-channel Gaussian fit to the real starts (means $(0.48, 0.50, 0.40)$, stds $(0.33, 0.30, 0.23)$), draw a constant per-trajectory drift velocity $(v_x, v_y, v_d)$ from Gaussians matching the real per-trajectory velocity stds $(0.008, 0.004, 0.001)$, and update each frame as
\begin{equation*}
    c_{t+1} = \mathrm{clamp}\!\left(c_t + v + \epsilon_t, 0, 1\right),
    \qquad
    \epsilon_t \sim \mathcal{N}\!\left(0, \sigma_\text{noise}^2\right),
\end{equation*}
with per-channel noise stds $\sigma_\text{noise} = (0.011, 0.004, 0.002)$. Each step samples $1$--$20$ trajectories (real mean $4.7$) and builds a prompt:
\begin{quote}
\ttfamily
"A GTA V street scene with pedestrians. The following represent pedestrian trajectories: [traj\_1], [traj\_2], ...".
\end{quote}
The trajectory embeddings $\{\langle traj_i \rangle \}$ are injected at the \texttt{[traj\_i]} positions and the frozen text encoder processes the full prompt. A symmetric decoder (${\sim}$12.8M parameters) reconstructs each trajectory from the text encoder output $\mathbf{c}_i$ at the corresponding position. The decoder consists of a linear expansion followed by three stride-2 transposed 1D convolutions with sigmoid activation, and receives $\mathbf{c}_i$ scaled by a learnable scalar that compensates for the magnitude change introduced by the text encoder's layers.

Although the encoder architecture includes a stochastic bottleneck (the reparameterization $\bz_i = \mu_i + \sigma_i \odot \epsilon$ is active during pretraining), we found empirically that removing the KL regularization term yields better trajectory reconstruction, so no KL penalty is applied. The variance $\sigma_i$ is therefore shaped only by the reconstruction gradient.

After pretraining, two changes are made: (1) the decoder is discarded, and (2) the stochastic sampling is replaced with the deterministic mean, $\bz_i = \mu_i$. $\mathrm{Enc}_\text{traj}$ is then frozen for all subsequent diffusion training.

Pretraining runs for 500K steps with AdamW (learning rate $3{\times}10^{-4}$, cosine decay, weight decay $10^{-2}$, gradient accumulation over 8 steps), using the loss $\mathcal{L}_\text{traj}$ (Eq.~\ref{eq:traj_vae_loss}) with $\lambda_\text{vel}=1$ and 1--20 trajectories per prompt.

\subsection{Appearance Encoder Details}
\label{app:app_encoder}

This section provides architectural details for the appearance encoder $\mathrm{Enc}_\text{app}$ introduced in Sec.~\ref{sec:appear_encoding}. Unlike the trajectory encoder, the appearance encoder is not pretrained: it is initialized randomly and trained jointly with the LoRA weights during diffusion fine-tuning.

\noindent \textbf{Architecture} \quad
The encoder takes the full first-frame VAE latent $V_0 \in \RR^{16 \times 60 \times 90}$ (at $720{\times}480$ resolution with VAE stride 8) and produces a dense appearance feature map $\bz^\text{app}$ through three 2D convolutions: one stride-2 layer ($16 \to 64$ channels, spatial $60{\times}90 \to 30{\times}45$) followed by two stride-1 refinement layers ($64 \to 64$), each with batch normalization and GELU. A $1{\times}1$ convolution head produces a per-pixel feature map of depth $D_\text{app} = 8$ at the compressed spatial resolution $30{\times}45$.

\noindent \textbf{Position sampling and projection} \quad
For each object $i$, the normalized starting position $(x_i^1, y_i^1)$ is mapped to the compressed grid and the corresponding $D_\text{app}$-dimensional feature vector is extracted via indexing. A single linear layer projects this to the text encoder dimension:
\begin{equation}
    e_{o_i} = W_\text{app} \, \bz^\text{app}[x_i^1, y_i^1, :] \in \RR^{4096}, \quad W_\text{app} \in \RR^{4096 \times 8}.
\end{equation}
The output is globally normalized to match the text encoder output statistics ($\text{std} \approx 0.15$ for T5-XXL, $\text{std} \approx 0.07$ for UMT5-XXL). The projection is initialized with $\mathcal{N}(0, 0.02^2)$ weights and zero bias.

\subsection{Additional Implementation Details}
\label{app:additional_impl}

The diffusion fine-tuning uses AdamW at learning rate $2{\times}10^{-4}$ in bf16 with gradient checkpointing, for 8{,}500 steps on CogVideoX (effective batch size 3 across 3 H100 GPUs, ${\sim}$57 hours) and 12{,}000 steps on WaN (effective batch size 2 across 2 H100 GPUs, ${\sim}$50 hours).

\section{Evaluation Protocol}
\label{app:evaluation_protocol}

\subsection{Training Data Composition}
\label{app:train_details}
The training data comprises 11{,}281 clips from MOTSynth (mean 4.2 pedestrians per clip, filtered to static-camera sequences), 8{,}236 from MoVi-Extended (mean 5.3 objects per clip), 2{,}943 from Pool (2--3 balls), and 3{,}040 from Football (up to 2 players and a ball).

\subsection{Inference Settings}
\label{app:inference}

Table~\ref{tab:inference_settings} details the inference configuration for each method. All baselines use their released default settings from their official code repositories.

\begin{table}[h]
\caption{Inference settings. All baselines use released defaults. MagicMotion uses bounding-box trajectory maps as ControlNet input. Solver abbreviations: CogV-DPM is the CogVideoX DPM scheduler; FM-Euler is flow-matching Euler; VP-SDE DPM++2M is the VP-SDE variant of DPM++2M from CogVideoX-SAT; FM-UniPC is flow-matching UniPC.}
\label{tab:inference_settings}
\centering
\footnotesize
\setlength{\tabcolsep}{2pt}
\begin{tabular}{lcccccc}
\toprule
& \textbf{Ours-Cog} & \textbf{Ours-Wan} & Tora & MagicMot. & Wan-Move & ATI \\
\midrule
Backbone & CogVideoX-5B & Wan2.1-14B & CogVideoX-5B & CogVideoX-5B & Wan2.1-14B & Wan2.1-14B \\
Mode & I2V & I2V & I2V & I2V & I2V & I2V \\
Resolution & $720{\times}480$ & $720{\times}480$ & $720{\times}480$ & $720{\times}480$ & $720{\times}480$ & $720{\times}480$ \\
Frames & 49 & 49 & 49 & 49 & 49 & 49 \\
Steps & 50 & 50 & 50 & 50 & 40 & 40 \\
Guidance & 6.0 & 5.0 & 6.0 & 6.0 & 5.0 & 5.0 \\
Solver & CogV-DPM & FM-Euler & VP-SDE DPM++2M & CogV-DPM & FM-UniPC & FM-UniPC \\
Traj.\ input & points & points & points & bboxes & points & points \\
\bottomrule
\end{tabular}
\end{table}

\subsection{DAVIS Static-Camera Evaluation Subset}
\label{app:davis}

To evaluate trajectory-conditioned generation on diverse real-world content beyond pedestrians, we evaluate on a subset of DAVIS 2017~\citep{pont2017davis}, a video object segmentation benchmark widely used for trajectory-conditioned evaluation~\citep{geng2024motion,wang2024levitor,posetraj,li2025magicmotion,wanmove}. We use the GT segmentation masks provided by DAVIS to extract trajectories, avoiding tracker noise.

\noindent \textbf{Static-camera filtering} \quad
Our method controls object motion but provides no signal about camera motion. In scenes with a moving camera, the apparent pixel-space displacement of an object reflects both its own motion and the camera's, so we cannot expect trajectory alignment without explicitly modeling the camera. Starting from the 90 sequences in the DAVIS 2017 train+val splits, we filter for static-camera scenes by computing background optical flow (Farneback~\citep{farneback2003two}) on sampled frame pairs using GT segmentation masks to isolate background pixels; sequences with mean background flow above 1.5~px are rejected. The remaining 13 sequences are processed as follows: (1) video stabilization via ECC alignment~\citep{evangelidis2008ecc} on background-only masks with Gaussian-smoothed cumulative transforms, center-cropped from $854{\times}480$ to $720{\times}480$; (2) trajectory extraction from per-frame median center-of-mass of GT segmentation masks, linearly interpolated through occlusion gaps and Gaussian-smoothed ($\sigma{=}2$); (3) depth estimation with Depth Anything V2~\citep{yang2024depth}. Sequences shorter than 49 frames are temporally interpolated via nearest-neighbor frame spreading.

\paragraph{Final dataset.}
The 13 scenes span 6 categories with 37 object trajectories total:

\begin{center}
\small
\begin{tabular}{llcl}
\toprule
Sequence & Category & Objects & Description \\
\midrule
bike-packing  & person     & 2 & person working on a bicycle \\
breakdance    & person     & 1 & breakdancer in a city square \\
crossing      & pedestrian & 3 & person, man, van at crosswalk \\
disc-jockey   & person     & 2 & DJ and turntable in dim room \\
dogs-jump     & animal     & 3 & 2 dogs, 1 person \\
drift-chicane & vehicle    & 1 & car on track \\
gold-fish     & animal     & 5 & goldfish in tank \\
judo          & sports     & 2 & two judokas grappling \\
planes-water  & vehicle    & 2 & planes on a wet runway \\
rallye        & vehicle    & 1 & rally car \\
schoolgirls   & pedestrian & 4 & 4 schoolgirls walking \\
sheep         & animal     & 5 & 2 sheep, 3 lambs in field \\
soccerball    & sports     & 1 & ball on grass \\
\bottomrule
\end{tabular}
\end{center}

All scenes use static cameras and are never seen during training. This dataset complements MOT17 (real-world but pedestrian-only) by testing generalization to animals, vehicles, sports, and other people-centric categories.

\subsection{Per-Video Captions}
\label{app:captions}

All baselines in Tables~\ref{tab:vin_results}--\ref{tab:ped_results} use per-video captions that describe the scene content. For the three ViN datasets (MoVi, Pool, Football), we use the ShareGPT4Video captions distributed with each dataset. For MOTSynth and MOT17, we generate scene-specific captions using Qwen2.5-VL-7B-Instruct~\citep{Qwen2.5-VL}, following the captioning methodology of Wan-Move~\citep{wanmove}.

This setup gives the baselines a textual advantage over our method. The per-video captions are produced by a video VLM (Qwen2.5-VL) that observes all frames of the clip, so they describe backgrounds, lighting, motion patterns, and even objects that are occluded in the first frame but become visible later. Our method follows the template from Eq.~\ref{eq:prompt} (``Scene where $o_0$ moves [traj$_0$] and $o_1$ moves [traj$_1$] \ldots''). The appearance encoder likewise operates on a single patch around each object's first-frame location, so when an object is occluded at the first frame its appearance token may encode the occluder rather than the target. Baselines therefore draw textual context from the entire video, while our conditioning is built exclusively from the first frame.

\noindent \textbf{Qwen2.5-VL prompt} \quad
We caption clips by sampling 16 frames per video and querying Qwen2.5-VL-7B-Instruct with the following instruction:
\begin{quote}
\ttfamily\footnotesize
Describe this video in detail. Focus on: (1) the scene setting and environment, (2) the camera perspective and any camera motion, (3) the main subjects and their appearance, (4) the motion and actions of each subject---direction, speed, interactions, (5) lighting conditions and visual style. Keep the description concise (2--4 sentences) and factual. Do not speculate about what happens outside the frame.
\end{quote}

\noindent \textbf{Example caption} \quad
For the MOT17 scene shown in the right panel of Figure~\ref{fig:qual_extra_mot17}, Qwen2.5-VL produces the following caption, used by all baselines on this scene:
\begin{quote}
\footnotesize\itshape
The video captures a scene set in an urban area with a paved walkway and a white car parked nearby. The camera is positioned at a high angle, providing a bird's-eye view of the scene. Three individuals are walking together; one person is dressed in black, another in a dark shirt and light-colored pants, and the third in a patterned outfit. They appear to be engaged in conversation as they walk side by side. The lighting suggests it might be daytime, with natural light illuminating the area. There is minimal camera movement, maintaining a steady focus on the group as they move across the frame.
\end{quote}
This caption conveys properties that require observing the full clip: the ``walking together'' group behaviour, the per-pedestrian clothing details, and scene elements such as the parked white car. For the same clip, our method's prompt is the template instantiation
\begin{quote}
\footnotesize\itshape
An outdoor scene with walking figures, where pedestrian walks [traj$_0$], pedestrian strolls [traj$_1$], pedestrian moves [traj$_2$].
\end{quote}
which conveys no scene, appearance, or interaction context beyond the first-frame VAE latent consumed by the appearance encoder.

\subsection{WaN Baseline Frame Count}
\label{app:81f_sanity}

ATI and Wan-Move are trained on 81-frame sequences, while our evaluation uses 49 frames. To reconcile this mismatch, we evaluate both baselines under two protocols: (1) generating 49 frames directly by patching a single hardcoded frame-count constant in each codebase, and (2) generating 81 frames at the native setting with trajectories padded to 81 frames by repeating the last position, then trimming the output to the first 49 frames for metrics. All other settings (checkpoints, seeds, captions, resolution) are identical between the two protocols.

Table~\ref{tab:81f_sanity} reports both variants across all six evaluation datasets. The results show no systematic advantage to either protocol: the 49-frame variant wins on PSNR on 5 of 6 datasets for both baselines, while the 81-frame variant is occasionally better on LPIPS, FVD, or EPE. Differences are generally small (PSNR within 1~dB, LPIPS within 0.03). We report the 49-frame results in the main tables.

\begin{table*}[h]
\caption{WaN-backbone baselines: 49-frame generation (reported in main tables) vs.\ native 81-frame generation trimmed to 49 frames. Trajectories for 81-frame runs are padded by repeating the last position. All other settings identical. Best per cell in \textbf{bold}.}
\label{tab:81f_sanity}
\centering
\footnotesize
\setlength{\tabcolsep}{2.5pt}
\begin{tabular}{ll c cccc cccc cccc}
\toprule
& & & \multicolumn{4}{c}{\textbf{MoVi}} & \multicolumn{4}{c}{\textbf{Pool}} & \multicolumn{4}{c}{\textbf{Football}} \\
\cmidrule(lr){4-7} \cmidrule(lr){8-11} \cmidrule(lr){12-15}
& Method & Frames & PSNR$\uparrow$ & LPIPS$\downarrow$ & FVD$\downarrow$ & EPE$\downarrow$
                   & PSNR$\uparrow$ & LPIPS$\downarrow$ & FVD$\downarrow$ & EPE$\downarrow$
                   & PSNR$\uparrow$ & LPIPS$\downarrow$ & FVD$\downarrow$ & EPE$\downarrow$ \\
\midrule
\multirow{2}{*}{\rotatebox{90}{\scriptsize ATI}}
& & 49 & \textbf{20.33} & .355 & 37.50 & \textbf{2.01} & \textbf{21.13} & .303 & 57.78 & \textbf{0.37} & \textbf{19.68} & \textbf{.250} & 46.32 & \textbf{1.33} \\
& & 81 & 20.02 & \textbf{.313} & \textbf{36.20} & 2.41 & 20.78 & \textbf{.284} & \textbf{51.62} & 0.46 & 19.57 & .259 & \textbf{44.47} & 1.73 \\
\midrule
\multirow{2}{*}{\rotatebox{90}{\scriptsize WM}}
& & 49 & \textbf{19.73} & \textbf{.331} & 40.38 & \textbf{2.18} & \textbf{22.94} & .191 & \textbf{33.98} & 0.29 & \textbf{18.36} & \textbf{.296} & 69.84 & \textbf{3.81} \\
& & 81 & 19.17 & .334 & \textbf{39.82} & 2.65 & 22.05 & \textbf{.186} & 38.10 & \textbf{0.27} & 17.94 & .311 & \textbf{68.80} & 4.57 \\
\bottomrule
\\[4pt]
\toprule
& & & \multicolumn{4}{c}{\textbf{MOTSynth}} & \multicolumn{4}{c}{\textbf{MOT17}} & \multicolumn{4}{c}{\textbf{DAVIS}} \\
\cmidrule(lr){4-7} \cmidrule(lr){8-11} \cmidrule(lr){12-15}
& Method & Frames & PSNR$\uparrow$ & LPIPS$\downarrow$ & FVD$\downarrow$ & EPE$\downarrow$
                   & PSNR$\uparrow$ & LPIPS$\downarrow$ & FVD$\downarrow$ & EPE$\downarrow$
                   & PSNR$\uparrow$ & LPIPS$\downarrow$ & FVD$\downarrow$ & EPE$\downarrow$ \\
\midrule
\multirow{2}{*}{\rotatebox{90}{\scriptsize ATI}}
& & 49 & \textbf{17.76} & \textbf{.395} & \textbf{62.32} & \textbf{8.61} & \textbf{17.05} & .416 & \textbf{65.35} & \textbf{4.28} & 14.68 & .443 & \textbf{104.90} & 14.02 \\
& & 81 & 17.32 & .413 & 67.13 & 9.88 & 16.70 & \textbf{.406} & 70.33 & 4.73 & \textbf{14.96} & \textbf{.425} & 105.71 & \textbf{11.15} \\
\midrule
\multirow{2}{*}{\rotatebox{90}{\scriptsize WM}}
& & 49 & \textbf{18.29} & \textbf{.354} & \textbf{54.29} & \textbf{3.60} & 16.27 & .405 & 60.20 & 3.41 & \textbf{15.86} & \textbf{.352} & \textbf{85.51} & 4.96 \\
& & 81 & 17.78 & .383 & 61.28 & 3.82 & \textbf{16.55} & \textbf{.379} & \textbf{58.70} & \textbf{3.33} & 15.45 & .355 & 88.15 & \textbf{4.22} \\
\bottomrule
\end{tabular}
\end{table*}

\clearpage

\section{MOTSynth-Multi}
\label{app:scaling}
\label{app:msm}

\noindent \textbf{Construction} \quad MOTSynth-Multi (Sec.~\ref{sec:scaling}) uses the synthetically trained models of Sec.~\ref{sec:experiments} and is drawn exclusively from held-out MOTSynth test scenes, none of which is used for training. Each bin contains 20 clips from 20 distinct scenes, with exactly 5, 10, or 15 controlled pedestrians, or 17--20 controlled pedestrians in the densest bin, where the number of eligible held-out scenes is the limiting factor. Pedestrians are counted within the $720 \!\times\! 480$ crop, and every controlled pedestrian is visible in the first frame. Clips are prepared and captioned exactly as the MOTSynth evaluation set of Sec.~\ref{sec:setup}.

\begin{figure*}[h]

\vspace{-8pt}
\centering
\setlength{\tabcolsep}{0.5pt}
\renewcommand{\arraystretch}{0.3}
\newcommand{\qimg}[1]{\includegraphics[width=0.155\textwidth]{#1}}
\newcommand{\framesarrow}[1]{%
  \begin{tikzpicture}[baseline=0pt]
    \draw[->, line width=0.5pt] (0,0) -- (0.465\textwidth, 0)
      node[midway, fill=white, inner sep=1pt] {\scriptsize #1};
  \end{tikzpicture}%
}
\begin{tabular}{@{}c ccc @{\hspace{6pt}} c ccc@{}}
& \multicolumn{3}{c}{\framesarrow{MOTSynth-Multi frames (20-pedestrian bin)}} & & \multicolumn{3}{c}{\framesarrow{MOTSynth-Multi frames (20-pedestrian bin)}} \\[1pt]
\rotatebox{90}{\scriptsize\hspace{8pt}GT} &
\qimg{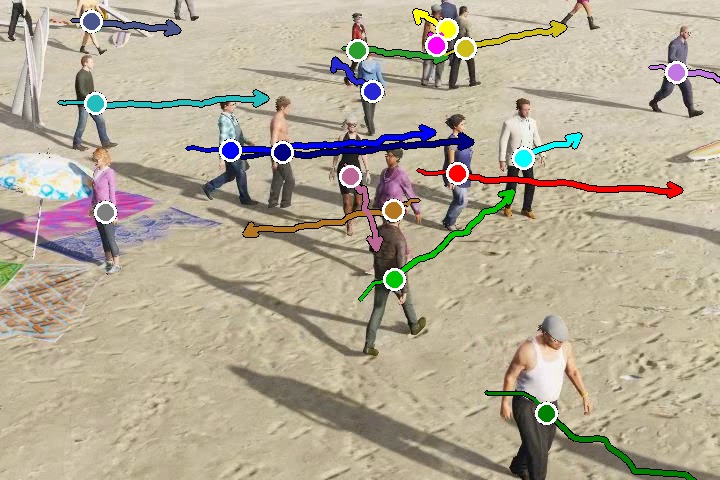} &
\qimg{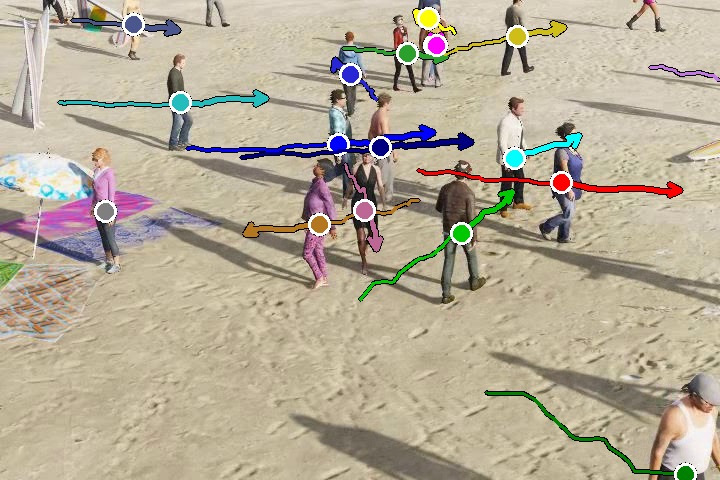} &
\qimg{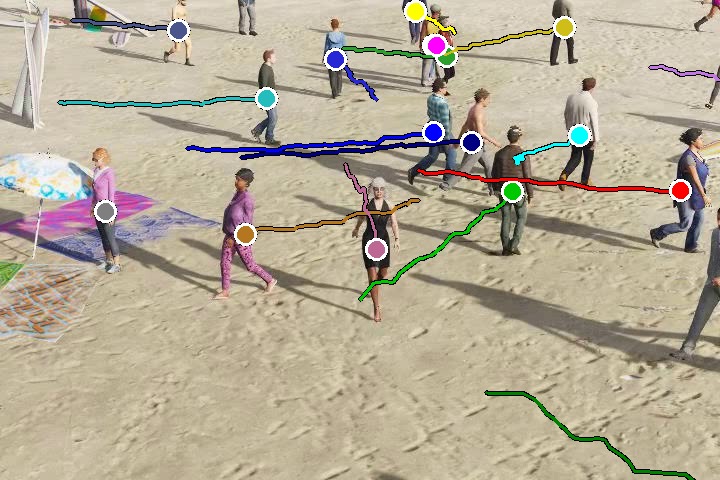} &
\rotatebox{90}{\scriptsize\hspace{8pt}GT} &
\qimg{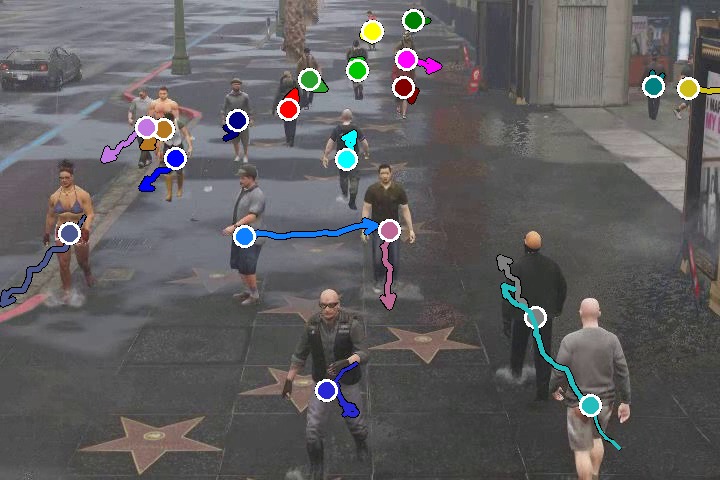} &
\qimg{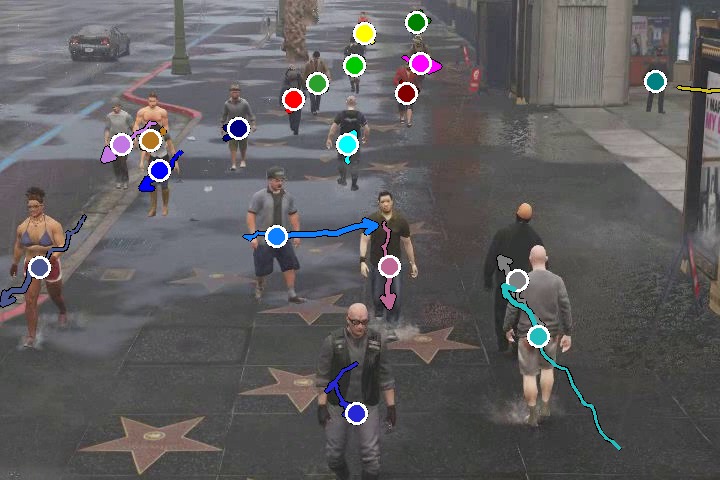} &
\qimg{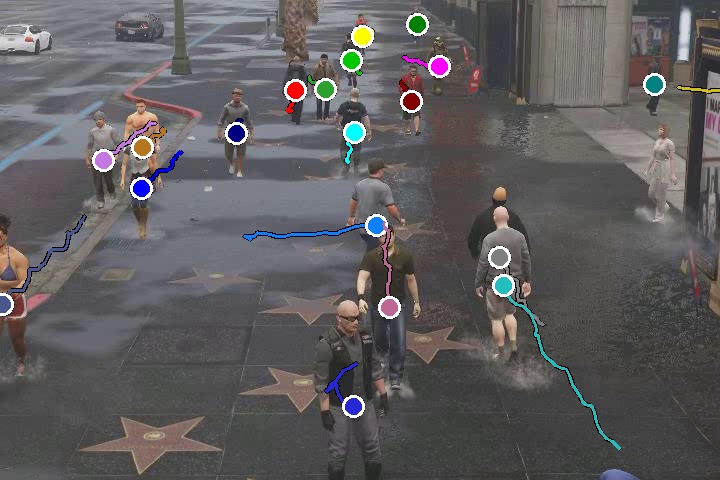} \\[0.5pt]
\rotatebox{90}{\scriptsize\hspace{4pt}Ours-CogV} &
\qimg{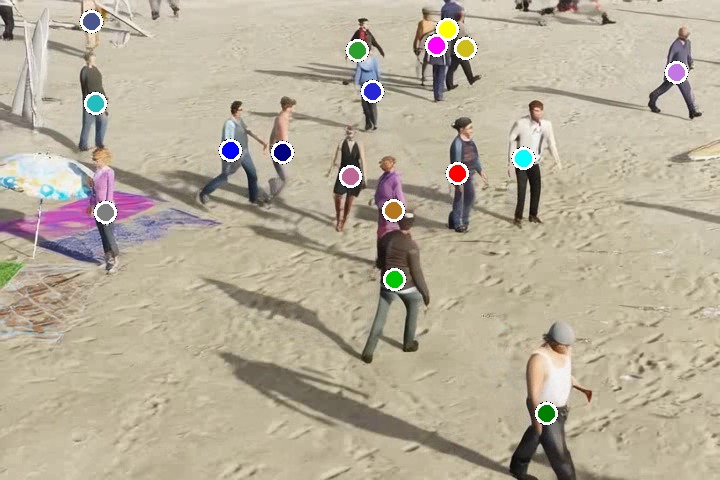} &
\qimg{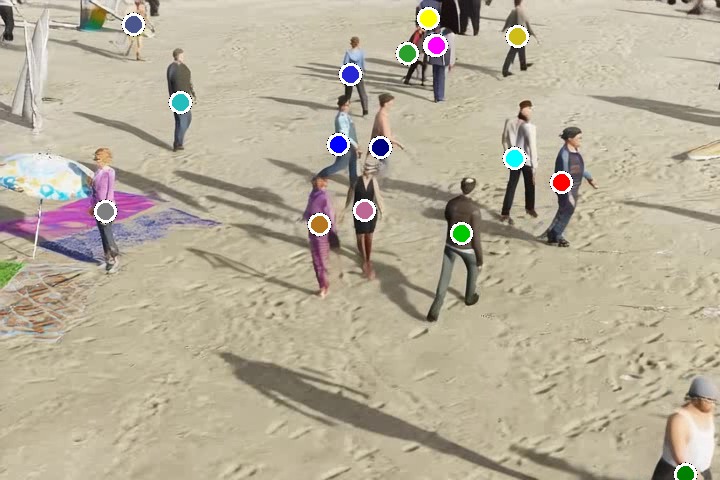} &
\qimg{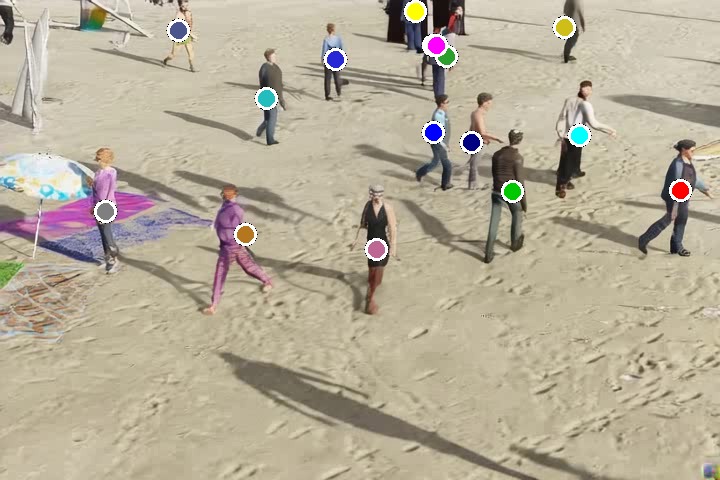} &
\rotatebox{90}{\scriptsize\hspace{4pt}Ours-WaN} &
\qimg{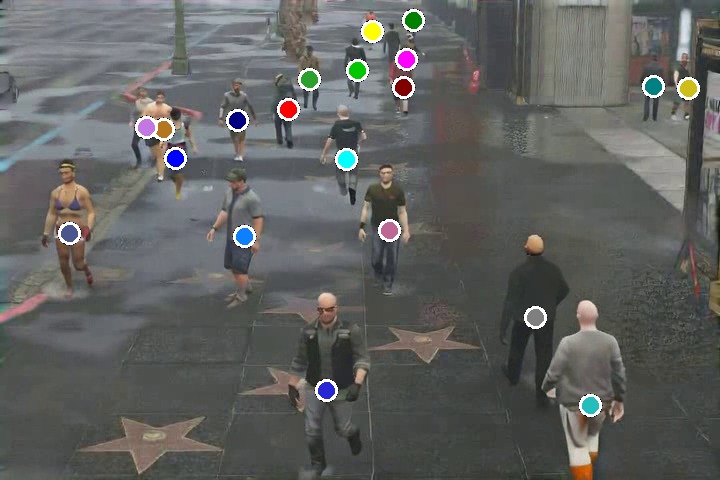} &
\qimg{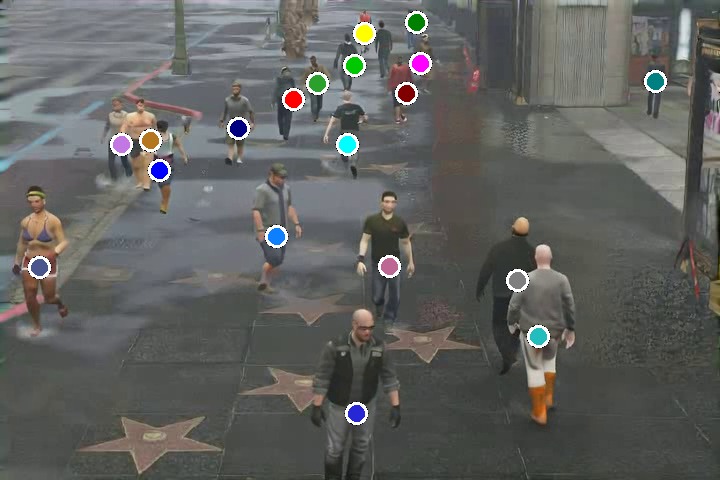} &
\qimg{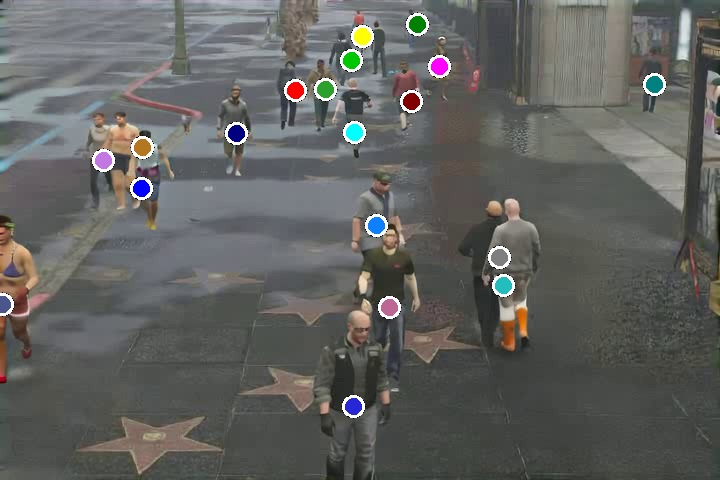} \\[0.5pt]
\rotatebox{90}{\scriptsize\hspace{4pt}MagicMotion} &
\qimg{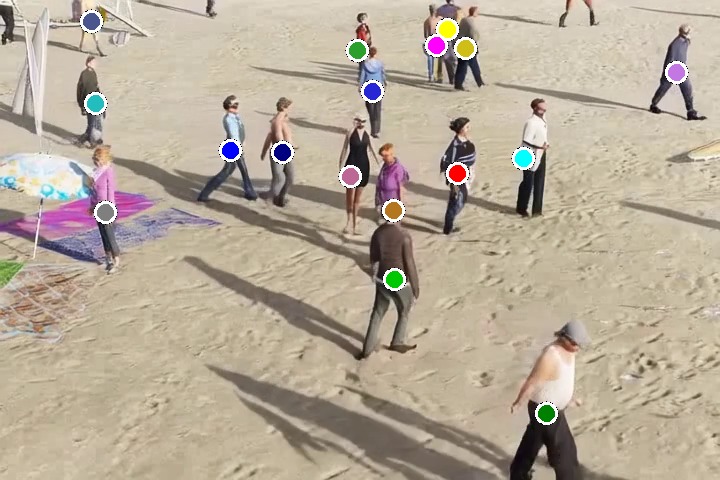} &
\qimg{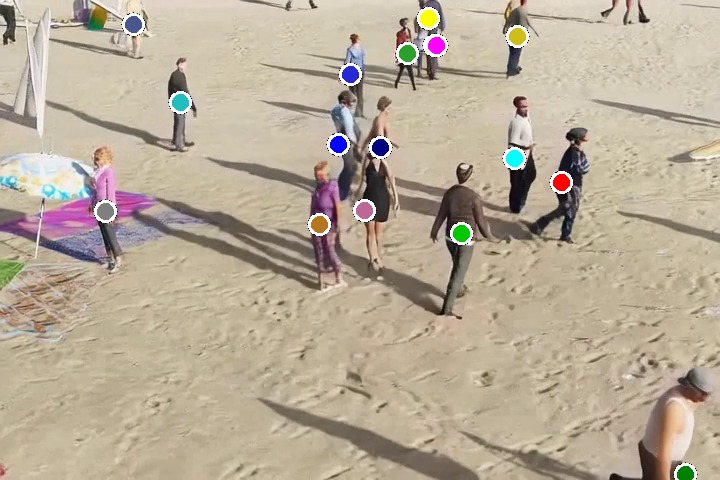} &
\qimg{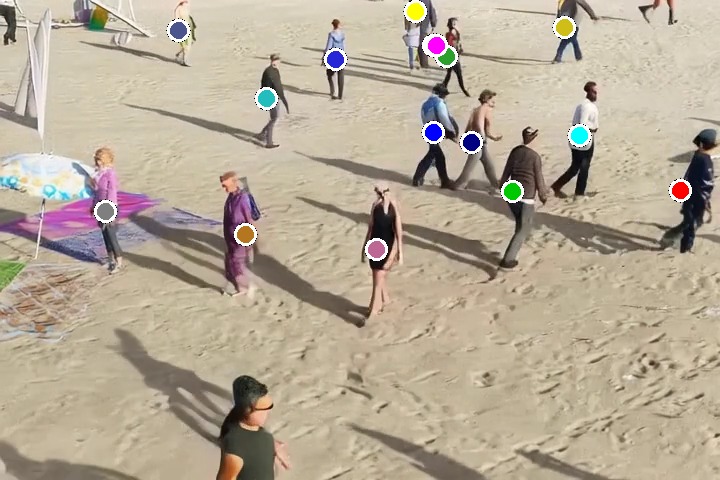} &
\rotatebox{90}{\scriptsize\hspace{8pt}ATI} &
\qimg{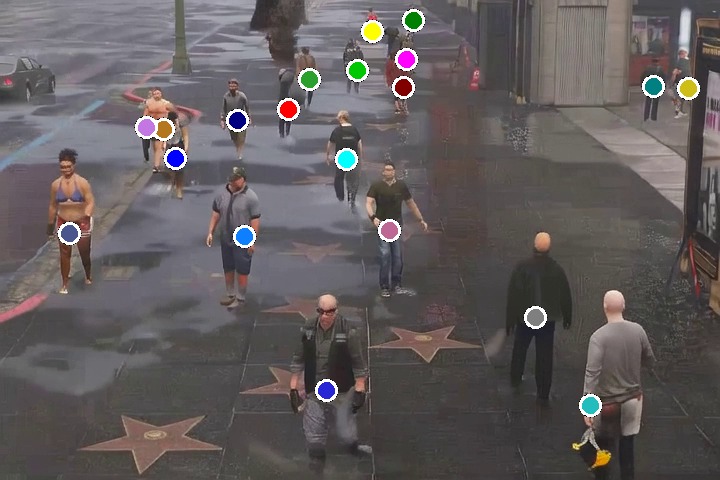} &
\qimg{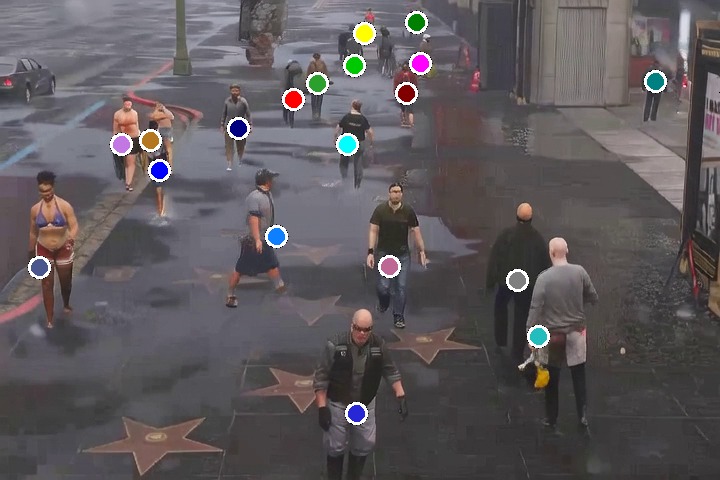} &
\qimg{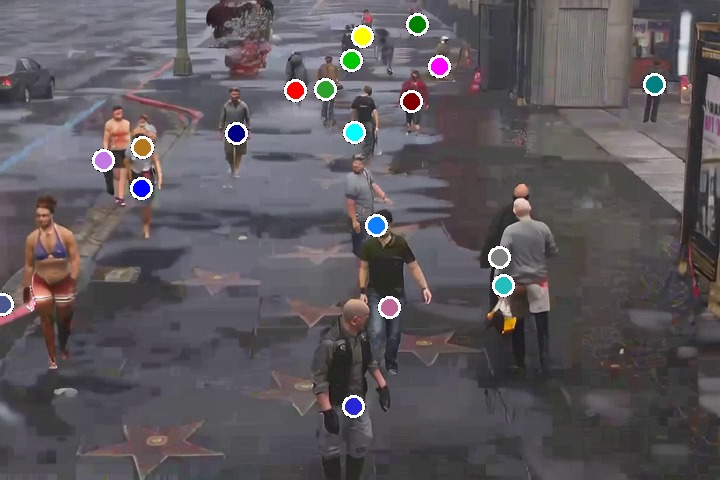} \\[0.5pt]
\rotatebox{90}{\scriptsize\hspace{8pt}Tora} &
\qimg{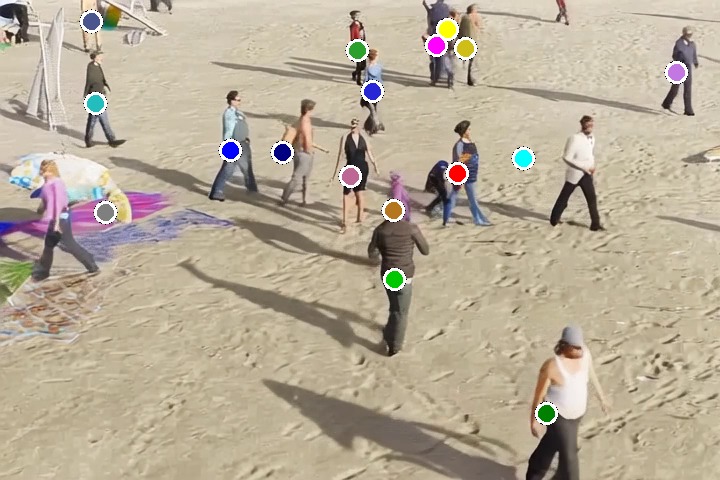} &
\qimg{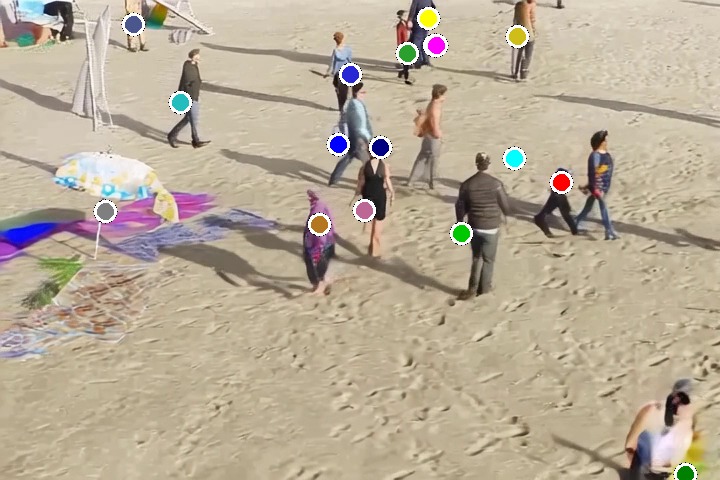} &
\qimg{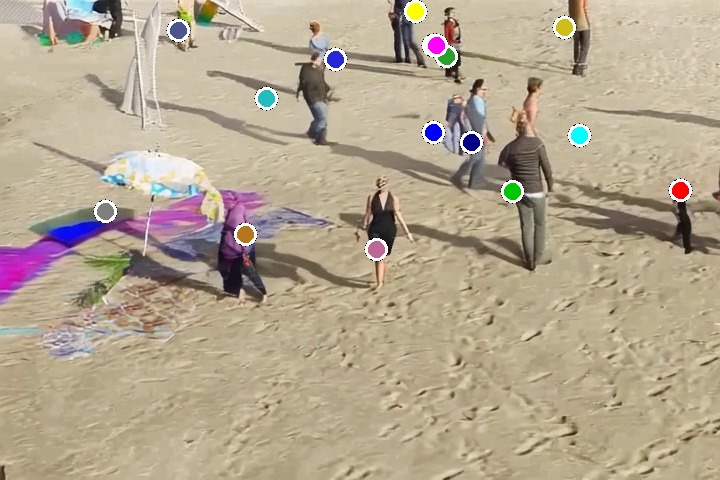} &
\rotatebox{90}{\scriptsize\hspace{4pt}Wan-Move} &
\qimg{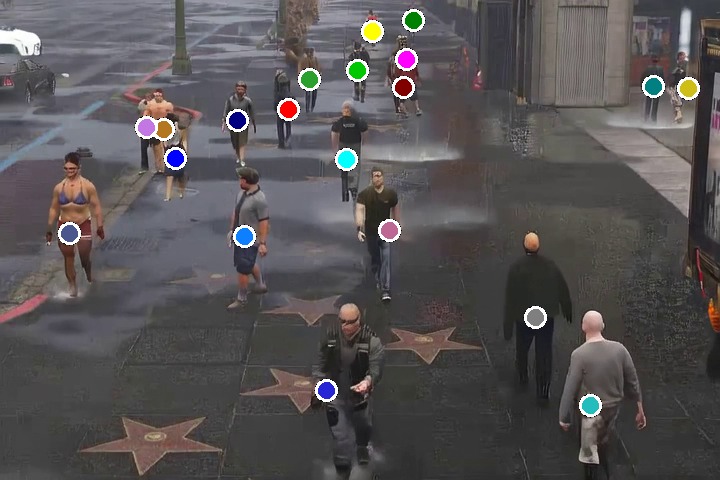} &
\qimg{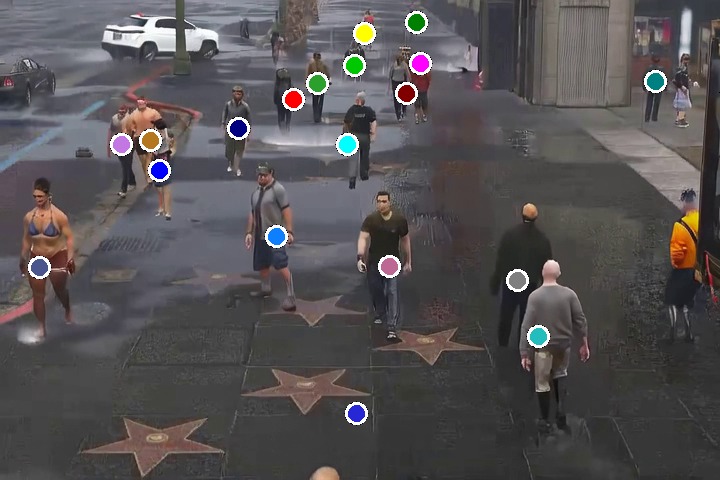} &
\qimg{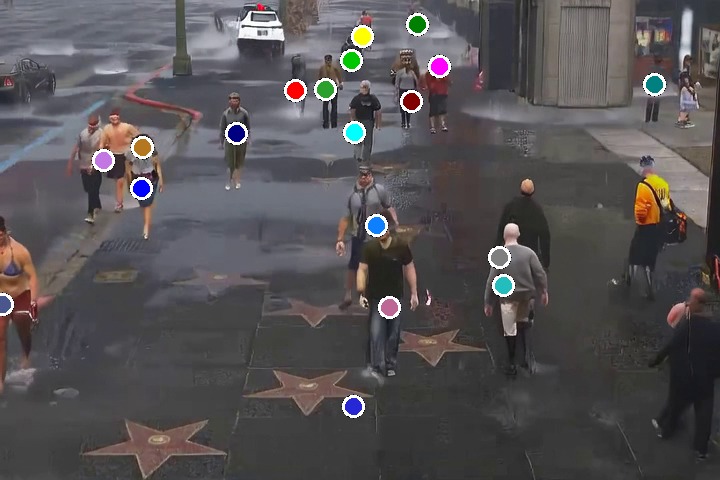} \\
\end{tabular}
\caption{\textbf{Qualitative comparison on the 20-pedestrian bin of MOTSynth-Multi with all baselines} (18 and 20 controlled pedestrians). Left: CogVideoX-5B against MagicMotion and Tora. Right: WaN~2.1-14B against ATI and Wan-Move. MagicMotion inserts a pedestrian that is not in the scene at the bottom of the last frame, and Tora loses several pedestrians, for example at the gray, teal and cyan dots. Wan-Move drops the blue-dot pedestrian in the foreground and inserts a pedestrian in an orange jacket on the right, and ATI produces a corrupted region at the top of the last frame. Our method keeps every pedestrian on its prescribed trajectory with both backbones.}
\label{fig:qual_msm}

\vspace{-14pt}
\end{figure*}

\begin{figure*}[t]

\vspace{-8pt}
\centering
\setlength{\tabcolsep}{0.5pt}
\renewcommand{\arraystretch}{0.3}
\newcommand{\qimg}[1]{\includegraphics[width=0.155\textwidth]{#1}}
\newcommand{\framesarrow}[1]{%
  \begin{tikzpicture}[baseline=0pt]
    \draw[->, line width=0.5pt] (0,0) -- (0.465\textwidth, 0)
      node[midway, fill=white, inner sep=1pt] {\scriptsize #1};
  \end{tikzpicture}%
}
\begin{tabular}{@{}c ccc @{\hspace{6pt}} c ccc@{}}
& \multicolumn{3}{c}{\framesarrow{MOTSynth-Multi frames (20-pedestrian bin)}} & & \multicolumn{3}{c}{\framesarrow{MOTSynth-Multi frames (15 pedestrians)}} \\[1pt]
\rotatebox{90}{\scriptsize\hspace{8pt}GT} &
\qimg{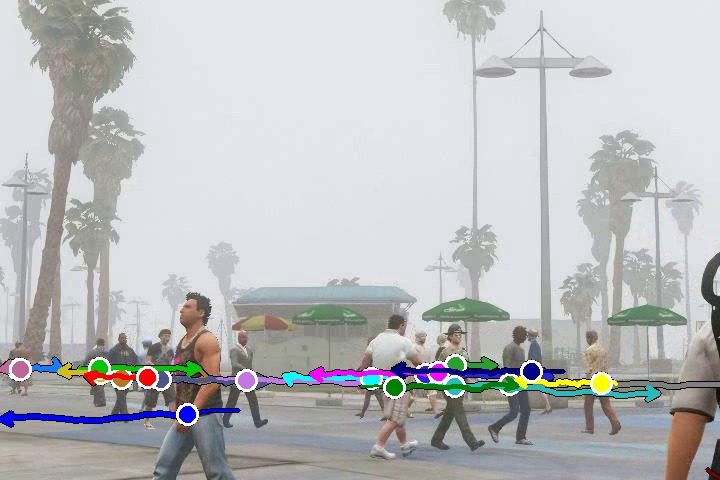} &
\qimg{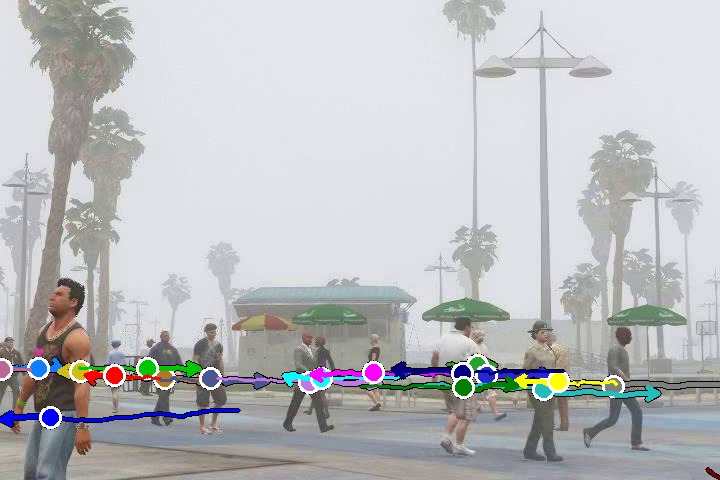} &
\qimg{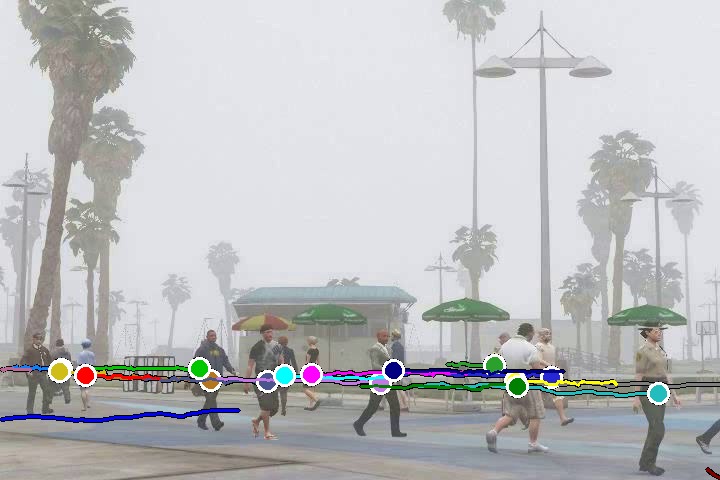} &
\rotatebox{90}{\scriptsize\hspace{8pt}GT} &
\qimg{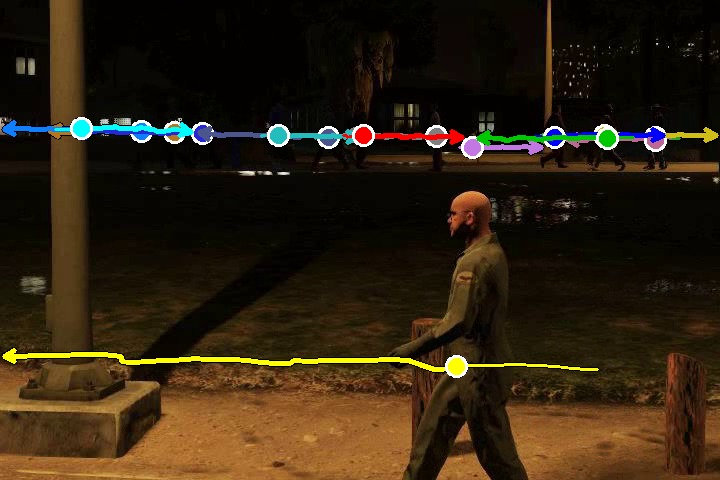} &
\qimg{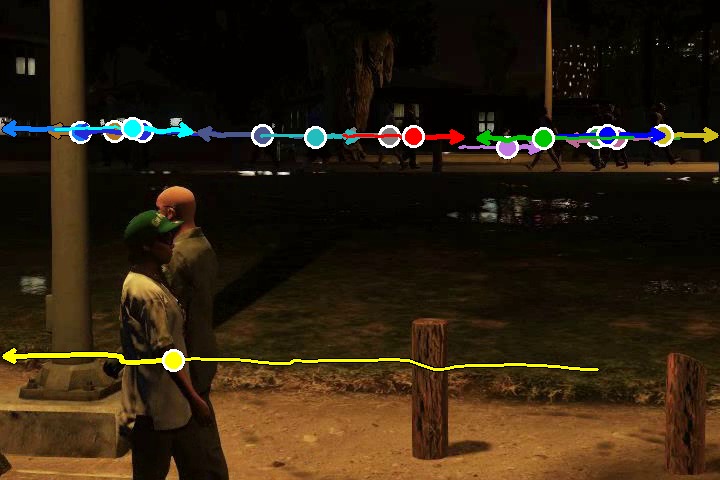} &
\qimg{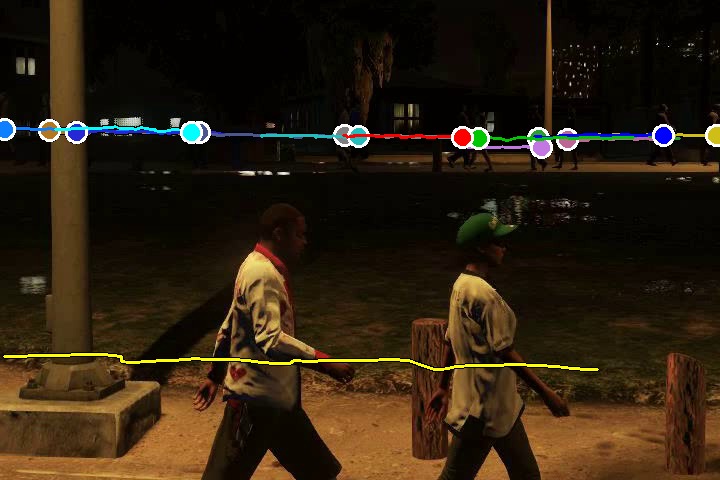} \\[0.5pt]
\rotatebox{90}{\scriptsize\hspace{4pt}Ours-CogV} &
\qimg{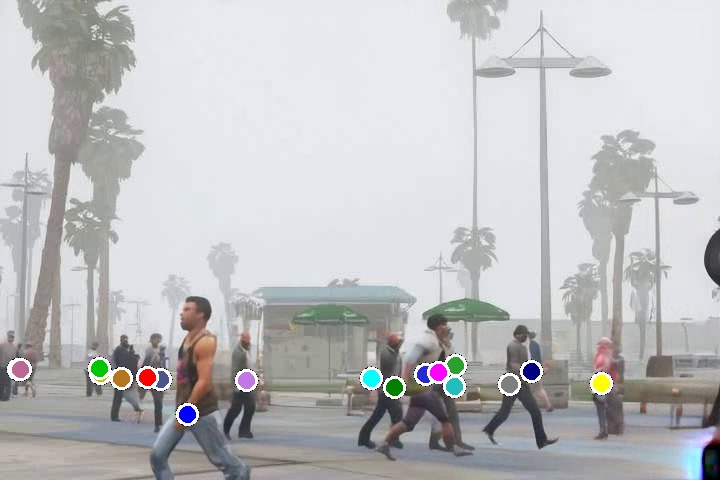} &
\qimg{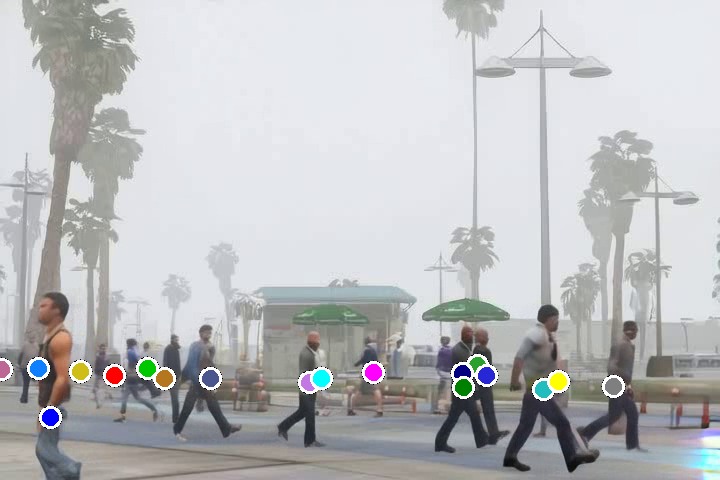} &
\qimg{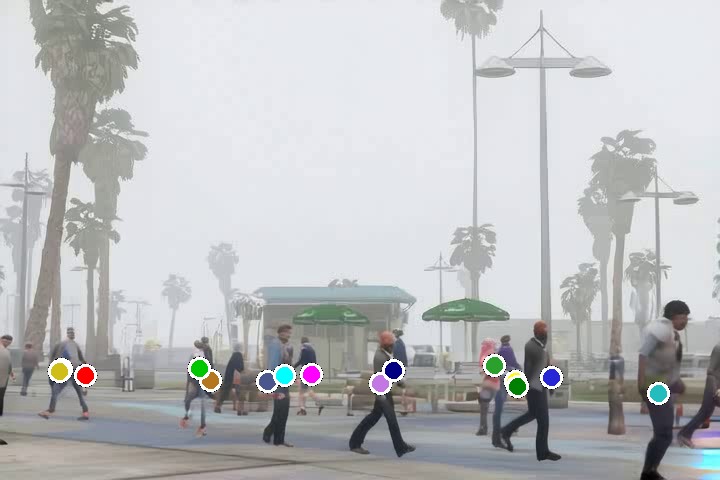} &
\rotatebox{90}{\scriptsize\hspace{4pt}Ours-WaN} &
\qimg{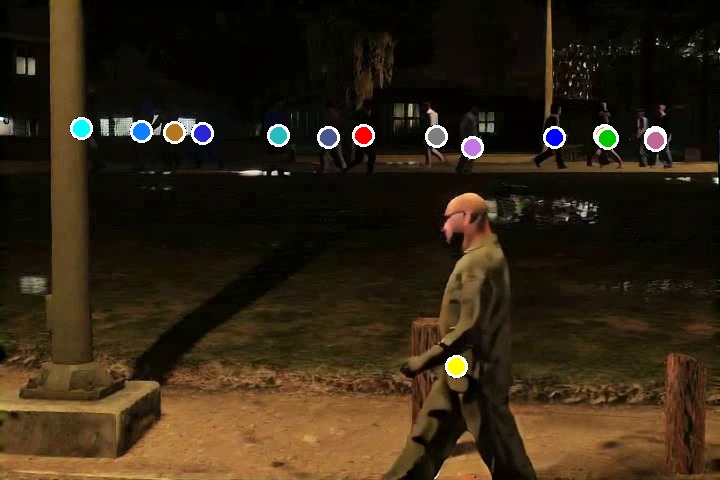} &
\qimg{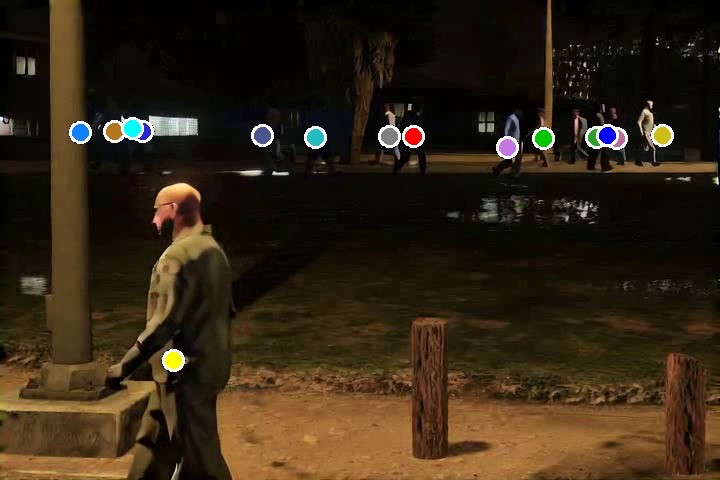} &
\qimg{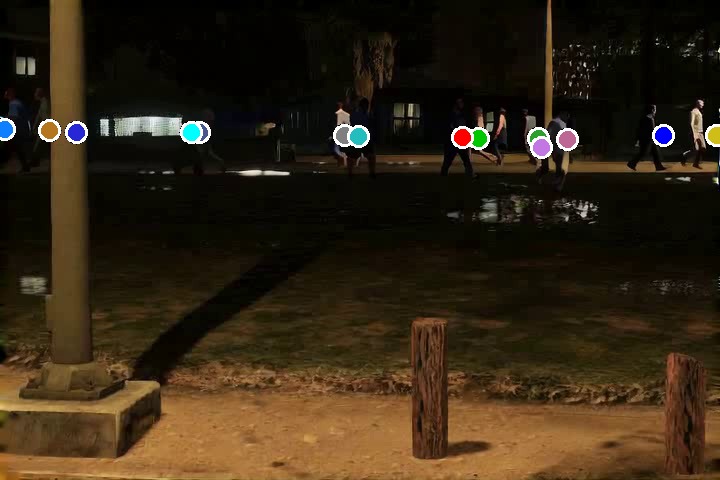} \\[0.5pt]
\rotatebox{90}{\scriptsize\hspace{4pt}MagicMotion} &
\qimg{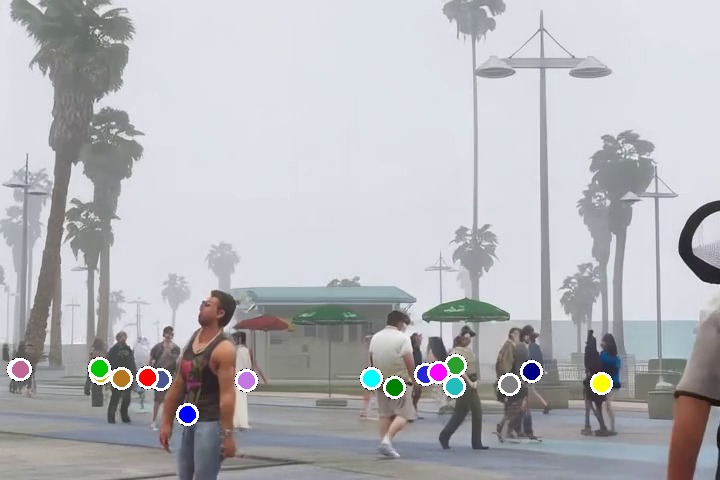} &
\qimg{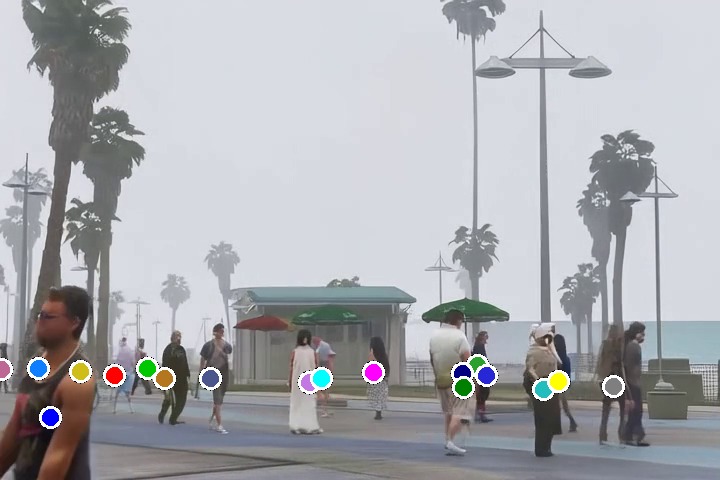} &
\qimg{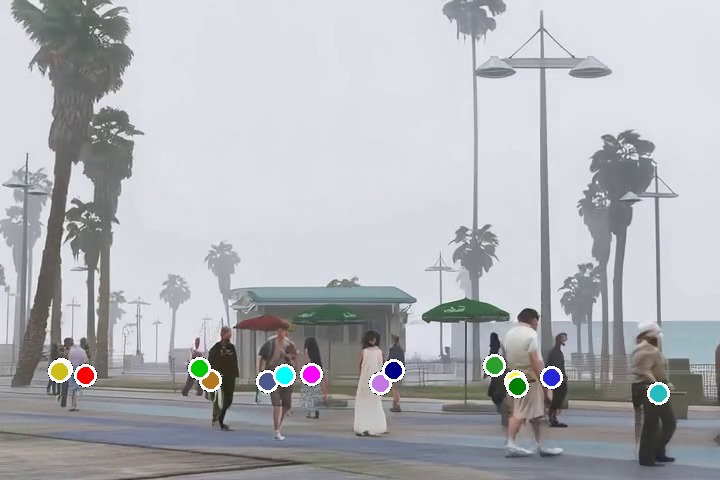} &
\rotatebox{90}{\scriptsize\hspace{8pt}ATI} &
\qimg{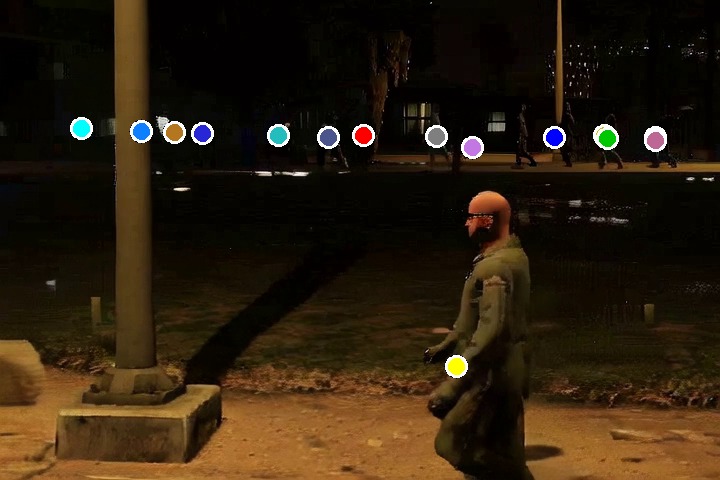} &
\qimg{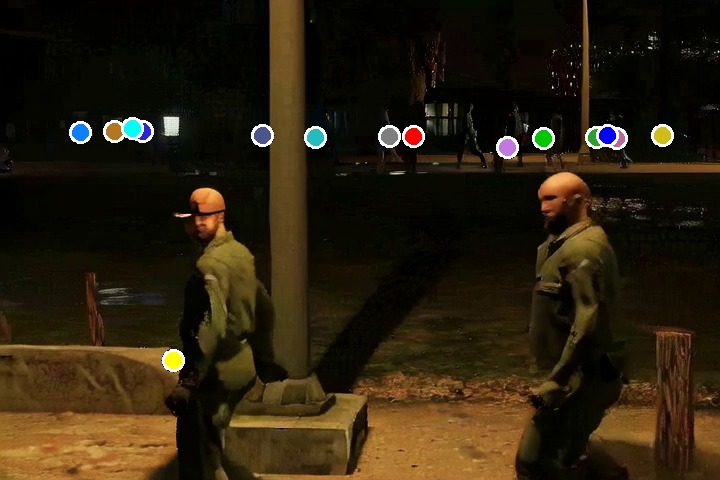} &
\qimg{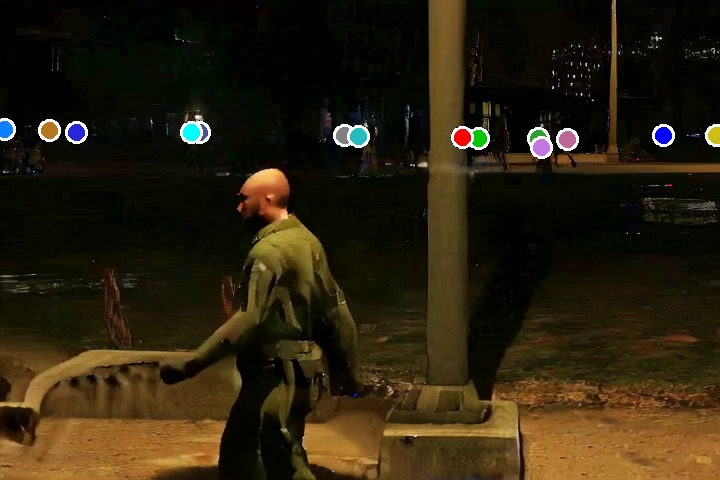} \\[0.5pt]
\rotatebox{90}{\scriptsize\hspace{8pt}Tora} &
\qimg{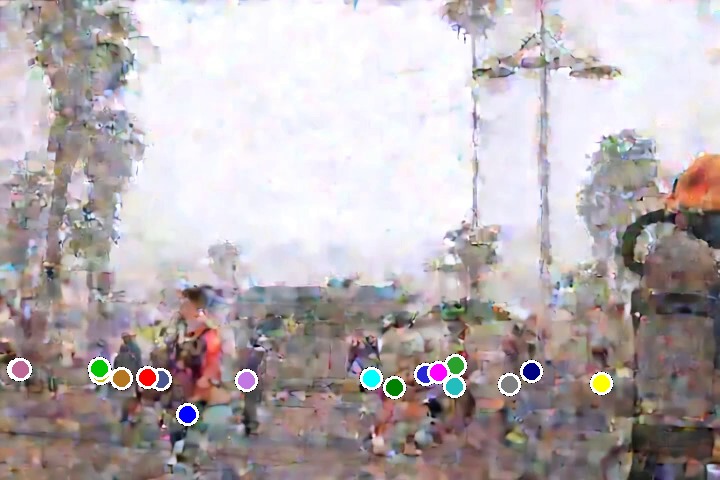} &
\qimg{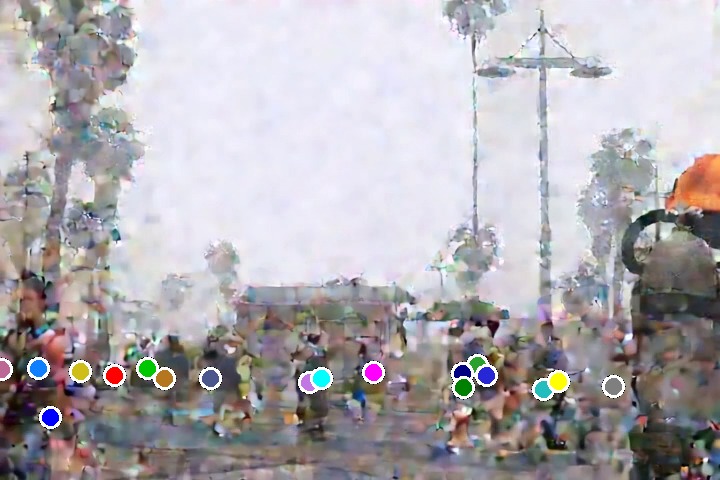} &
\qimg{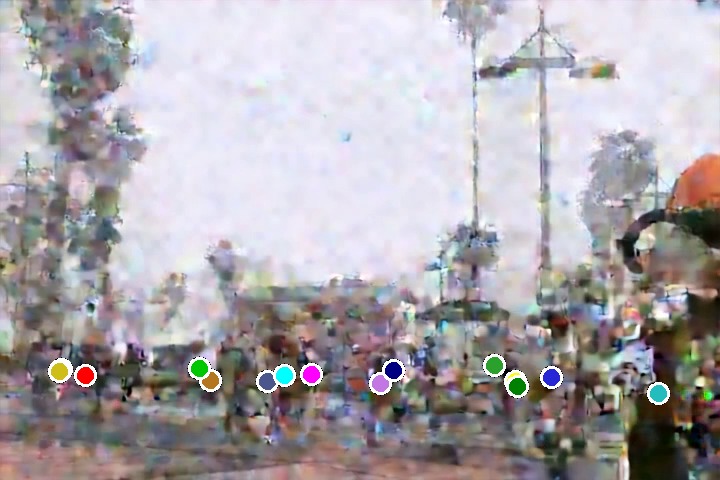} &
\rotatebox{90}{\scriptsize\hspace{4pt}Wan-Move} &
\qimg{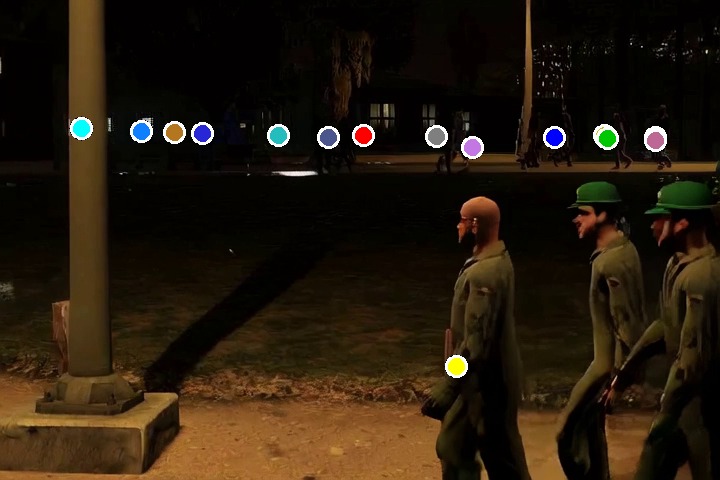} &
\qimg{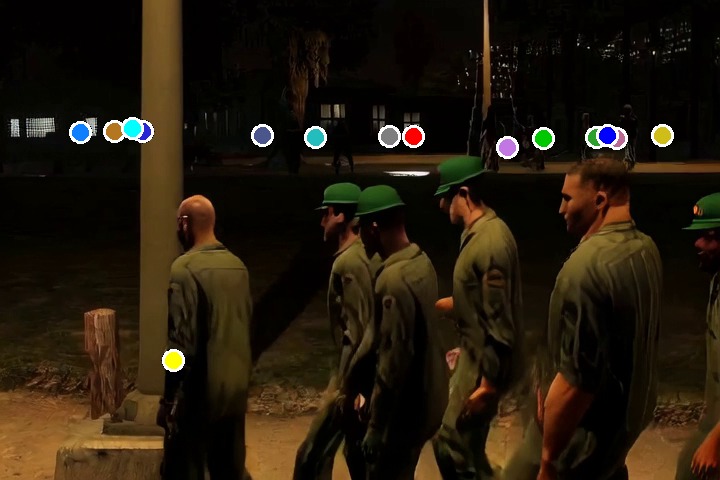} &
\qimg{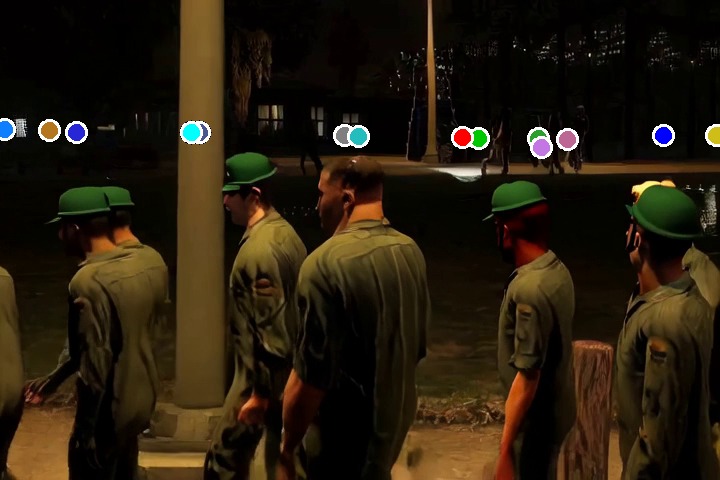} \\
\end{tabular}
\caption{\textbf{Additional qualitative comparison on MOTSynth-Multi.} Left: CogVideoX-5B on a scene from the 20-pedestrian bin (19 pedestrians). Right: WaN~2.1-14B on a 15-pedestrian scene. Tora degrades the image quality of the crowded promenade, while MagicMotion and our method preserve it. On the right, the tracked pedestrian in the foreground walks out of the frame; ATI keeps him in view, and Wan-Move fills the scene with copies of him that do not correspond to any trajectory, while our method lets him exit on his prescribed path.}
\label{fig:qual_msm_extra}

\vspace{-14pt}
\end{figure*}

\noindent \textbf{Results} \quad Table~\ref{tab:scaling} reports all four metrics. EPE decreases for every method as the count grows, because pedestrians in dense crowds move less, which likely also contributes to the rising PSNR of the baselines. With CogVideoX, MagicMotion attains a lower FVD in the densest bin and a lower EPE in the sparsest. Figures~\ref{fig:qual_msm} and~\ref{fig:qual_msm_extra} show scenes from the two densest bins.

\clearpage

\section{Extended Ablations}
\label{app:ablation_extra}

\subsection{Additive and Leave-One-Out Ablation on Pool and Football}
\label{app:lattice}

Table~\ref{tab:lattice_pf} completes Table~\ref{tab:ablation_combined} with the two remaining in-distribution datasets. On Pool, the full model again achieves the best PSNR and EPE, and the trajectory token is unusually informative: on its own it already reaches 31.60\,dB, since billiard motion is fully described by the image-plane trajectories of a few balls. On Football, every configuration with attention localization lies within 1.5\,dB PSNR of the others and the bare backbone already reaches an EPE of 0.05, so this dataset discriminates little between the components, and attention localization alone gives the best LPIPS and FVD.

\begin{table}[h]
\caption{\textbf{Additive and leave-one-out ablation on Pool and Football} (WaN~2.1-14B, configurations as in Table~\ref{tab:ablation_combined}). Best in bold.}
\label{tab:lattice_pf}
\centering
\footnotesize
\setlength{\tabcolsep}{3.2pt}
\begin{tabular}{rl cccc cccc cccc}
\toprule
& & & & & & \multicolumn{4}{c}{\textbf{Pool}} & \multicolumn{4}{c}{\textbf{Football}} \\
\cmidrule(lr){7-10} \cmidrule(lr){11-14}
\# & Configuration & A & T & P & d & PSNR$\uparrow$ & LPIPS$\downarrow$ & FVD$\downarrow$ & EPE$\downarrow$ & PSNR$\uparrow$ & LPIPS$\downarrow$ & FVD$\downarrow$ & EPE$\downarrow$ \\
\midrule
1 & Backbone only &  &  &  &  & 27.01 & .158 & 48.66 & 0.66 & 23.55 & .145 & 36.22 & 0.05 \\
2 & Attention only & \cmark &  &  &  & 29.69 & .065 & 43.92 & 0.09 & 24.66 & \textbf{.133} & \textbf{22.12} & \textbf{0.04} \\
\midrule
3 & w/o trajectory token & \cmark &  & \cmark &  & 29.20 & .101 & 45.91 & 0.11 & 23.33 & .143 & 31.03 & 0.05 \\
4 & Trajectory token only &  & \cmark &  & \cmark & 31.60 & .075 & \textbf{27.75} & 0.10 & 23.78 & .159 & 31.64 & 0.06 \\
\midrule
5 & w/o attention localization &  & \cmark & \cmark & \cmark & 29.46 & .113 & 37.47 & 0.17 & 23.02 & .192 & 58.19 & 0.15 \\
6 & w/o appearance token & \cmark & \cmark &  & \cmark & 32.45 & .057 & 38.36 & \textbf{0.08} & \textbf{24.76} & .143 & 26.02 & \textbf{0.04} \\
7 & w/o depth channel & \cmark & \cmark & \cmark &  & 32.37 & .054 & 38.25 & 0.09 & 24.32 & .146 & 27.67 & 0.06 \\
\midrule
8 & \textbf{Ours} & \cmark & \cmark & \cmark & \cmark & \textbf{33.07} & \textbf{.049} & 31.34 & \textbf{0.08} & 24.27 & .141 & 24.24 & 0.05 \\
\bottomrule
\end{tabular}
\end{table}

\subsection{Design Choices}
\label{app:design}

Table~\ref{tab:design} ablates two design choices of the full model. First, we drop the bounding-box loss of Eq.~\ref{eq:diffusion_loss} by setting $\alpha{=}0$. This lowers PSNR by 0.2--1.0\,dB and increases FVD and EPE on MoVi, MOTSynth, and Pool. On Football, where all configurations perform similarly (Appendix~\ref{app:lattice}), PSNR changes by less than 0.2\,dB. Second, we insert the trajectory token after the text encoder, where the appearance token is inserted, instead of before it. This degrades every dataset sharply, for example by 4.6\,dB PSNR on MoVi and 9.8\,dB on Pool, with FVD two to four times higher. The trajectory encoder is pretrained so that its embeddings remain informative after passing through the text encoder (Sec.~\ref{sec:traj_encoding}), so the token must be inserted before it.

\begin{table}[h]
\caption{\textbf{Design choices} (WaN~2.1-14B) on the four in-distribution datasets. Best per dataset in bold.}
\label{tab:design}
\centering
\footnotesize
\setlength{\tabcolsep}{2.5pt}
\begin{tabular}{l cccc cccc}
\toprule
\multicolumn{1}{c}{} & \multicolumn{4}{c}{\textbf{MoVi}} & \multicolumn{4}{c}{\textbf{MOTSynth}} \\
\cmidrule(lr){2-5} \cmidrule(lr){6-9}
Configuration & PSNR$\uparrow$ & LPIPS$\downarrow$ & FVD$\downarrow$ & EPE$\downarrow$ & PSNR$\uparrow$ & LPIPS$\downarrow$ & FVD$\downarrow$ & EPE$\downarrow$ \\
\midrule
Bounding-box loss $\alpha{=}0$ & 28.21 & .147 & 29.23 & 0.63 & 21.06 & .252 & 45.54 & 2.34 \\
Trajectory token after text encoder & 24.60 & .242 & 76.88 & 1.69 & 17.39 & .534 & 132.4 & 3.37 \\
\textbf{Ours} & \textbf{29.24} & \textbf{.131} & \textbf{23.39} & \textbf{0.57} & \textbf{21.39} & \textbf{.244} & \textbf{41.43} & \textbf{2.11} \\
\midrule
\multicolumn{1}{c}{} & \multicolumn{4}{c}{\textbf{Pool}} & \multicolumn{4}{c}{\textbf{Football}} \\
\cmidrule(lr){2-5} \cmidrule(lr){6-9}
Configuration & PSNR$\uparrow$ & LPIPS$\downarrow$ & FVD$\downarrow$ & EPE$\downarrow$ & PSNR$\uparrow$ & LPIPS$\downarrow$ & FVD$\downarrow$ & EPE$\downarrow$ \\
\midrule
Bounding-box loss $\alpha{=}0$ & 32.86 & \textbf{.046} & 32.70 & 0.09 & \textbf{24.44} & \textbf{.136} & 29.04 & \textbf{0.04} \\
Trajectory token after text encoder & 23.27 & .159 & 91.23 & 0.22 & 17.86 & .358 & 97.82 & 0.33 \\
\textbf{Ours} & \textbf{33.07} & .049 & \textbf{31.34} & \textbf{0.08} & 24.27 & .141 & \textbf{24.24} & 0.05 \\
\bottomrule
\end{tabular}
\end{table}

\subsection{Seed Variance of the Leave-One-Out Ablation}
\label{app:seeds}

Each cell of Table~\ref{tab:ablation_combined} is a single-seed estimate. To verify that the component effects exceed sampling noise, we repeat the leave-one-out configurations on MoVi and MOTSynth with nine seeds, the reported seed and eight new ones, changing only the initial noise (Table~\ref{tab:seeds}). The reported single-seed values lie within 2.1 standard deviations of the nine-seed means. On MoVi, every component effect exceeds the seed standard deviation at least fivefold on every metric. Even the smallest PSNR gap, from removing the appearance token, is 0.75\,dB against a standard deviation of at most 0.10\,dB. On MOTSynth, removing attention localization or the appearance token lowers PSNR well beyond seed noise, by 2.1 and 0.7\,dB. Removing the trajectory token or the depth channel shows mainly in EPE, which rises from 2.15 to 2.49 and 2.32, consistent with their role of encoding motion and depth. The PSNR effect of the trajectory token on MOTSynth (0.08\,dB) is within seed noise.

\begin{table}[h]
\caption{\textbf{Leave-one-out ablation over nine seeds} (WaN~2.1-14B), mean\,$\pm$\,standard deviation. Best in bold.}
\label{tab:seeds}
\centering
\scriptsize
\setlength{\tabcolsep}{2.4pt}
\begin{tabular}{l cccc cccc}
\toprule
& \multicolumn{4}{c}{\textbf{MoVi}} & \multicolumn{4}{c}{\textbf{MOTSynth}} \\
\cmidrule(lr){2-5} \cmidrule(lr){6-9}
Configuration & PSNR$\uparrow$ & LPIPS$\downarrow$ & FVD$\downarrow$ & EPE$\downarrow$ & PSNR$\uparrow$ & LPIPS$\downarrow$ & FVD$\downarrow$ & EPE$\downarrow$ \\
\midrule
\textbf{Ours} & \textbf{29.31}{\tiny$\pm$.10} & \textbf{.130}{\tiny$\pm$.001} & \textbf{23.47}{\tiny$\pm$.26} & \textbf{0.56}{\tiny$\pm$.01} & \textbf{21.46}{\tiny$\pm$.08} & \textbf{.244}{\tiny$\pm$.002} & \textbf{41.03}{\tiny$\pm$.69} & \textbf{2.15}{\tiny$\pm$.05} \\
w/o trajectory token & 27.92{\tiny$\pm$.07} & .144{\tiny$\pm$.002} & 26.65{\tiny$\pm$.42} & 0.66{\tiny$\pm$.01} & 21.38{\tiny$\pm$.06} & .254{\tiny$\pm$.002} & 42.58{\tiny$\pm$.71} & 2.49{\tiny$\pm$.10} \\
w/o appearance token & 28.56{\tiny$\pm$.09} & .144{\tiny$\pm$.001} & 27.87{\tiny$\pm$.40} & 0.66{\tiny$\pm$.01} & 20.78{\tiny$\pm$.09} & .258{\tiny$\pm$.002} & 45.34{\tiny$\pm$.52} & 2.19{\tiny$\pm$.05} \\
w/o depth channel & 27.52{\tiny$\pm$.06} & .145{\tiny$\pm$.001} & 27.84{\tiny$\pm$.46} & 0.63{\tiny$\pm$.01} & 21.34{\tiny$\pm$.05} & .250{\tiny$\pm$.002} & 43.09{\tiny$\pm$.46} & 2.32{\tiny$\pm$.10} \\
w/o attention localization & 22.42{\tiny$\pm$.07} & .318{\tiny$\pm$.003} & 40.55{\tiny$\pm$1.1} & 1.95{\tiny$\pm$.05} & 19.35{\tiny$\pm$.09} & .317{\tiny$\pm$.004} & 56.01{\tiny$\pm$.80} & 3.27{\tiny$\pm$.14} \\
\bottomrule
\end{tabular}
\end{table}

\subsection{Depth as an Optional Input}
\label{app:depth}

In our main evaluation, the depth channel $d_i(t)$ is derived from the ground-truth clip, as during training (Sec.~\ref{sec:setup}). To quantify how much the method relies on it, Table~\ref{tab:depth} compares each model with the same model whose trajectory tokens are re-encoded with depth set to zero, on both backbones and all six datasets. Removing depth costs 1--1.7\,dB PSNR on MoVi and MOT17. On Football, MOTSynth, and DAVIS the two variants are within 0.2\,dB PSNR, and with CogVideoX the depth-free variant even attains a slightly lower EPE on several datasets. Depth is therefore beneficial but optional, and our real-data models of Appendix~\ref{app:sav} use no depth at all.

\begin{table}[h]
\caption{\textbf{Effect of the depth channel} at the checkpoints of the main tables. Better value per dataset and backbone in bold.}
\label{tab:depth}
\centering
\footnotesize
\setlength{\tabcolsep}{2.5pt}
\begin{tabular}{ll cccc cccc}
\toprule
& & \multicolumn{4}{c}{\textbf{WaN 14B}} & \multicolumn{4}{c}{\textbf{CogV}} \\
\cmidrule(lr){3-6} \cmidrule(lr){7-10}
Dataset & Depth & PSNR$\uparrow$ & LPIPS$\downarrow$ & FVD$\downarrow$ & EPE$\downarrow$ & PSNR$\uparrow$ & LPIPS$\downarrow$ & FVD$\downarrow$ & EPE$\downarrow$ \\
\midrule
\multirow{2}{*}{MoVi-Extended} & with & \textbf{29.24} & \textbf{.131} & \textbf{23.39} & \textbf{0.57} & \textbf{29.77} & \textbf{.108} & \textbf{14.98} & 0.53 \\
& without & 27.56 & .145 & 27.66 & 0.62 & 28.84 & .118 & 15.07 & \textbf{0.52} \\
\cmidrule(l){1-10}
\multirow{2}{*}{Pool} & with & \textbf{33.07} & \textbf{.049} & \textbf{31.34} & \textbf{0.08} & \textbf{35.64} & \textbf{.025} & \textbf{8.79} & 0.07 \\
& without & 32.37 & .054 & 38.25 & 0.09 & 30.43 & .035 & 11.14 & \textbf{0.06} \\
\cmidrule(l){1-10}
\multirow{2}{*}{Football} & with & 24.27 & \textbf{.141} & \textbf{24.24} & \textbf{0.05} & 24.79 & \textbf{.090} & 13.87 & \textbf{0.03} \\
& without & \textbf{24.32} & .146 & 27.67 & 0.06 & \textbf{24.85} & .095 & \textbf{12.83} & \textbf{0.03} \\
\cmidrule(l){1-10}
\multirow{2}{*}{MOTSynth} & with & \textbf{21.39} & \textbf{.244} & \textbf{41.43} & \textbf{2.11} & 22.34 & .232 & \textbf{39.70} & 2.32 \\
& without & 21.32 & .250 & 43.11 & 2.28 & \textbf{22.50} & \textbf{.230} & 40.51 & \textbf{2.21} \\
\cmidrule(l){1-10}
\multirow{2}{*}{MOT17} & with & \textbf{19.94} & \textbf{.277} & \textbf{51.46} & \textbf{2.17} & \textbf{21.59} & \textbf{.265} & \textbf{53.92} & 2.37 \\
& without & 18.96 & .299 & 56.21 & 2.61 & 20.58 & .270 & 54.73 & \textbf{2.16} \\
\cmidrule(l){1-10}
\multirow{2}{*}{DAVIS} & with & \textbf{18.30} & \textbf{.301} & 88.37 & 2.31 & 18.39 & .267 & \textbf{89.39} & 2.43 \\
& without & 18.25 & \textbf{.301} & \textbf{88.27} & \textbf{2.30} & \textbf{18.42} & \textbf{.266} & 92.35 & \textbf{2.40} \\
\bottomrule
\end{tabular}
\end{table}

\clearpage

\section{Attention Localization Step by Step}
\label{app:algorithm}

Algorithm~\ref{alg:attn_loc} details the trajectory heatmap construction and the column replacement of Sec.~\ref{sec:xattn_replace} for explicit cross-attention. Its only inputs beyond the attention tensors are the prescribed trajectories and the position of each object token in the prompt: the column index $q_i$ is the position of the category token $o_i$ in the tokenized prompt of Eq.~\ref{eq:prompt}, and the initial location $(x_i^1, y_i^1)$ used by the appearance encoder is the first entry of $\tau_i$. Because $\sigma$ is defined in pixels, each Gaussian is rendered at video resolution and then resampled to the latent grid. The heatmaps are computed once per sample and shared by all heads, layers, and denoising steps. Each object occupies a single, independent column, so the cost of the replacement grows with the number of objects rather than with the video resolution. For joint self-attention, the column replacement is computed with the two-SDPA formulation of Eq.~\ref{eq:two_sdpa}.

\begin{algorithm}[h]

\caption{Attention localization at one cross-attention layer}
\label{alg:attn_loc}
\begin{algorithmic}[1]
\REQUIRE trajectories $\{(x_i(t), y_i(t))\}_{t=1}^{T}$ with per-frame visibility, for $i = 1, \dots, N$
\REQUIRE object-token columns $\{q_i\}_{i=1}^{N}$; standard deviation $\sigma$ (pixels)
\REQUIRE queries $\bQ_{\mathrm{vid}} \in \RR^{M \times d}$; text keys and values $\bK_{\mathrm{txt}}, \bV_{\mathrm{txt}} \in \RR^{L \times d}$
\ENSURE localized attention output $\bO' \in \RR^{M \times d}$
\item[] \textit{Heatmap construction} \hfill $\triangleright$ once per sample
\FOR{$i = 1$ \TO $N$}
  \FOR{each frame $t$ in which object $i$ is visible}
    \STATE render $g_i^t$, an isotropic Gaussian of std.\ $\sigma$ at $\big(x_i(t)\, W_p,\ y_i(t)\, H_p\big)$ on the $H_p \times W_p$ grid
    \STATE resize $g_i^t$ bilinearly to the $h \times w$ latent grid and normalize it to sum to one
  \ENDFOR
  \STATE interpolate $\{g_i^t\}$ linearly from $T$ video frames to $F$ latent frames; renormalize each frame
  \STATE set the latent frames in which object $i$ is absent to zero
  \STATE $\bh_i \leftarrow \mathrm{vec}\big(\{g_i^f\}_{f=1}^{F}\big) \in \RR^{M}$, \quad $M = F h w$
\ENDFOR
\item[] \vspace{-2pt}\textit{Column replacement} \hfill $\triangleright$ every layer, head, and denoising step
\STATE $\bA \leftarrow \mathrm{softmax}\big(d^{-1/2}\, \bQ_{\mathrm{vid}} \bK_{\mathrm{txt}}^{\top}\big) \in \RR^{M \times L}$
\FOR{$i = 1$ \TO $N$}
  \STATE $\bA[:, q_i] \leftarrow \bh_i$
\ENDFOR
\STATE $\bA[m, :] \leftarrow \bA[m, :] \,/ \sum_{k=1}^{L} \bA[m, k]$ \ for $m = 1, \dots, M$ \hfill $\triangleright$ Eq.~\ref{eq:col_replace}
\STATE \textbf{return} $\bO' \leftarrow \bA \bV_{\mathrm{txt}}$
\end{algorithmic}

\end{algorithm}

\end{document}